\documentclass[letterpaper]{article} 
\usepackage{aaai25}  
\usepackage{times}  
\usepackage{helvet}  
\usepackage{courier}  
\usepackage[hyphens]{url}  
\usepackage{graphicx} 
\usepackage{natbib}  
\usepackage{caption} 
\usepackage{algorithm}
\usepackage{algorithmicx}

\usepackage{amssymb}
\usepackage{amsmath}
\usepackage[scr=boondoxo,scrscaled=1.05]{mathalfa}
\usepackage{algpseudocode}
\usepackage{multirow}
\usepackage{bigstrut}
\usepackage{booktabs}

\usepackage{newfloat}
\usepackage{listings}
\DeclareCaptionStyle{ruled}{labelfont=normalfont,labelsep=colon,strut=off} 
\floatstyle{ruled}
\newfloat{listing}{tb}{lst}{}
\floatname{listing}{Listing}
\title{OTLRM: Orthogonal Learning-based Low-Rank Metric for Multi-Dimensional Inverse Problems}
\author{
    Xiangming Wang\textsuperscript{\rm 1},
    Haijin Zeng\textsuperscript{\rm 2}\thanks{Co-corresponding author},
    Jiaoyang Chen\textsuperscript{\rm 1},\\
    Sheng Liu\textsuperscript{\rm 3},
    Yongyong Chen\textsuperscript{\rm 1}\footnotemark[1],
    Guoqing Chao\textsuperscript{\rm 4}
}
\affiliations{
    \textsuperscript{\rm 1} Harbin Institute of Technology (Shenzhen),
    \textsuperscript{\rm 2} Gent University \\
    \textsuperscript{\rm 3} University of Electronic Science and Technology of China, 
    \textsuperscript{\rm 4} Harbin Institute of Technology (Weihai)\\
    Code: \url{https://github.com/xianggkl/OTLRM} \quad Email: xmwang28@gmail.com
}

\begin{document}

\maketitle

\begin{abstract}
In real-world scenarios, complex data such as multispectral images and multi-frame videos inherently exhibit robust low-rank property. 
This property is vital for multi-dimensional inverse problems, such as tensor completion, spectral imaging reconstruction, and multispectral image denoising. 
Existing tensor singular value decomposition (t-SVD) definitions rely on hand-designed or pre-given transforms, which lack flexibility for defining tensor nuclear norm (TNN). 
The TNN-regularized optimization problem is solved by the singular value thresholding (SVT) operator, which leverages the t-SVD framework to obtain the low-rank tensor. 
However, it is quite complicated to introduce SVT into deep neural networks due to the numerical instability problem in solving the derivatives of the eigenvectors.
In this paper, we introduce a novel data-driven generative low-rank t-SVD model based on the learnable orthogonal transform, which can be naturally solved under its representation. 
Prompted by the linear algebra theorem of the Householder transformation, our learnable orthogonal transform is achieved by constructing an endogenously orthogonal matrix adaptable to neural networks, optimizing it as arbitrary orthogonal matrices. 
Additionally, we propose a low-rank solver as a generalization of SVT, which utilizes an efficient representation of generative networks to obtain low-rank structures. 
Extensive experiments highlight its significant restoration enhancements.
\end{abstract}

\section{Introduction}

Real-world multi-dimensional data, such as multispectral images (MSIs), videos, and Magnetic Resonance Imaging (MRI) data, are usually affected by unpredictable factors during capture and transmission.
To reconstruct the original tensors, many tasks are extended under different observations, such as Tensor Completion (TC) \cite{qin2022low,mai2022deep}, spectral imaging \cite{cai2022degradation}, and spectral denoising \cite{wang2022learning}.

Current approaches have achieved outstanding results benefiting from the low-rank nature of the original data tensor.
However, the tensor rank is still not well defined.
Various definitions of the tensor rank are not deterministic and all have specific advantages and limitations.
For example, the CANDECOMP PARAFAC (CP) rank \cite{carroll1970analysis} is defined as the number of rank-$1$ tensors obtained by the CP decomposition.
Since computing the CP decomposition of a tensor is NP-hard and finding its accurate convex approximation is challenging, CP rank may not be appropriate to give solutions tailored to practical application areas.
Additionally, Tucker rank \cite{tucker1966some,sun2022fast} is based on the tensor unfolding scheme, which unfolds the tensor into matrices along different dimensions.
As a result, it leads to a broken structure of the original tensor, and such an unfolding describes the correlation between only one mode of the tensor and all other modes (one mode versus the rest), which may bring undesirable results to the tensor reconstruction.
Overall, how to define an appropriate tensor rank in different tensor decompositions is a problem worth discussing and analyzing.

With the tensor-tensor product (t-product) \cite{kernfeld2015tensor} gradually becoming a comprehensive approach for tensor multiplication, the tensor Singular Value Decomposition (t-SVD) \cite{zhang2014novel, lu_TNN, shengLiu_revisiting} which constructs the original tensor as the t-product of two orthogonal tensors and an $f$-diagonal tensor, has received widespread attention and research.
Explicitly, the t-product is defined in a transform domain based on an arbitrary invertible linear transform.
Given a tensor $\mathcal{X} \in \mathbb{R}^{n_1 \times n_2 \times n_3}$ and a transform matrix $\mathbf{L} \in \mathbb{R}^{n_3 \times n_3}$, the transformed tensor $L(\mathcal{X})$ can be formulated as
\begin{equation}
    L(\mathcal{X}) = \mathcal{X} \times_3 \mathbf{L},
\end{equation}
where $\times_3$ denotes the mode-3 tensor product.
Following the definition, t-SVD can be defined in the transform domain, which captures the low-rank structure of the tensor.
\textit{Based on some fixed linear invertible transforms such as Discrete Fourier Transform (DFT) \cite{zhang2016exact} and Discrete Cosine Transform (DCT) \cite{Lu_2021_ICCV}, current transforms have theoretical guarantees, yet are fixed and not data-adaptive, which cannot fit different data instances well.}

Under various t-SVD definitions of the transform domain, there are many algorithms and models for solving the low-rank problem.
Since directly minimizing the tensor tubal rank of a tensor is an NP-hard problem, the tensor nuclear norm (TNN) \cite{zhang2014novel} was proposed as its convex approximation, which is
\begin{equation}
    \Vert \mathcal{X} \Vert_{\mathbf{L},*} := \sum_{i=1}^{n_3} \Vert L(\mathcal{X})(:,:,i) \Vert_*,
\end{equation}
where $\Vert \cdot \Vert_*$ is the matrix nuclear norm and $\Vert \cdot \Vert_{\mathbf{L},*}$ is the tensor nuclear norm based on the transform $\mathbf{L}$.
To solve the matrix nuclear norm, the singular value thresholding (SVT) operator was proposed \cite{Caisvt}, which can be formulated as:
\begin{equation}
    \mathrm{SVT}_{\gamma}(\mathbf{X}) = \mathbf{U}\mathbf{S}_{\gamma}\mathbf{V}^{\mathbf{T}},
\end{equation}
where $\mathbf{S}_{\gamma} = max\{ \mathbf{S} - \gamma, 0 \}$.
With the assistance of SVT, recent models utilizing optimization algorithms \cite{wang2022learning, qin2022low} and deep unfolding networks \cite{mai2022deep} have demonstrated promising outcomes. However, training neural networks based on SVD \cite{Ionescu_2015_ICCV, wang2023transformed} poses challenges due to its complexity and numerical instability when handling derivatives of eigenvectors. This can result in sub-optimal convergence during the training process. Despite the robust differentiability of eigenvectors, the partial derivatives may become large in cases where two eigenvalues are equal or nearly equal \cite{WangRobustSVD}.

To achieve a data-adaptive and theoretically invertible transform, fostering compatibility with deep networks, we introduce a novel approach: the learnable Orthogonal Transform-induced generative Low-Rank t-SVD Model (OTLRM).
Compared with existing t-SVD methods, our OTLRM has two merits: 
(1) \textbf{Orthogonality and data adaptability.} We construct the learnable orthogonal transform $\mathbf{L}$ by several learnable Householder transformations, possessing inherent orthogonality and adaptability. It enables the adjustment of the transform for each dataset while maintaining a theoretical guarantee for the requirement of ``arbitrary invertible linear transform". While current t-SVD methods (such as DTNN \cite{kong2021tensor} and Tensor $Q$-rank \cite{DTNN_jiang}) can only accommodate one or the other. (2) \textbf{Generative framework.} Under the t-SVD representation, OTLRM can directly generate the expected tensor with the guidance of the observations within the DNN optimization framework. While others decompose the target tensor by SVD and truncate the singular values.
Especially, we introduce a dense rank estimation operator, which stores and enriches the rank information of each band in the transform domain.
Primary contributions are outlined as follows:

\begin{itemize}
    \item Prompted by the linear algebra theorem of the Householder transformation, we construct a learnable endogenously orthogonal transform into the neural network.
    Different from the predefined orthogonal transform, the proposed learnable endogenously orthogonal transform can be naturally embedded in the neural network, which has more flexible data-adaptive capability.
    \item With the endogenously orthogonal transform, we propose a low-rank solver as a generalization of SVT, which utilizes efficient t-SVD representation to obtain low-rank structures. 
    In contrast to SVT-induced low-rank algorithms, OTLRM is solved by gradient descent-based algorithms within the DNN optimization framework.
    \item We conduct extensive experiments on several tasks: TC, snapshot compressive imaging, spectral denoising, under three types of datasets: MSIs, videos and MRI data.
    Abundant and superior experimental results validate the effectiveness of our method.
\end{itemize}

\section{Related works}
\noindent\textbf{Transform-based t-SVD.}
Inspired by TNN \cite{zhang2014novel} in the Fourier domain, many t-SVD algorithms have been proposed.
Specifically, Zhang et al. proposed the DFT-based t-SVD method \cite{zhang2016exact}.
Zhou et al. \cite{zhou2019bayesian} introduced the Bayesian version of tensor tubal rank to automatically determine the tensor rank.
The weighted TNN \cite{mu2020weighted} is also proposed to distinguish different tensor singular values under Fast Fourier Transform (FFT).
Considering that t-product is defined on any invertible linear transforms, other transforms are explored.
For example, Lu et al. \cite{lu2019low} proposed the DCT-induced TNN with theoretical guarantees for exact recovery. 
While Song et al. \cite{TTNN_song} employed unitary transform matrices captured from the original datasets.
However, fixed transforms can not be inherently suitable for current data instances.
Therefore, Jiang et al. \cite{DTNN_jiang} proposed a dictionary-based TNN (DTNN), which constructed a learnable data-adaptive dictionary as the transform.
Additionally, Kong et al. \cite{kong2021tensor} gave a definition of the tensor rank with a data-dependent transform, based on the learnable matrix $\mathbf{Q}$.
Recently, Luo et al. \cite{lrtf} induced DNN as the transform and generated the transformed tensor by gradient descent algorithms.
Due to the nonlinear transform used in the DNN, it does not fulfill the theoretical requirements for arbitrary invertible linear transforms.

\noindent\textbf{SVT in Neural Network.}
Cai et al. proposed the SVT algorithm for matrix completion \cite{Caisvt}.
To implement the SVT algorithm into the deep learning network, Ionescu et al. gave a sound mathematical apparatus and derived the gradient propagation formulas for SVD in deep networks \cite{Ionescu_2015_ICCV,IonescuTraining}. 
However, although robustly calculating the derivatives of the SVD gradient propagation is direct, it becomes numerically unstable during the calculation of certain partial derivatives.
Wang et al. introduced a Taylor expansion-based approach to compute the eigenvector gradients \cite{WangRobustSVD} and Song et al. induced the orthogonality loss to improve the generalization abilities and training stability of the SVD \cite{SongImproSVD}.
These approaches involve integrating the matrix-based SVT directly into deep neural networks, which still introduces complex SVD to solve low-rank problems.

\section{Notations and Preliminaries}
\subsection{Notations}
In this paper, scalars, vectors, matrices, and tensors are denoted respectively as lowercase letters, e.g. $a$, boldface lowercase letters, e.g. $\mathbf{a}$, boldface capital letters, e.g. $\mathbf{A}$ and boldface Calligraphy letters, e.g. $\mathcal{A}$. $\mathcal{A}^{(i)}$ represents the $i$-th frontal slice of the tensor $\mathcal{A}$.

\subsection{Orthogonal Transform}
Orthogonal transform could maintain the original orthogonality between vectors without losing fine-level details. Based on the following linear algebra theorem, an orthogonal matrix $\mathbf{L}$ is constructed, which satisfies $\mathbf{L}^{\mathbf{T}}\mathbf{L}=\mathbf{L}\mathbf{L}^{\mathbf{T}}=\mathbf{I}$ and $\mathbf{I}$ is the identity matrix.

\noindent\textbf{Theorem 1.} \cite{uhlig2001constructive} \textit{Every real orthogonal $n \times n$ matrix $\mathbf{L}$ is the product of at most $n$ real orthogonal Householder transformations. And this also is true for complex unitary $\mathbf{L}$ and complex Householders.}

Following the lines of Theorem 1, we can randomly generate a parameter matrix $\mathbf{W} \in \mathbb{R}^{n \times n}$ containing $n$ column vectors $\mathbf{w}_{i} \in \mathbb{R}^{n \times 1}$, and construct $n$ orthogonal Householder transformations as below:
\begin{equation}
\label{F}
    \mathbf{F}_i = \mathbf{I} - 2 \frac{\mathbf{w}_i\mathbf{w}_i^{\mathbf{T}}}{\Vert \mathbf{w}_i \Vert^{2}},
\end{equation}
where $1 \leq i \leq n$.
Then the orthogonal matrix $\mathbf{L}$ is represented as the product of the $n$ orthogonal Householder transformations $\mathbf{F}_i$:
\begin{equation}
\label{h_equation}
    \mathbf{L} = \mathbf{F}_1 \mathbf{F}_2 \cdots \mathbf{F}_{n}\\
    = \prod\limits_{i=1}^{n}(\mathbf{I} - 2 \frac{\mathbf{w}_i\mathbf{w}_i^{\mathbf{T}}}{\Vert \mathbf{w}_i \Vert^{2}}),
\end{equation}
and it is worth noting that $\mathbf{W}$ is the parameter matrix to be optimized and $\mathbf{L}$ is the endogenous orthogonal matrix which is another form of $\mathbf{W}$ based on the Householder transformation operations of Eq. \eqref{F} and Eq. \eqref{h_equation}.

In contrast to approaches that attain orthogonality by incorporating an orthogonality metric into the loss function, matrices formed through Householder transformations possess inherent orthogonality and learnability.
This intrinsic property eliminates the need for external orthogonal constraints and ensures identical learnable parameters as a conventional transform matrix. 
Specifically, for optimizing the orthogonal transform, this chain multiplication enables the loss to be derived parallelly for each column vector $\mathbf{w}_i$.
And all the optimization can be automatically solved by the built-in differentiation engine in PyTorch, facilitating straightforward integration into deep neural networks.

\subsection{$L$-transformed Tensor Singular Value Decomposition}
Distinguished from matrix SVD, t-SVD is based on the tensor-tensor product, which is defined directly in a transform domain with an arbitrary invertible linear transform $\mathbf{L}$ \cite{kernfeld2015tensor}.

\noindent\textbf{Definition 1. Mode-$3$ tensor-matrix product \cite{kolda2009tensor}}
    For any third-order tensor $\mathcal{A} \in \mathbb{R}^{n_1 \times n_2 \times n_3}$ with  a matrix $\mathbf{U} \in \mathbb{R}^{n \times n_3}$, the mode-$3$ tensor-matrix product is defined as 
    \begin{equation}
    \label{mode3_equation}
            \hat{\mathcal{A}}=\mathcal{A}\times _3\mathbf{U}\Leftrightarrow \hat{\mathbf{A}}_{\left( 3 \right)}=\mathbf{UA}_{\left(3 \right)},     
    \end{equation}
    where  $\hat{\mathcal{A}} \in \mathbb{R}^{n_1 \times n_2\times n}$,  $\mathbf{A}_{(3)}$ and $\hat{\mathbf{A}}_{(3)}$ are mode-$3$ matricization of $\mathcal{A}$ and $\hat{\mathcal{A}}$,  respectively. 

\noindent\textbf{Definition 2. Tensor-Tensor face-wise product \cite{kernfeld2015tensor}}
    Given two tensors $\mathcal{A} \in \mathbb{R}^{n_1 \times \ell \times n_3}$ and $\mathcal{B} \in \mathbb{R}^{\ell \times n_2 \times n_3}$, the face-wise product of $\mathcal{A}$ and $\mathcal{B}$ is defined as
\begin{equation}
    (\mathcal{A} \triangle \mathcal{B})^{(i)} = \mathcal{A}^{(i)} \mathcal{B}^{(i)}, 
\end{equation}
where $\mathcal{A}^{(i)}$ is $i$-th frontal slice of   $\mathcal{A}$. 

\noindent\textbf{Definition 3. Tensor-tensor product in $\mathbf{L}$-transform domain \cite{kernfeld2015tensor}}
    Define $\ast_{L}$ as $\mathbb{R}^{n_1 \times \ell \times n_3} \times \mathbb{R}^{\ell \times n_2 \times n_3} \rightarrow \mathbb{R}^{n_1 \times n_2 \times n_3}$, we have the tensor-tensor product:
\begin{equation}
\label{t-product_equation}
    \mathcal{A} \ast_{L} \mathcal{B} = L^{-1}(L(\mathcal{A}) \triangle L(\mathcal{B})), 
\end{equation}
where $L(\mathcal{A}) = \mathcal{A} \times _3 \mathbf{L}$ and $L^{-1}(\cdot)$ is the inverse transform operator of $L(\cdot)$.

\noindent\textbf{Definition 4. Special tensors \cite{braman2010third,kilmer2013third}}
\textbf{Identity tensor}: for an identity tensor $\mathcal{I}$, its every frontal slice is an identity matrix. 
\textbf{$f$-diagonal tensor}: for an $f$-diagonal tensor $\mathcal{S}$, its every frontal slice is a diagonal matrix.
\textbf{Orthogonal tensor}: if the tensor $\mathcal{U}$ satisfies that $\mathcal{U}^{\mathbf{T}} \ast_L \mathcal{U} = \mathcal{U} \ast_L \mathcal{U}^{\mathbf{T}} = \mathcal{I}$, it is called orthogonal tensor.
\textbf{Semi-orthogonal tensor}: if the tensor $\mathcal{U}$ satisfies $\mathcal{U}^{\mathbf{T}} \ast_{L} \mathcal{U} = \mathcal{I}$, it is called semi-orthogonal tensor.

Subsequently, the definition of the t-SVD is expressed in terms of the $\mathbf{L}$-transform and its inverse.

\noindent\textbf{Lemma 1. Tensor singular value decomposition (t-SVD) \cite{braman2010third,kilmer2011factorization,kilmer2013third}}
    Given a tensor $\mathcal{X} \in \mathbb{R}^{n_1 \times n_2 \times n_3}$, the t-SVD can be formulated as
\begin{equation}
    \mathcal{X} = \mathcal{U} \ast_{L} \mathcal{S} \ast_{L} \mathcal{V}^{\mathbf{T}},
\end{equation}
where $\mathcal{U} \in \mathbb{R}^{n_1 \times n_1 \times n_3}$, $\mathcal{V} \in \mathbb{R}^{n_2 \times n_2 \times n_3}$ are orthogonal tensors and $\mathcal{S} \in \mathbb{R}^{n_1 \times n_2 \times n_3}$ is $f$-diagonal.

\noindent\textbf{Definition 5. Tensor tubal-rank \cite{zhang2014novel}}
    The tensor tubal-rank $r$ of the target tensor $\mathcal{X}$ is defined as the number of the non-zero singular tubes of the $f$-diagonal tensor $S$, which is
\begin{equation}
    rank(\mathcal{X}) = \#\{i, \mathcal{S}(i,i,:) \neq 0\}.
\end{equation}
And an alternative definition is that the tenor tubal-rank of $\mathcal{X}$ is the largest rank of every frontal slice of the $L$-transformed tensor $L(\mathcal{X})$.

\noindent\textbf{Remark 1. Skinny t-SVD \cite{kilmer2013third,zhang2016exact}}
\label{remark_skinnySVD}
Given a tensor $\mathcal{X} \in \mathbb{R}^{n_1 \times n_2 \times n_3}$ which has tensor tubal-rank $r$, it’s more efficient to compute the skinny t-SVD. 
And the decomposition can be reformulated as
\begin{equation}
    \mathcal{X} = \mathcal{U} \ast_{L} \mathcal{S} \ast_{L} \mathcal{V}^{\mathbf{T}},
\end{equation}
where $\mathcal{U} \in \mathbb{R}^{n_1 \times r \times n_3}$, $\mathcal{V} \in \mathbb{R}^{n_2 \times r \times n_3}$ are semi-orthogonal tensors, $\mathcal{S} \in \mathbb{R}^{r \times r \times n_3}$ is $f$-diagonal and especially, $\mathcal{U}^{\mathbf{T}} \ast_{L} \mathcal{U} = \mathcal{I}$, $\mathcal{V}^{\mathbf{T}} \ast_{L} \mathcal{V} = \mathcal{I}$.

\noindent\textbf{Lemma 2. Tensor Singular Value Thresholding (t-SVT) \cite{wang2022learning}}
    Let $\mathcal{X}=\mathcal{U} \ast_{L} \mathcal{S} \ast_{L} \mathcal{V}^{\mathbf{T}}$ be the t-SVD for tensor $\mathcal{X} \in \mathbb{R}^{n_1 \times n_2 \times n_3}$.
    The t-SVT operator $\delta$ is:
    \begin{equation}
        \delta_{\gamma}(\mathcal{X})=\mathcal{U} \ast_{L} \mathcal{S}_{\gamma} \ast_{L} \mathcal{V}^{\mathbf{T}},
    \end{equation}
    where $\mathcal{S}_{\gamma}=L^{-1}(\max\{ L(\mathcal{S}) - \gamma, 0 \})$ and $\gamma$ is the threshold value which controls the degree of the rank restriction.

\section{Learnable Endogenous Orthogonal Transform based Generative t-SVD Low-rank Model}
Suppose that $\mathcal{Y} \in \mathbb{R}^{n_1 \times n_2 \times n_3}$ denotes the unknown pure tensor (real scene or video), $\mathcal{X} \in \mathbb{R}^{n_1 \times n_2 \times n_3}$ is the desired low-rank tensor and $\mathbf{H}(\cdot)$ is defined as the capture operation that obtains the source data (hence, $\mathbf{H}(\mathcal{Y})$ is the observed measurement).
Based on the definition of transposition and orthogonality of the transform of the t-product, the desired low-rank tensor can be generated by the skinny low-rank t-SVD representation, which is $\mathcal{X} = \mathcal{U} \ast_{L} \mathcal{S} \ast_{L} \mathcal{V}^{\mathbf{T}}$.
With the endogenous orthogonal transform $\mathbf{L} \in \mathbb{R}^{n_3 \times n_3}$ (and the parameter matrix to be optimized is $\mathbf{W}$) which naturally satisfies $\mathbf{L}^{\mathbf{T}}\mathbf{L}=\mathbf{L}\mathbf{L}^{\mathbf{T}}=\mathbf{I}$, our generative t-SVD low-rank optimization model can be formulated as
\begin{equation}
    \min_{\mathcal{U}, \mathcal{V}, \mathcal{S}, \mathbf{W}} \phi(\mathbf{H}(\mathcal{X}), \mathbf{H}(\mathcal{Y})), \text{s.t.} \mathcal{X} = \mathcal{U} \ast_{L} \mathcal{S} \ast_{L} \mathcal{V}^{\mathbf{T}},
\end{equation}
where $\phi(\cdot)$ is the fidelity loss function which can be adjusted according to the target application, $\mathcal{U} \in \mathbb{R}^{n_1 \times r \times n_3}$, $\mathcal{V} \in \mathbb{R}^{n_2 \times r \times n_3}$ and $\mathcal{S} \in \mathbb{R}^{r \times r \times n_3}$. And explicitly, $\ast_{L}$ denotes the orthogonal transform $\mathbf{L}$-based tensor-tensor product according to Eq. \eqref{h_equation} and Eq. \eqref{t-product_equation}.

Since $\mathcal{S}$ is an $f$-diagonal tensor, we construct the rank tensor by a transformed matrix $\mathbf{S} \in \mathbb{R}^{n_3 \times r}$ instead, with the diagonalization $Diag(\cdot)$.
It is reformulated as (please refer to the supplementary material for details):
\begin{equation}
\label{equation_diag}
    \min_{\mathcal{U}, \mathcal{V}, \mathbf{S}, \mathbf{W}} \phi(\mathbf{H}( L^{-1}(L(\mathcal{U}) \triangle Diag(\mathbf{S}) \triangle L(\mathcal{V})^{\mathbf{T}}) ), \mathbf{H}(\mathcal{Y})).
\end{equation}

\subsection{Dense Rank Estimation Operator}
Traditionally, t-SVT is adopted to truncate and restrict the rank matrix by the soft thresholding operator.
However, controlling the low-rank degree of the generated tensor $\mathcal{X}$ by simply adjusting the rank $r$ and the threshold value $\gamma$ is coarse, which does not well exploit the data adaptability in DNNs.
Inspired by the threshold contraction and to find the most suitable tensor tubal rank, we inject a DNN-induced dense rank estimation operator $\rho(\cdot)$ into the optimal solution of the factor $\mathcal{S}$, which captures and enriches the rank information.
Specifically, $\rho(\cdot)$ can be viewed as a rank information extractor.
In this paper, we use the fully connected layer for experiments.
Given the rank matrix $\mathbf{S}$ as input, the dense rank estimation operator can be formulated as:
\begin{equation}
\label{rank_estimation_operator}
    \rho(\mathbf{S}) = LReLU \cdots (LReLU(\mathbf{S} \times_3 \mathbf{G}_1) \times_3 \cdots) \times_3 \mathbf{G}_k,
\end{equation}
where $k$ denotes the number of the layers, $LReLU$ is the LeakyReLU \cite{he2015delving} function and each $\mathbf{G}$ is the learnable rank feature matrix.
The optimization model is reformulated as
\begin{equation}
\label{*_equation}
    \min_{\mathcal{U}, \mathcal{V}, \mathbf{S}, \mathbf{W}} \phi(\mathbf{H}( L^{-1}(L(\mathcal{U}) \triangle Diag(\rho(\mathbf{S})) \triangle L(\mathcal{V})^{\mathbf{T}}) ), \mathbf{H}(\mathcal{Y})).
\end{equation}

\subsection{Orthogonal Total Variation}
To improve the generative capability and get a more suitable transform, we adopt the Orthogonal Total Variation (OTV) constraint for $\Theta = \{ L(\mathcal{U}), L(\mathcal{V}), \theta_{L^{-1}} \}$, where $\theta_{L^{-1}}$ denotes the weight matrix of the $L^{-1}(\cdot)$ module, aiming to enhance the local smoothness prior structure, which is
\begin{equation}
    \mathrm{OTV}(\Theta) = \Vert \nabla_x L(\mathcal{U}) \Vert_{\mathscr{l}_1} + \Vert \nabla_y L(\mathcal{V})^{\mathbf{T}} \Vert_{\mathscr{l}_1} + \Vert \nabla_x \theta_{L^{-1}} \Vert_{\mathscr{l}_1}.
\end{equation}
As a result, the final optimization model is
\begin{equation}
\label{final_equation}
\begin{aligned}
    \min_{\mathcal{U}, \mathcal{V}, \mathbf{S}, \mathbf{W}} &\phi(\mathbf{H}( L^{-1}(L(\mathcal{U}) \triangle Diag(\rho(\mathbf{S})) \triangle L(\mathcal{V})^{\mathbf{T}}) ), \mathbf{H}(\mathcal{Y}))
    \\ &+ \lambda \mathrm{OTV}(\Theta),
\end{aligned}
\end{equation}
where $\lambda$ is a trade-off parameter, and the complete algorithm is depicted in Algorithm \ref{ours_algorithm}.
\begin{algorithm}[H]
	\renewcommand\arraystretch{1.0}
	\caption{The Proposed OTLRM Algorithm. } 
	\begin{algorithmic}[1] 
		\renewcommand{\algorithmicrequire}{\textbf{Input:}}
		\renewcommand{\algorithmicensure}{\textbf{Output:}}
		\Require
		The coarse estimated rank $r$, hyperparameter $\lambda$, and the maximum iteration $t_{max}$. 
		\Ensure The reconstructed tensor $\mathcal{X}\in\mathbb{R}^{n_1\times n_2\times n_3}$.
		\State  {\textbf{Initialization:}} The iteration $t = 0$. 
		\While {$t<t_{max}$}
            \State Compute $\mathbf{L}$ via Eq. \eqref{h_equation}; 
            \State Compute $\hat{\mathcal{U}}$ via $\hat{\mathcal{U}} = L(\mathcal{U})$;
            \State Compute $\hat{\mathcal{V}}$ via $\hat{\mathcal{V}} = L(\mathcal{V})$;
            \State Compute $\hat{\mathcal{S}}$ via $\hat{S}=Diag(\rho(\mathbf{S}))$;
            \State Compute the loss  via Eq. \eqref{final_equation};
            \State Perform gradient backpropagation;
		\EndWhile \\
        Get the final low-rank tensor $\mathcal{X} = L^{-1}(\hat{\mathcal{U}} \triangle \hat{\mathcal{S}} \triangle \hat{\mathcal{V}}^{\mathbf{T}})$.
	\end{algorithmic}
	\label{ours_algorithm}
\end{algorithm}

\section{Applications for The Proposed Model}
\subsection{Tensor Completion}
Given the unknown pure tensor $\mathcal{Y} \in \mathbb{R}^{n_1 \times n_2 \times n_3}$, tensor completion aims to recover underlying data from the observed entries $\Omega = \{ (i_1,i_2,i_3) | \zeta_{i_1,i_2,i_3} = 1 \}$, which follows the Bernoulli sampling scheme $\Omega \sim \mathbf{Ber}(p)$.
And $p$ is the probability of taking target Bernoulli variables with independent and identically distributed.
Based on the above definition, $\mathbf{H}(\cdot)$ can be specified as $\mathbf{H}_{\Omega}(\cdot) : \mathbb{R}^{n_1 \times n_2 \times n_3} \rightarrow \mathbb{R}^{n_1 \times n_2 \times n_3}$, which keeps the entries in $\Omega$ fixed and sets the rest zero.
The loss function of $\phi(\cdot)$ can be formulated as
\begin{equation}
\label{loss_tc_equation}
    \phi(\mathcal{X}, \mathcal{Y}) = \Vert \mathbf{H}_{\Omega}(\mathcal{X}) - \mathbf{H}_{\Omega}(\mathcal{Y}) \Vert^2_F.
\end{equation}

\subsection{MSI reconstruction in CASSI system}
Coded aperture snapshot compressive imaging (CASSI), which aims at scanning scenes with spatial and spectral dimensions, has achieved impressive performance. With a coded aperture and a disperser, it encodes and shifts each band of the original scene $\mathcal{Y} \in \mathbb{R}^{n_1 \times n_2 \times n_3}$ with known mask $\mathbf{M} \in \mathbb{R}^{n_1 \times n_2}$ and later blends all the bands to generate a 2-D measurement $\mathbf{X} \in \mathbb{R}^{n_1 \times (n_2 + d \times (n_3 - 1))}$, where $d$ is the shift step. Considering the measurement noise $\mathbf{N} \in \mathbb{R}^{n_1 \times (n_2 + d \times (n_3 - 1))}$ generated in the coding system, the whole process can be formulated as
\begin{equation}
    \mathbf{X} = \sum_{k = 1}^{n_3} shift(\mathcal{Y}(:,:,k) \odot \mathbf{M}) + \mathbf{N}.
\end{equation}
For convenience, by the definition of the $\mathbf{H}(\cdot)$, the above operations can be simplified to the following formula:
\begin{equation}
    \mathbf{X} = \mathbf{H}(\mathcal{Y}) + \mathbf{N}, 
\end{equation}
where for MSI reconstruction, the $\mathbf{H}(\cdot)$ operator is defined as $\mathbb{R}^{n_1 \times n_2 \times n_3} \rightarrow \mathbb{R}^{n_1 \times (n_2 + d \times (n_3 - 1))}$ in CASSI application.
And the loss function of $\phi(\cdot)$ can be formulated as
\begin{equation}
\label{loss_cassi_equation}
    \phi(\mathbf{X}, \mathcal{Y}) = \Vert \mathbf{X} - \mathbf{H}(\mathcal{Y}) \Vert^2_F.
\end{equation}
\subsection{MSI Denoising}
The purpose of MSI denoising is to recover clean MSIs from noise-contaminated observations.
In that case, the $\mathbf{H}(\cdot)$ operator can be defined as a noise-adding operation.
The loss function of $\phi(\cdot)$ can be formulated as
\begin{equation}
\label{loss_msidenoising_equation}
    \phi(\mathcal{X}, \mathcal{Y}) = \Vert \mathcal{X} - \mathbf{H}(\mathcal{Y}) \Vert_{\mathscr{l}_1}.
\end{equation}
\subsection{Computational Complexity Analysis}
Suppose that the target tensor $\mathcal{X} \in \mathbb{R}^{n_1 \times n_2 \times n_3}$ and the orthogonal transform $\mathbf{L} \in \mathbb{R}^{n_3 \times n_3}$.
For the process of the transform construction, the computational complexity is $O((n_3)^4)$.
For the transformed tensors $\hat{\mathcal{U}},\hat{\mathcal{V}}$, the computational complexity is $O(n_1 r (n_3)^2) + O(r n_2 (n_3)^2)$.
For the $\hat{\mathcal{S}}$, given the number of DNN layers $k$, the computational complexity is $O(n_3 r^2 k)$.
And for the product of the $\hat{\mathcal{U}},\hat{\mathcal{V}},\hat{\mathcal{S}}$, the computational complexity is $O(n_3(n_1 r^2 n_2))$.
Thus, the computational complexity of our method is $O((n_3)^4) + O(n_1 r (n_3)^2) + O(r n_2 (n_3)^2) + O(n_3 r^2 k) + O(n_3(n_1 r^2 n_2))$.
Due to the fact that $r << \min\{n_1,n_2\}$, it can be simplified as $O( (n_3)^4 + (n_1+n_2) r (n_3)^2 + n_1 n_2 n_3 r^2 )$.

\section{Experiments}

\subsection{Datasets and Settings}
In this section, we introduce the datasets (videos, MSIs and MRI data) and the experiment settings used in three multi-dimensional inverse problems.
Please refer to the supplementary materials for MRI completion experiments.

\noindent\textbf{Overall Settings:}
All the experiments are implemented in PyTorch and conducted on one NVIDIA GeForce RTX 3090 GPU with 20GB RAM.
Adam \cite{adam} is used to optimize Eq. \eqref{*_equation} and Eq. \eqref{final_equation}.
For the rank estimation module $\rho(\cdot)$, we used a two-layer DNN (composed of two linear transforms and a LeakyReLU \cite{he2015delving} function) in all experiments (the $k$ in Eq. \eqref{rank_estimation_operator} is 2).
For the initialization of the learnable parameters in the model, we follow the typical kaiming initialization \cite{he2015delving} and please refer to the supplementary material for the initialization ablation experiments.
It is noteworthy that to enhance the model's data adaptability, we employ the endogenously orthogonal transforms $\mathbf{L}_1$ and $\mathbf{L}_2$ to operate on $\mathcal{U}$ and $\mathcal{V}$, respectively, while $\mathbf{L}_3$ serves as the inverse.

\noindent\textbf{Tensor Completion:}
The evaluation encompasses four MSIs sourced from the CAVE database~\footnote{The data is available at \url{https://www.cs.columbia.edu/CAVE/databases/multispectral/}}, namely \emph{Balloon}, \emph{Beer}, \emph{Pompom} and \emph{Toy}, along with two videos obtained from the NTT database~\footnote{The data is available at \url{http://www.brl.ntt.co.jp/people/akisato/saliency3.html}.}, labeled \emph{Bird} and \emph{Horse}. Each MSI is resized to dimensions of $256 \times 256 \times 31$, while the videos have a spatial size of $288 \times 352$. In our experiments, we utilize the initial 30 frames of each video.
Regarding the MSI datasets, we employ sampling rates (SRs) of 0.05, 0.10, and 0.15, while for the video datasets, the sampling rates are set to 0.10, 0.15, and 0.20.

\noindent\textbf{MSI Reconstruction in CASSI system:}
We select five datasets (\emph{scene01}-\emph{scene05}, $256 \times 256 \times 31$) in KAIST~\footnote{ The data is available at \url{https://vclab.kaist.ac.kr/siggraphasia2017p1/kaistdataset.html}} \cite{choi2017high} for simulation.
The shift step $d$ is 2, which means that the measurement is of size $256 \times 310$.

\noindent\textbf{MSI Denoising:}
We select three scenes (\emph{scene01}, \emph{scene02} and \emph{scene10}) in KAIST \cite{choi2017high} database (all of size $256 \times 256 \times 31$) for testing.
And the noise cases are set for Gaussian noise with standard deviations $0.2$ and $0.3$.

\subsection{Comparisons with State-of-the-Arts}
In this section. we compare our method with state-of-the-arts.
The \textbf{best} and \underline{second-best} are highlighted in bold and underlined, respectively.
\textbf{OTLRM*} represents the model without $\mathrm{OTV}$ loss, and \textbf{OTLRM} denotes the whole model.

\noindent\textbf{Evaluation Metrics:}
For numerical comparison, we use peak signal-to-noise ratio (PSNR) and structural similarity (SSIM) as metrics in all three problems.
And feature similarity index measure (FSIM) is added for MSI denoising problem.
The higher PSNR, SSIM, and FSIM the better.

\noindent\textbf{Tensor Completion:}
Especially, we compare with other state-of-the-art methods, including TNN \cite{lu_TNN}, TQRTNN \cite{TQRTNN}, UTNN \cite{TTNN_song}, DTNN \cite{DTNN_jiang}, LS2T2NN \cite{liu2023learnable} and HLRTF \cite{lrtf}.

Table \ref{msitableresult} and \ref{videotableresult} shows the numerical results of our method and the state-of-the-arts with 1 to 4 dB improvement in PSNR, especially in low sampling rate.
Compared to the traditional TNN-based methods, our model demonstrates excellent results of learnable orthogonal transforms and the generative t-SVD model, which maintains flexible data-adaption low-rank property.
From Figures 1 and 2 in Appendix, we can observe that with $\mathrm{OTV}$ loss our method is smoother and cleaner than other SOTA methods. 

\begin{table*}[htb]
        \centering
        \caption{Evaluation PSNR, SSIM and Time on \emph{CAVE} dataset of \textbf{tensor completion} results by different methods for \textbf{MSIs} under different SRs. Top Left:\emph{Balloons}, Bottom Left:\emph{Beer}; Top Right:\emph{Pompom}, Bottom Right:\emph{Toy}.}

        \def\arraystretch{1.0}
        \setlength{\tabcolsep}{1.7pt}
\scalebox{0.7}{
    \begin{tabular}{cc|ccccccc|}
    \bottomrule[0.15em]
    \multirow{2}[2]{*}{\textbf{Method}} & \multirow{2}[2]{*}{\textbf{Reference}} & \multicolumn{2}{c}{\textbf{SR=0.05}} & \multicolumn{2}{c}{\textbf{SR=0.10}} & \multicolumn{2}{c}{\textbf{SR=0.15}} & \multirow{2}[2]{*}{\textbf{Time (s)}} \\
          &  & \textbf{PSNR}$\uparrow$  & \textbf{SSIM}$\uparrow$  & \textbf{PSNR}$\uparrow$  & \textbf{SSIM}$\uparrow$  & \textbf{PSNR}$\uparrow$  & \textbf{SSIM}$\uparrow$  & \\
    \hline
    Observed & None & 13.53  & 0.12  & 13.76  & 0.15   & 14.01  & 0.18  &  None \\
           TNN\cite{lu_TNN} & TPAMI2019 & 31.69  & 0.87 & 36.27  & 0.94  & 39.66  & \underline{0.97}  & \textbf{12} \\
           TQRTNN \cite{TQRTNN}&TCI2021& 31.68  & 0.87  & 36.26  & 0.94 & 39.66  & \underline{0.97}  & 27 \\
           UTNN\cite{TTNN_song} &NLAA2020  & 33.78  & 0.92  & 39.28  & 0.97  & 43.20  & \textbf{0.99} & 234\\
          DTNN\cite{DTNN_jiang} & TNNLS2023 & 35.44  & 0.94   & 40.51  & \underline{0.98}  & 44.22  & \textbf{0.99} & 417 \\
           LS2T2NN  \cite{liu2023learnable} & TCSVT2023 & \underline{38.35} & \underline{0.97}  & \underline{42.99} & \textbf{0.99} & 45.58 & \textbf{0.99} & 54 \\
           HLRTF\cite{lrtf} & CVPR2022 & 37.25   & 0.96  & 42.43   & \underline{0.98}  & 45.65   & \textbf{0.99} &  \underline{22}\\
          \textbf{OTLRM*}& Ours &  37.64  & 0.95 & 42.68  & \underline{0.98} & \underline{46.03}  & \textbf{0.99} & 117\\
           \textbf{OTLRM} & Ours & \textbf{40.20}  & \textbf{0.98} &\textbf{44.34} & \textbf{0.99} & \textbf{46.92}   & \textbf{0.99} & 126\\
    \hline
    Observed & None & 9.65  & 0.02  & 9.89  & 0.03 & 10.13  & 0.04  &  None \\
          TNN \cite{lu_TNN}& TPAMI2019   & 32.41  & 0.88  & 37.81  & 0.96 & 41.35  & \underline{0.98}  & \textbf{13} \\
          TQRTNN\cite{TQRTNN}&TCI2021  & 32.46  & 0.88  & 37.57  & 0.95   & 41.28  & \underline{0.98}  & 33 \\
          UTNN \cite{TTNN_song}&NLAA2020  & 35.62  & 0.95 & 41.04  & \underline{0.98}  & 44.66  & \textbf{0.99} & 171  \\
          DTNN\cite{DTNN_jiang}  &TNNLS2023  & 35.95  & 0.95   & 41.29  & 0.97  & 45.05  & \textbf{0.99} & 415\\
         LS2T2NN \cite{liu2023learnable}&TCSVT2023 & 38.67 & 0.96  & 43.36 & \underline{0.98} & 46.25 & \underline{0.98} & 87 \\
           HLRTF\cite{lrtf} & CVPR2022  & 37.20  & 0.95  & 42.76  & \underline{0.98} & 45.73 & \textbf{0.99} & \underline{21} \\
          \textbf{OTLRM*} &Ours&   \underline{39.16} & \underline{0.97} & \underline{43.53}  & \textbf{0.99} & \underline{46.92}  & \textbf{0.99} & 111 \\
          \textbf{OTLRM}& Ours & \textbf{41.74}  & \textbf{0.98} & \textbf{45.09}  & \textbf{0.99}  & \textbf{47.52} & \textbf{0.99} & 118 \\
    \toprule[0.15em]
    \end{tabular}%
    } 
\hspace{-2.5mm}
\scalebox{0.7}{
    \begin{tabular}{ccccccc}
    \bottomrule[0.15em]
     \multicolumn{2}{c}{\textbf{SR=0.05}} & \multicolumn{2}{c}{\textbf{SR=0.10}} & \multicolumn{2}{c}{\textbf{SR=0.15}} & \multirow{2}[2]{*}{\textbf{Time (s)}} \\
        \textbf{PSNR}$\uparrow$  & \textbf{SSIM} $\uparrow$ & \textbf{PSNR}$\uparrow$  & \textbf{SSIM}$\uparrow$  & \textbf{PSNR}$\uparrow$  & \textbf{SSIM}$\uparrow$  & \\
    \hline
     11.66  & 0.05   & 11.89  & 0.08  & 12.14  & 0.11  &  None \\
            23.34  & 0.56   & 28.39  & 0.76  & 31.74  & 0.85  & \textbf{13} \\
            23.33  & 0.56 & 28.39  & 0.76  & 31.70  & 0.85  & 33  \\
             25.32  & 0.66  & 31.34  & 0.86  & 35.62  & 0.93  & 150  \\
            28.05  & 0.78  & 32.48  & 0.89  & 36.17  & 0.94  & 409   \\
            \underline{31.75} & \underline{0.87} & 36.57 & 0.95 & 39.23 & \underline{0.97} & 220 \\
            30.04  & 0.82 & \underline{37.65}  & \underline{0.96} & \underline{40.92}  & \textbf{0.98} & \underline{21} \\
           28.99 & 0.74 & 35.12  & 0.91 & 39.44  & 0.96 & 108 \\
           \textbf{35.42}  &  \textbf{0.93} & \textbf{39.41}  & \textbf{0.97}  & \textbf{42.20} & \textbf{0.98} & 125 \\
    \hline
    11.17  & 0.25   & 11.41  & 0.29   & 11.66  & 0.32  &  None \\
             27.38  & 0.82  & 31.45  & 0.90  & 34.42  & 0.94  & \textbf{13}  \\
            27.12  & 0.81  & 31.43  & 0.90  & 34.37  & 0.94  & 33  \\
            29.40  & 0.87  & 35.00  & 0.95  & 39.05  & \underline{0.98} & 190\\
            30.34  & 0.90  & 35.38  & 0.96 & 39.33  & 0.97  & 411 \\
           32.12 & 0.91  & 36.56 & 0.96  & 40.23  & \underline{0.98} & 142 \\
           \underline{33.00}  & \underline{0.92}   & \underline{39.01}  & \underline{0.97} & \underline{42.95} & \textbf{0.99} & \underline{21} \\
            32.63 & 0.91  & 38.41  & \underline{0.97}  & 42.46  & \underline{0.98} & 113 \\
           \textbf{36.24}  & \textbf{0.97} & \textbf{41.11}  & \textbf{0.99} & \textbf{44.14} & \textbf{0.99} & 128 \\
    \toprule[0.15em]
    \end{tabular}%
    } 
	\label{msitableresult}
\end{table*}

\begin{table*}[htb]
        \centering
        \caption{Evaluation PSNR, SSIM and Time on \emph{NTT} dataset of \textbf{tensor completion} results by different methods for \textbf{videos} under different SRs. Left:\emph{Bird}, Right: \emph{Horse}.}

        \def\arraystretch{1.0}
        \setlength{\tabcolsep}{1.9pt}
\scalebox{0.7}{
    \begin{tabular}{cc|ccccccc|}
    \bottomrule[0.15em]
   \multirow{2}[2]{*}{\textbf{Method}} &\multirow{2}[2]{*}{\textbf{Reference}} & \multicolumn{2}{c}{\textbf{SR=0.10}} & \multicolumn{2}{c}{\textbf{SR=0.15}} & \multicolumn{2}{c}{\textbf{SR=0.20}}& \multirow{2}[2]{*}{\textbf{Time (s)}}\bigstrut[t]\\
          &   & \textbf{PSNR}$\uparrow$  & \textbf{SSIM}$\uparrow$  & \textbf{PSNR}$\uparrow$  & \textbf{SSIM}$\uparrow$  & \textbf{PSNR}$\uparrow$  & \textbf{SSIM}$\uparrow$  &  \bigstrut[b]\\
    \hline
    Observed & None & 7.74  & 0.02  & 7.98  & 0.03  & 8.25  & 0.04  &   None  \bigstrut[t]\\
           TNN\cite{lu_TNN}&TPAMI2019& 27.22  & 0.76  & 29.28  & 0.83 & 31.08  & 0.87  & \textbf{17} \\
           TQRTNN\cite{TQRTNN} &TCI2021& 27.41  & 0.77 & 29.55  & 0.83  & 31.33  & 0.88  & 58  \\
           UTNN\cite{TTNN_song}&NLAA2020  & 27.83  & 0.79  & 30.11  & 0.85 & 31.95  & 0.89  & 156  \\
           DTNN\cite{DTNN_jiang}  &TNNLS2023 & 28.55  & 0.83  & 30.60  & 0.88  & 32.51  & 0.91  & 503  \\
           LS2T2NN \cite{liu2023learnable}&TCSVT2023& \underline{30.40} & \underline{0.85} & \underline{33.38} & \underline{0.91} & \underline{35.55} & \underline{0.94} & 63  \\
           HLRTF \cite{lrtf} &CVPR2022& 26.25  & 0.70  & 28.50  & 0.78 & 31.27 & 0.86 & \underline{22} \\
           \textbf{OTLRM}  &Ours &\textbf{34.24}  & \textbf{0.93} & \textbf{36.04}  & \textbf{0.95} & \textbf{37.44} & \textbf{0.96} & 124
\bigstrut[b] \\
    \toprule[0.15em]
    \end{tabular}%
    }
\hspace{-2.5mm}
\scalebox{0.7}{
    \begin{tabular}{ccccccc}
    \bottomrule[0.15em]
    \multicolumn{2}{c}{\textbf{SR=0.10}} & \multicolumn{2}{c}{\textbf{SR=0.15}} & \multicolumn{2}{c}{\textbf{SR=0.20}} & \multirow{2}[2]{*}{\textbf{Time (s)}} \bigstrut[t]\\
          \textbf{PSNR}$\uparrow$  & \textbf{SSIM}$\uparrow$  & \textbf{PSNR}$\uparrow$  & \textbf{SSIM}$\uparrow$  & \textbf{PSNR}$\uparrow$  & \textbf{SSIM}$\uparrow$  &  \bigstrut[b]\\
    \hline
     6.53  & 0.01  & 6.78  & 0.02  & 7.04  & 0.02  &  None  \bigstrut[t]\\
          26.81  & 0.66  & 28.42  & 0.73 & 29.79  & 0.79  & \textbf{17}\\
          26.93  & 0.67  & 28.64  & 0.74 & 29.95  & 0.79  & 51  \\
          27.27  & 0.68   & 29.00  & 0.76   & 30.30  & 0.81  & 146  \\
          27.52  & 0.74  & 29.42  & 0.80  & 30.98  & 0.85  & 580\\
          \underline{29.48} & \underline{0.76}  & \underline{31.44} & \underline{0.83} & \underline{33.45} & \underline{0.88} & 83 \\
          25.73  & 0.60  & 26.37  & 0.63  & 28.61 & 0.76 & \underline{23} \\
          \textbf{30.64}  & \textbf{0.83} & \textbf{32.20}  & \textbf{0.87}  & \textbf{33.71} & \textbf{0.90} & 125
\bigstrut[b] \\
    \toprule[0.15em]
    \end{tabular}%
    }
	\label{videotableresult}
\end{table*}

\noindent\textbf{MSI Reconstruction in CASSI system:}
We compare our method with DeSCI\cite{desci}, $\lambda$-Net\cite{miao2019net}, TSA-Net\cite{meng2020end}, HDNet\cite{hu2022hdnet}, DGSMP\cite{huang2021deep}, ADMM-Net\cite{admm-net}, GAP-Net\cite{meng2023deep}, PnP-CASSI \cite{zheng2021deep}, DIP-HSI\cite{meng2021self} and HLRTF\cite{lrtf}.

Table \ref{tab:simu} exhibits clear advantages over other self-supervised and zero-shot methods, even having a slight increase of at least 0.34 dB on average over supervised methods.
Figure 6 in Appendix shows the visual performance of \emph{scene03} and notes that the observed image is $256 \times 310$ and resized to $256 \times 256$ for a neat presentation.
It can be observed that our method demonstrates a clearer result with sharper grain boundaries.

\begin{table*}[htb]
	\caption{Comparisons between the proposed model and SOTA methods on 5 simulation scenes (\emph{scene01}$\sim$\emph{scene05}) in KAIST.}

        \def\arraystretch{1}
        \setlength{\tabcolsep}{2pt}
	\centering
	\resizebox{0.95\textwidth}{!}
	{
		\centering
		\begin{tabular}{ccc|cccccccccccc}
			\bottomrule[0.15em]
   \multirow{2}[2]{*}{\textbf{Method}} &\multirow{2}[2]{*}{\textbf{Category}} & \multirow{2}[2]{*}{\textbf{Reference}} & \multicolumn{2}{c}{\textbf{scene01}} & \multicolumn{2}{c}{\textbf{scene02}} & \multicolumn{2}{c}{\textbf{scene03}}& \multicolumn{2}{c}{\textbf{scene04}}&\multicolumn{2}{c}{\textbf{scene05}}&\multicolumn{2}{c}{\textbf{Avg}}\\
   &  & & \textbf{PSNR}$\uparrow$  & \textbf{SSIM}$\uparrow$  & \textbf{PSNR}$\uparrow$  & \textbf{SSIM}$\uparrow$  & \textbf{PSNR}$\uparrow$  & \textbf{SSIM}$\uparrow$  &  \textbf{PSNR}$\uparrow$  & \textbf{SSIM}$\uparrow$  & \textbf{PSNR}$\uparrow$  & \textbf{SSIM}$\uparrow$  & \textbf{PSNR}$\uparrow$  & \textbf{SSIM}$\uparrow$\\
   \midrule[0.1em]
			DeSCI\cite{desci}
			& Model
               & TPAMI2019
			&28.38 & 0.80
			&26.00 & 0.70
			&23.11 & 0.73
			&28.26 & 0.86
			&25.41 & 0.78
			&26.23 & 0.77
			\\
			$\lambda$-Net\cite{miao2019net}
			& CNN (Supervised)
               & ICCV2019
			&30.10&0.85
			&28.49&0.81
			&27.73&0.87
			&37.01&0.93
			&26.19&0.82
			&29.90&0.86
			\\
			TSA-Net\cite{meng2020end}
			& CNN (Supervised)
               & ECCV2020
			&32.31&0.89
			&31.03&0.86
			&32.15&0.92
			&37.95&0.96
			&29.47&0.88
			&32.58&0.90
			\\
			HDNet\cite{hu2022hdnet}
			& Transformer (Supervised)
               & CVPR2022
			&\underline{34.96}&\textbf{0.94}
			&\textbf{35.64}&\textbf{0.94}
			&\underline{35.55}&\underline{0.94}
			&41.64&\textbf{0.98}
			&32.56&\textbf{0.95}
			&\underline{36.07}&\textbf{0.95}
			\\
			DGSMP\cite{huang2021deep}
			& Deep Unfolding (Supervised)
               & CVPR2021
			&33.26&\underline{0.92}
			&32.09&\underline{0.90}
			&33.06&0.93
			&40.54&0.96
			&28.86&0.88
			&33.56&0.92
			\\
			ADMM-Net\cite{admm-net}
			& Deep Unfolding (Supervised)
               & ICCV2019
			&34.03&\underline{0.92}
			&\underline{33.57}&\underline{0.90}
			&34.82&0.93
			&39.46&\underline{0.97}
			&31.83&0.92
			&34.74&\underline{0.93}
           \\
			GAP-Net\cite{meng2023deep}
			& Deep Unfolding (Supervised)
               & IJCV2023
			&33.63&0.91
			&33.19&\underline{0.90}
			&33.96&0.93
			&39.14&\underline{0.97}
			&31.44&0.92
			&34.27&\underline{0.93}
			\\
			PnP-CASSI\cite{zheng2021deep}
			& PnP (Zero-Shot)
               & PR2021
			&29.09&0.80
			&28.05&0.71
			&30.15&0.85
			&39.17&0.94
			&27.45&0.80
			&30.78&0.82
			\\
			DIP-HSI\cite{meng2021self}
			& PnP (Zero-Shot)
               & ICCV2021
			&31.32&0.86
			&25.89&0.70
			&29.91&0.84
			&38.69&0.93
			&27.45&0.80
			&30.65&0.82
			\\
			HLRTF\cite{lrtf}
                &Tensor Network (Self-Supervised)
                & CVPR2022
			&34.56&0.91
			&33.37&0.87
			&\underline{35.55}&\underline{0.94}
			&\underline{43.56}&\textbf{0.98}
			&\underline{33.08}&\underline{0.93}
			&36.02&\underline{0.93}
                \\
			\bf OTLRM
			& Tensor Network (Self-Supervised)
               & Ours
			&\textbf{35.03}&0.91
			&32.90&0.82
			&\textbf{36.25}&\textbf{0.95}
			&\textbf{44.75}&\textbf{0.98}
			&\textbf{33.13}&0.92
			&\textbf{36.41}&0.92
			\\
			\toprule[0.15em]
		\end{tabular}
	}
	\label{tab:simu}
\end{table*}

\noindent\textbf{MSI Denoising:}
For MSI denoising, we select ten SOTA methods which contain NonLRMA\cite{NonLRMA}, TLRLSSTV\cite{TLR_LSSTV}, LLxRGTV\cite{LLxRGTV}, 3DTNN\cite{3DTNN_FW}, LRTDCTV\cite{LRTDCTV}, E3DTV\cite{E3DTV}, DIP\cite{dip}, DDRM\cite{ddrm}, DDS2M\cite{dds2m} and HLRTF \cite{lrtf}.

Table \ref{table:kaist} shows the performance with three metrics.
Figure 11 in Appendix shows the denoising results of \emph{scene02} with Gaussian noise $\mathcal{N}(0, 0.3)$.
Compared with model-based methods, our OTLRM can deeply capture the low-rankness and help improve the abilities of MSI denoising within the DNN framework.
For diffusion-based methods and deep prior-induced PnP methods, our self-supervised method can also enhance the performance and achieve comparable results.
Compared with the tensor network-based method HLRTF, our method is more informative and smoother.

\begin{table*}[htb]
        \centering
        \caption{PSNR, SSIM, FSIM and Time on \emph{KAIST} dataset. Left: \emph{case}: $\mathcal{N}(0, 0.2)$-\emph{scene10}, Right: \emph{case}: $\mathcal{N}(0, 0.3)$-\emph{scene01}.}

        \def\arraystretch{1.0}
        \setlength{\tabcolsep}{1pt}
		\scalebox{0.65}{
\begin{tabular}{cccccc|}
\toprule[0.15em]
\textbf{Method} & \textbf{Reference} & \textbf{PSNR} $\uparrow $& \textbf{SSIM} $\uparrow$ & \textbf{FSIM} $\uparrow$ & \textbf{Time (s)} \\
\midrule[0.1em]
Noisy & None & 16.18 & 0.12 & 0.401 & None \\
NonLRMA~\cite{NonLRMA} &TGRS 2017 & 21.26 & 0.41 & 0.803 & \underline{11} \\
TLRLSSTV\ ~\cite{TLR_LSSTV} &TGRS 2021& 24.88 & 0.53 & 0.767 & 76 \\
LLxRGTV~\cite{LLxRGTV} &SP 2021& \underline{31.15} & 0.80 & 0.917 & 38 \\
3DTNN~\cite{3DTNN_FW} &TGRS 2019& 28.04 & 0.78 & 0.881 & 20 \\
LRTDCTV~\cite{LRTDCTV} &JSTAR 2023& 25.95 & 0.66 & 0.816 & 43 \\
E3DTV~\cite{E3DTV} &TIP 2020& 30.34 & \textbf{0.87} & \underline{0.926} & \textbf{10} \\
DIP~\cite{dip} &ICCVW 2019& 24.18 & 0.61 & 0.825 & 72 \\
DDRM~\cite{ddrm} & NeurIPS 2022& 29.41 & \textbf{0.87} & 0.922  & 20 \\
DDS2M~\cite{dds2m} & ICCV 2023&\textbf{32.80} & 0.79 & 0.895 & 354 \\
HLRTF~\cite{lrtf} &CVPR2022& 30.42 & \underline{0.83} & \textbf{0.940} & 23 \\
\bf
OTLRM & Ours  & 30.82 & 0.72 & 0.881 & 45 \\
\bottomrule[0.15em]
\end{tabular}}
\hspace{-2.5mm}
\scalebox{0.65}{
\begin{tabular}{cccccc}
\toprule[0.15em]
\textbf{Method} &\textbf{Reference}&\textbf{PSNR} $\uparrow $& \textbf{SSIM} $\uparrow$ & \textbf{FSIM} $\uparrow$ & \textbf{Time (s)} \\
\midrule[0.1em]
Noisy    &None& 12.98 & 0.06  & 0.320  & None \\
NonLRMA~\cite{NonLRMA}  &TGRS 2017& 20.30 & 0.36  & 0.772  & \underline{11} \\
TLRLSSTV~\cite{TLR_LSSTV} & TGRS 2021& 22.82 & 0.41  & 0.689 & 76 \\
LLxRGTV~\cite{LLxRGTV}  &SP 2021& 27.64 & 0.68  & 0.868  & 38 \\
3DTNN~\cite{3DTNN_FW} & TGRS 2019& 26.04 & 0.72  & 0.848 & 20 \\
LRTDCTV~\cite{LRTDCTV}  &JSTAR 2023& 24.59 & 0.53  & 0.739 & 42 \\
E3DTV~\cite{E3DTV}    &TIP 2020& 28.36 & \textbf{0.82}  & \underline{0.900}  & \textbf{9} \\
DIP~\cite{dip} & ICCVW 2019 &20.06 & 0.41 & 0.798 & 74 \\
DDRM~\cite{ddrm}  & NeurIPS 2022&27.81  & \underline{0.79} & 0.893 & 23 \\
DDS2M~\cite{dds2m} & ICCV 2023&30.08  & 0.67  & 0.834 & 318 \\
HLRTF~\cite{lrtf}& CVPR2022 &\underline{30.34} & 0.69 & 0.874 & 25 \\
\bf 
OTLRM & Ours & \textbf{32.56} & \textbf{0.82} & \textbf{0.902} &47 \\
\bottomrule[0.15em]
\end{tabular}}
\label{table:kaist}
\end{table*}

\subsection{Ablation Analysis}
\label{ablation}
In this section, we do ablation experiments on the hyperparameter tensor rank $r$, the trade-off parameter $\lambda$, the number $k$ of the DNN layers in the rank estimation module $\rho(\cdot)$ and the visual results of the weight matrices in $\rho(\cdot)$ respectively to demonstrate the advantages of our model.
Additionally, in traditional skinny t-SVD algorithm \ref{remark_skinnySVD}, $\mathcal{U}$ and $\mathcal{V}$ are semi-orthogonal tensors.
However, our model does not introduce semi-orthogonal constraints on these tensors.
We explain this in the below ablation analysis.
What's more, for the advantages of the orthogonal transform $\mathbf{L}$, we also compare our OTLRM with two typical methods, DTNN \cite{kong2021tensor} and Tensor $Q$-rank \cite{DTNN_jiang}.

For ablation settings, MSI \emph{Balloons} (of size $256 \times 256 \times 31$) in CAVE datasets is selected for TC.
In Figures 12 and 13 in Appendix, the solid blue line denotes the PSNR value and the dotted red line represents the SSIM value.

\noindent\textbf{Effect of $r$:}
The tensor rank $r \in [1, \min\{n_1,n_2\}]$ which is an integer characterizes the low-rankness of the generated tensor $\mathcal{X} \in \mathbb{R}^{n_1 \times n_2 \times n_3}$.
In the ablation experiment, we choose the rank $r$ ranging from 10 to 100 at intervals of 10.
From Figure 12 in Appendix, when the rank $r$ is low, the performance is less effective due to the lost information of the original tensor.
When the rank is too high, the result tensor is not well guaranteed to be low-rank, which leads to a sub-optimal performance.
Thus, given $n = \min\{n_1,n_2\}$, the rank $r$ can be set between $[n/20 ,n/5]$.

\begin{table}[htb]
        \centering
        \caption{Ablation study for the effect of $k$ in the rank estimation module $\rho(\cdot)$ with MSI \emph{Balloons}.}

        \def\arraystretch{1.0}
        \setlength{\tabcolsep}{3pt}
\scalebox{0.75}{
    \begin{tabular}{c|cccccc}
    \bottomrule[0.15em]
   \multirow{2}[2]{*}{$k$-\textbf{layers}} & \multicolumn{2}{c}{\textbf{SR=0.05}} & \multicolumn{2}{c}{\textbf{SR=0.10}} & \multicolumn{2}{c}{\textbf{SR=0.15}}\bigstrut[t]\\
          & \textbf{PSNR}$\uparrow$  & \textbf{SSIM}$\uparrow$  & \textbf{PSNR}$\uparrow$  & \textbf{SSIM}$\uparrow$  & \textbf{PSNR}$\uparrow$  & \textbf{SSIM}$\uparrow$\bigstrut[b]\\
    \hline
    0 &35.82&0.94&40.39&\underline{0.98}&41.92&\underline{0.98} \bigstrut[t]\\
    1 &38.72&0.97&43.10&\textbf{0.99}&45.12&\textbf{0.99} \bigstrut[t]\\
    2 &\textbf{40.20}&\textbf{0.98}&\underline{44.34}&\textbf{0.99}&\textbf{46.92}&\textbf{0.99} \bigstrut[t]\\
    3 &\underline{40.15}&\textbf{0.98}&\textbf{44.56}&\textbf{0.99}&\underline{46.85}&\textbf{0.99} \bigstrut[t]\\
    \toprule[0.15em]
    \end{tabular}%
    }
\label{table:ablation_k}
\end{table}

\noindent\textbf{Effect of $\lambda$:}
The hyperparameter $\lambda$ which controls the $\mathrm{OTV}$ loss is set using a strategy of sampling one point every order of magnitude from 1e-1 to 1e-10.
From Figure 13 in Appendix, we can find that when $\lambda$ is large, the performance is decreased, probably because too strong the $\mathrm{OTV}$ loss aggravates the local similarities and loses the detailed features.
Our method remains effective when $\lambda$ falls between 1e-7 and 1e-9.
Thus, our model is easy to tune for the best $\lambda$.

\noindent\textbf{Effect of $k$:}
In all experiments, we simply use a two-layer DNN to represent 
the rank estimation module $\rho(\cdot)$.
And we analyze the effect of $\rho(\cdot)$ on the results by adjusting the number $k$ of layers of the DNN.
Here, $k = 1$ indicates that only one LeakyReLU activation function is added.
From Table \ref{table:ablation_k}, deeper DNN may have the potential for better results due to their better fitting ability.

\noindent\textbf{Effect of $\rho(\cdot)$:}
For the rank information extractor $\rho(\cdot)$, we adopted a two-layer deep neural network that contains one LeakyReLU layer sandwiched between two linear layers as a simple implementation.
The visualization results of the two weight matrices are shown in Figure 15 in Appendix.
It can be seen that the weight matrices learn different and rich rank information for different dimensions and different positions of the rank matrix $\mathbf{S}$ with low-rank property.

\noindent\textbf{Analysis for The Semi-orthogonality of $\mathcal{U}$ and $\mathcal{V}$:}
Following the skinny t-SVD algorithm, the semi-orthogonality of the $\mathcal{U}$ and $\mathcal{V}$ is naturally and strictly guaranteed by the process of SVD.
However, our generative t-SVD model is solved based on the gradient descent-based algorithm, which means that we do not need the SVD.
To verify the validity of the semi-orthogonality of $\mathcal{U}$ and $\mathcal{V}$, we construct a semi-orthogonal loss term for ablation experiments, which is:
\begin{equation}
    \Psi(\mathcal{U}, \mathcal{V}) = \beta (\Vert \mathcal{U}^{\mathbf{T}} \ast_{L} \mathcal{U} - \mathcal{I} \Vert^2_{F} + \Vert \mathcal{V}^{\mathbf{T}} \ast_{L} \mathcal{V} - \mathcal{I} \Vert^2_{F}).
\end{equation}
Figure 14 in Appendix shows the completion results with \emph{SR=0.05} under different penalty coefficient $\beta$.
It can be observed that the results show a decreasing trend as $\beta$ continues to rise.
Therefore, we relax the semi-orthogonal constraints on $\mathcal{U}$ and $\mathcal{V}$ to enhance the performance.

\noindent\textbf{Comparisons on Orthogonal Transform $\mathbf{L}$:}
Table 3 in Appendix shows the evaluation results of the above methods.
\textbf{(1)} The transform of DTNN is a redundant dictionary with constraints on each column of the dictionary to have an F-norm of one. 
It may not technically be called orthogonal, which makes its inverse transform difficult to calculate. 
To obtain the transform $\mathbf{L} \in \mathbb{R}^{n_3 \times d}$, the computational complexity of DTNN is $O((n_3)^3d^2)$ and ours is $O((n_3)^4)$ (running time of the whole model \textbf{415s} vs \textbf{118s}).
\textbf{(2)} Tensor $Q$-rank imposes constraints on the transform process with a “two-step” strategy. 
This involves first finding an appropriate transform based on certain selection criteria and then incorporating it into the reconstruction process. 
This ``separate" strategy may not well capture the intrinsic characteristics of the data. 
In contrast, our learnable transform can be directly embedded and inherently achieves orthogonality.

\section{Conclusion}
In this paper, we proposed a learnable orthogonal transform-induced generative low-rank framework based on t-SVD for multi-dimensional tensor recovery, possessing the orthogonal and learnable transform which enables flexible data-adaptive capability while maintains theoretical guarantee for ``arbitrary invertible linear transform".
Constructed by a series of Householder transformation units, this transform can be learnable and seamlessly integrated into the neural network with endogenous orthogonality.
Compared to traditional solutions of t-SVD, our generative t-SVD representation model can naturally maintain the low-rank structure and be solved by gradient descent-based algorithms.
Comprehensive experiments verify the effectiveness of our method in MSIs and videos for three practical problems.

\section{Acknowledgment}

This work was supported in part by the National Natural Science Foundation of China under Grants 62106063 and 62276079, by the Guangdong Natural Science Foundation under Grant 2022A1515010819, and by National Natural Science Foundation of China Joint Fund Project Key Support Project under U22B2049.

\bibliography{aaai25}

\begin{thebibliography}{54}
\providecommand{\natexlab}[1]{#1}

\bibitem[{Braman(2010)}]{braman2010third}
Braman, K. 2010.
\newblock Third-order tensors as linear operators on a space of matrices.
\newblock \emph{Linear Algebra and its Applications}, 433(7): 1241--1253.

\bibitem[{Cai, Cand\`{e}s, and Shen(2010)}]{Caisvt}
Cai, J.-F.; Cand\`{e}s, E.~J.; and Shen, Z. 2010.
\newblock A Singular Value Thresholding Algorithm for Matrix Completion.
\newblock \emph{SIAM Journal on Optimization}, 20(4): 1956--1982.

\bibitem[{Cai et~al.(2022)Cai, Lin, Wang, Yuan, Ding, Zhang, Timofte, and Gool}]{cai2022degradation}
Cai, Y.; Lin, J.; Wang, H.; Yuan, X.; Ding, H.; Zhang, Y.; Timofte, R.; and Gool, L.~V. 2022.
\newblock Degradation-aware unfolding half-shuffle transformer for spectral compressive imaging.
\newblock In \emph{Proceedings of the Advances in Neural Information Processing Systems}, 37749--37761.

\bibitem[{Carroll and Chang(1970)}]{carroll1970analysis}
Carroll, J.~D.; and Chang, J.-J. 1970.
\newblock Analysis of individual differences in multidimensional scaling via an N-way generalization of “Eckart-Young” decomposition.
\newblock \emph{Psychometrika}, 35(3): 283--319.

\bibitem[{Chen et~al.(2017)Chen, Guo, Wang, Wang, Peng, and He}]{NonLRMA}
Chen, Y.; Guo, Y.; Wang, Y.; Wang, D.; Peng, C.; and He, G. 2017.
\newblock Denoising of hyperspectral images using nonconvex low rank matrix approximation.
\newblock \emph{IEEE Transactions on Geoscience and Remote Sensing}, 55(9): 5366--5380.

\bibitem[{Choi et~al.(2017)Choi, Kim, Gutierrez, Jeon, and Nam}]{choi2017high}
Choi, I.; Kim, M.; Gutierrez, D.; Jeon, D.; and Nam, G. 2017.
\newblock High-quality hyperspectral reconstruction using a spectral prior.
\newblock Technical report.

\bibitem[{He et~al.(2015)He, Zhang, Ren, and Sun}]{he2015delving}
He, K.; Zhang, X.; Ren, S.; and Sun, J. 2015.
\newblock Delving deep into rectifiers: Surpassing human-level performance on imagenet classification.
\newblock In \emph{Proceedings of the IEEE International Conference on Computer Vision}, 1026--1034.

\bibitem[{Hu et~al.(2022)Hu, Cai, Lin, Wang, Yuan, Zhang, Timofte, and Van~Gool}]{hu2022hdnet}
Hu, X.; Cai, Y.; Lin, J.; Wang, H.; Yuan, X.; Zhang, Y.; Timofte, R.; and Van~Gool, L. 2022.
\newblock Hdnet: High-resolution dual-domain learning for spectral compressive imaging.
\newblock In \emph{Proceedings of the IEEE/CVF Conference on Computer Vision and Pattern Recognition}, 17542--17551.

\bibitem[{Huang et~al.(2021)Huang, Dong, Yuan, Wu, and Shi}]{huang2021deep}
Huang, T.; Dong, W.; Yuan, X.; Wu, J.; and Shi, G. 2021.
\newblock Deep gaussian scale mixture prior for spectral compressive imaging.
\newblock In \emph{Proceedings of the IEEE/CVF Conference on Computer Vision and Pattern Recognition}, 16216--16225.

\bibitem[{Ionescu, Vantzos, and Sminchisescu(2015{\natexlab{a}})}]{Ionescu_2015_ICCV}
Ionescu, C.; Vantzos, O.; and Sminchisescu, C. 2015{\natexlab{a}}.
\newblock Matrix Backpropagation for Deep Networks With Structured Layers.
\newblock In \emph{Proceedings of the IEEE International Conference on Computer Vision}, 2965--2973.

\bibitem[{Ionescu, Vantzos, and Sminchisescu(2015{\natexlab{b}})}]{IonescuTraining}
Ionescu, C.; Vantzos, O.; and Sminchisescu, C. 2015{\natexlab{b}}.
\newblock Training deep networks with structured layers by matrix backpropagation.
\newblock \emph{arXiv preprint arXiv:1509.07838}.

\bibitem[{Jiang et~al.(2023)Jiang, Zhao, Zhang, and Ng}]{DTNN_jiang}
Jiang, T.-X.; Zhao, X.-L.; Zhang, H.; and Ng, M.~K. 2023.
\newblock Dictionary Learning With Low-Rank Coding Coefficients for Tensor Completion.
\newblock \emph{IEEE Transactions on Neural Networks and Learning Systems}, 34(2): 932--946.

\bibitem[{Kawar et~al.(2022)Kawar, Elad, Ermon, and Song}]{ddrm}
Kawar, B.; Elad, M.; Ermon, S.; and Song, J. 2022.
\newblock Denoising diffusion restoration models.
\newblock In \emph{Proceedings of the Advances in Neural Information Processing Systems}, 23593--23606.

\bibitem[{Kernfeld, Kilmer, and Aeron(2015)}]{kernfeld2015tensor}
Kernfeld, E.; Kilmer, M.; and Aeron, S. 2015.
\newblock Tensor--tensor products with invertible linear transforms.
\newblock \emph{Linear Algebra and its Applications}, 485: 545--570.

\bibitem[{Kilmer et~al.(2013)Kilmer, Braman, Hao, and Hoover}]{kilmer2013third}
Kilmer, M.~E.; Braman, K.; Hao, N.; and Hoover, R.~C. 2013.
\newblock Third-order tensors as operators on matrices: A theoretical and computational framework with applications in imaging.
\newblock \emph{SIAM Journal on Matrix Analysis and Applications}, 34(1): 148--172.

\bibitem[{Kilmer and Martin(2011)}]{kilmer2011factorization}
Kilmer, M.~E.; and Martin, C.~D. 2011.
\newblock Factorization strategies for third-order tensors.
\newblock \emph{Linear Algebra and its Applications}, 435(3): 641--658.

\bibitem[{Kingma and Ba(2015)}]{adam}
Kingma, D.~P.; and Ba, J.~L. 2015.
\newblock Adam: A Method for Stochastic Optimization.
\newblock In \emph{Proceedings of The International Conference on Learning Representations}.

\bibitem[{Kolda and Bader(2009)}]{kolda2009tensor}
Kolda, T.~G.; and Bader, B.~W. 2009.
\newblock Tensor decompositions and applications.
\newblock \emph{SIAM Review}, 51(3): 455--500.

\bibitem[{Kong, Lu, and Lin(2021)}]{kong2021tensor}
Kong, H.; Lu, C.; and Lin, Z. 2021.
\newblock Tensor Q-rank: new data dependent definition of tensor rank.
\newblock \emph{Machine Learning}, 110(7): 1867--1900.

\bibitem[{Liu et~al.(2023)Liu, Leng, Zhao, Zeng, Wang, and Yang}]{liu2023learnable}
Liu, S.; Leng, J.; Zhao, X.-L.; Zeng, H.; Wang, Y.; and Yang, J.-H. 2023.
\newblock Learnable Spatial-Spectral Transform-Based Tensor Nuclear Norm for Multi-Dimensional Visual Data Recovery.
\newblock \emph{IEEE Transactions on Circuits and Systems for Video Technology}.

\bibitem[{Liu et~al.(2024)Liu, Zhao, Leng, Li, Yang, and Chen}]{shengLiu_revisiting}
Liu, S.; Zhao, X.-L.; Leng, J.; Li, B.-Z.; Yang, J.-H.; and Chen, X. 2024.
\newblock Revisiting High-Order Tensor Singular Value Decomposition From Basic Element Perspective.
\newblock \emph{IEEE Transactions on Signal Processing}, 72: 4589--4603.

\bibitem[{Liu et~al.(2018)Liu, Yuan, Suo, Brady, and Dai}]{desci}
Liu, Y.; Yuan, X.; Suo, J.; Brady, D.~J.; and Dai, Q. 2018.
\newblock Rank minimization for snapshot compressive imaging.
\newblock \emph{IEEE Transactions on Pattern Analysis and Machine Intelligence}, 41(12): 2990--3006.

\bibitem[{Lu(2021)}]{Lu_2021_ICCV}
Lu, C. 2021.
\newblock Transforms Based Tensor Robust PCA: Corrupted Low-Rank Tensors Recovery via Convex Optimization.
\newblock In \emph{Proceedings of the IEEE/CVF International Conference on Computer Vision}, 1145--1152.

\bibitem[{Lu et~al.(2019)Lu, Feng, Chen, Liu, Lin, and Yan}]{lu_TNN}
Lu, C.; Feng, J.; Chen, Y.; Liu, W.; Lin, Z.; and Yan, S. 2019.
\newblock Tensor robust principal component analysis with a new tensor nuclear norm.
\newblock \emph{IEEE Transactions on Pattern Analysis and Machine Intelligence}, 42(4): 925--938.

\bibitem[{Lu, Peng, and Wei(2019)}]{lu2019low}
Lu, C.; Peng, X.; and Wei, Y. 2019.
\newblock Low-rank tensor completion with a new tensor nuclear norm induced by invertible linear transforms.
\newblock In \emph{Proceedings of the IEEE/CVF Conference on Computer Vision and Pattern Recognition}, 5996--6004.

\bibitem[{Luo et~al.(2022)Luo, Zhao, Meng, and Jiang}]{lrtf}
Luo, Y.; Zhao, X.-L.; Meng, D.; and Jiang, T.-X. 2022.
\newblock HLRTF: Hierarchical Low-Rank Tensor Factorization for Inverse Problems in Multi-Dimensional Imaging.
\newblock In \emph{Proceedings of the IEEE/CVF Conference on Computer Vision and Pattern Recognition}, 19303--19312.

\bibitem[{Ma et~al.(2019)Ma, Liu, Shou, and Yuan}]{admm-net}
Ma, J.; Liu, X.-Y.; Shou, Z.; and Yuan, X. 2019.
\newblock Deep tensor admm-net for snapshot compressive imaging.
\newblock In \emph{Proceedings of the IEEE International Conference on Computer Vision}.

\bibitem[{Mai, Lam, and Lee(2022)}]{mai2022deep}
Mai, T. T.~N.; Lam, E.~Y.; and Lee, C. 2022.
\newblock Deep Unrolled Low-Rank Tensor Completion for High Dynamic Range Imaging.
\newblock \emph{IEEE Transactions on Image Processing}, 31: 5774--5787.

\bibitem[{Meng, Ma, and Yuan(2020)}]{meng2020end}
Meng, Z.; Ma, J.; and Yuan, X. 2020.
\newblock End-to-end low cost compressive spectral imaging with spatial-spectral self-attention.
\newblock In \emph{Proceedings of the European Conference on Computer Vision}, 187--204. Springer.

\bibitem[{Meng et~al.(2021)Meng, Yu, Xu, and Yuan}]{meng2021self}
Meng, Z.; Yu, Z.; Xu, K.; and Yuan, X. 2021.
\newblock Self-supervised neural networks for spectral snapshot compressive imaging.
\newblock In \emph{Proceedings of the IEEE International Conference on Computer Vision}, 2622--2631.

\bibitem[{Meng, Yuan, and Jalali(2023)}]{meng2023deep}
Meng, Z.; Yuan, X.; and Jalali, S. 2023.
\newblock Deep unfolding for snapshot compressive imaging.
\newblock \emph{International Journal of Computer Vision}, 131(11): 2933--2958.

\bibitem[{Miao et~al.(2019)Miao, Yuan, Pu, and Athitsos}]{miao2019net}
Miao, X.; Yuan, X.; Pu, Y.; and Athitsos, V. 2019.
\newblock l-net: Reconstruct hyperspectral images from a snapshot measurement.
\newblock In \emph{Proceedings of the IEEE/CVF International Conference on Computer Vision}, 4059--4069.

\bibitem[{Miao et~al.(2023)Miao, Zhang, Zhang, and Tao}]{dds2m}
Miao, Y.; Zhang, L.; Zhang, L.; and Tao, D. 2023.
\newblock Dds2m: Self-supervised denoising diffusion spatio-spectral model for hyperspectral image restoration.
\newblock In \emph{Proceedings of the IEEE/CVF International Conference on Computer Vision}, 12086--12096.

\bibitem[{Mu et~al.(2020)Mu, Wang, Lu, Zhang, and Qi}]{mu2020weighted}
Mu, Y.; Wang, P.; Lu, L.; Zhang, X.; and Qi, L. 2020.
\newblock Weighted tensor nuclear norm minimization for tensor completion using tensor-\protect{SVD}.
\newblock \emph{Pattern Recognition Letters}, 130: 4--11.

\bibitem[{Peng et~al.(2020)Peng, Xie, Zhao, Wang, Yee, and Meng}]{E3DTV}
Peng, J.; Xie, Q.; Zhao, Q.; Wang, Y.; Yee, L.; and Meng, D. 2020.
\newblock Enhanced 3DTV regularization and its applications on \protect{HSI} denoising and compressed sensing.
\newblock \emph{IEEE Transactions on Image Processing}, 29: 7889--7903.

\bibitem[{Qin et~al.(2022)Qin, Wang, Zhang, Wang, Luo, and Huang}]{qin2022low}
Qin, W.; Wang, H.; Zhang, F.; Wang, J.; Luo, X.; and Huang, T. 2022.
\newblock Low-rank high-order tensor completion with applications in visual data.
\newblock \emph{IEEE Transactions on Image Processing}, 31: 2433--2448.

\bibitem[{Sidorov and Yngve~Hardeberg(2019)}]{dip}
Sidorov, O.; and Yngve~Hardeberg, J. 2019.
\newblock Deep hyperspectral prior: Single-image denoising, inpainting, super-resolution.
\newblock In \emph{Proceedings of the IEEE/CVF International Conference on Computer Vision Workshops}, 0--0.

\bibitem[{Song, Ng, and Zhang(2020)}]{TTNN_song}
Song, G.; Ng, M.~K.; and Zhang, X. 2020.
\newblock Robust tensor completion using transformed tensor singular value decomposition.
\newblock \emph{Numerical Linear Algebra with Applications}, 27(3): e2299.

\bibitem[{Song, Sebe, and Wang(2022)}]{SongImproSVD}
Song, Y.; Sebe, N.; and Wang, W. 2022.
\newblock Improving covariance conditioning of the \protect{SVD} meta-layer by orthogonality.
\newblock In \emph{Proceedings of the European Conference on Computer Vision}, 356--372. Springer.

\bibitem[{Sun, Vong, and Wang(2022)}]{sun2022fast}
Sun, Y.; Vong, C.~M.; and Wang, S. 2022.
\newblock Fast \protect{AUC} Maximization Learning Machine With Simultaneous Outlier Detection.
\newblock \emph{IEEE Transactions on Cybernetics}.

\bibitem[{Tucker(1966)}]{tucker1966some}
Tucker, L.~R. 1966.
\newblock Some mathematical notes on three-mode factor analysis.
\newblock \emph{Psychometrika}, 31(3): 279--311.

\bibitem[{Uhlig(2001)}]{uhlig2001constructive}
Uhlig, F. 2001.
\newblock Constructive ways for generating (generalized) real orthogonal matrices as products of (generalized) symmetries.
\newblock \emph{Linear Algebra and its Applications}, 332: 459--467.

\bibitem[{Wang et~al.(2023)Wang, Li, Bai, Jin, Zhou, and Zhao}]{wang2023transformed}
Wang, A.; Li, C.; Bai, M.; Jin, Z.; Zhou, G.; and Zhao, Q. 2023.
\newblock Transformed Low-Rank Parameterization Can Help Robust Generalization for Tensor Neural Networks.
\newblock \emph{arXiv preprint arXiv:2303.00196}.

\bibitem[{Wang et~al.(2022{\natexlab{a}})Wang, Wang, Hong, Roy, and Chanussot}]{wang2022learning}
Wang, M.; Wang, Q.; Hong, D.; Roy, S.~K.; and Chanussot, J. 2022{\natexlab{a}}.
\newblock Learning tensor low-rank representation for hyperspectral anomaly detection.
\newblock \emph{IEEE Transactions on Cybernetics}, 53(1): 679--691.

\bibitem[{Wang et~al.(2022{\natexlab{b}})Wang, Dang, Hu, Fua, and Salzmann}]{WangRobustSVD}
Wang, W.; Dang, Z.; Hu, Y.; Fua, P.; and Salzmann, M. 2022{\natexlab{b}}.
\newblock Robust Differentiable \protect{SVD}.
\newblock \emph{IEEE Transactions on Pattern Analysis and Machine Intelligence}, 44(9): 5472--5487.

\bibitem[{Wu et~al.(2021)Wu, Li, Li, and Wu}]{TQRTNN}
Wu, F.; Li, Y.; Li, C.; and Wu, Y. 2021.
\newblock A Fast Tensor Completion Method Based on Tensor \text{QR} Decomposition and Tensor Nuclear Norm Minimization.
\newblock \emph{IEEE Transactions on Computational Imaging}, 7: 1267--1277.

\bibitem[{Zeng et~al.(2023)Zeng, Huang, Chen, Luong, and Philips}]{LRTDCTV}
Zeng, H.; Huang, S.; Chen, Y.; Luong, H.; and Philips, W. 2023.
\newblock All of Low-rank and Sparse: A Recast Total Variation Approach to Hyperspectral Denoising.
\newblock \emph{IEEE Journal of Selected Topics in Applied Earth Observations and Remote Sensing}.

\bibitem[{Zeng and Xie(2021)}]{LLxRGTV}
Zeng, H.; and Xie, X. 2021.
\newblock Hyperspectral image denoising via global spatial-spectral total variation regularized nonconvex local low-rank tensor approximation.
\newblock \emph{Signal Processing}, 178: 107805.

\bibitem[{Zeng et~al.(2020)Zeng, Xie, Cui, Yin, and Ning}]{TLR_LSSTV}
Zeng, H.; Xie, X.; Cui, H.; Yin, H.; and Ning, J. 2020.
\newblock Hyperspectral image restoration via global L 1-2 spatial--spectral total variation regularized local low-rank tensor recovery.
\newblock \emph{IEEE Transactions on Geoscience and Remote Sensing}, 59(4): 3309--3325.

\bibitem[{Zhang and Aeron(2016)}]{zhang2016exact}
Zhang, Z.; and Aeron, S. 2016.
\newblock Exact tensor completion using t-SVD.
\newblock \emph{IEEE Transactions on Signal Processing}, 65(6): 1511--1526.

\bibitem[{Zhang et~al.(2014)Zhang, Ely, Aeron, Hao, and Kilmer}]{zhang2014novel}
Zhang, Z.; Ely, G.; Aeron, S.; Hao, N.; and Kilmer, M. 2014.
\newblock Novel methods for multilinear data completion and de-noising based on tensor-\protect{SVD}.
\newblock In \emph{Proceedings of the IEEE conference on computer vision and pattern recognition}, 3842--3849.

\bibitem[{Zheng et~al.(2021)Zheng, Liu, Meng, Qiao, Tong, Yang, Han, and Yuan}]{zheng2021deep}
Zheng, S.; Liu, Y.; Meng, Z.; Qiao, M.; Tong, Z.; Yang, X.; Han, S.; and Yuan, X. 2021.
\newblock Deep plug-and-play priors for spectral snapshot compressive imaging.
\newblock \emph{Photonics Research}.

\bibitem[{Zheng et~al.(2019)Zheng, Huang, Zhao, Jiang, Ma, and Ji}]{3DTNN_FW}
Zheng, Y.-B.; Huang, T.-Z.; Zhao, X.-L.; Jiang, T.-X.; Ma, T.-H.; and Ji, T.-Y. 2019.
\newblock Mixed noise removal in hyperspectral image via low-fibered-rank regularization.
\newblock \emph{IEEE Transactions on Geoscience and Remote Sensing}, 58(1): 734--749.

\bibitem[{Zhou and Cheung(2019)}]{zhou2019bayesian}
Zhou, Y.; and Cheung, Y.-M. 2019.
\newblock Bayesian low-tubal-rank robust tensor factorization with multi-rank determination.
\newblock \emph{IEEE Transactions on Pattern Analysis and Machine Intelligence}, 43(1): 62--76.

\end{thebibliography}

\appendix

\section{Supplementary Material}

In this supplementary material, we first present the proofs of Theorem 1 and the derivation of the final low-rank tensor in Algorithm 1, then make some additions to the experiments mentioned in the main manuscript, and we also provide the results for several ablation experiments to verify the efficiency of the proposed method.

\section{Proofs and Derivation}

\subsection{Proofs of Theorem 1:}
The proofs of Theorem 1 in the main paper are below \cite{uhlig2001constructive}.
The four previous lemmas with their proofs are given and \textbf{Proof A.5} is the main proof process.

\noindent\textbf{Lemma A.1} The Householder matrix $\mathbf{F} = \mathbf{I} - 2\mathbf{w}\mathbf{w}^{\mathbf{T}}$ is symmetric and orthogonal.

\noindent\textbf{Proof A.1} Since $\mathbf{F} = \mathbf{I} - 2\mathbf{w}\mathbf{w}^{\mathbf{T}}$, $(\mathbf{w}\mathbf{w}^{\mathbf{T}})^{\mathbf{T}} = \mathbf{w}\mathbf{w}^{\mathbf{T}}$, and $\mathbf{w}^{\mathbf{T}} \mathbf{w} = 1$, we can conclude that
\begin{equation}
    \mathbf{F}^{\mathbf{T}} = (\mathbf{I} - 2\mathbf{w}\mathbf{w}^{\mathbf{T}})^{\mathbf{T}} = \mathbf{I} - 2\mathbf{w}\mathbf{w}^{\mathbf{T}} = \mathbf{F}.
\end{equation}
Therefore, $\mathbf{F}$ is symmetric.
\begin{equation}
\begin{aligned}
    \mathbf{F}\mathbf{F}^{\mathbf{T}} &= (\mathbf{I} - 2\mathbf{w}\mathbf{w}^{\mathbf{T}})(\mathbf{I} - 2\mathbf{w}\mathbf{w}^{\mathbf{T}})^{\mathbf{T}} \\
&= (\mathbf{I} - 2\mathbf{w}\mathbf{w}^{\mathbf{T}})(\mathbf{I} - 2\mathbf{w}\mathbf{w}^{\mathbf{T}}) \\
&= \mathbf{I} - 4\mathbf{w}\mathbf{w}^{\mathbf{T}} + 4\mathbf{w}\mathbf{w}^{\mathbf{T}} = \mathbf{I}.
\end{aligned}    
\end{equation}
Therefore, $\mathbf{F}$ is orthogonal.

\noindent\textbf{Lemma A.2} Given any two non-zero vectors $\mathbf{x}$ and $\mathbf{y}$ with the same 2-norm, there exists a Householder transformation $\mathbf{F}$ satisfying $\mathbf{F}\mathbf{x} = \mathbf{y}$.

\noindent\textbf{Proof A.2} We can construct the Householder matrix with vector $\mathbf{w} = \frac{\mathbf{x} - \mathbf{y}}{\|\mathbf{x} - \mathbf{y}\|_2}$ and 
\begin{equation}
    \mathbf{F} = \mathbf{I} - 2 \frac{(\mathbf{x} - \mathbf{y})(\mathbf{x} - \mathbf{y})^{\mathbf{T}}}{\|\mathbf{x} - \mathbf{y}\|_2^2},
\end{equation}
then
\begin{equation}
    \mathbf{F}\mathbf{x} = \mathbf{x} - 2 \frac{(\mathbf{x} - \mathbf{y})(\mathbf{x} - \mathbf{y})^{\mathbf{T}}}{\|\mathbf{x} - \mathbf{y}\|_2^2} \mathbf{x} = \mathbf{y}.
\end{equation}

\noindent\textbf{Lemma A.3 (QR factorization)} A rectangular matrix $\mathbf{A} \in \mathbb{R}^{n \times n}$ can be factored into a product of an orthogonal matrix $\mathbf{Q} \in \mathbb{R}^{n \times n}$ and an upper triangular matrix $\mathbf{R} \in \mathbb{R}^{n \times n}$: $\mathbf{A} = \mathbf{Q}\mathbf{R}$, where $\mathbf{Q}$ is the product of $n-1$ orthogonal Householder matrices.

\noindent\textbf{Proof A.3} The matrix $\mathbf{A}$ can be written in block form as $\mathbf{A} = \begin{bmatrix} \mathbf{a}_1 & \mathbf{a}_2 & \dots & \mathbf{a}_n \end{bmatrix}$, where $\mathbf{a}_i = \begin{bmatrix} a_{i1} & a_{i2} & \dots & a_{in} \end{bmatrix}^{\mathbf{T}}$.
Using Lemma A.2, the vector $\mathbf{a}_1 = \begin{bmatrix} a_{11} & a_{21} & \dots & a_{n1} \end{bmatrix}^{\mathbf{T}}$ can be transformed to $\begin{bmatrix} \alpha & 0 & \dots & 0 \end{bmatrix}^{\mathbf{T}}$ with a Householder transformation $\mathbf{F}_1$. 
We denote the non-zero value in the vector as $\alpha$ for convenience. 
This process can be formulated as:
\begin{equation}
\begin{aligned}
    \mathbf{F}_1 \mathbf{A} &= \mathbf{F}_1 \begin{bmatrix} \mathbf{a}_1 & \dots & \mathbf{a}_n \end{bmatrix} \\
    &= 
\begin{bmatrix}
\alpha & a_{12} & a_{13} & \dots & a_{1n} \\
0 & a_{22} & a_{23} & \dots & a_{2n} \\
0 & 0 & a_{33} & \dots & a_{3n} \\
\vdots & \vdots & \vdots & \ddots & \vdots \\
0 & 0 & 0 & \dots & a_{nn}
\end{bmatrix}.
\end{aligned}
\end{equation}

By repeating the above process on the first column of the matrix $\begin{bmatrix}
a_{22} & a_{2n} \\
\vdots & \ddots \\
a_{n2} & a_{nn}
\end{bmatrix}$ with a Householder transformation $\mathbf{F}_2$, the vector $\begin{bmatrix} a_{22} & a_{32} & \dots & a_{n2} \end{bmatrix}^{\mathbf{T}}$ can be transformed to $\begin{bmatrix} \alpha & 0 & \dots & 0 \end{bmatrix}^{\mathbf{T}}$.
Therefore, we conclude that
\begin{equation}
    \mathbf{F}_2 \mathbf{F}_1 \mathbf{A} = 
\begin{bmatrix}
\alpha & \alpha & a_{13} & \dots & a_{1n} \\
0 & \alpha & a_{23} & \dots & a_{2n} \\
0 & 0 & a_{33} & \dots & a_{3n} \\
\vdots & \vdots & \vdots & \ddots & \vdots \\
0 & 0 & 0 & \dots & a_{nn}
\end{bmatrix}.
\end{equation}

By repeating the above process for a total of $n-1$ times, we can obtain an upper triangular matrix:
\begin{equation}
    \mathbf{F}_{n-1} \mathbf{F}_{n-2} \dots \mathbf{F}_1 \mathbf{A} = 
\begin{bmatrix}
\alpha & \alpha & \dots & \alpha \\
0 & \alpha & \dots & \alpha \\
\vdots & \vdots & \ddots & \vdots \\
0 & 0 & \dots & \alpha
\end{bmatrix} = \mathbf{R}.
\end{equation}

Considering that each Householder transformation $\mathbf{F}_i$ is orthogonal and symmetric, we conclude that
\begin{equation}
    \mathbf{A} = \mathbf{F}_1 \mathbf{F}_2 \dots \mathbf{F}_{n-1} \mathbf{R} = \mathbf{Q} \mathbf{R}.
\end{equation}

\noindent\textbf{Lemma A.4} If an $n \times n$ matrix $\mathbf{A}$ is not only upper triangular but also orthogonal, then $\mathbf{A}$ is a diagonal matrix.

\noindent\textbf{Proof A.4} The $n \times n$ matrix $\mathbf{A}$ can be written as a block form $\mathbf{A} = 
\begin{bmatrix}
\mathbf{A}_1 & \mathbf{A}_2 \\
0 & a_{nn}
\end{bmatrix}$, where $\mathbf{A}_1 \in \mathbb{R}^{(n-1) \times (n-1)}$, $\mathbf{A}_2 \in \mathbb{R}^{(n-1) \times 1}$. 
Since $\mathbf{A}$ is orthogonal, we have
\begin{equation}
\begin{aligned}
    \mathbf{A}\mathbf{A}^{\mathbf{T}} &= 
\begin{bmatrix}
\mathbf{A}_1 & \mathbf{A}_2 \\
0 & a_{nn}
\end{bmatrix}
\begin{bmatrix}
\mathbf{A}_1^{\mathbf{T}} & 0 \\
\mathbf{A}_2^{\mathbf{T}} & a_{nn}
\end{bmatrix} \\
&=
\begin{bmatrix}
\mathbf{A}_1\mathbf{A}_1^{\mathbf{T}} + \mathbf{A}_2\mathbf{A}_2^{\mathbf{T}} & \mathbf{A}_2 a_{nn} \\
\mathbf{A}_2^{\mathbf{T}} a_{nn} & a_{nn}^2
\end{bmatrix}
= \mathbf{I}.
\end{aligned}
\end{equation}

Therefore, $a_{nn} = \pm 1$, $\mathbf{A}_2 = 0$, and $\mathbf{A}_1\mathbf{A}_1^{\mathbf{T}} = \mathbf{I} \in \mathbb{R}^{(n-1) \times (n-1)}$.
The matrix block $\mathbf{A}_1$ is not only upper triangular but also orthogonal. By repeating the above process for a total of $n$ times, we can conclude that $\mathbf{A}$ is diagonal with diagonal entries equal to $\pm 1$.

\noindent\textbf{Theorem A.5} Every real orthogonal $n \times n$ matrix $\mathbf{A}$ is the product of at most $n$ real orthogonal Householder transformations.

\noindent\textbf{Proof A.5} With Lemma A.3, we can upper triangularize the given real orthogonal matrix $\mathbf{A}$ as:
\begin{equation}
    \mathbf{F}_{n-1}\mathbf{F}_{n-2} \cdots \mathbf{F}_1\mathbf{A} = \mathbf{R}.
\end{equation}

Since $\mathbf{R}$ is both upper triangular and orthogonal as a product of orthogonal matrices, according to Lemma A.4, $\mathbf{R}$ is diagonal with diagonal entries equal to $\pm 1$. By constraining the entry of $\mathbf{R}$ to be positive when constructing the QR factorization in Lemma A.3, we have $r_{11} = r_{22} = \cdots = r_{n-1,n-1} = 1$.

If the last diagonal entry $r_{nn} = -1$, by setting $\mathbf{F}_n = \mathbf{I}_n - 2\mathbf{e}_n\mathbf{e}_n^{\mathbf{T}}$, we can obtain that
\begin{equation}
    \mathbf{F}_n\mathbf{F}_{n-1} \cdots \mathbf{F}_1\mathbf{A} = \mathbf{F}_n\mathbf{R} = \mathbf{I}.
\end{equation}

As each Householder matrix $\mathbf{F}_i$ is its own inverse (Lemma A.1), we conclude that
\begin{equation}
    \mathbf{A} = \mathbf{F}_1\mathbf{F}_2 \cdots \mathbf{F}_n.
\end{equation}

If the last diagonal entry $r_{nn} = 1$, then $\mathbf{R} = \mathbf{I}$. Because
\begin{equation}
    \mathbf{F}_{n-1}\mathbf{F}_{n-2} \cdots \mathbf{F}_1\mathbf{A} = \mathbf{R},
\end{equation}
we can conclude that 
\begin{equation}
    \mathbf{A} = \mathbf{F}_1 \mathbf{F}_2 \cdots \mathbf{F}_{n-1}.
\end{equation}

Thus, every real orthogonal \( n \times n \) matrix \( \mathbf{A} \) can be written as the product of at most \( n \) Householder matrices.

\subsection{Derivation for Algorithm 1:}
We provide the derivation of the final low-rank tensor in Algorithm 1 of the main paper. 
For the classical t-SVT operator, it aims to solve the following optimization problem: 
\begin{equation}
    \begin{aligned} \label{eqeq001}
        \underset{\mathcal{X}}{\min}\ \gamma \lVert \mathcal{X} \rVert _*+\frac{1}{2}\lVert \mathcal{X}-\mathcal{Y} \rVert _{F}^{2}, 
    \end{aligned}
\end{equation}
where $\| \mathcal{X}\|_*$  is the tensor nuclear paradigm based on the Fourier transform.
Therefore, we have
\begin{equation}
    \begin{aligned}
        \underset{\mathcal{X}}{\min}\ \frac{1}{n_3}\sum_{k=1}^{n_3}{\left( \gamma \lVert \mathbf{\hat{X}}_k \rVert _*+\frac{1}{2}\lVert \mathbf{\hat{X}}_k-\mathbf{\hat{Y}}_k \rVert _{F}^{2} \right)}, 
    \end{aligned}
\end{equation}
\begin{equation}
    \underset{\mathcal{X}}{\min}\ \frac{1}{n_3}\sum_{k=1}^{n_3}{\left( \gamma \lVert \mathbf{\hat{X}}_k \rVert _* + \frac{1}{2}\lVert \mathbf{\hat{X}}_k-\mathbf{\hat{U}}_k\mathbf{\hat{S}}_k\mathbf{\hat{V}}_k^{\mathbf{T}} \rVert \right)},
\end{equation}
where $\hat{\mathcal{X}} = \mathcal{X} \times _3 \mathbf{L} = L(\mathcal{X})$, $\hat{\mathcal{Y}} = \mathcal{Y} \times _3 \mathbf{L} = L(\mathcal{Y})$, $\hat{\mathbf{X}}_k$ is $k$-th slice matrix of $\hat{\mathcal{X}}$, $\hat{\mathbf{Y}}_k$ is $k$-th slice matrix of $\hat{\mathcal{Y}}$, and $\mathbf{\hat{U}}_k, \mathbf{\hat{S}}_k, $ , and $\mathbf{\hat{V}}_k $ are SVD of $\hat{\mathbf{X}}_k$. 
Based on this derivation, we have that the solution of low-rank matrix $\hat{\mathbf{X}}_k$ is 
\begin{equation}
    \begin{aligned}
        \mathbf{\hat{X}}_k=\mathbf{\hat{U}}_k \max \left( \mathbf{\hat{S}}_k-\gamma ,0 \right) \mathbf{\hat{V}}_k^{\mathbf{T}}. 
    \end{aligned}
\end{equation}
Thus, the solution of low-rank tensor $\mathcal{X}$ in \eqref{eqeq001} is 
\begin{equation}
    \mathcal{X}=L^{-1}\left( L\left( \mathcal{U} \right) \bigtriangleup \max \left( L\left( \mathcal{S} \right) -\gamma ,0 \right) \bigtriangleup L\left( \mathcal{V}^{\mathbf{T}} \right) \right) . 
\end{equation}
In this paper, the transform matrix of $\| \mathcal{X}\|_*$ is learnable orthogonal transform $\mathbf{L}$ via Householder transformation, which can also follow the above derivations due to the property of the arbitrary invertible linear transform in t-SVD \cite{kong2021tensor}.
Within the generative framework of parameterized t-SVD, since $\mathcal{S}$ is a simple $f$-diagonal tensor, we construct the rank tensor by a learnable transformed matrix $\mathbf{S} \in \mathbb{R}^{n_3 \times r}$ instead.
And to find a more suitable tensor rank, we inject the DNN-induced dense rank estimation operator $\rho(\cdot)$ which captures and enriches the rank information.
Above equation can be reformulated as:
\begin{equation}
    \mathcal{X} = L^{-1}( L(\mathcal{U}) \bigtriangleup Diag(\rho(\mathbf{S})) \bigtriangleup L(\mathcal{V}^{\mathbf{T}}) ).
\end{equation}


\section{Results for the applications}
\label{r_three_app}
\subsection{Tensor Completion}
\begin{figure*}[htb]
	\footnotesize
	\setlength{\tabcolsep}{1pt}
	\begin{center}
            \scalebox{0.48}{
		\begin{tabular}{ccccccccccc}
			\includegraphics[width=0.2\linewidth]{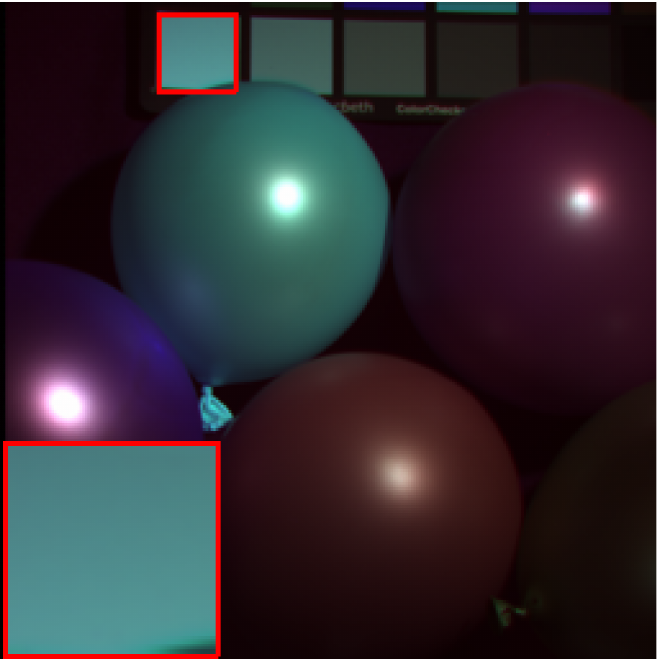}&
			\includegraphics[width=0.2\linewidth]{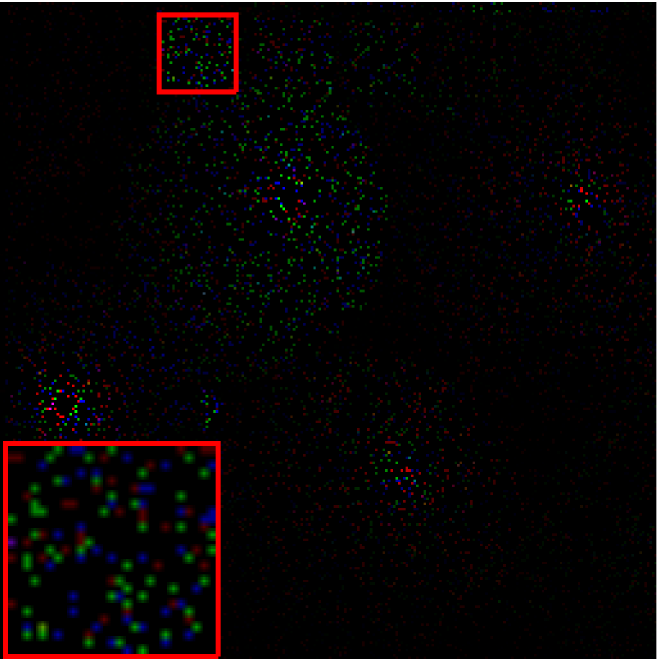}&
			\includegraphics[width=0.2\linewidth]{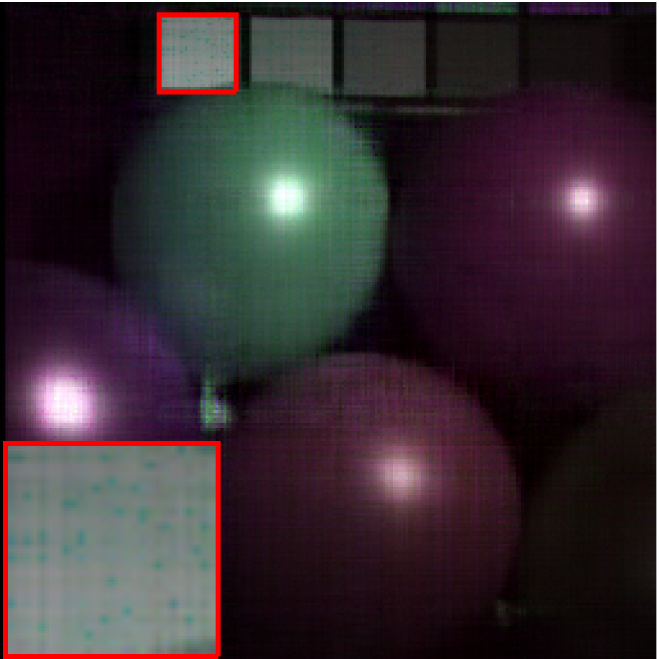}&
			\includegraphics[width=0.2\linewidth]{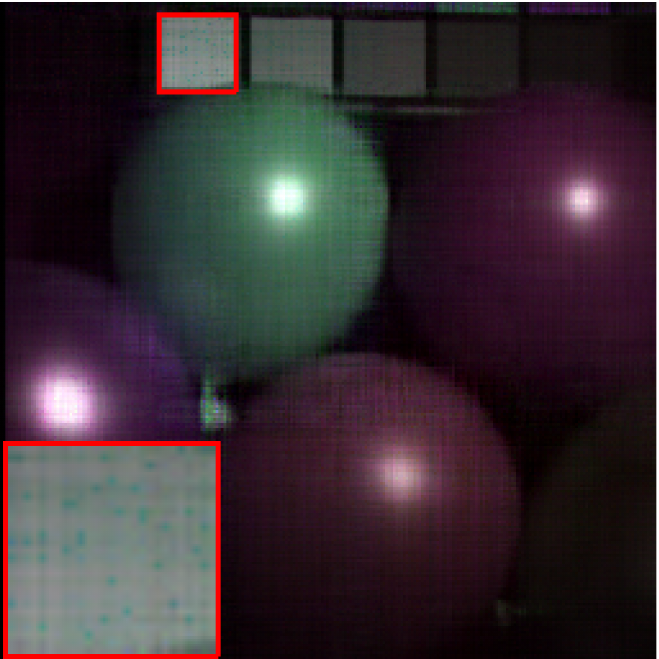}&
			\includegraphics[width=0.2\linewidth]{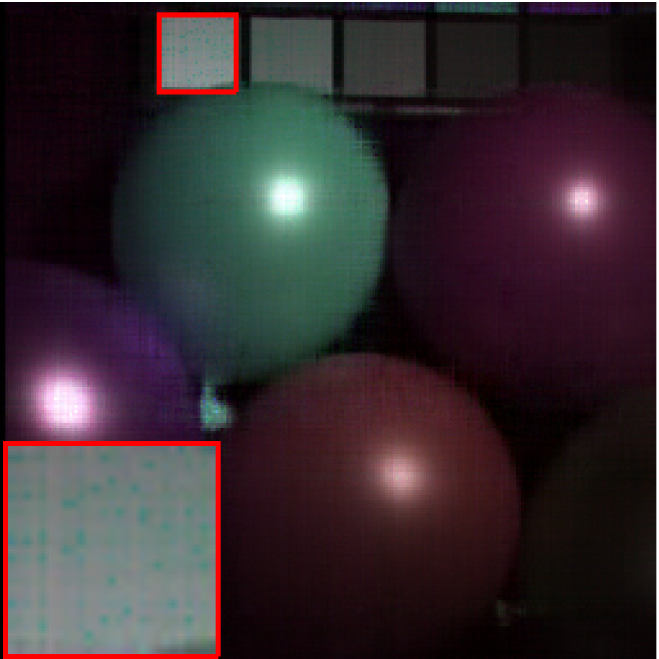}&
			\includegraphics[width=0.2\linewidth]{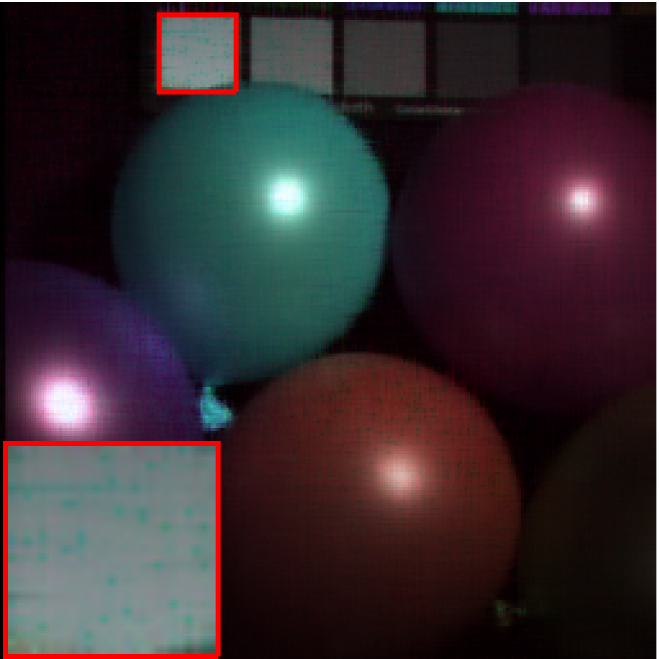}&
			\includegraphics[width=0.2\linewidth]{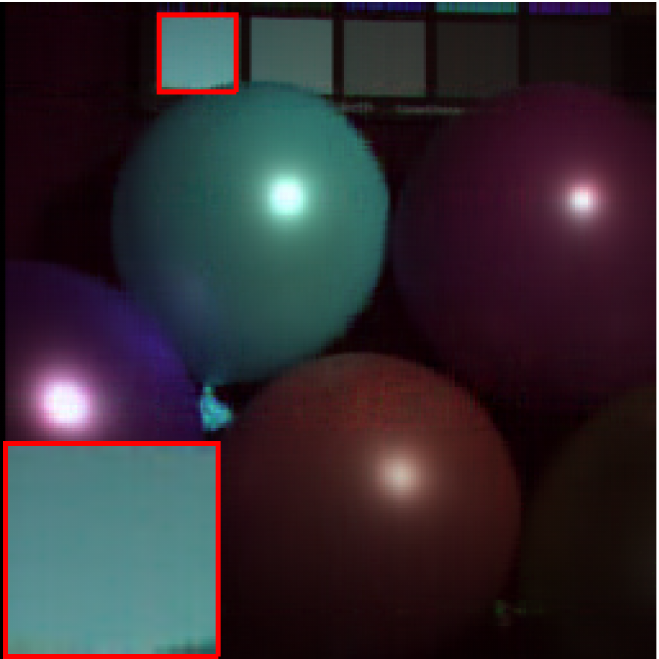}&
            \includegraphics[width=0.2\linewidth]{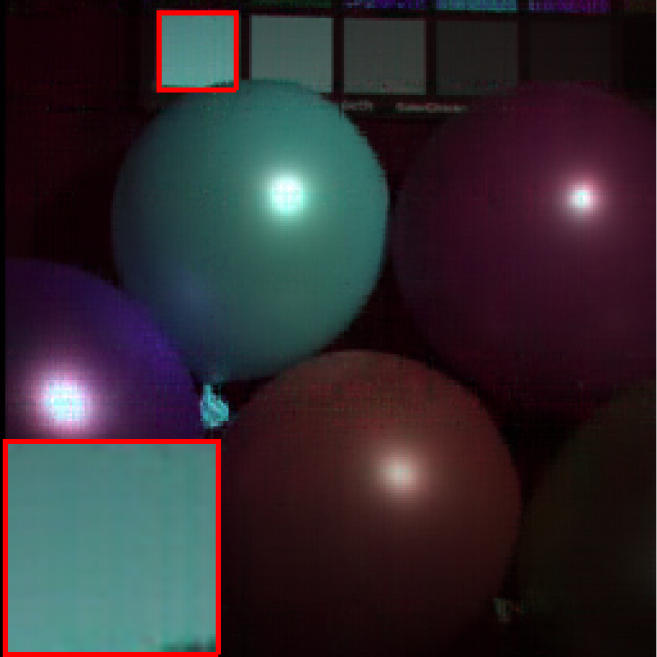}&
            \includegraphics[width=0.2\linewidth]{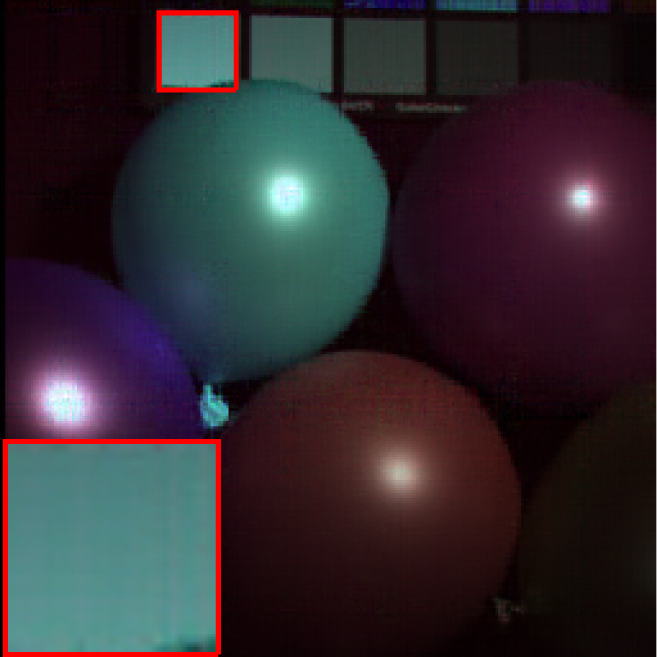}&
            \includegraphics[width=0.2\linewidth]{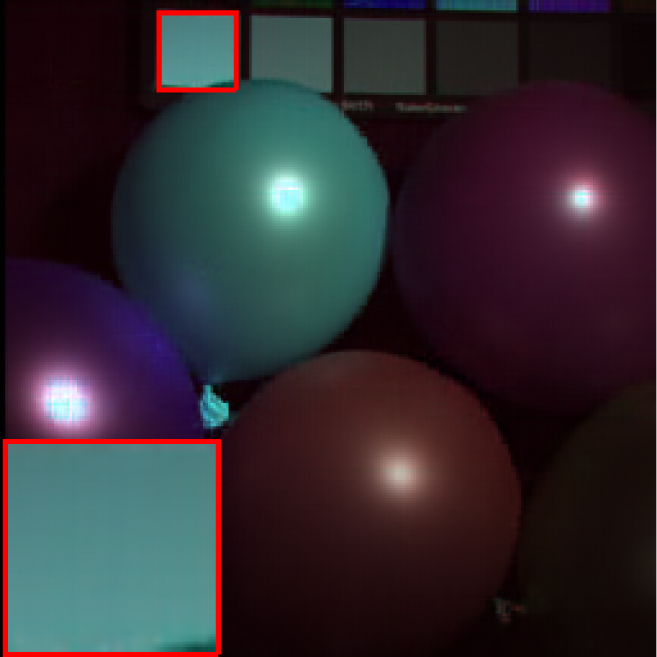}\\
			
			\includegraphics[width=0.2\linewidth]{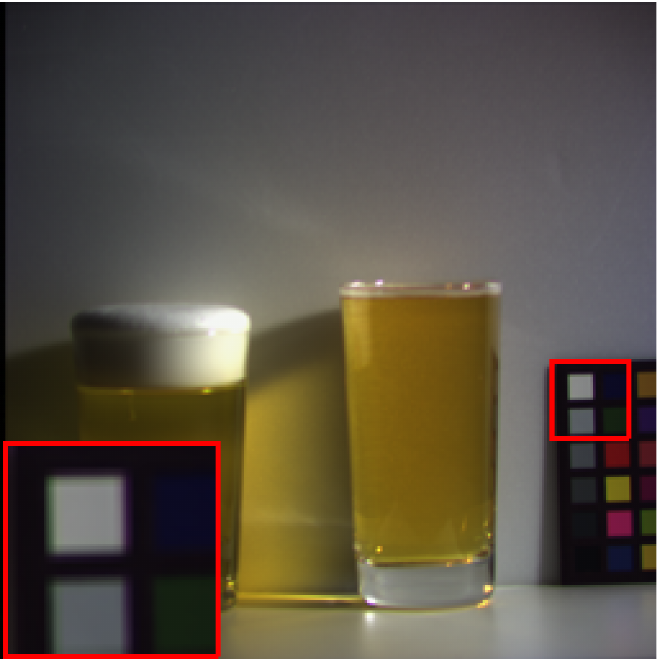}&
			\includegraphics[width=0.2\linewidth]{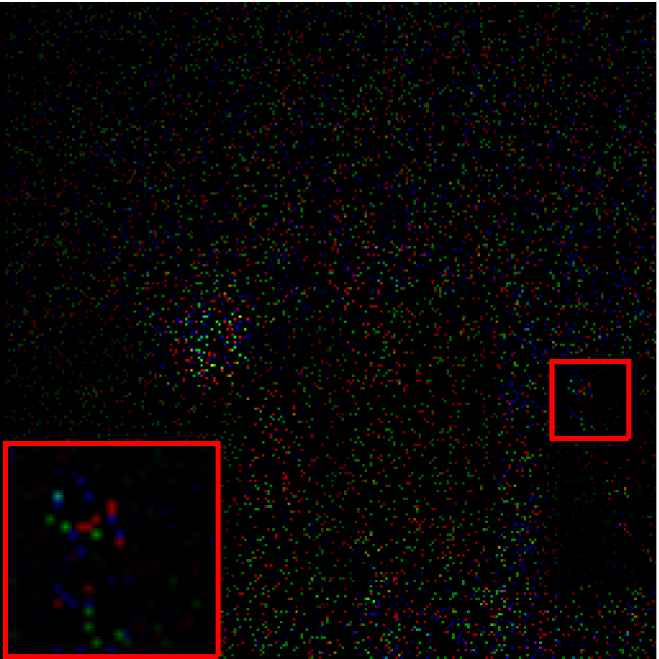}&
			\includegraphics[width=0.2\linewidth]{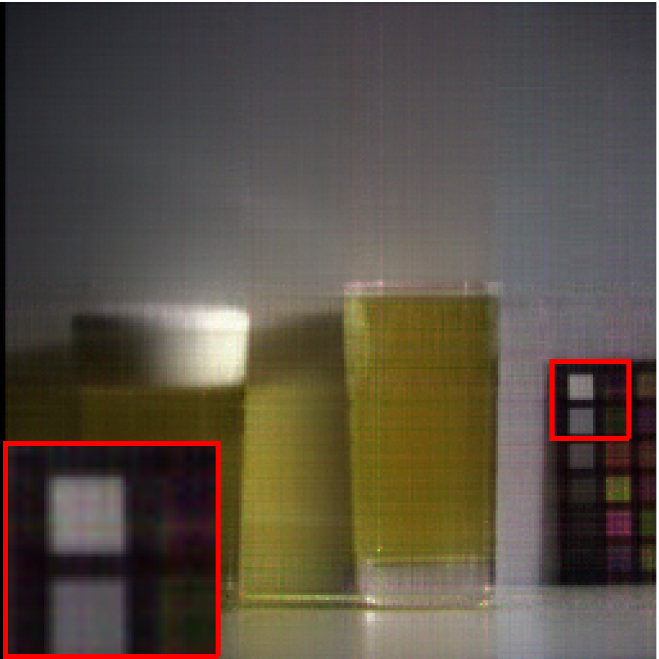}&
			\includegraphics[width=0.2\linewidth]{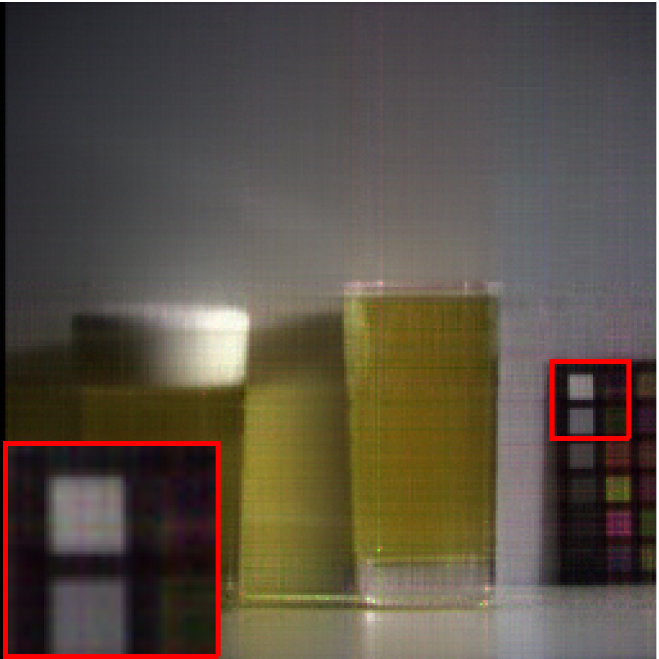}&
			\includegraphics[width=0.2\linewidth]{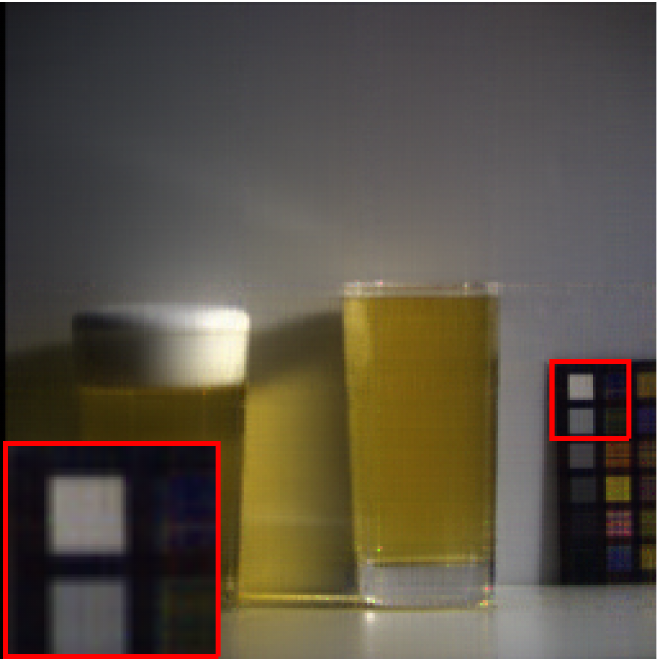}&
			\includegraphics[width=0.2\linewidth]{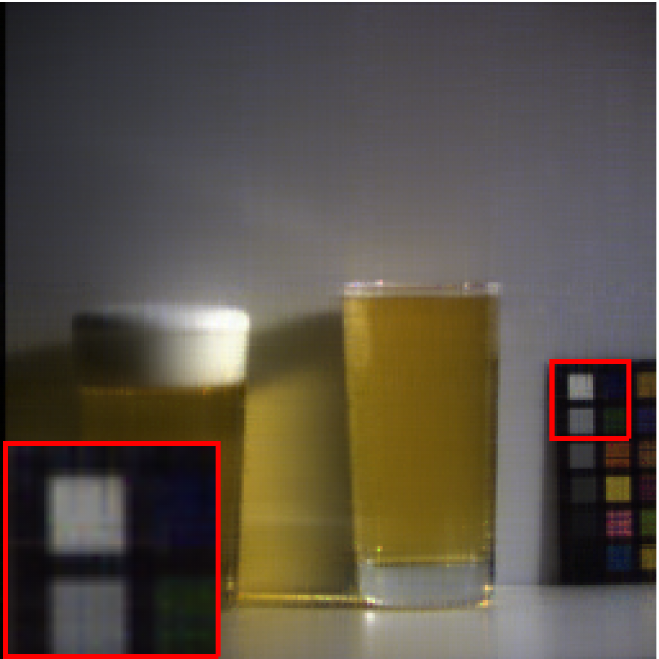}&
			\includegraphics[width=0.2\linewidth]{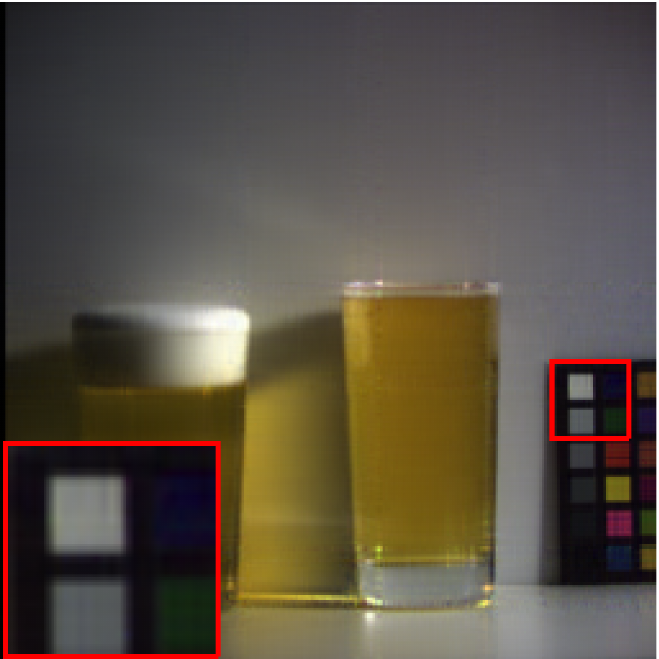}&
            \includegraphics[width=0.2\linewidth]{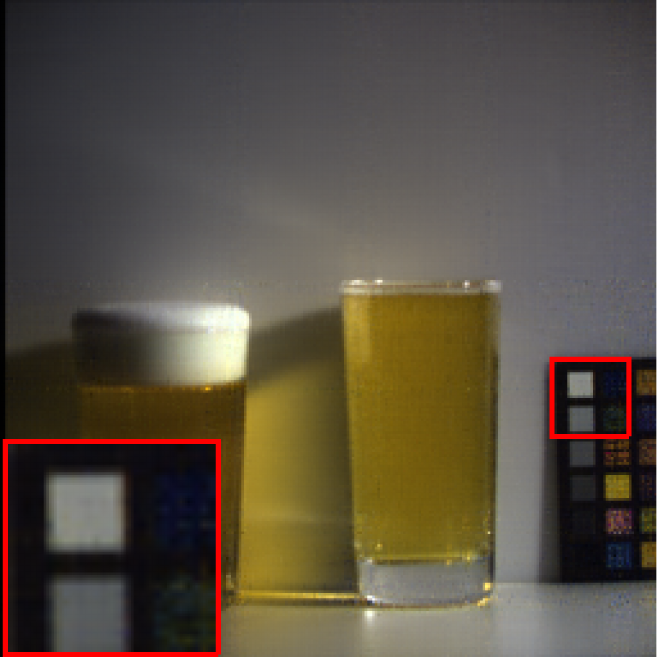}&
            \includegraphics[width=0.2\linewidth]{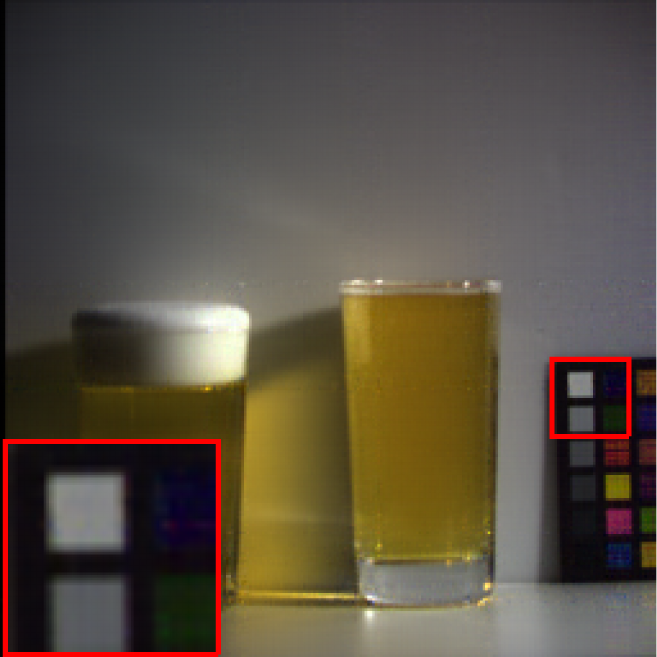}&
            \includegraphics[width=0.2\linewidth]{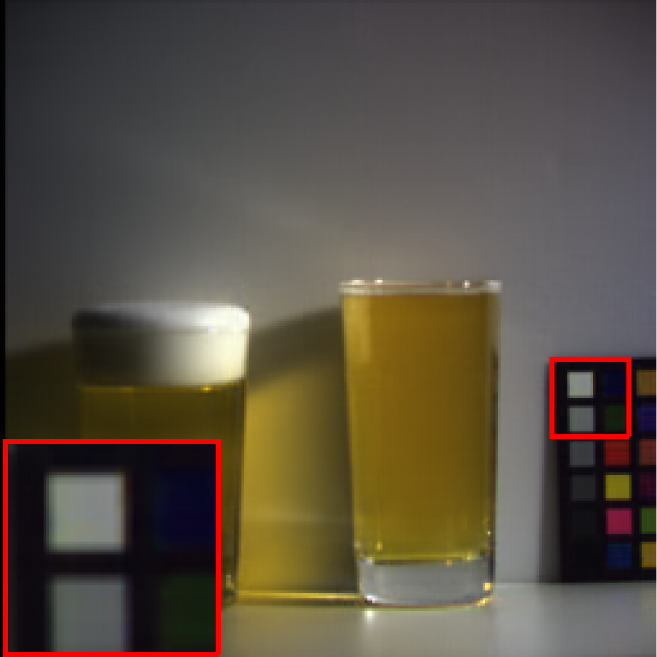}\\

			\includegraphics[width=0.2\linewidth]{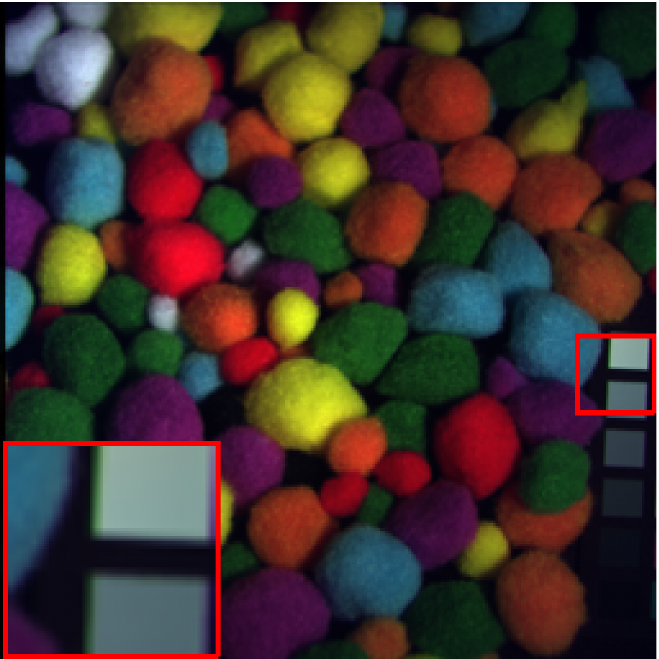}&
			\includegraphics[width=0.2\linewidth]{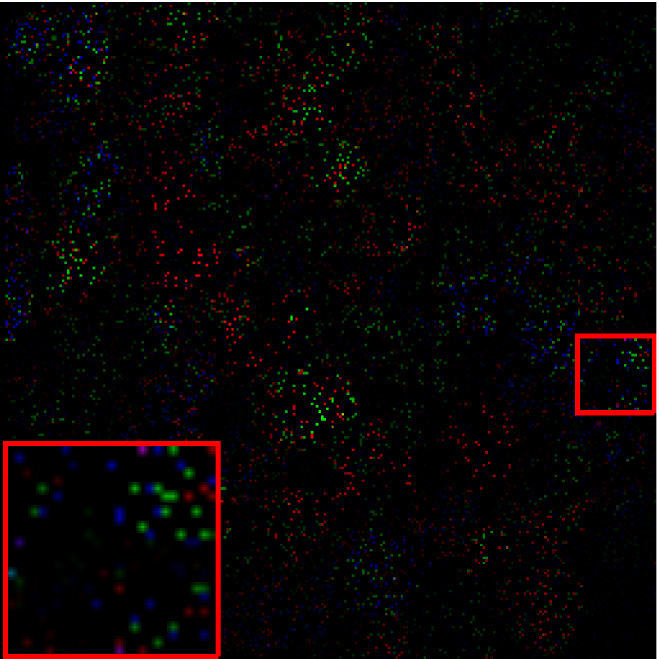}&
			\includegraphics[width=0.2\linewidth]{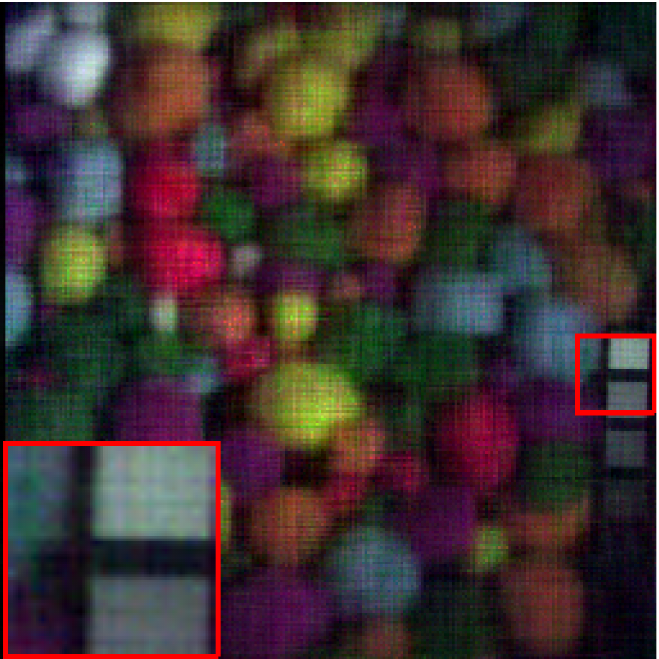}&
			\includegraphics[width=0.2\linewidth]{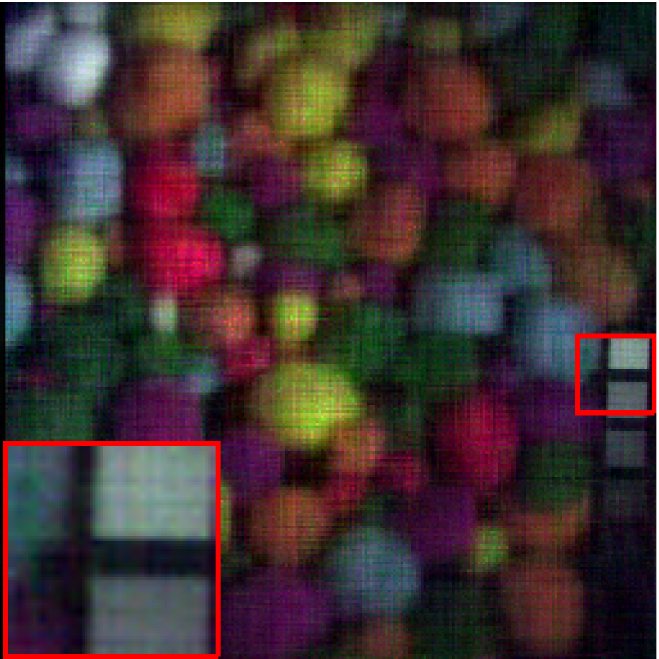}&
			\includegraphics[width=0.2\linewidth]{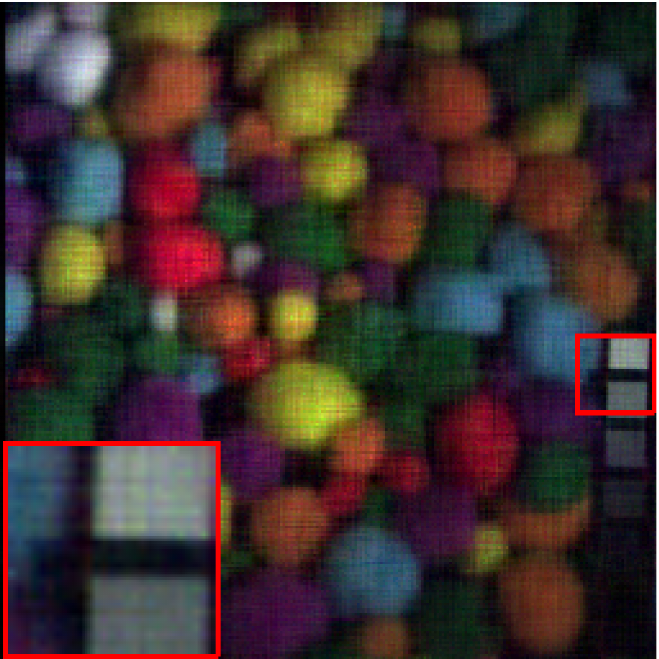}&
			\includegraphics[width=0.2\linewidth]{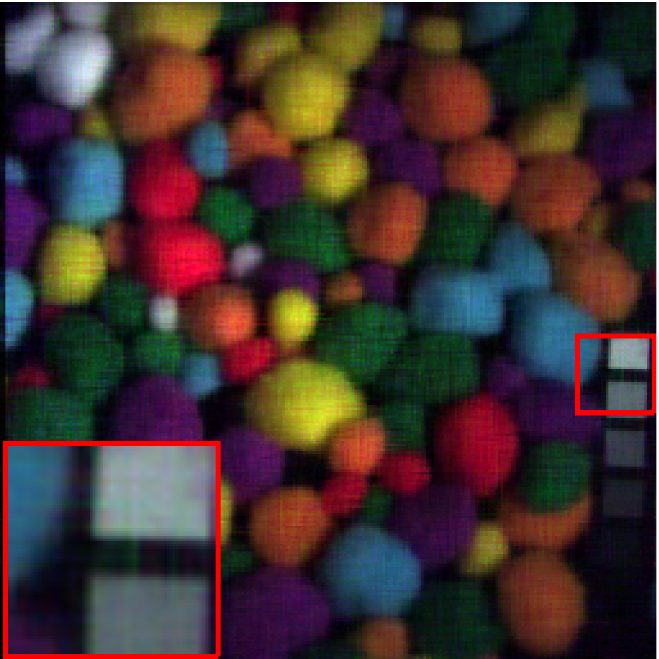}&
			\includegraphics[width=0.2\linewidth]{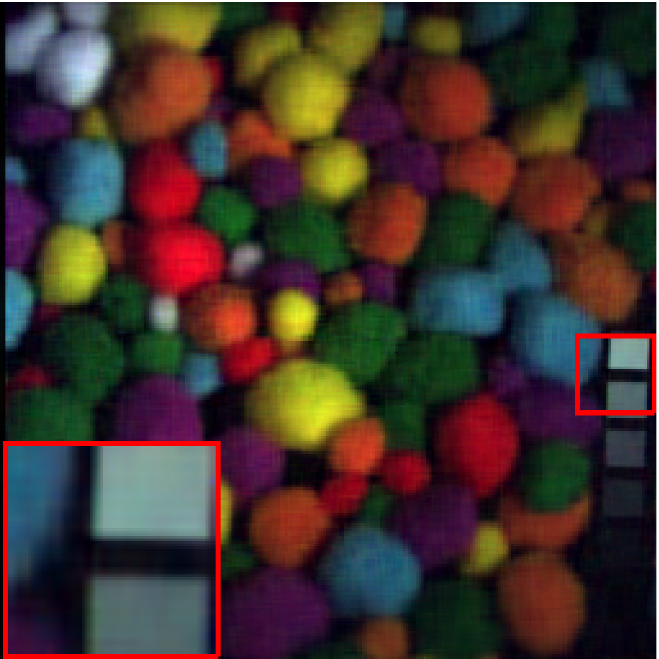}&
            \includegraphics[width=0.2\linewidth]{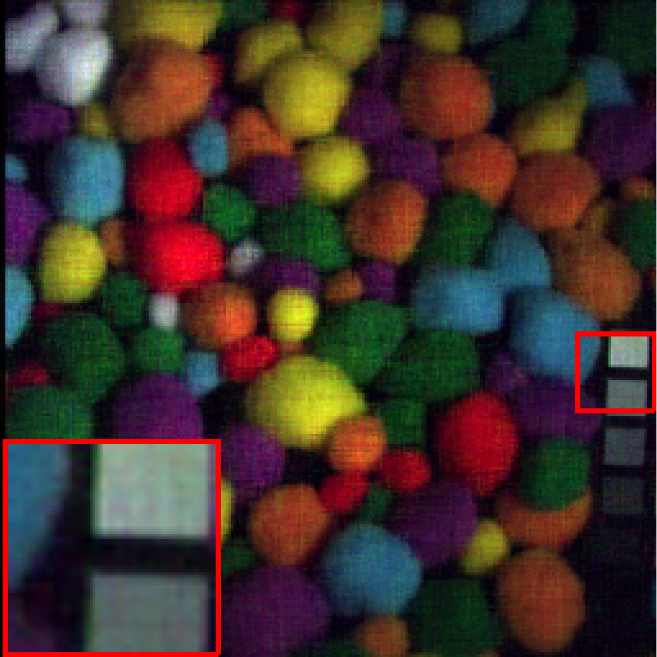}&
            \includegraphics[width=0.2\linewidth]{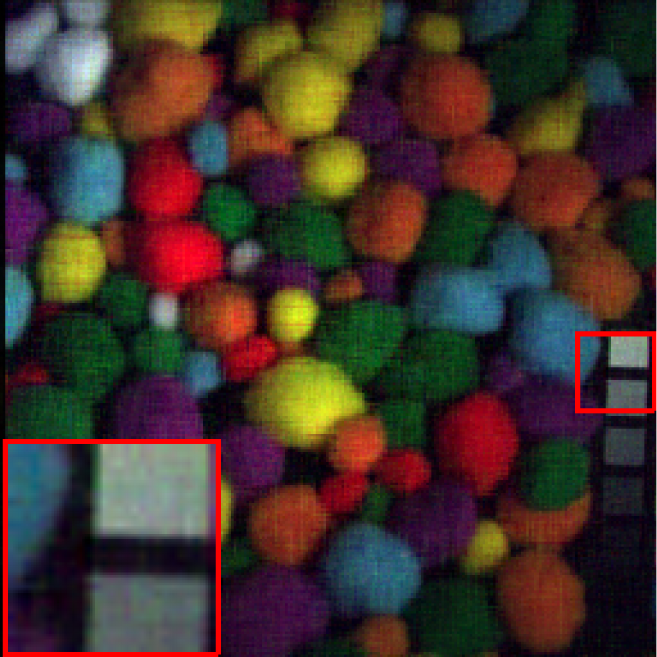}&
            \includegraphics[width=0.2\linewidth]{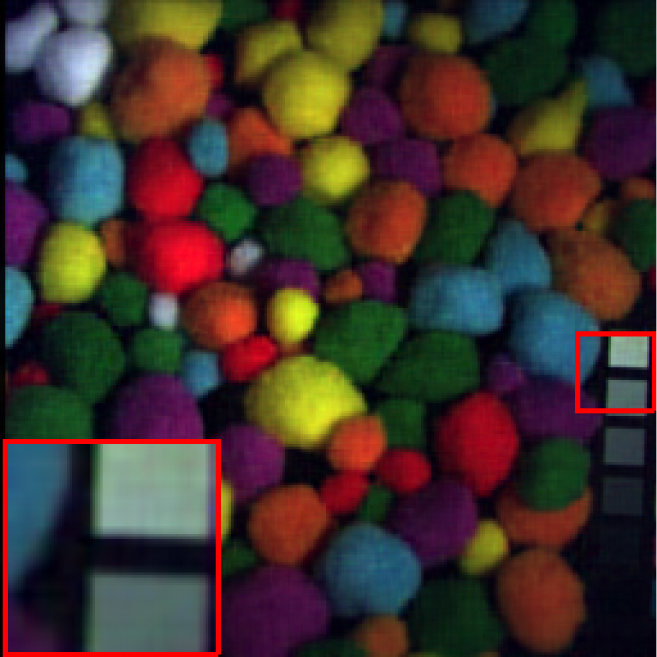}\\
			
			\includegraphics[width=0.2\linewidth]{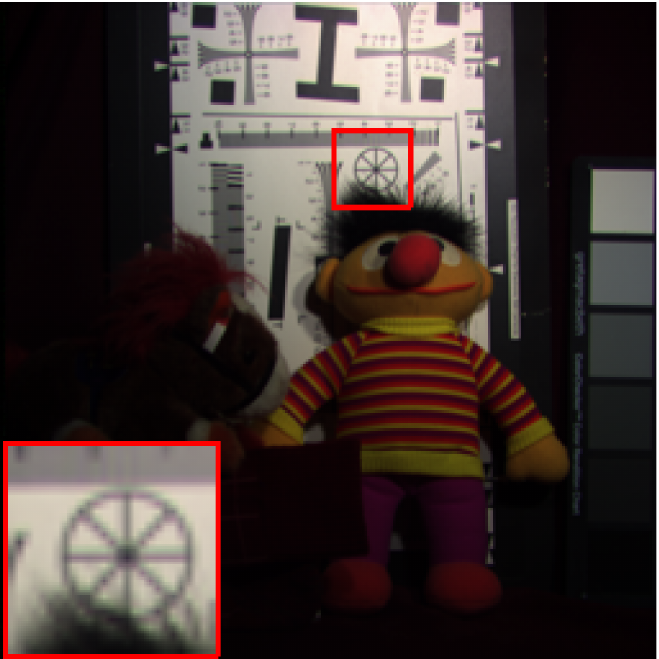}&
			\includegraphics[width=0.2\linewidth]{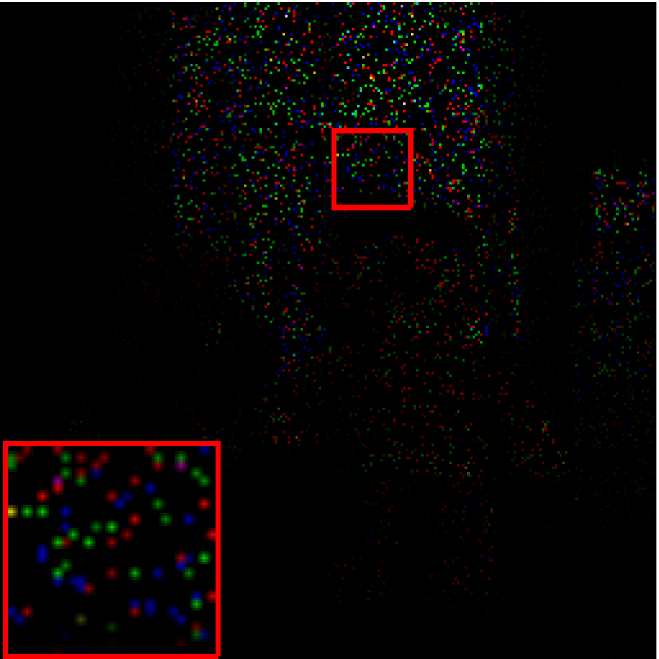}&
			\includegraphics[width=0.2\linewidth]{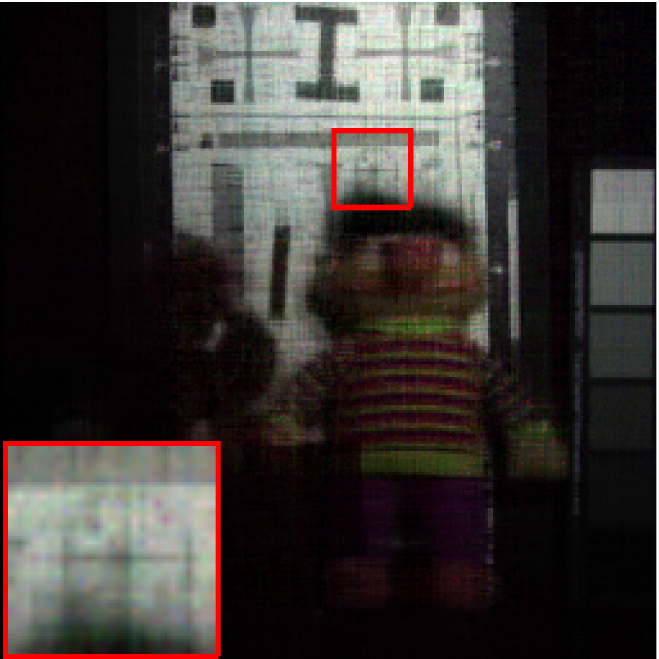}&
			\includegraphics[width=0.2\linewidth]{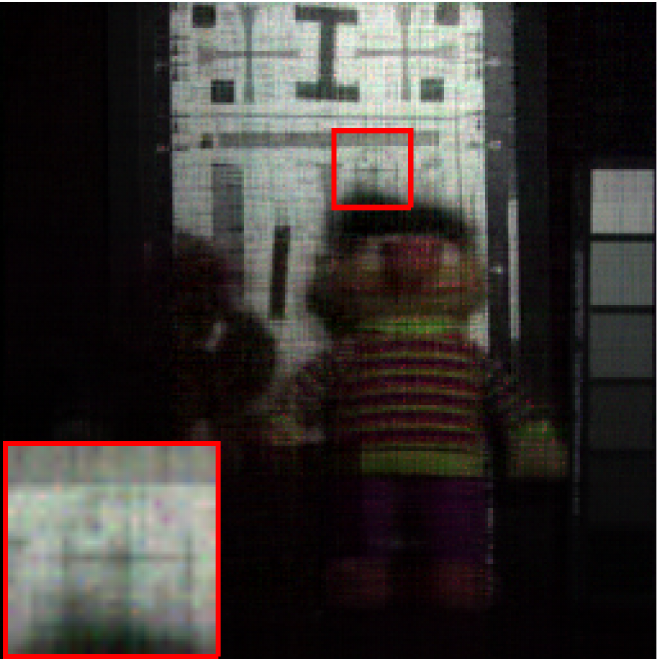}&
			\includegraphics[width=0.2\linewidth]{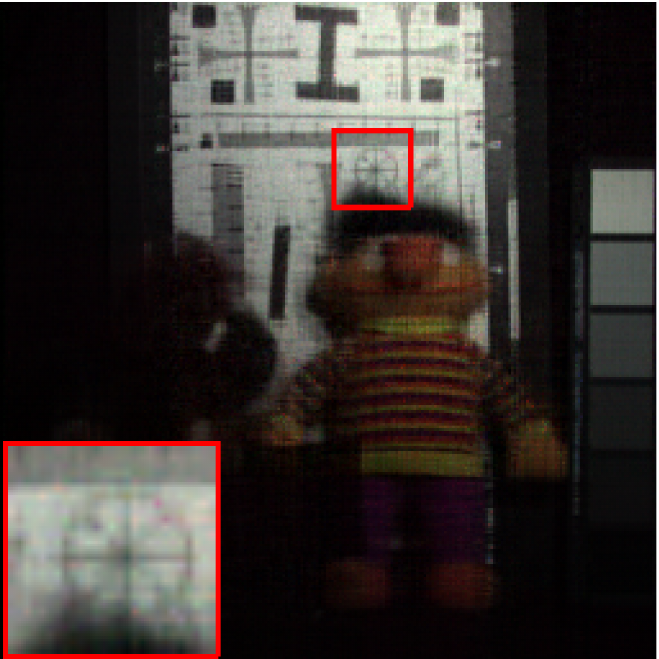}&
			\includegraphics[width=0.2\linewidth]{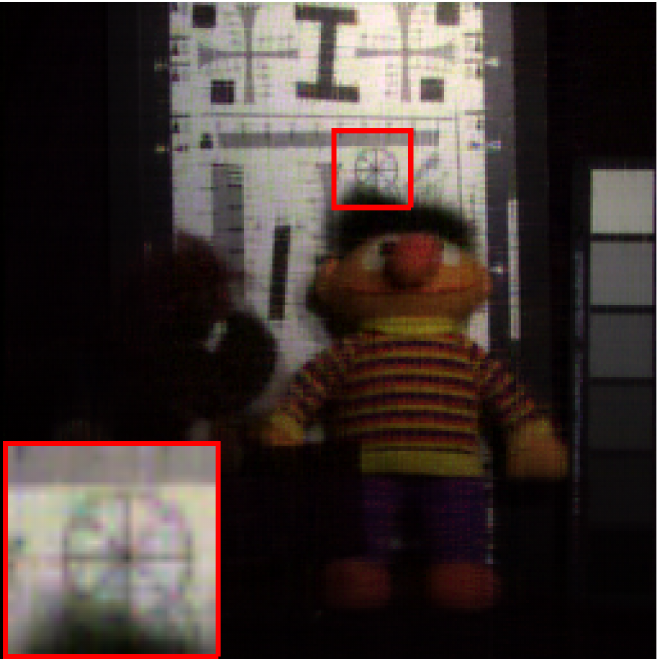}&
			\includegraphics[width=0.2\linewidth]{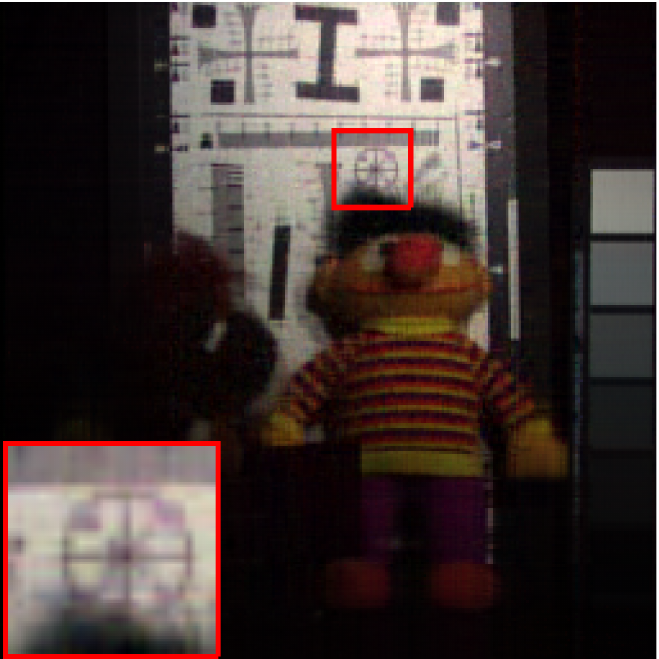}&
            \includegraphics[width=0.2\linewidth]{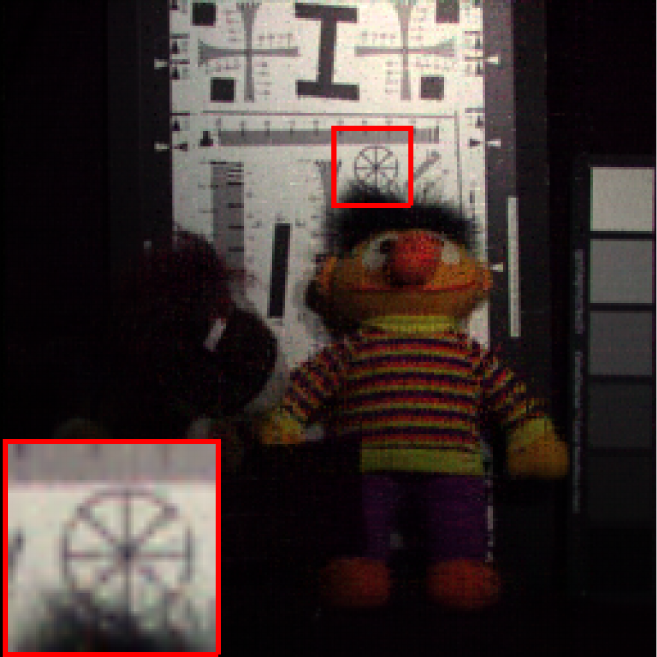}&
            \includegraphics[width=0.2\linewidth]{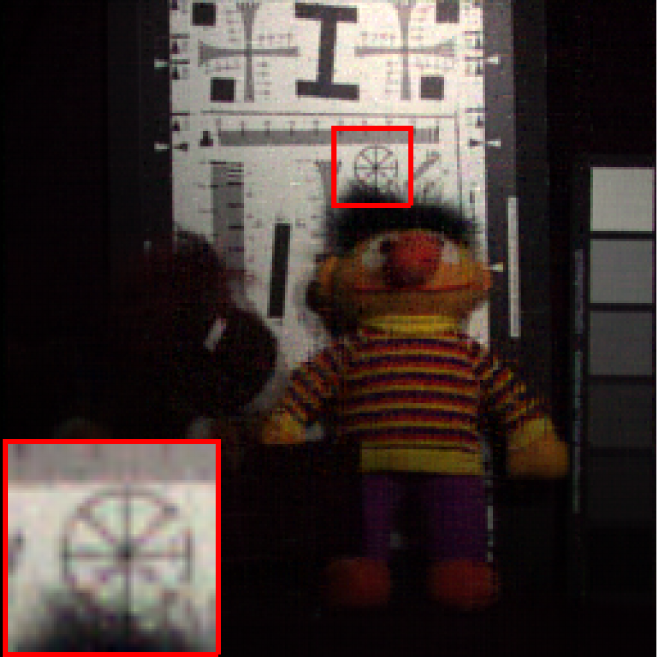}&
            \includegraphics[width=0.2\linewidth]{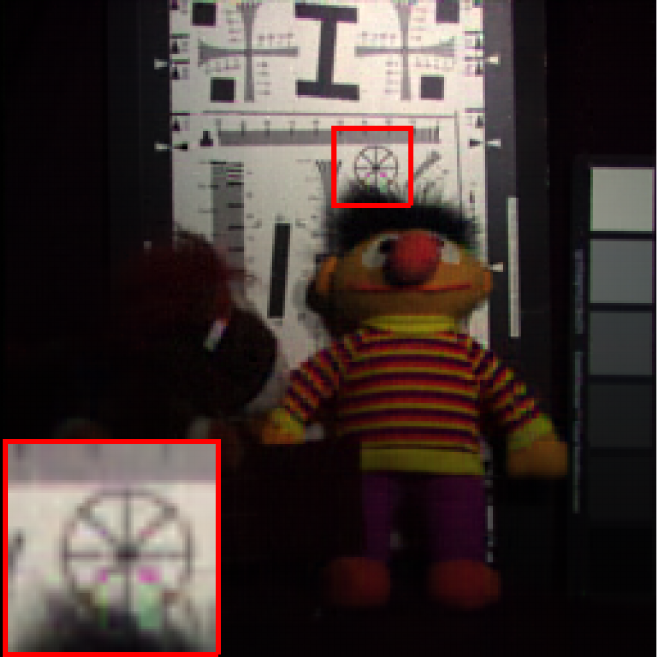}\\
			
			Original & Observed & TNN &TQRTNN  & UTNN & DTNN & LS2T2NN & HLRTF & OTLRM* & OTLRM\\
               && \cite{lu_TNN} &\cite{TQRTNN}&\cite{TTNN_song}&\cite{DTNN_jiang}&\cite{liu2023learnable}&\cite{lrtf}&&
		\end{tabular}} 

		\caption{ The selected pseudo-color images of recovery results by different methods on MSIs under \emph{SR=0.05}.  From top to bottom: \emph{Balloon},  \emph{Beer}, \emph{Pompom}, and \emph{Toy}. }
		\label{MSIimshow05}
	\end{center}
\end{figure*}

\begin{figure*}[htb]
	\footnotesize
	\setlength{\tabcolsep}{1pt}
	\begin{center}
             \scalebox{0.52}{
		\begin{tabular}{cccccccccc}
			\includegraphics[width=0.21\textwidth]{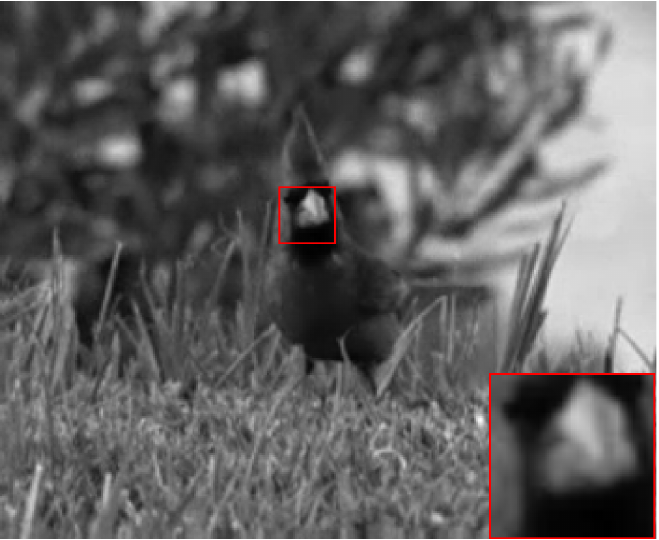} &
			\includegraphics[width=0.21\textwidth]{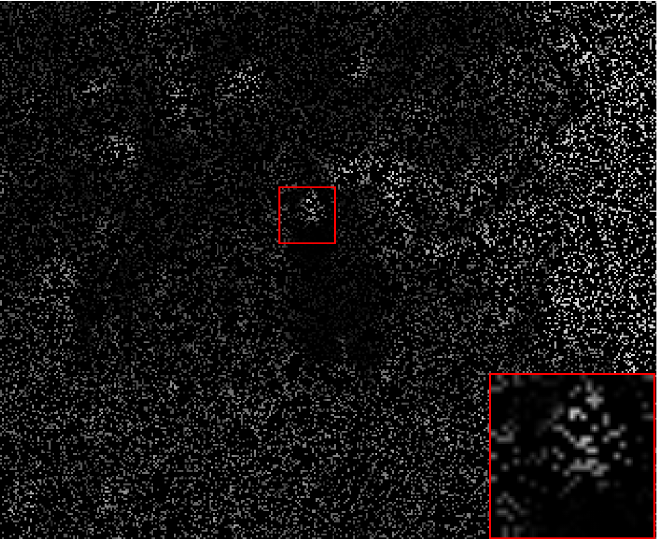}&
			\includegraphics[width=0.21\textwidth]{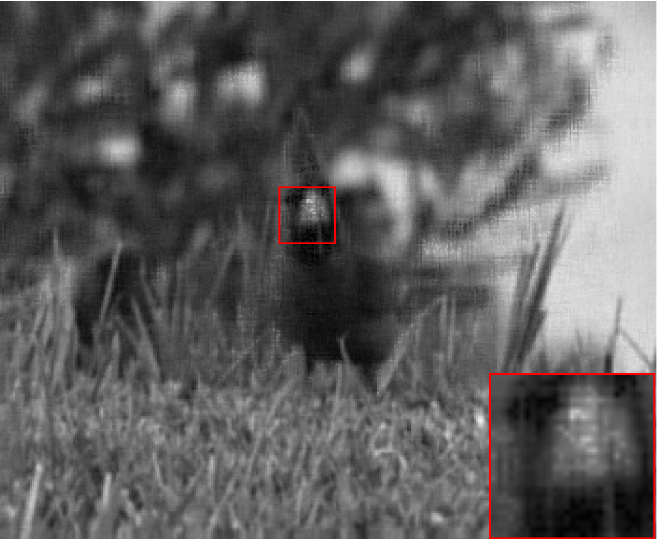}&
			\includegraphics[width=0.21\textwidth]{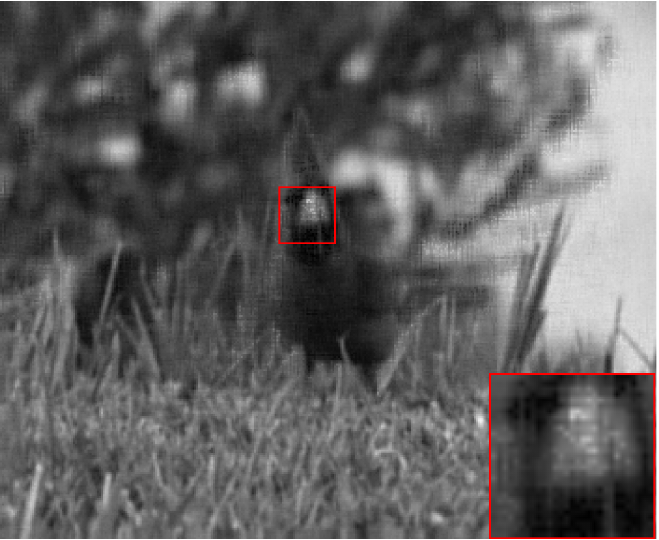}&
			\includegraphics[width=0.21\textwidth]{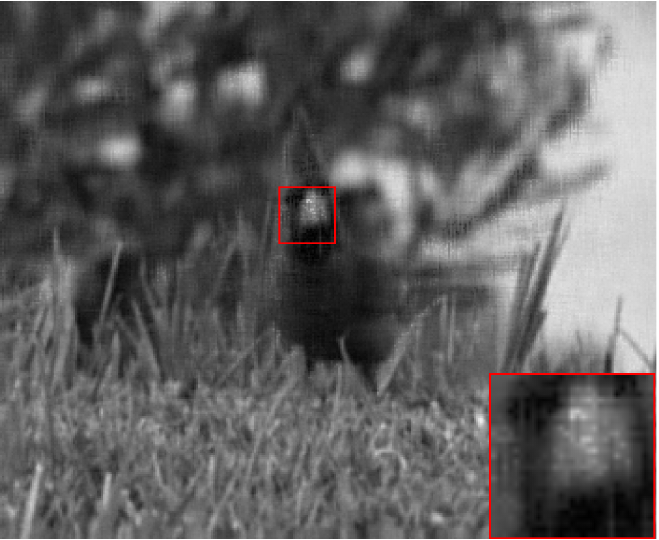}&
			\includegraphics[width=0.21\textwidth]{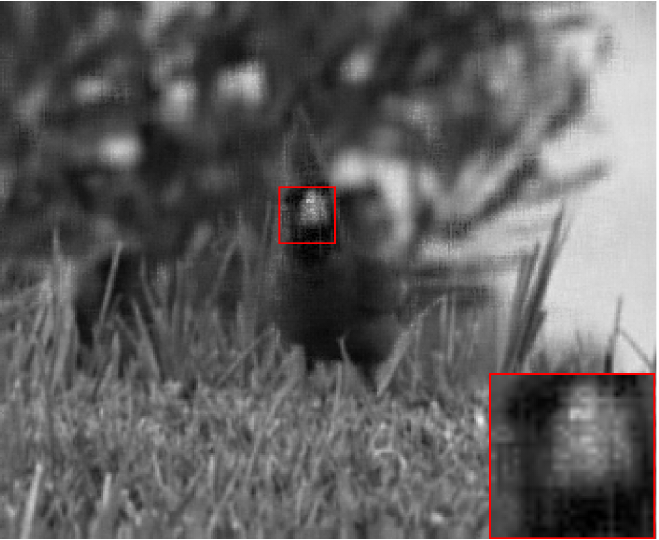}&
			\includegraphics[width=0.21\textwidth]{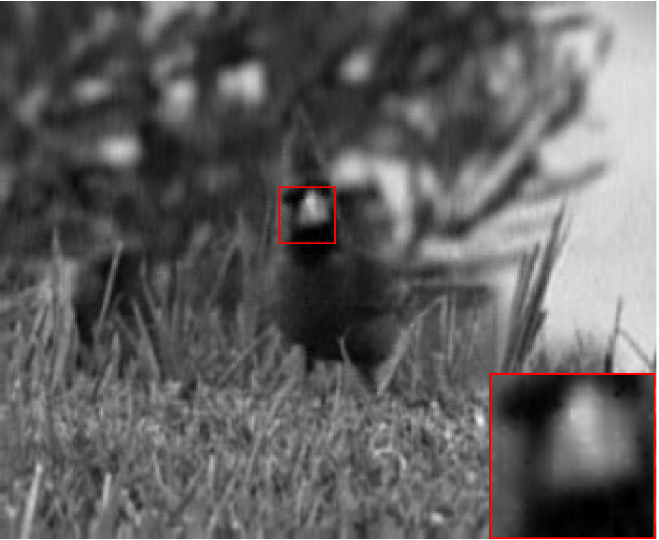}&
                \includegraphics[width=0.21\textwidth]{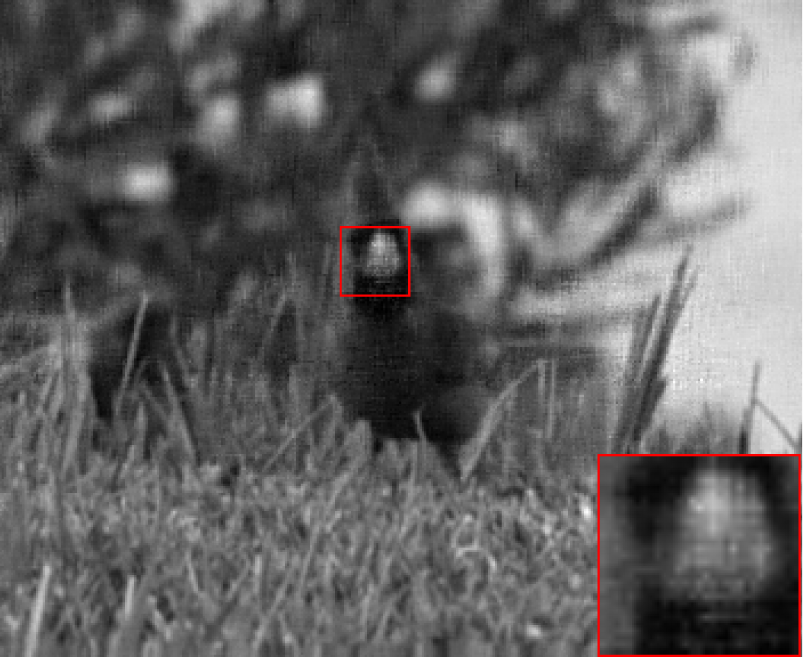}&
                \includegraphics[width=0.21\textwidth]{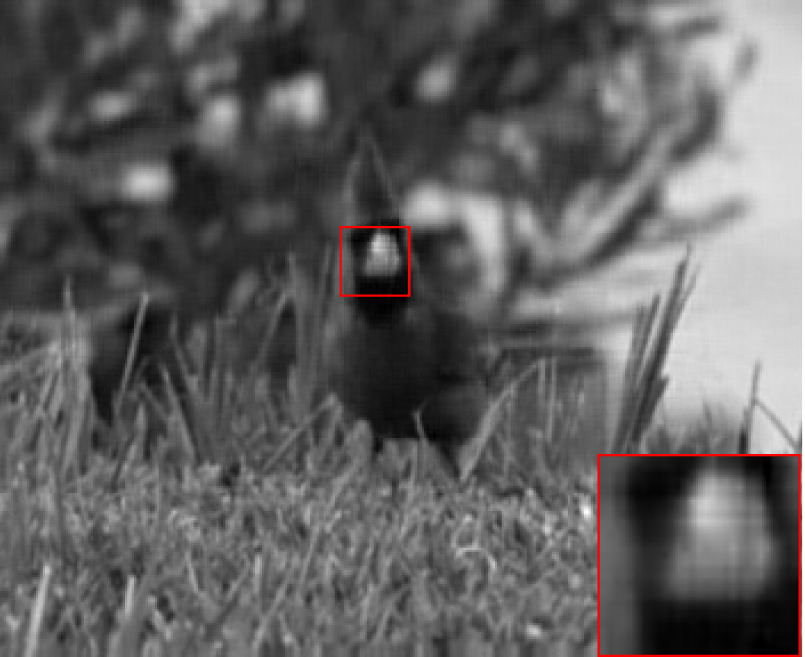}\\
			\includegraphics[width=0.21\textwidth]{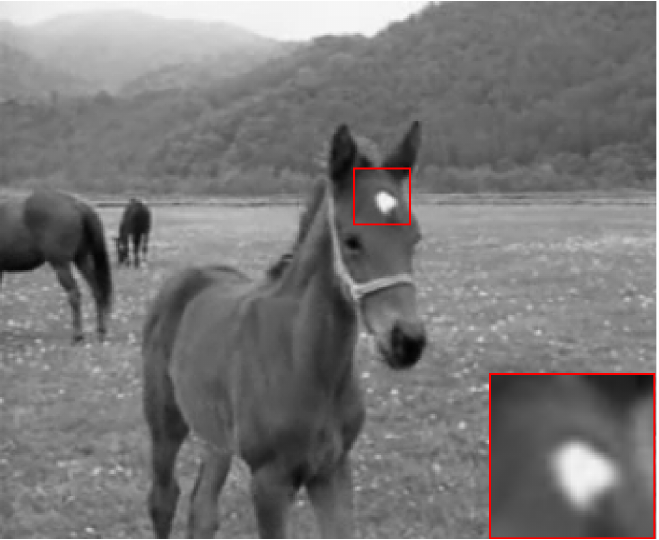} &
			\includegraphics[width=0.21\textwidth]{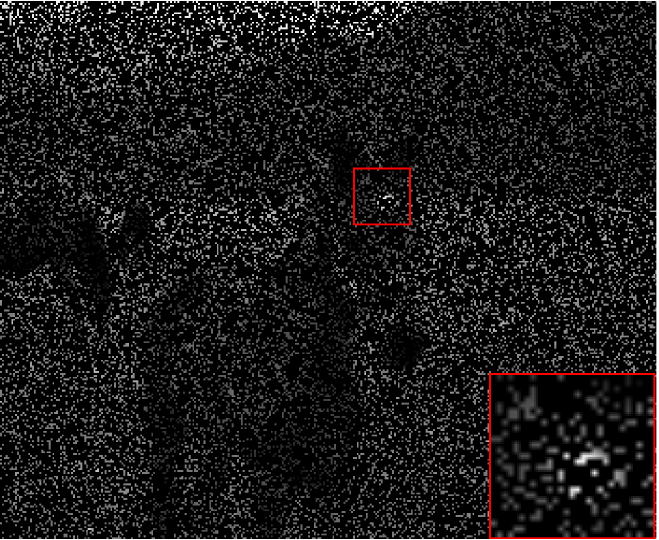}&
			\includegraphics[width=0.21\textwidth]{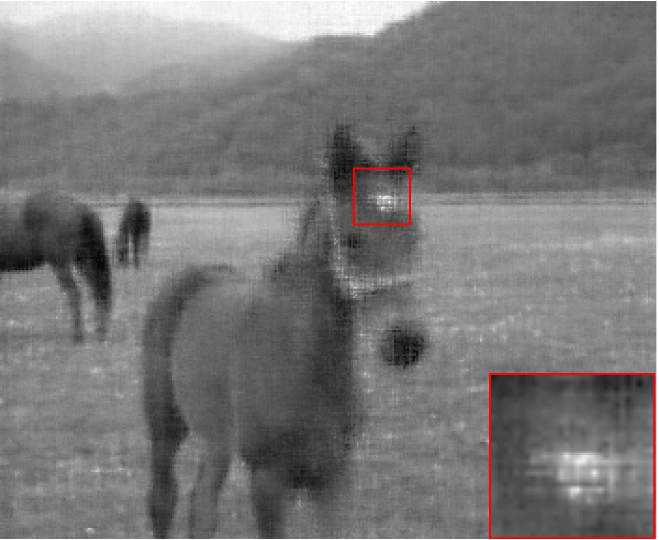}&
			\includegraphics[width=0.21\textwidth]{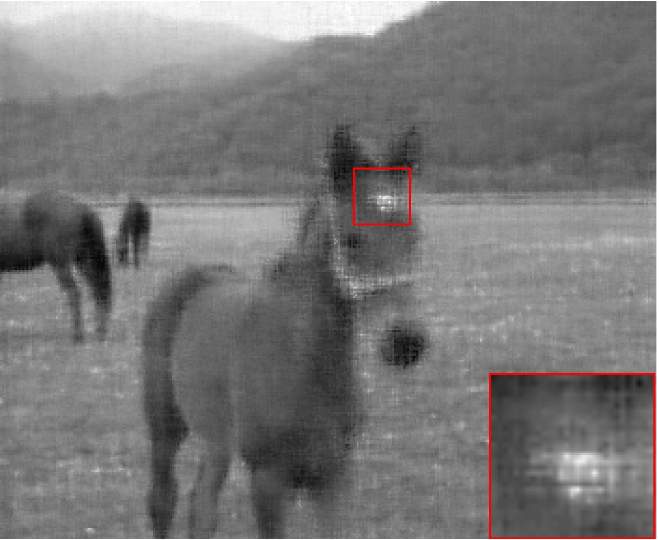}&
			\includegraphics[width=0.21\textwidth]{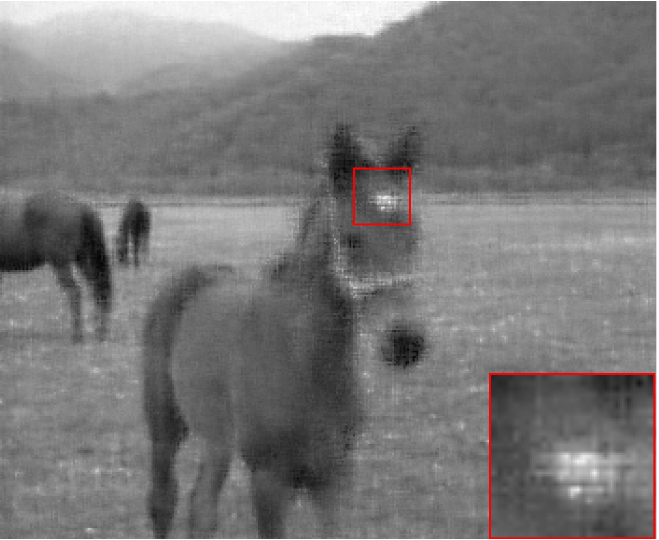}&
			\includegraphics[width=0.21\textwidth]{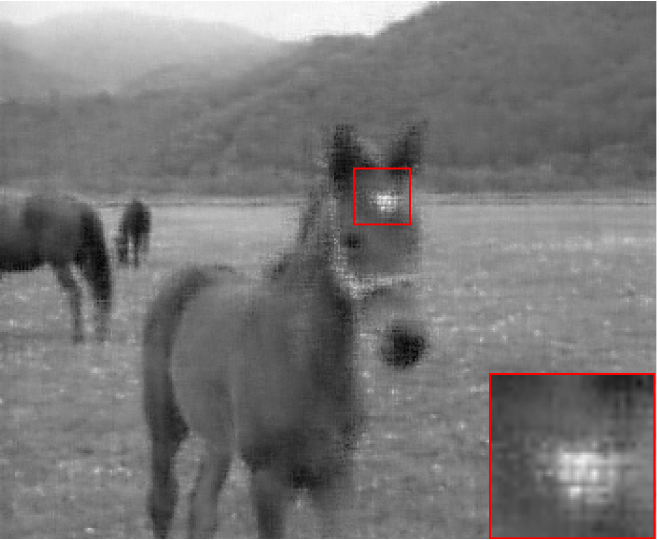}&
			\includegraphics[width=0.21\textwidth]{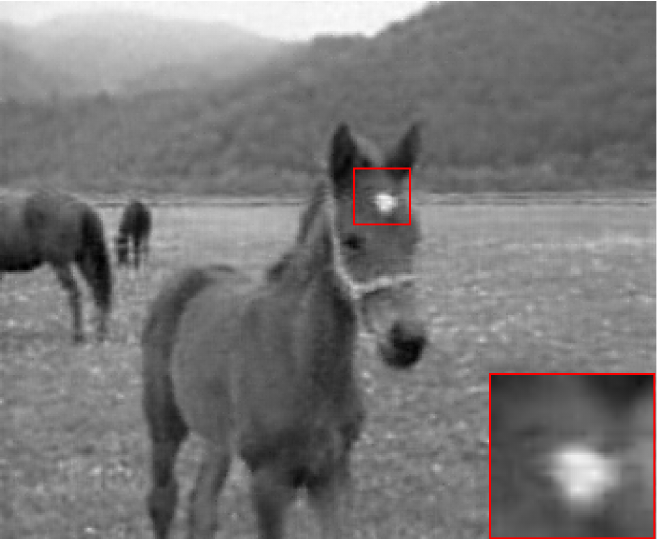}&
                \includegraphics[width=0.21\textwidth]{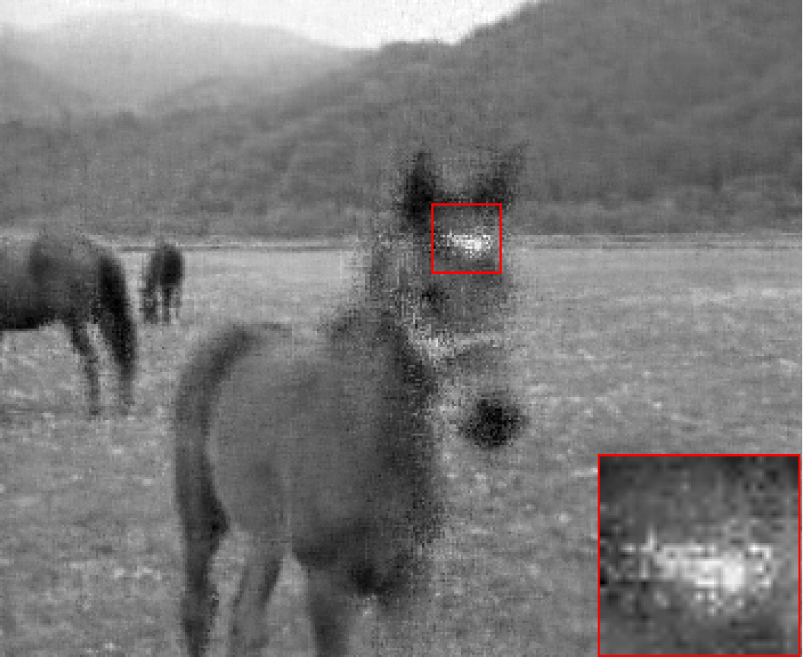}&
                \includegraphics[width=0.21\textwidth]{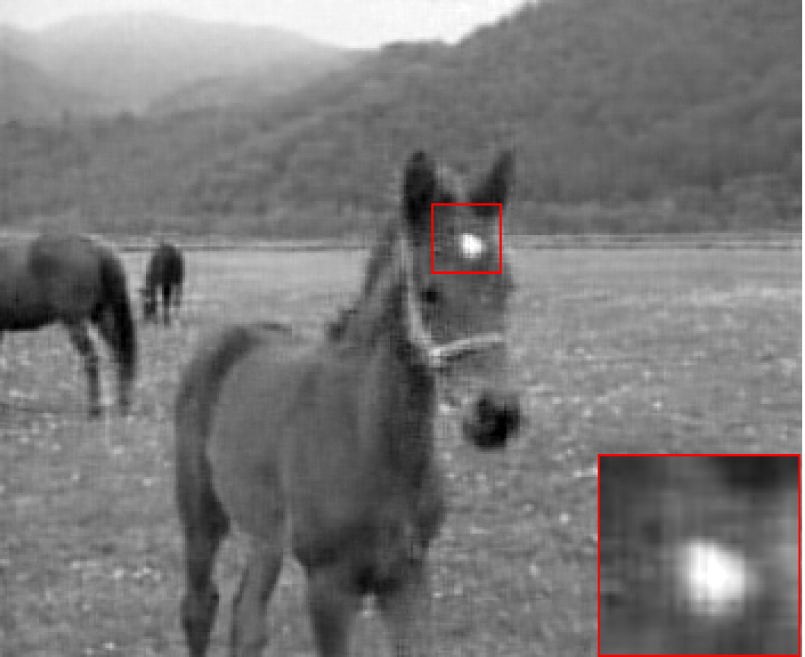}\\
			 \small Original &\small Observed & TNN &TQRTNN  & UTNN& DTNN& LS2T2NN& HLRTF& OTLRM\\
            && \cite{lu_TNN}&\cite{TQRTNN}&\cite{TTNN_song}&\cite{DTNN_jiang}&\cite{liu2023learnable}&\cite{lrtf}&&
            
		\end{tabular}}

		\caption{ Selected band of recovery results by different methods on videos under \emph{SR=0.20}.  From top to bottom: \emph{Bird} and \emph{Horse}. }
		\label{imagevideofig}
	\end{center}
\end{figure*}

We compare with other state-of-the-art methods, including a t-SVD baseline method TNN \cite{lu_TNN}, a tensor QR method TQRTNN \cite{TQRTNN}, a unitary transform-based method UTNN \cite{TTNN_song}, a dictionary encoding-based method DTNN \cite{DTNN_jiang}, a learnable redundant transform-based method LS2T2NN \cite{liu2023learnable} and a generative tensor factorization based method HLRTF \cite{lrtf}.
Compared to the traditional TNN-based methods, our model demonstrates excellent results of learnable orthogonal transformations and the generative t-SVD model, which maintains flexible data-adaption low-rank property.
From Figure \ref{MSIimshow05} and \ref{imagevideofig}, we can observe that with $\mathrm{OTV}$ loss our method is smoother and cleaner than other SOTA methods.

For \emph{Balloons}, \emph{Beer}, \emph{Pompom} and \emph{Toy} on \emph{CAVE} dataset, we evaluate the PSNR and SSIM under \emph{SR=0.20} in Table \ref{msitableresult_02}.
And the selected pseudo-color images are shown in Figure \ref{MSIimshow05_sr0.2}.
We select the \emph{Feathers} on \emph{CAVE} dataset under \emph{SR=0.05}, \emph{SR=0.10}, \emph{SR=0.15} and \emph{SR=0.20}.
Table \ref{msitableresult_feathers} and Figure \ref{MSIimshow05_feathers} demonstrate the numerical results and visual presentation under different SRs.
Figure \ref{imagevideofig_sr0.1} illustrates the results of video recovery under \emph{SR=0.10}.

\begin{table*}[htb]
        \centering
        \caption{Evaluation PSNR, SSIM on \emph{CAVE} dataset of \textbf{tensor completion} results by different methods on \textbf{MSIs} under \emph{SR=0.20}.}

        \def\arraystretch{1.0}
        \setlength{\tabcolsep}{1pt}
\scalebox{0.85}{
    \begin{tabular}{cc|cccccccc}
    \bottomrule[0.15em]
    \multirow{2}[2]{*}{\textbf{Method}} & \multirow{2}[2]{*}{\textbf{Reference}} & \multicolumn{2}{c}{\textbf{Balloons}} & \multicolumn{2}{c}{\textbf{Beer}} & \multicolumn{2}{c}{\textbf{Pompom}} & \multicolumn{2}{c}{\textbf{Toy}} \\
          &  & \textbf{PSNR}$\uparrow$  & \textbf{SSIM}$\uparrow$  & \textbf{PSNR}$\uparrow$  & \textbf{SSIM}$\uparrow$  & \textbf{PSNR}$\uparrow$  & \textbf{SSIM}$\uparrow$  & 
          \textbf{PSNR}$\uparrow$  & \textbf{SSIM}$\uparrow$\\
    \hline
    Observed & None & 14.27  & 0.20  & 10.40  & 0.05  & 12.40  & 0.13  &  11.92 & 0.36 \\
           TNN\cite{lu_TNN} & TPAMI 2019 & 42.28 & \underline{0.98} & 43.99  & \textbf{0.99}  & 34.39 & 0.91 & 36.68 & \underline{0.96} \\
           TQRTNN \cite{TQRTNN}&TCI 2021& 42.30  & \underline{0.98}  & 43.88  & \textbf{0.99} & 34.41  & 0.91  & 36.75 & \underline{0.96} \\
           UTNN\cite{TTNN_song} &NLAA 2020  & 46.40  & \textbf{0.99} & 47.18 & \textbf{0.99} &38.97  & 0.96 & 42.07 & \textbf{0.99}\\
          DTNN\cite{DTNN_jiang} & TNNLS 2023 & 47.42 & \textbf{0.99} & 48.47 & \textbf{0.99} & 39.64 & \underline{0.96} & 42.89 & \textbf{0.99}\\
           LS2T2NN  \cite{liu2023learnable} & TCSVT 2023 & \underline{47.97} & \textbf{0.99}  & \underline{48.55} & \textbf{0.99} & 41.30 & \underline{0.98} & 42.99 & \textbf{0.99}\\
           HLRTF\cite{lrtf} & CVPR 2022 & 47.50 & \textbf{0.99}  & 47.53 & \textbf{0.99} & \underline{43.31} & \textbf{0.99} & \underline{44.24} & \textbf{0.99}\\
           \textbf{OTLRM} & Ours & \textbf{48.53} & \textbf{0.99} & \textbf{48.61} & \textbf{0.99} & \textbf{44.24} & \textbf{0.99} & \textbf{46.94} & \textbf{0.99}\\
    \toprule[0.15em]
    \end{tabular}%
    } 
	\label{msitableresult_02}
\end{table*}

\begin{figure*}[htb]
	\footnotesize
	\setlength{\tabcolsep}{1pt}
	\begin{center}
            \scalebox{0.55}{
		\begin{tabular}{cccccccccc}
			\includegraphics[width=0.2\linewidth]{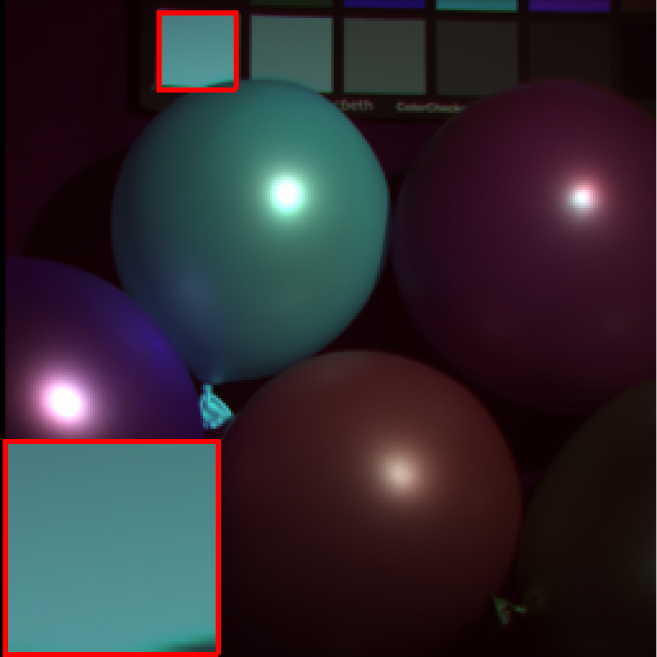}&
			\includegraphics[width=0.2\linewidth]{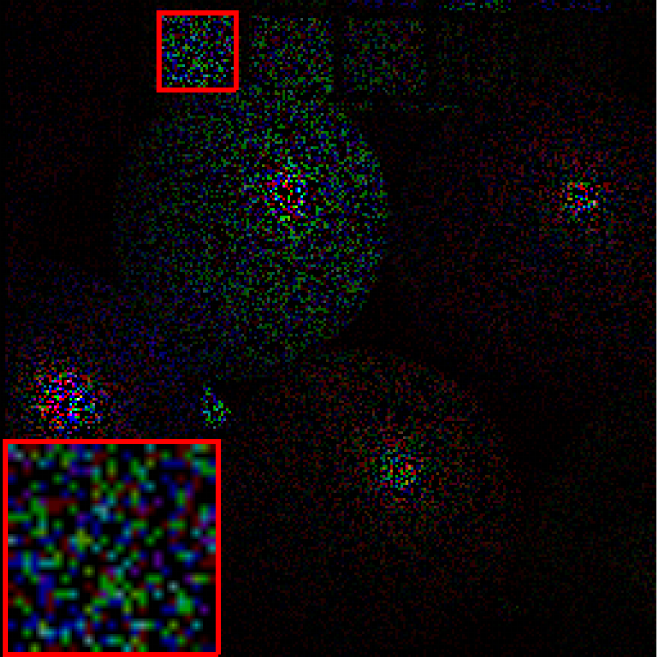}&
			\includegraphics[width=0.2\linewidth]{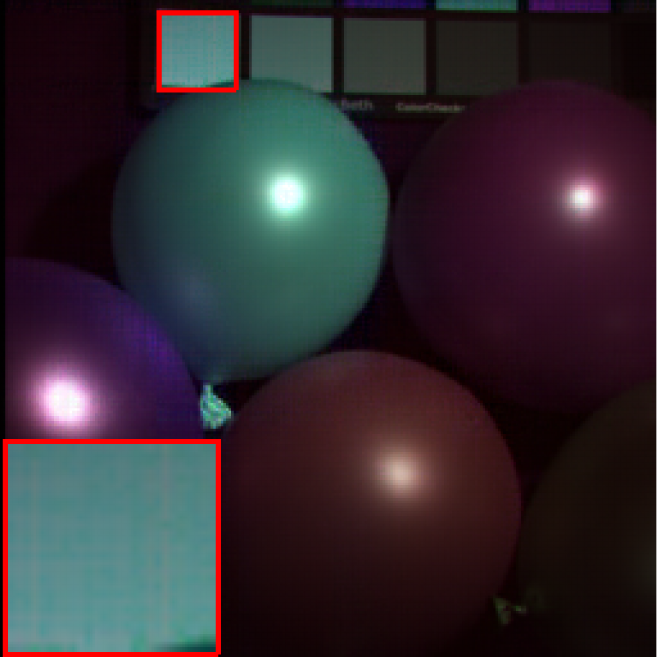}&
			\includegraphics[width=0.2\linewidth]{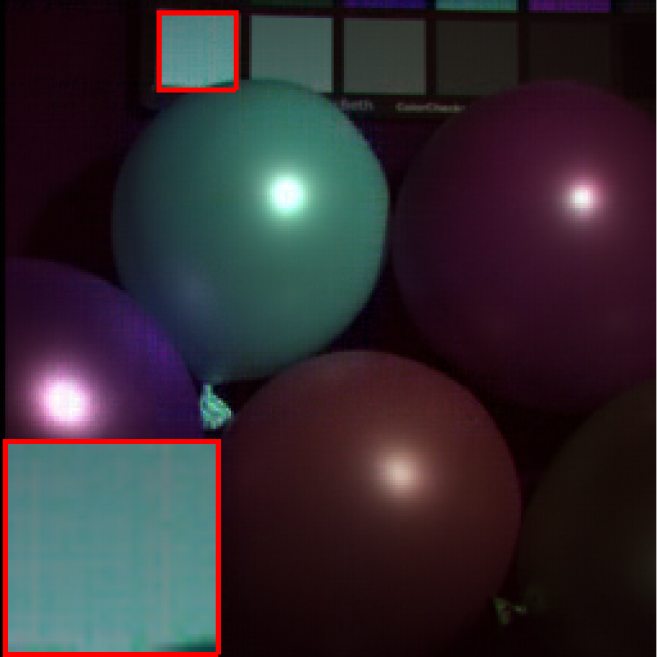}&
			\includegraphics[width=0.2\linewidth]{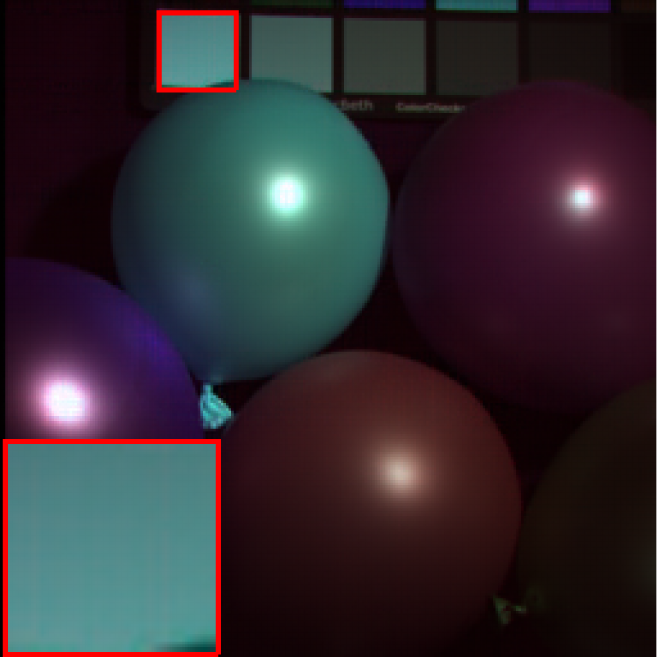}&
			\includegraphics[width=0.2\linewidth]{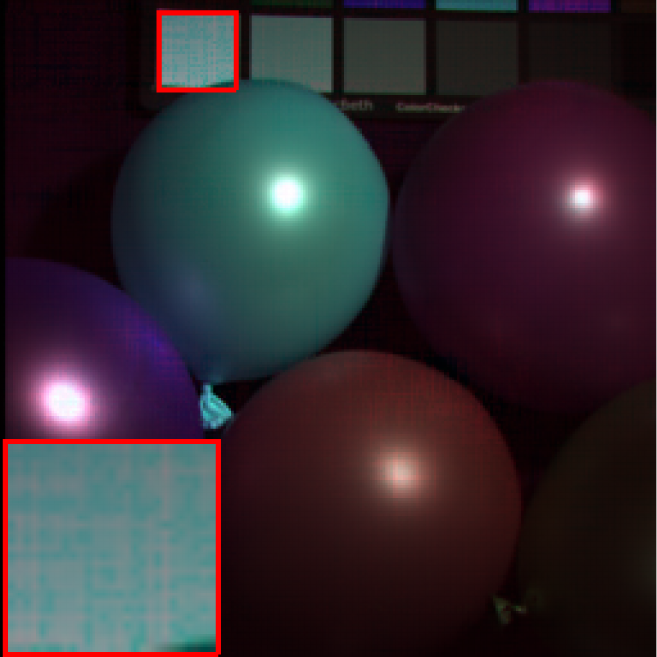}&
			\includegraphics[width=0.2\linewidth]{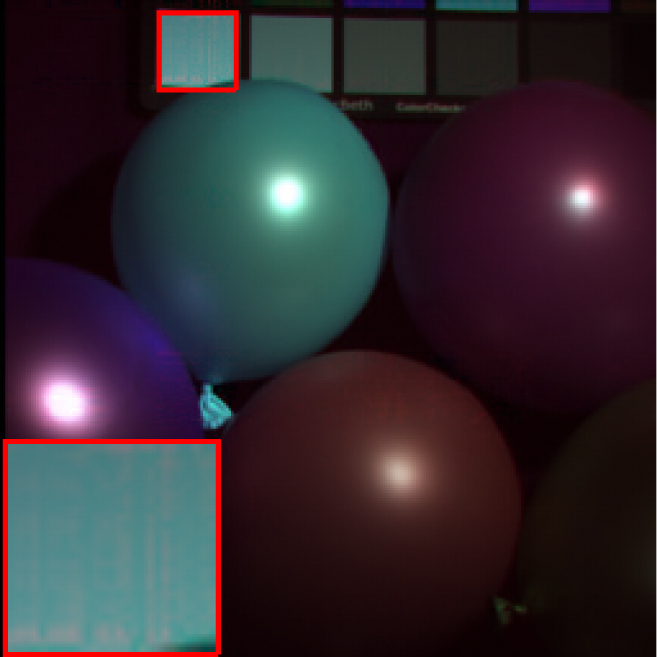}&
            \includegraphics[width=0.2\linewidth]{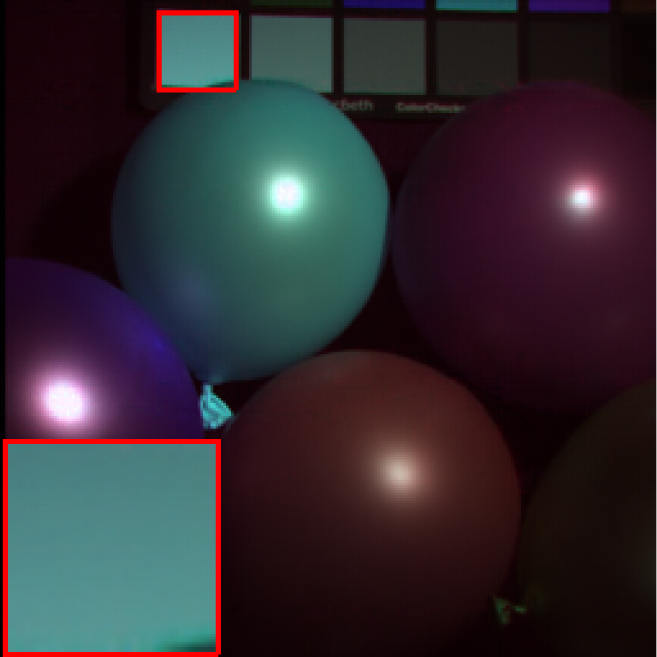}&
            \includegraphics[width=0.2\linewidth]{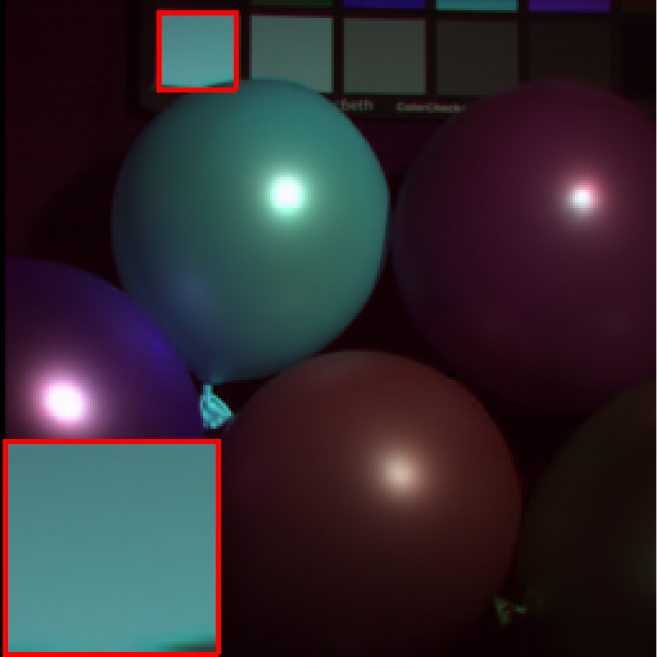}\\
			
			\includegraphics[width=0.2\linewidth]{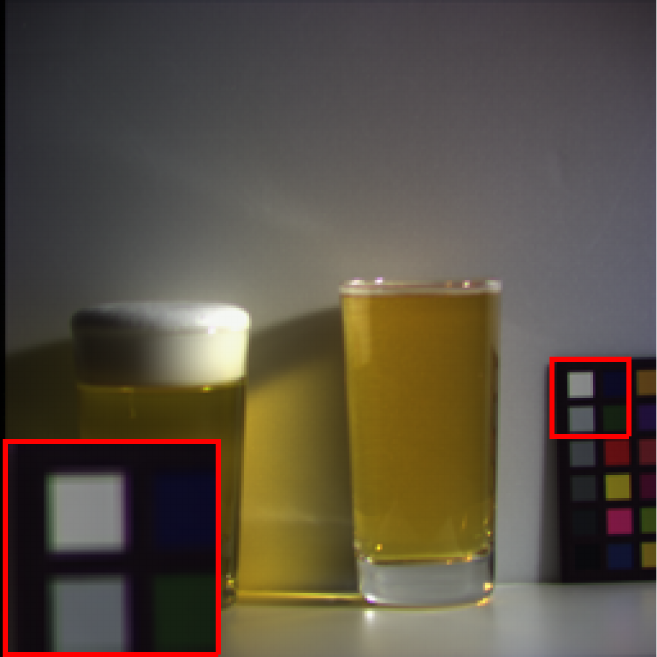}&
			\includegraphics[width=0.2\linewidth]{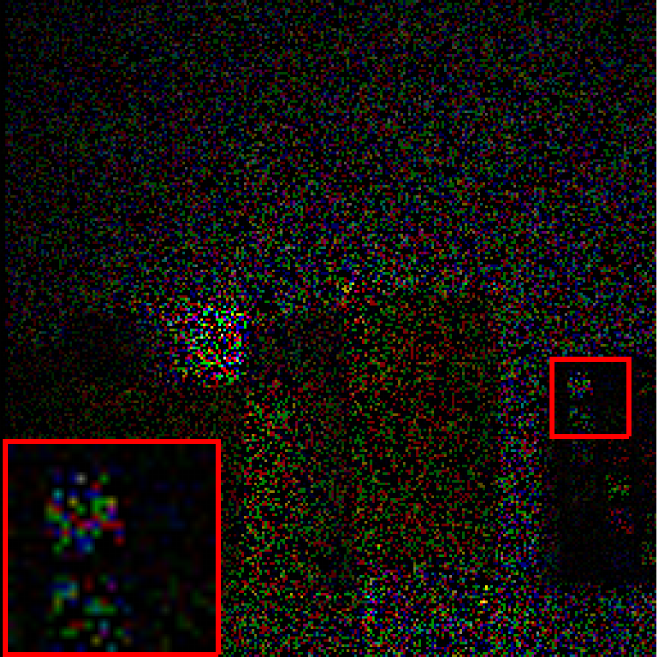}&
			\includegraphics[width=0.2\linewidth]{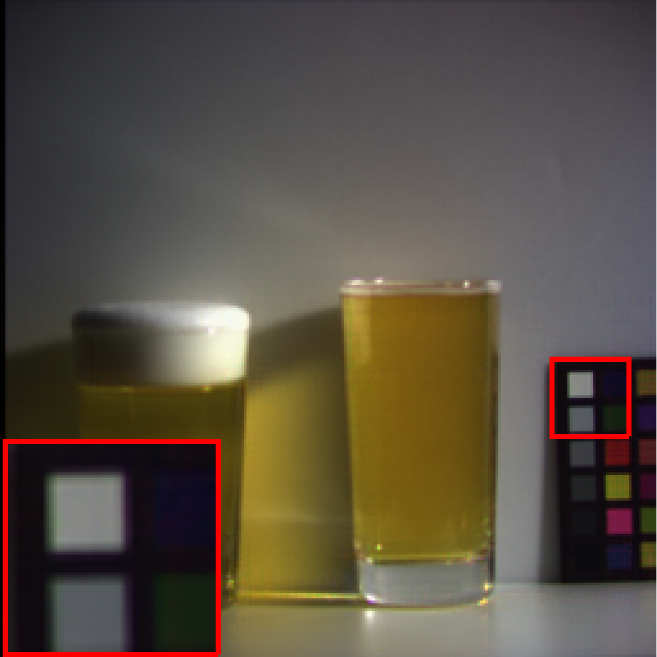}&
			\includegraphics[width=0.2\linewidth]{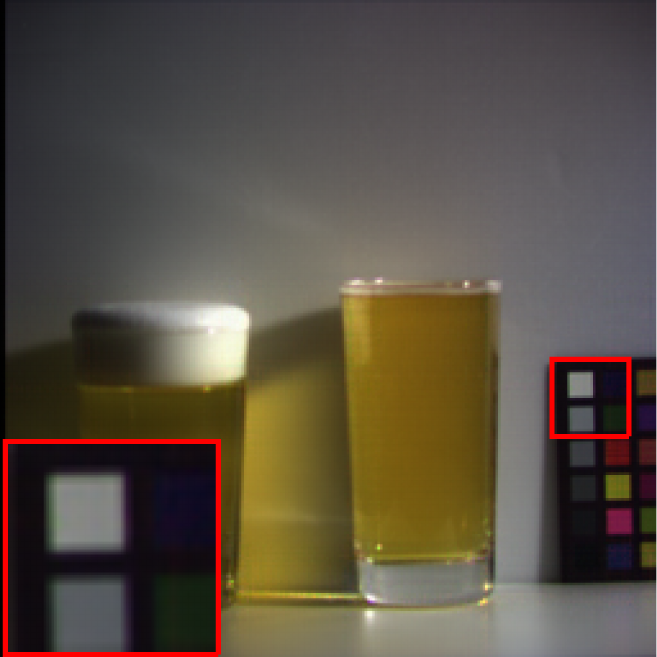}&
			\includegraphics[width=0.2\linewidth]{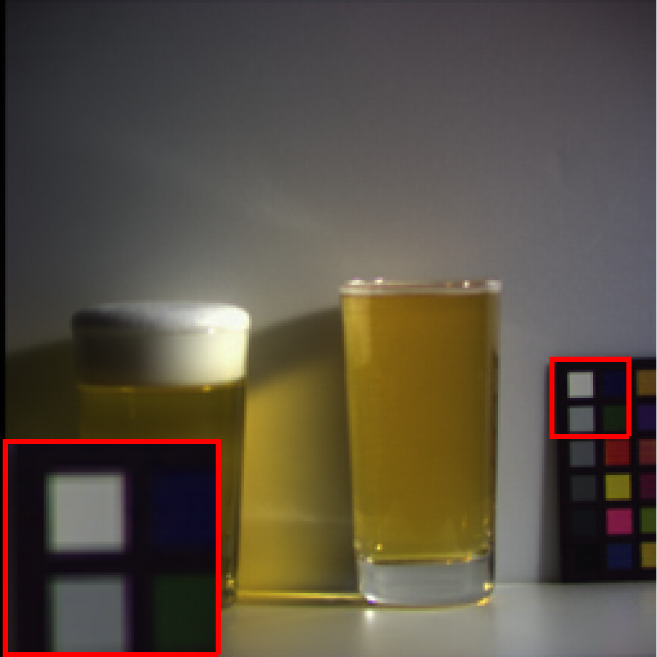}&
			\includegraphics[width=0.2\linewidth]{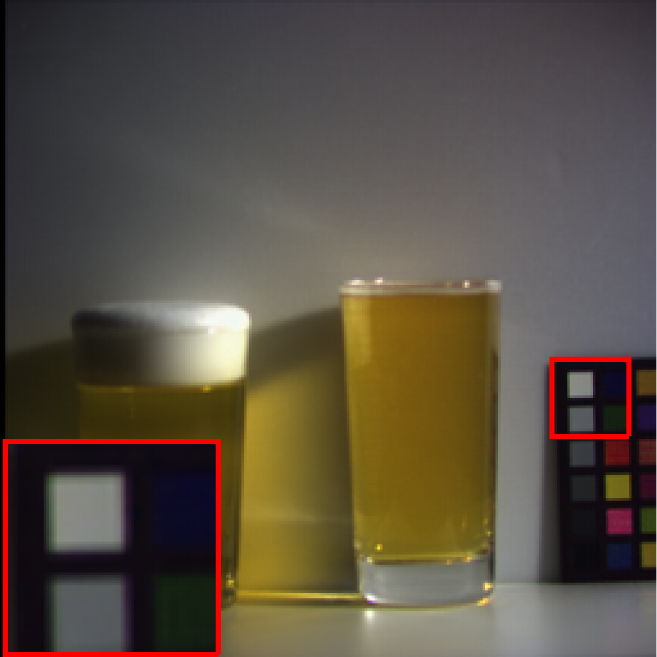}&
			\includegraphics[width=0.2\linewidth]{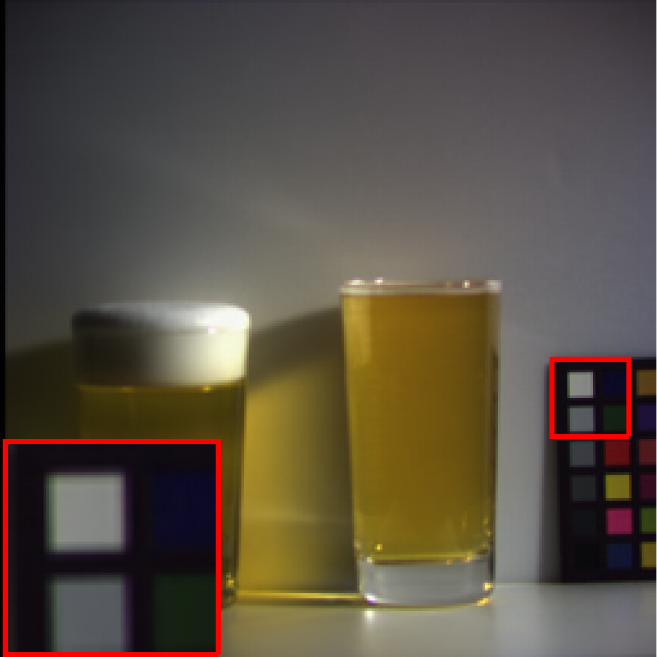}&
            \includegraphics[width=0.2\linewidth]{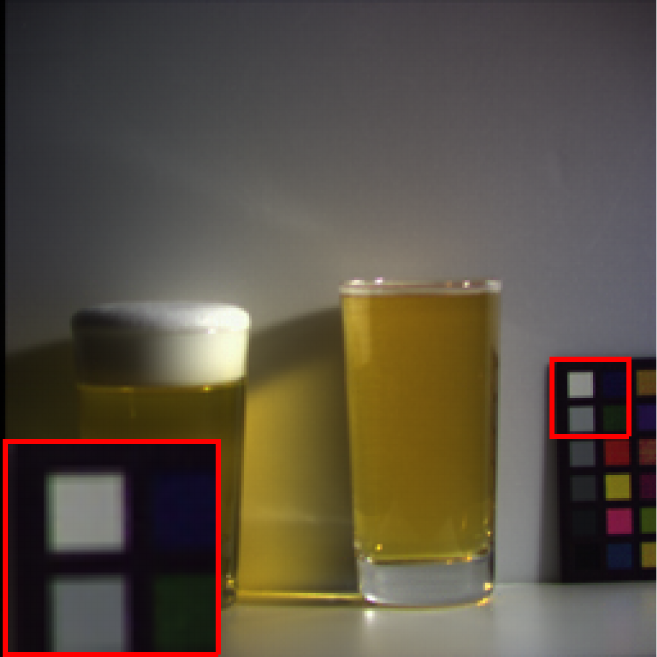}&
            \includegraphics[width=0.2\linewidth]{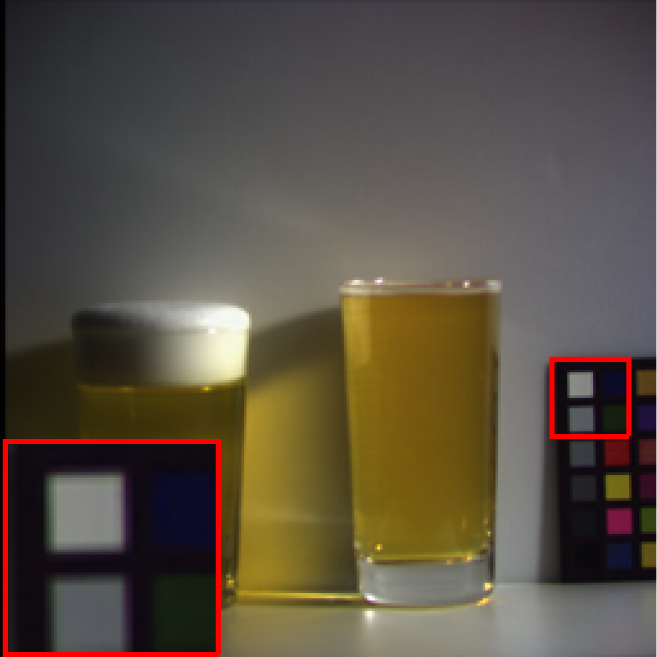}\\
			
			\includegraphics[width=0.2\linewidth]{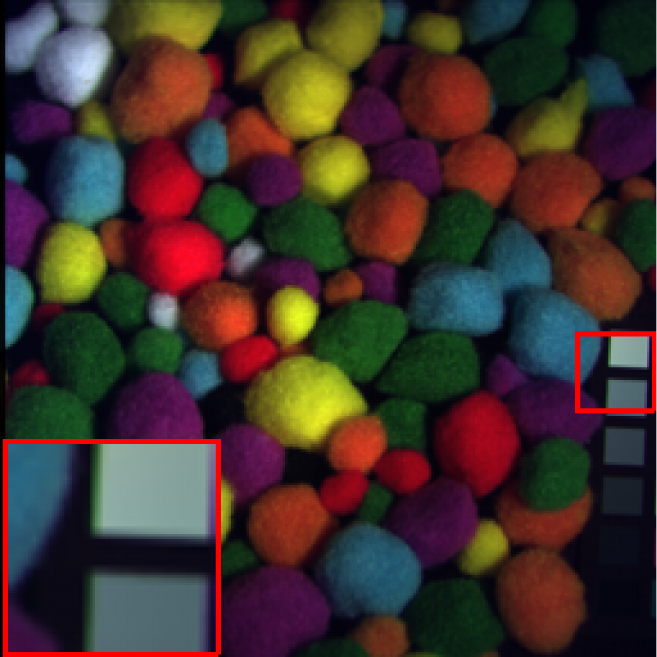}&
			\includegraphics[width=0.2\linewidth]{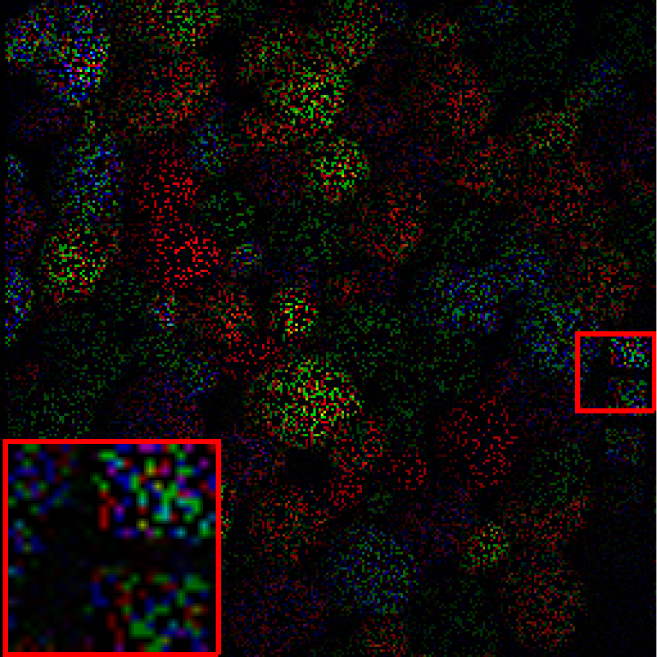}&
			\includegraphics[width=0.2\linewidth]{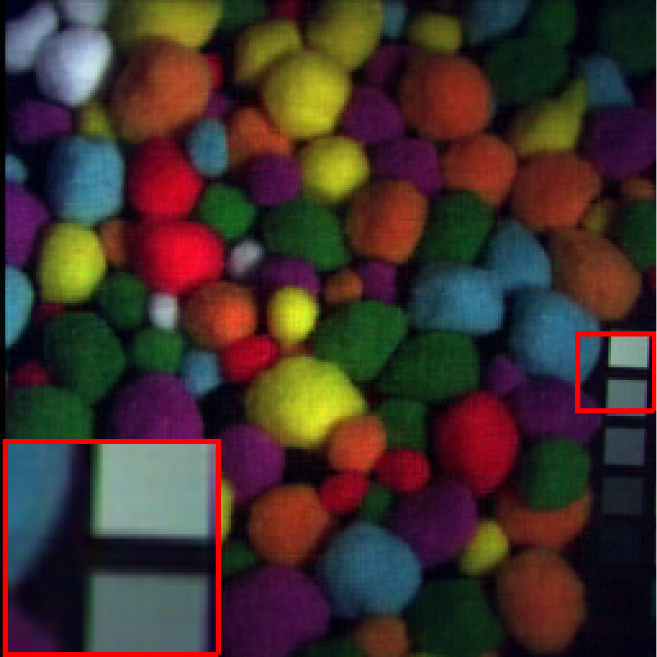}&
			\includegraphics[width=0.2\linewidth]{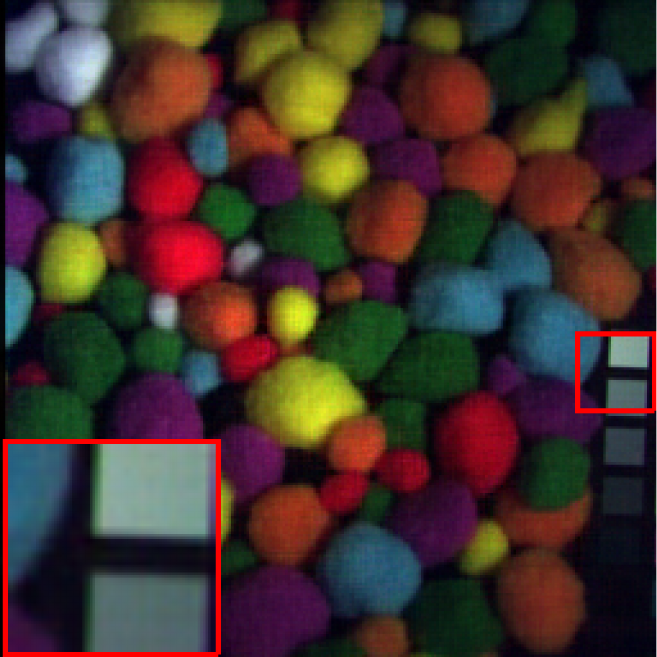}&
			\includegraphics[width=0.2\linewidth]{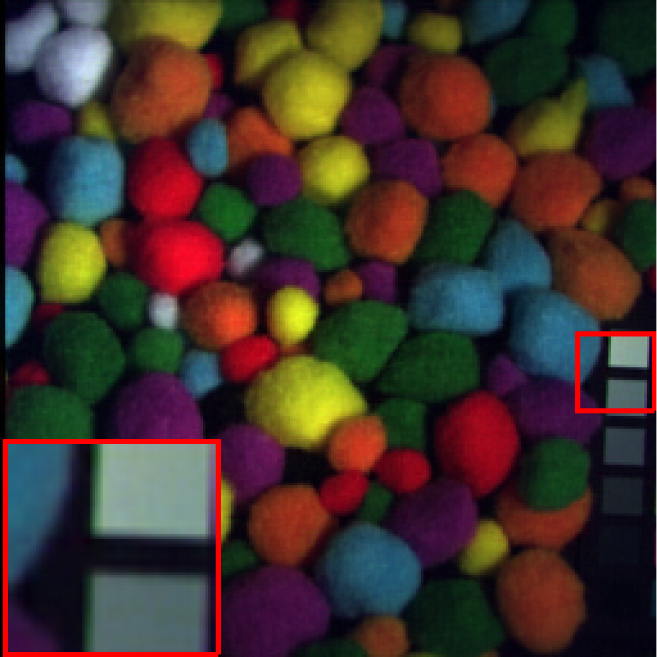}&
			\includegraphics[width=0.2\linewidth]{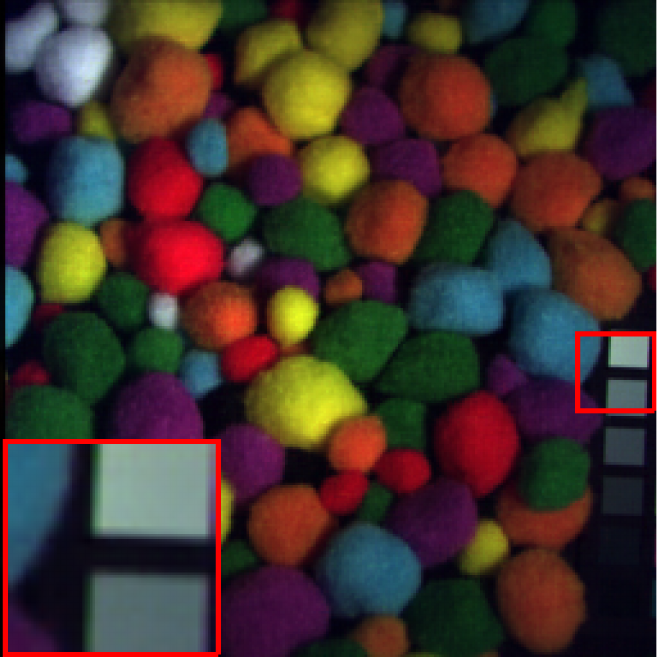}&
			\includegraphics[width=0.2\linewidth]{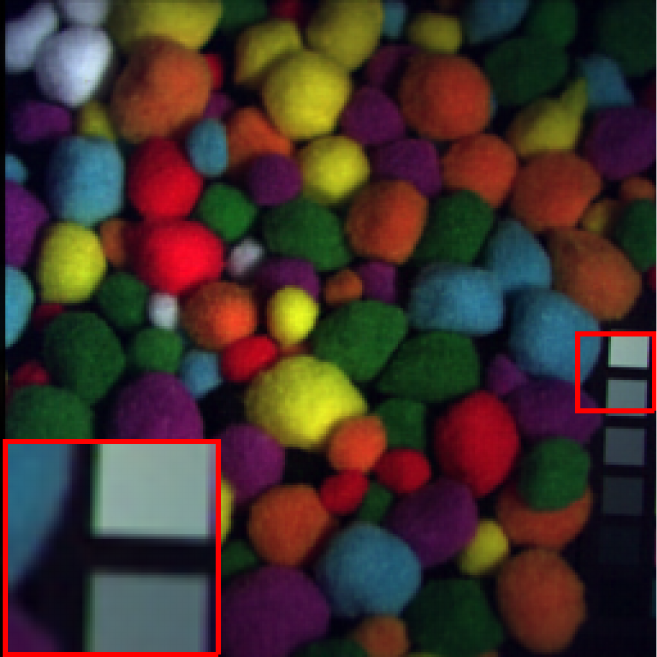}&
            \includegraphics[width=0.2\linewidth]{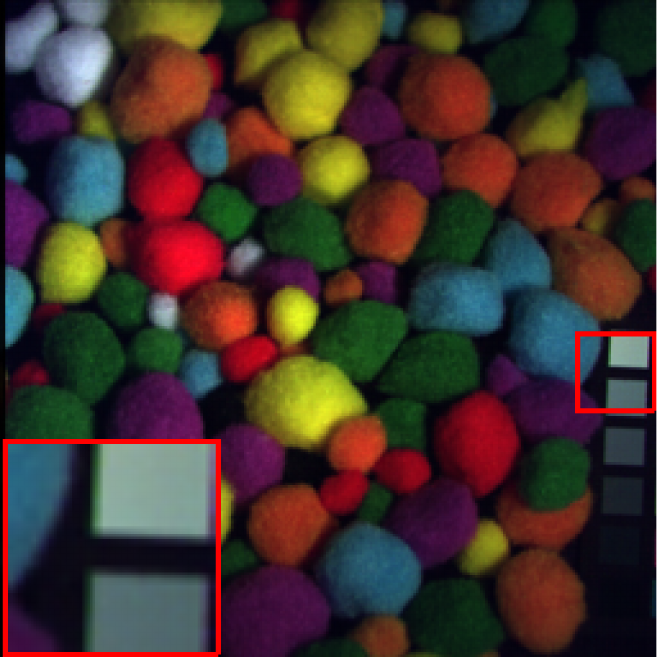}&
            \includegraphics[width=0.2\linewidth]{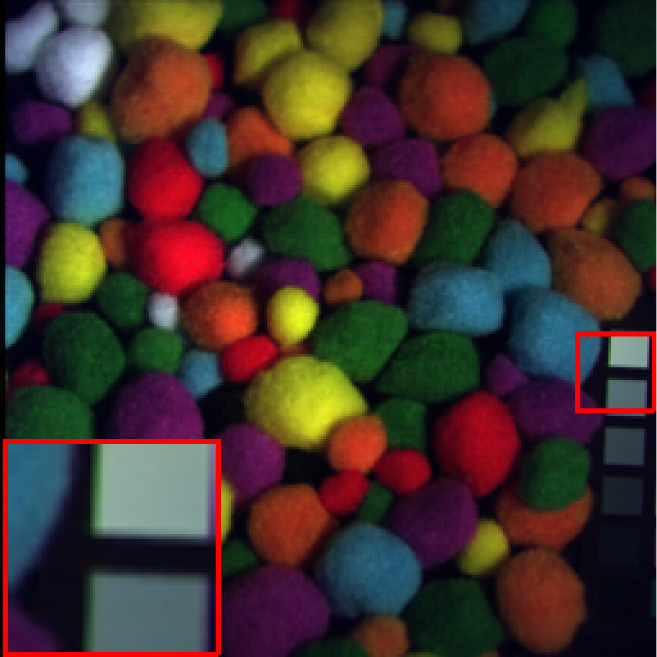}\\
			
			\includegraphics[width=0.2\linewidth]{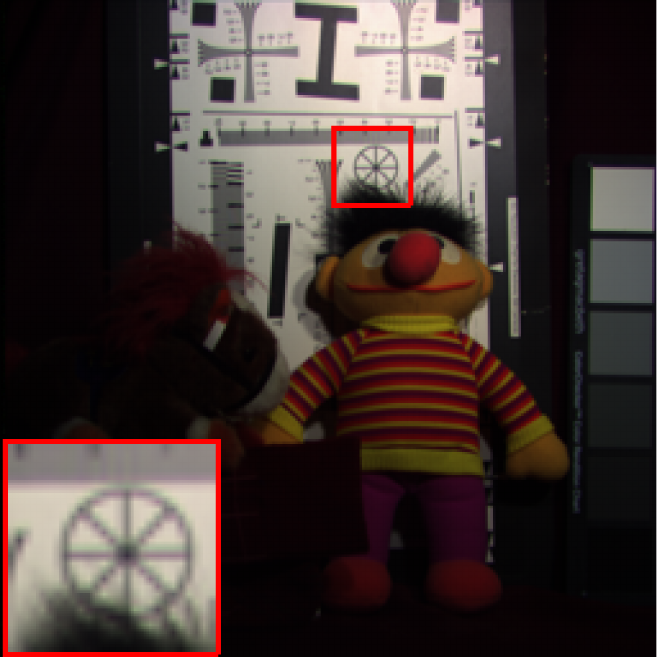}&
			\includegraphics[width=0.2\linewidth]{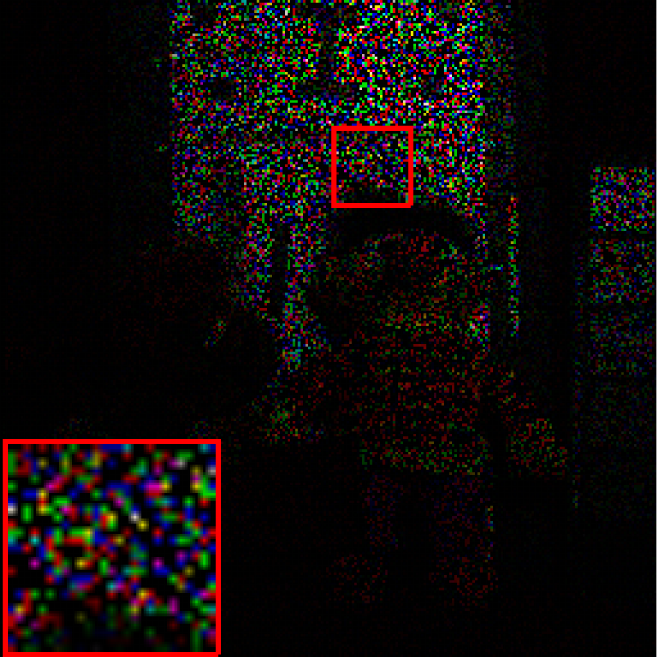}&
			\includegraphics[width=0.2\linewidth]{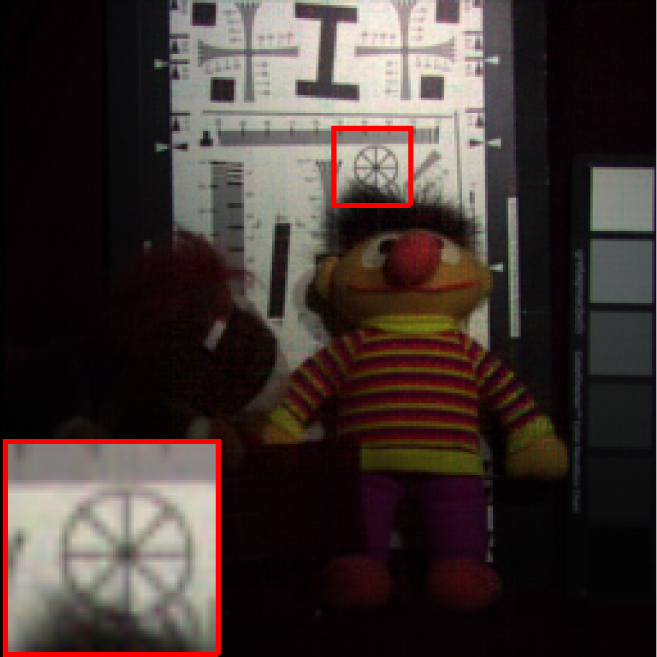}&
			\includegraphics[width=0.2\linewidth]{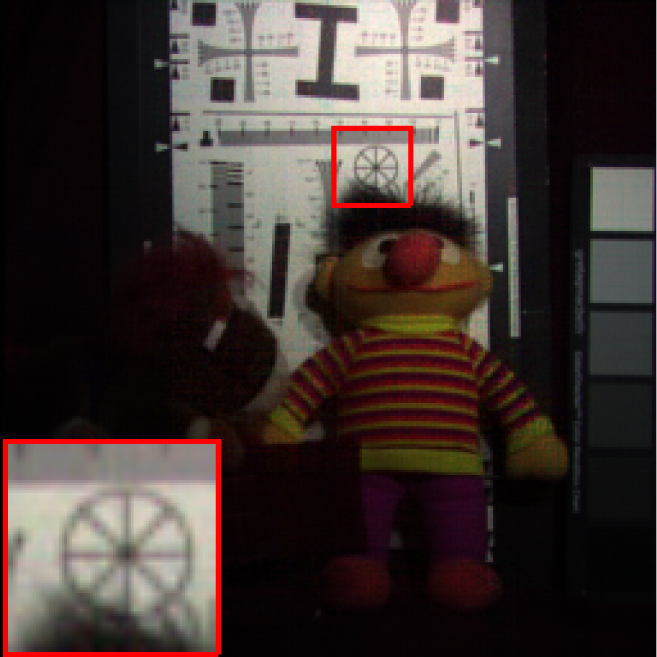}&
			\includegraphics[width=0.2\linewidth]{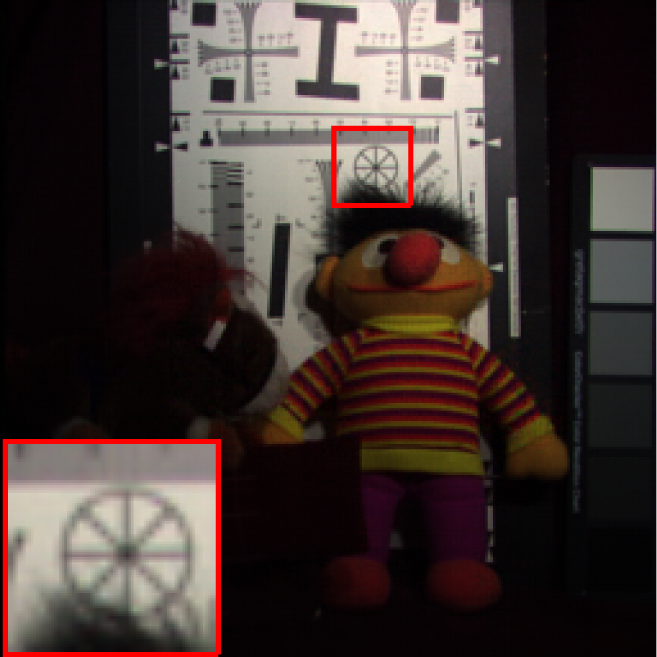}&
			\includegraphics[width=0.2\linewidth]{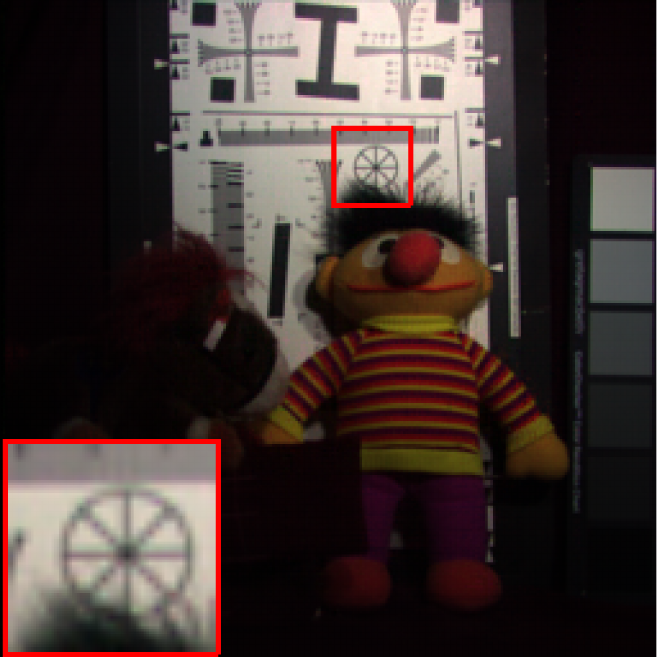}&
			\includegraphics[width=0.2\linewidth]{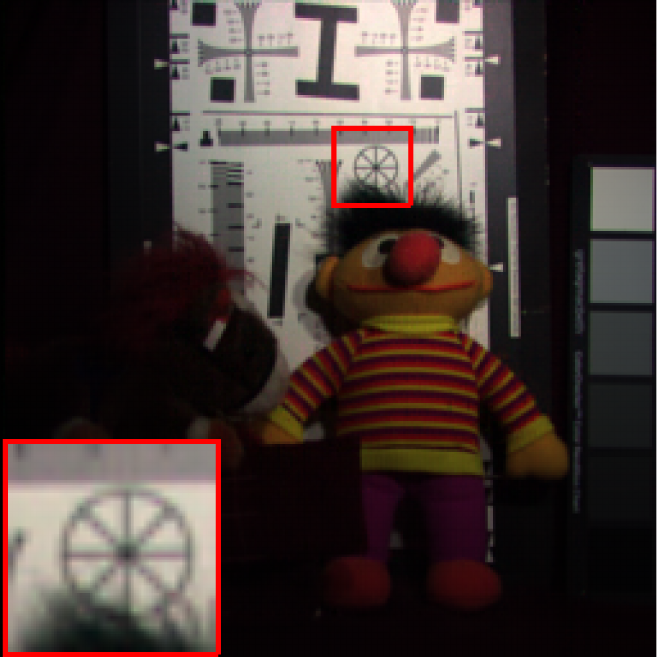}&
            \includegraphics[width=0.2\linewidth]{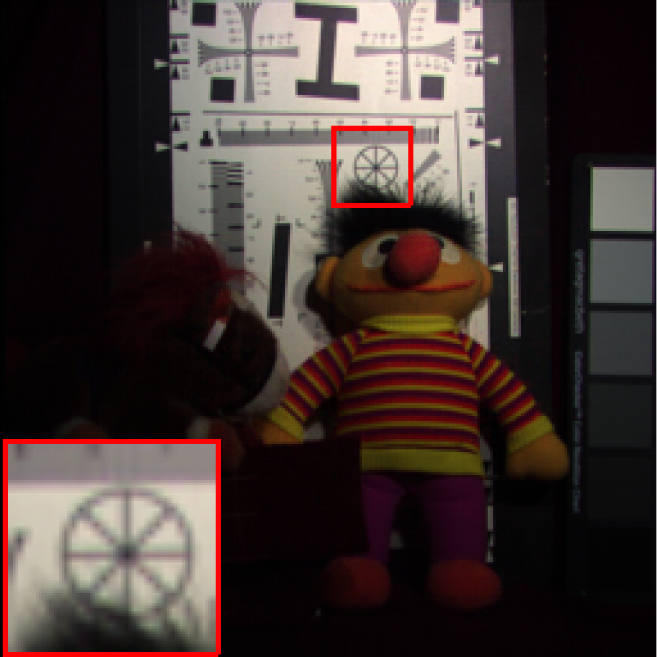}&
            \includegraphics[width=0.2\linewidth]{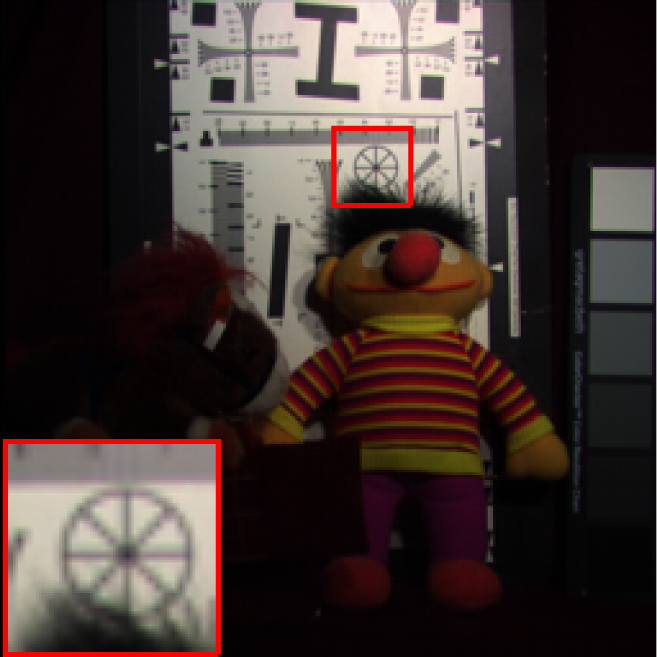}\\
			
			Original & Observed & TNN &TQRTNN  & UTNN & DTNN & LS2T2NN & HLRTF & OTLRM\\
               && \cite{lu_TNN} &\cite{TQRTNN}&\cite{TTNN_song}&\cite{DTNN_jiang}&\cite{liu2023learnable}&\cite{lrtf}&
		\end{tabular}} 

		\caption{ The selected pseudo-color images of recovery results by different methods on MSIs under \emph{SR=0.20}.  From top to bottom: \emph{Balloon},  \emph{Beer}, \emph{Pompom}, and \emph{Toy}. }

		\label{MSIimshow05_sr0.2}
	\end{center}
\end{figure*}

\begin{table*}[htb]
        \centering
        \caption{Evaluation on \emph{CAVE} of \textbf{tensor completion} results by different methods on \textbf{MSI} \emph{Feathers} under different SRs.}

        \def\arraystretch{1.0}
        \setlength{\tabcolsep}{2pt}
\scalebox{0.8}{
    \begin{tabular}{cc|cccccccc}
    \bottomrule[0.15em]
    \multirow{2}[2]{*}{\textbf{Method}} & \multirow{2}[2]{*}{\textbf{Reference}} & \multicolumn{2}{c}{\textbf{SR=0.05}} & \multicolumn{2}{c}{\textbf{SR=0.10}} & \multicolumn{2}{c}{\textbf{SR=0.15}} & \multicolumn{2}{c}{\textbf{SR=0.20}} \\
          &  & \textbf{PSNR}$\uparrow$  & \textbf{SSIM}$\uparrow$  & \textbf{PSNR}$\uparrow$  & \textbf{SSIM}$\uparrow$  & \textbf{PSNR}$\uparrow$  & \textbf{SSIM}$\uparrow$  & 
          \textbf{PSNR}$\uparrow$  & \textbf{SSIM}$\uparrow$\\
    \hline
    Observed & None & 13.35  & 0.18  & 13.59  & 0.22 & 13.83  & 0.26  & 14.10  & 0.29\\
           TNN\cite{lu_TNN} & TPAMI 2019 & 27.59  & 0.76  & 31.98  & 0.88 & 34.79  & 0.93 & 37.00  & 0.95\\
           TQRTNN \cite{TQRTNN}&TCI 2021& 27.57  & 0.76 & 31.94  & 0.88  & 34.78  & 0.93 & 37.07  & 0.95\\
           UTNN\cite{TTNN_song} &NLAA 2020  & 28.30  & 0.82  & 33.33  & 0.92 & 36.88  & 0.96 & 39.81  & \underline{0.98}\\
          DTNN\cite{DTNN_jiang}& TNNLS 2023 & 29.02  & 0.83 & 33.32  & 0.92 & 36.95  & 0.96 & 39.89  & 0.97\\
           LS2T2NN  \cite{liu2023learnable} & TCSVT 2023& 30.99 & 0.88  & 35.39 & 0.95 & 38.45 & 0.97 & 41.34 & \underline{0.98} \\
           HLRTF\cite{lrtf} & CVPR 2022 & \underline{32.31}  & \underline{0.90} & \underline{37.89}  & \underline{0.96}  & \underline{41.38} & \underline{0.98} & \underline{43.86} & \textbf{0.99}\\
           \textbf{OTLRM} & Ours & \textbf{34.95} & \textbf{0.95} & \textbf{39.96}  & \textbf{0.98} & \textbf{42.73} & \textbf{0.99}  & \textbf{45.34} & \textbf{0.99}\\
    \toprule[0.15em]
    \end{tabular}%
    } 
	\label{msitableresult_feathers}
\end{table*}

\begin{figure*}[htb]
	\footnotesize
	\setlength{\tabcolsep}{1pt}
	\begin{center}

            \scalebox{0.51}{
		\begin{tabular}{cccccccccc}
			\includegraphics[width=0.2\linewidth]{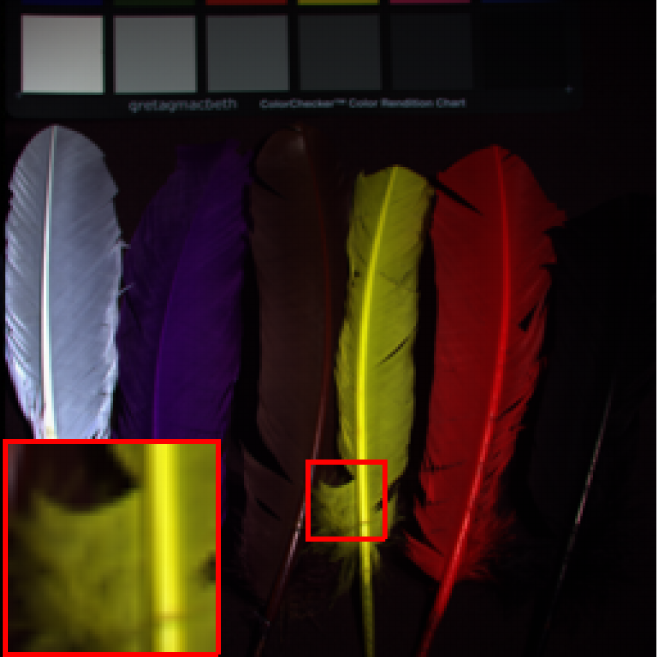}&
			\includegraphics[width=0.2\linewidth]{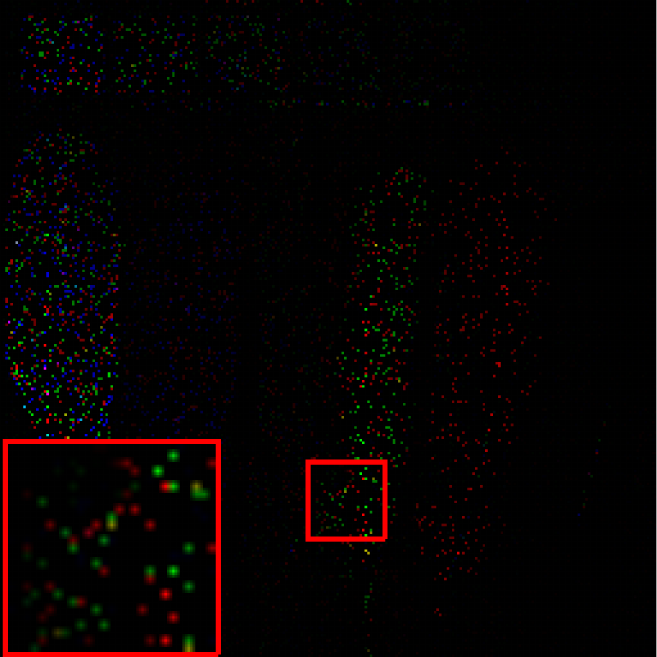}&
			\includegraphics[width=0.2\linewidth]{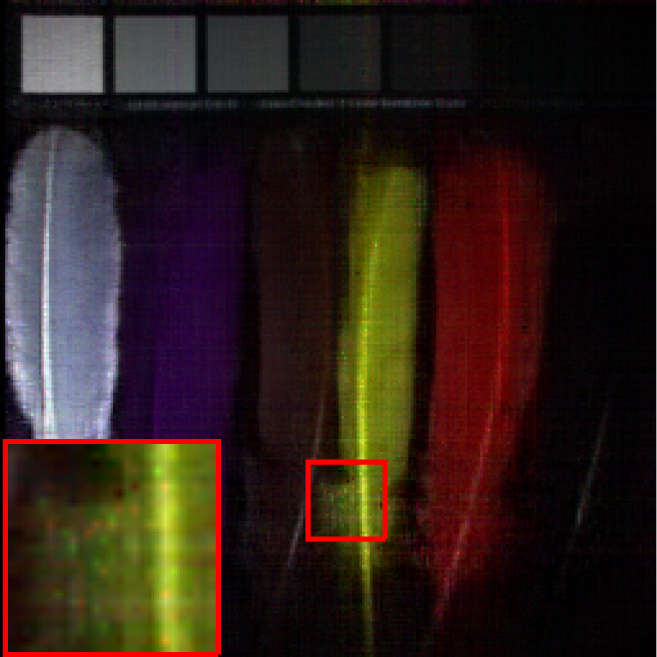}&
			\includegraphics[width=0.2\linewidth]{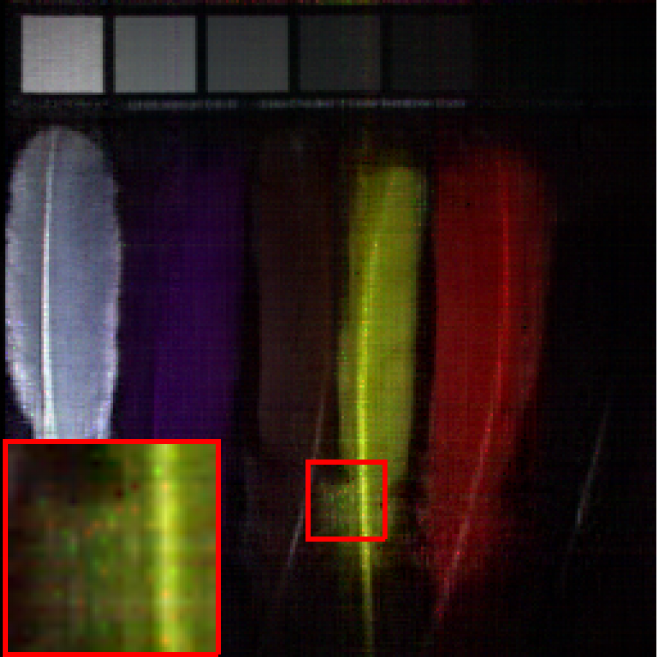}&
			\includegraphics[width=0.2\linewidth]{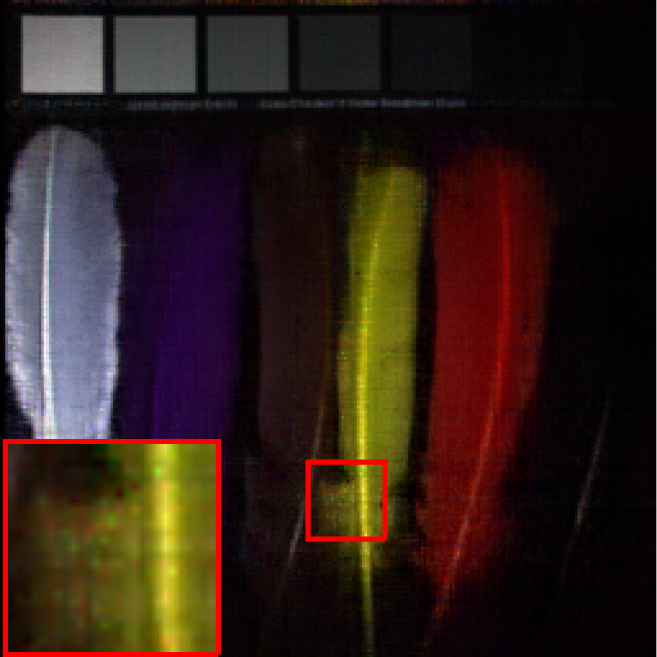}&
			\includegraphics[width=0.2\linewidth]{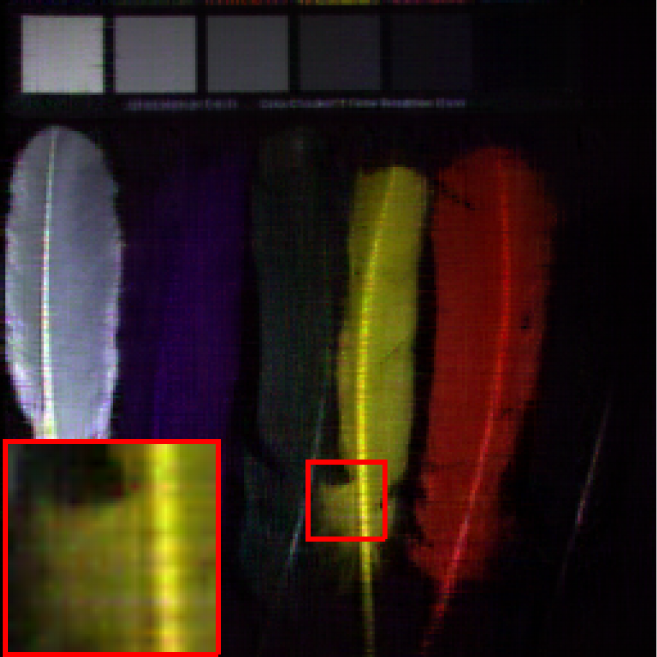}&
			\includegraphics[width=0.2\linewidth]{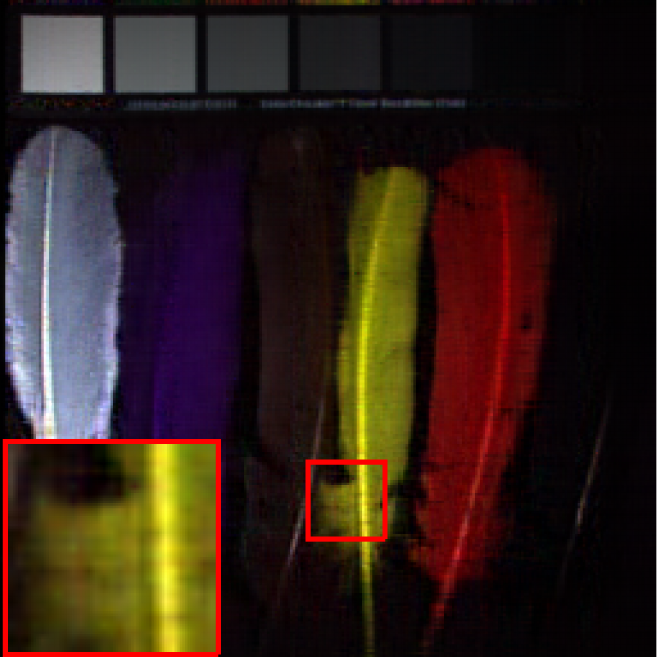}&
            \includegraphics[width=0.2\linewidth]{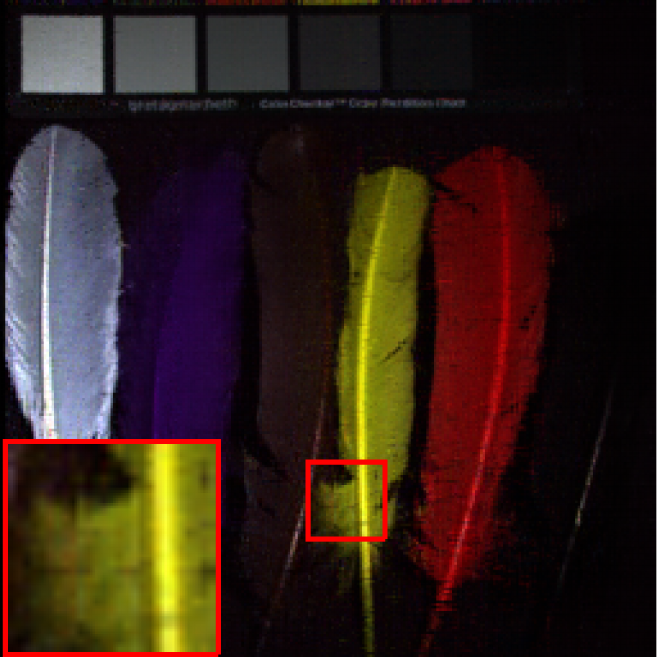}&
            \includegraphics[width=0.2\linewidth]{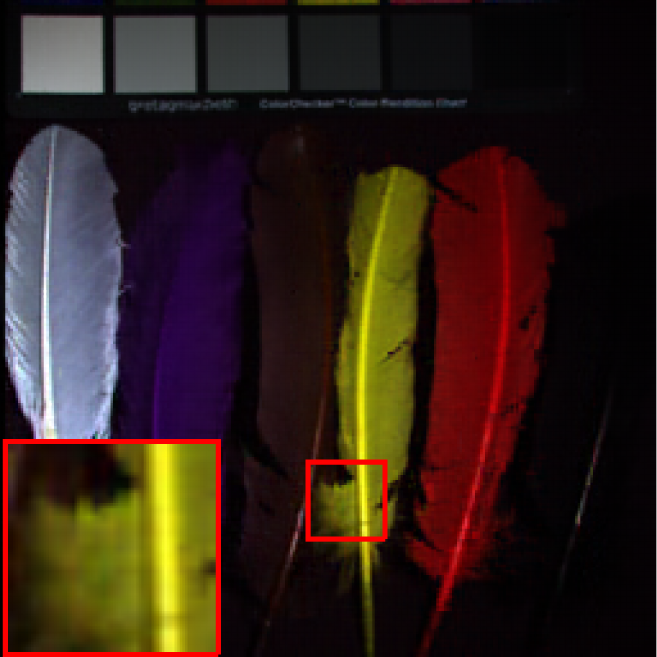}\\
			
			\includegraphics[width=0.2\linewidth]{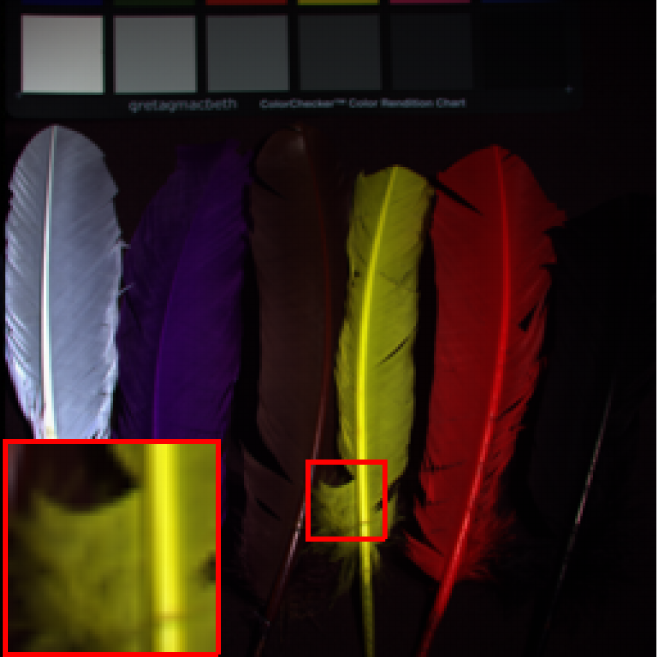}&
			\includegraphics[width=0.2\linewidth]{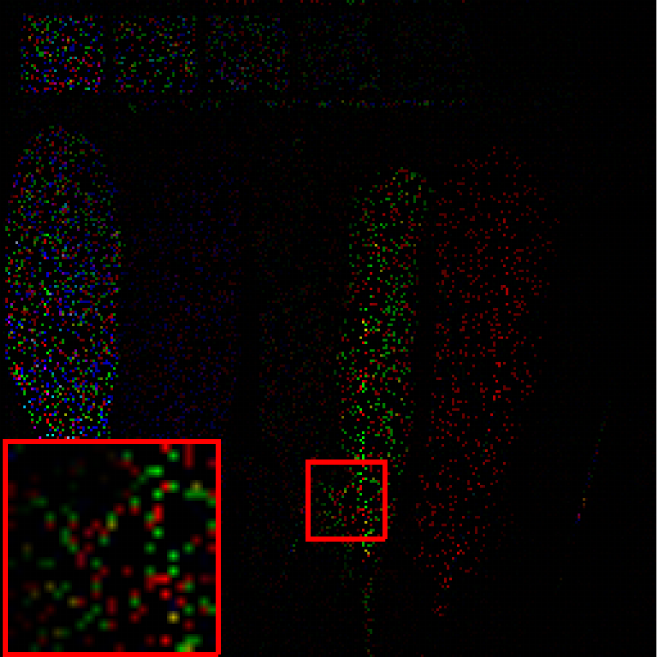}&
			\includegraphics[width=0.2\linewidth]{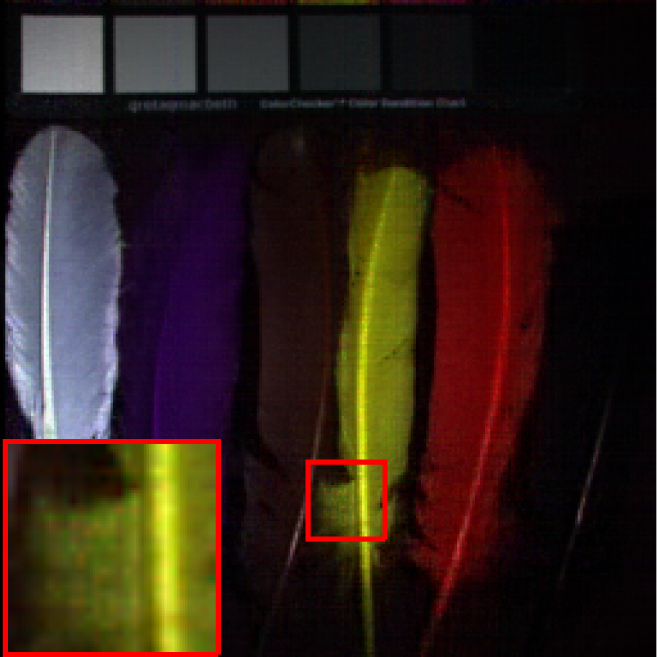}&
			\includegraphics[width=0.2\linewidth]{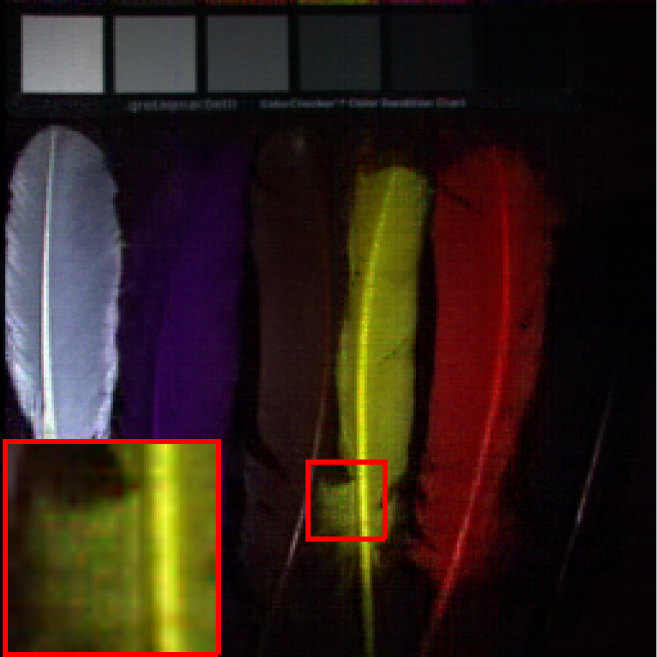}&
			\includegraphics[width=0.2\linewidth]{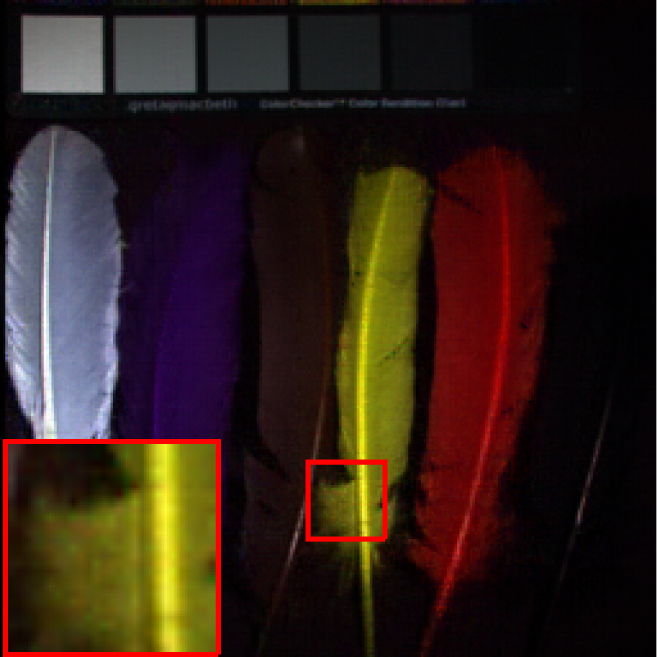}&
			\includegraphics[width=0.2\linewidth]{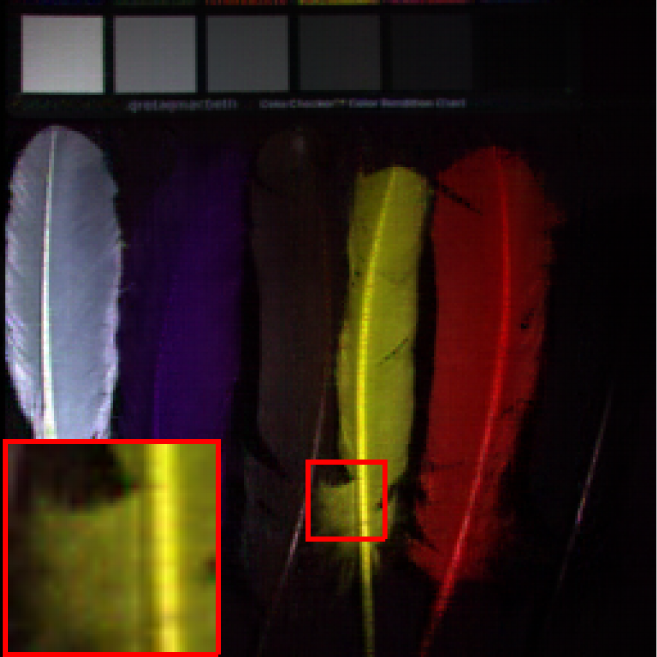}&
			\includegraphics[width=0.2\linewidth]{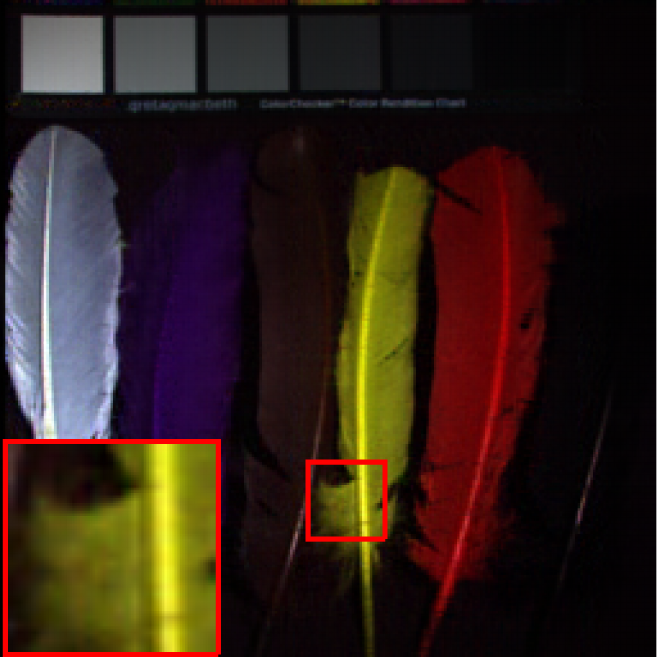}&
            \includegraphics[width=0.2\linewidth]{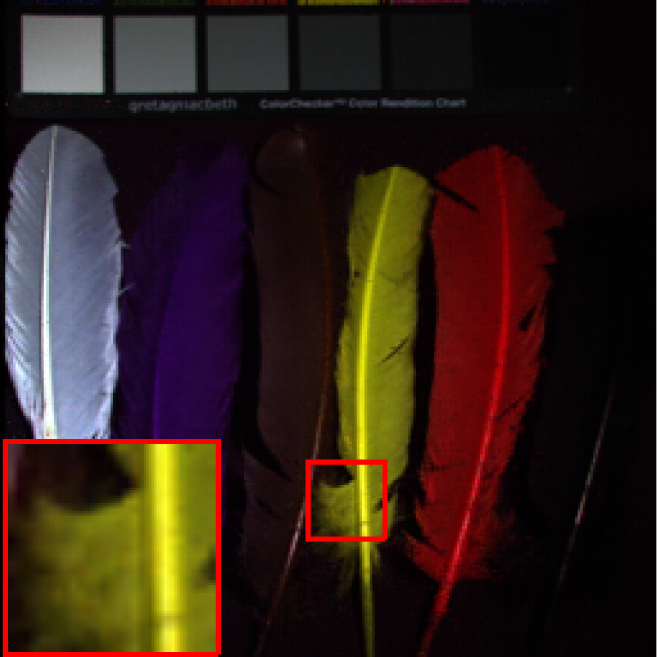}&
            \includegraphics[width=0.2\linewidth]{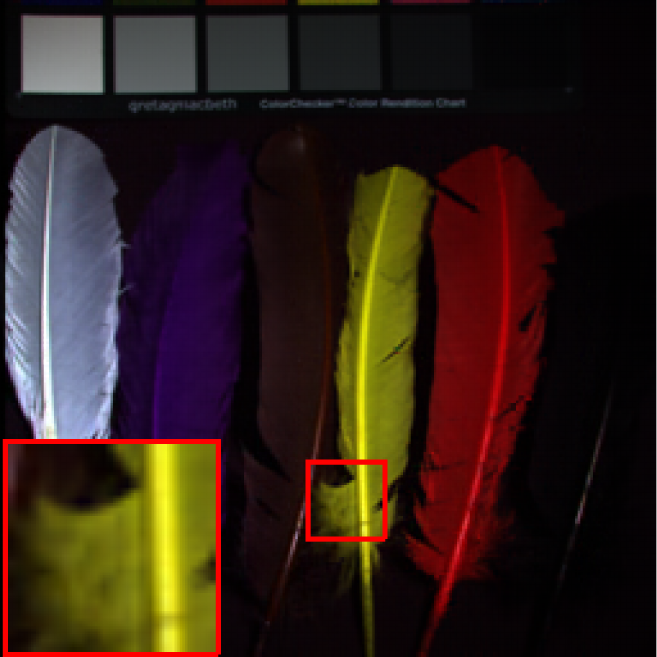}\\
			
			\includegraphics[width=0.2\linewidth]{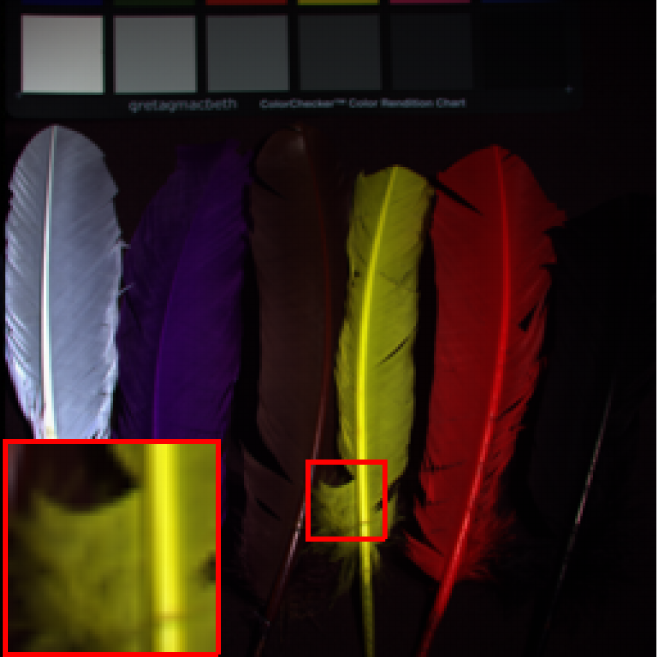}&
			\includegraphics[width=0.2\linewidth]{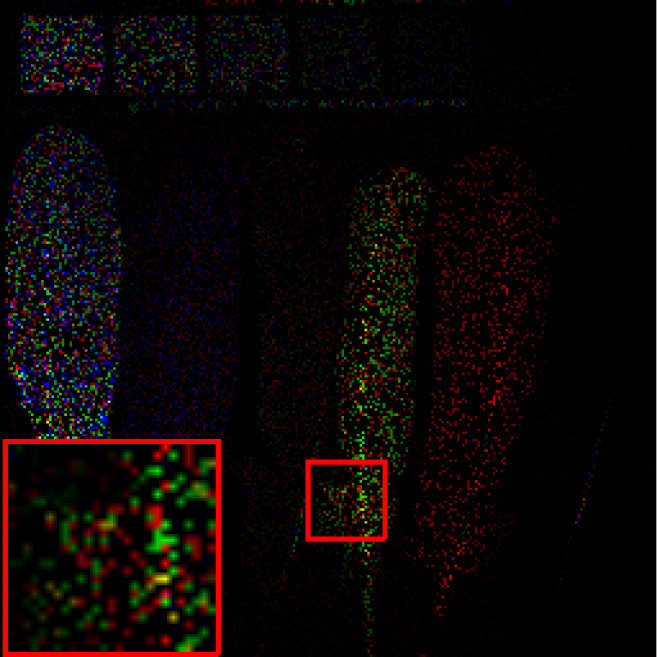}&
			\includegraphics[width=0.2\linewidth]{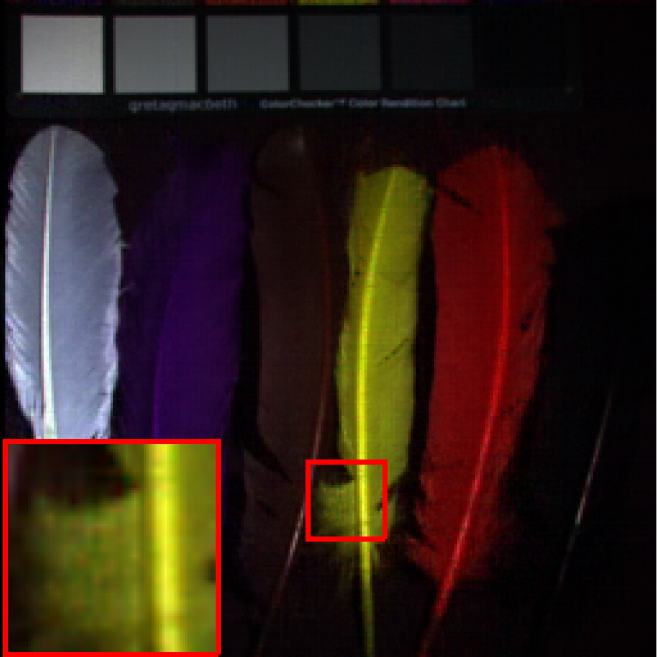}&
			\includegraphics[width=0.2\linewidth]{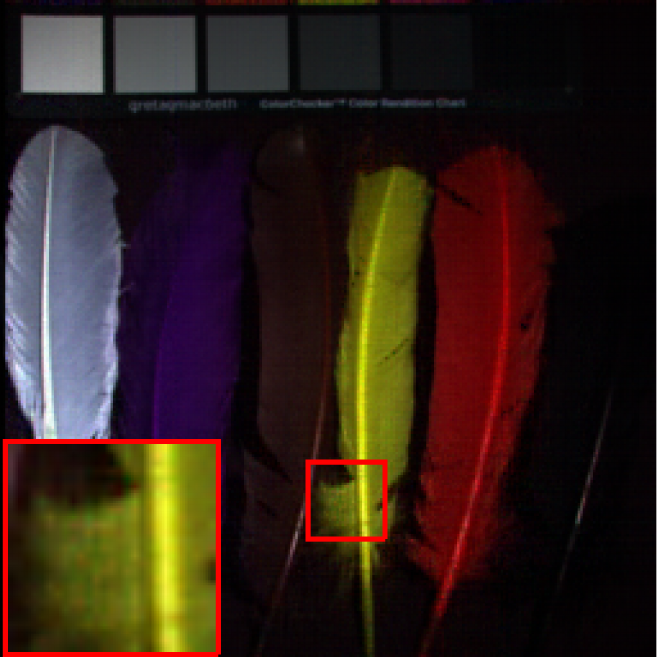}&
			\includegraphics[width=0.2\linewidth]{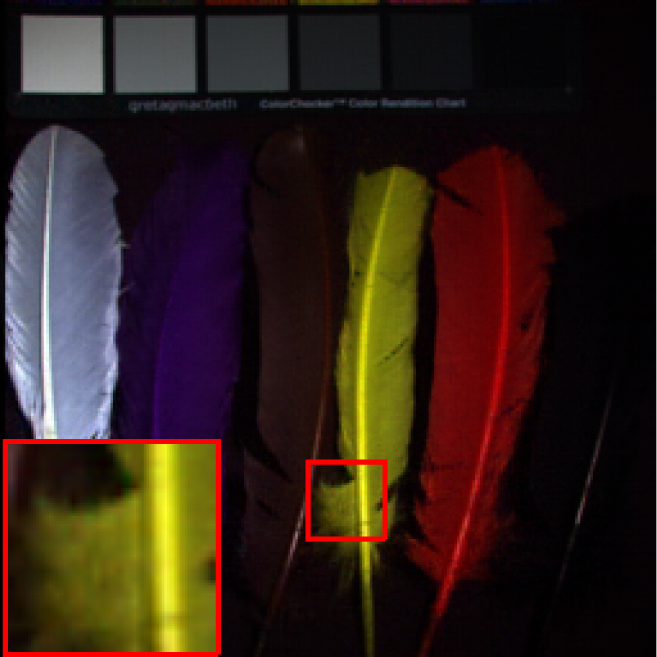}&
			\includegraphics[width=0.2\linewidth]{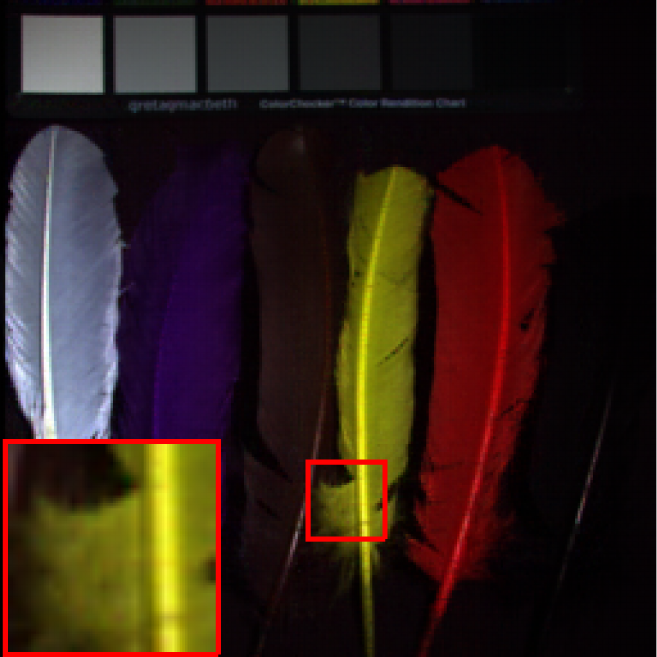}&
			\includegraphics[width=0.2\linewidth]{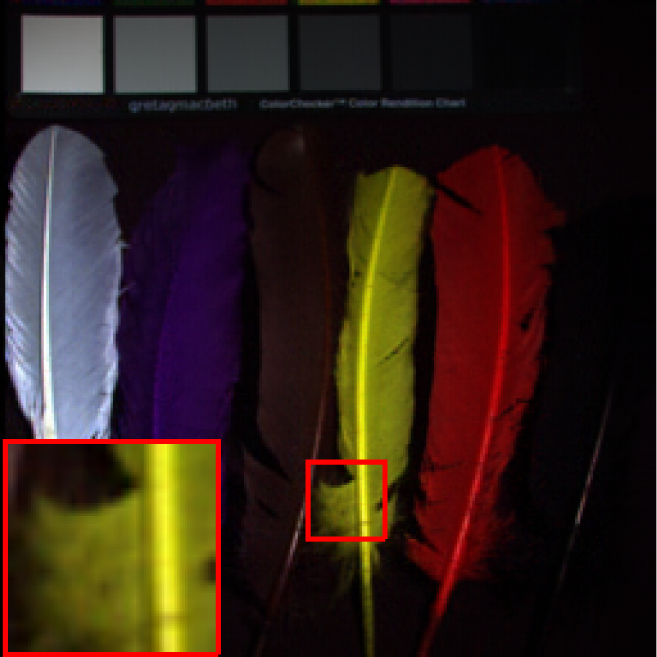}&
            \includegraphics[width=0.2\linewidth]{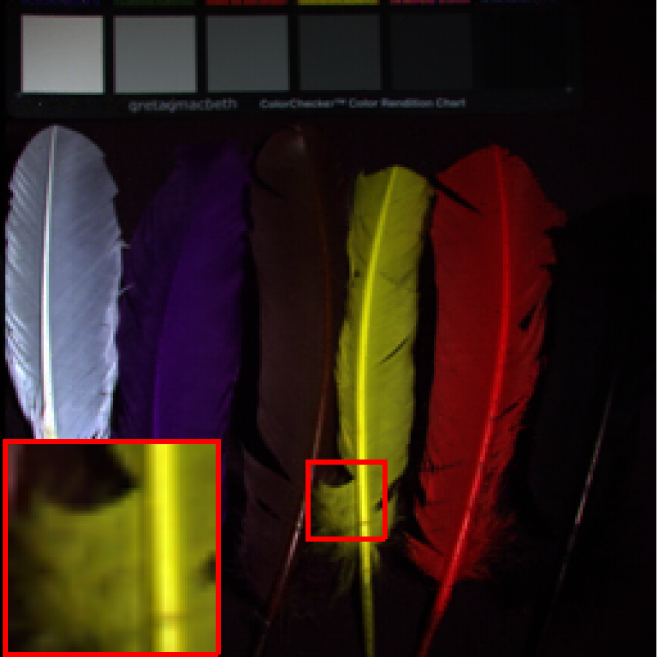}&
            \includegraphics[width=0.2\linewidth]{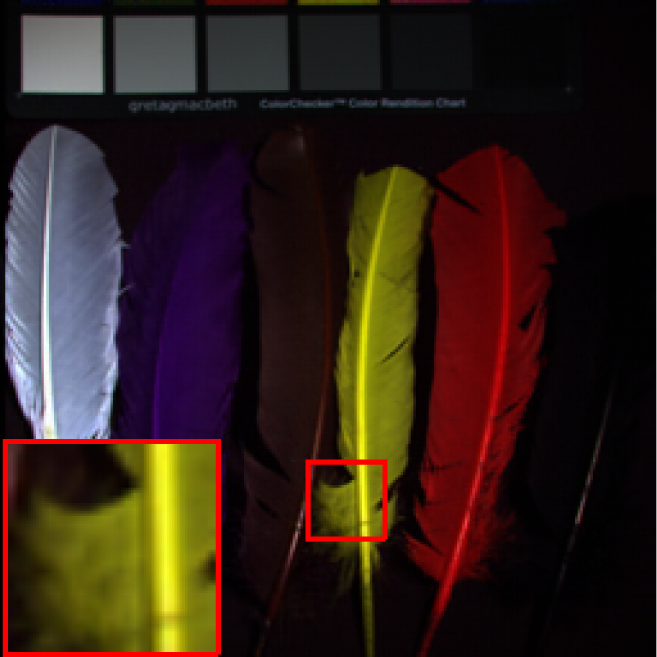}\\
			
			\includegraphics[width=0.2\linewidth]{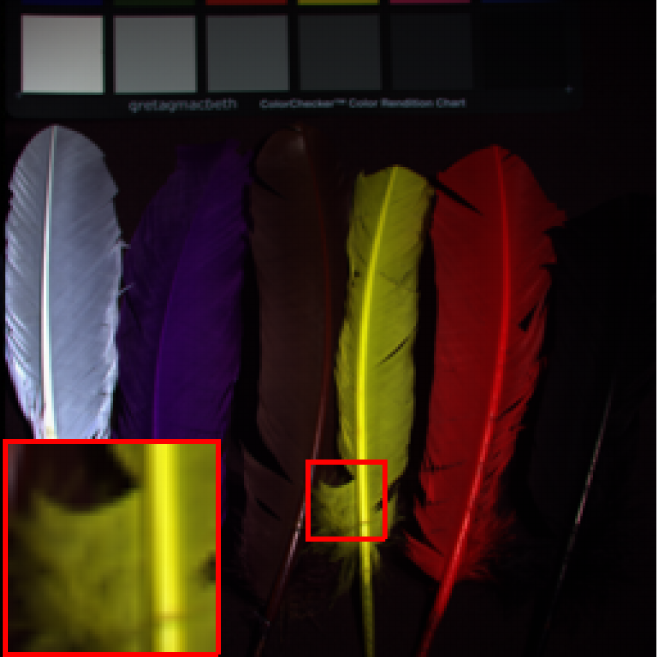}&
			\includegraphics[width=0.2\linewidth]{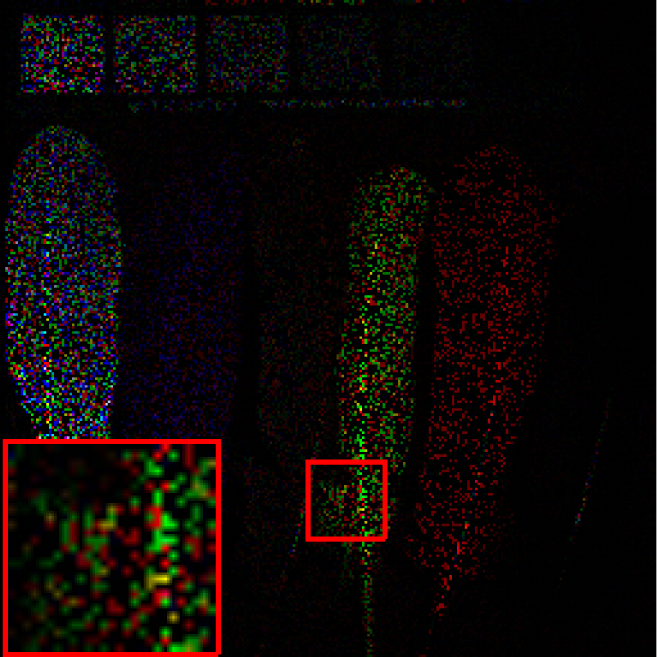}&
			\includegraphics[width=0.2\linewidth]{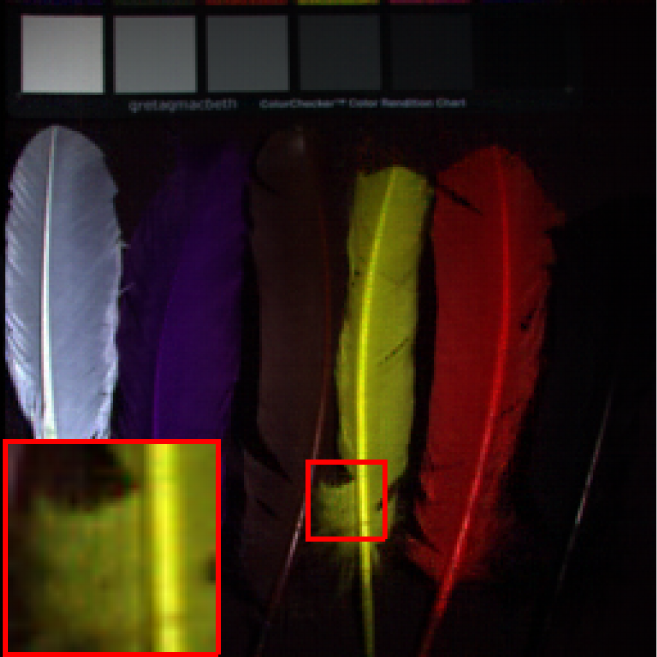}&
			\includegraphics[width=0.2\linewidth]{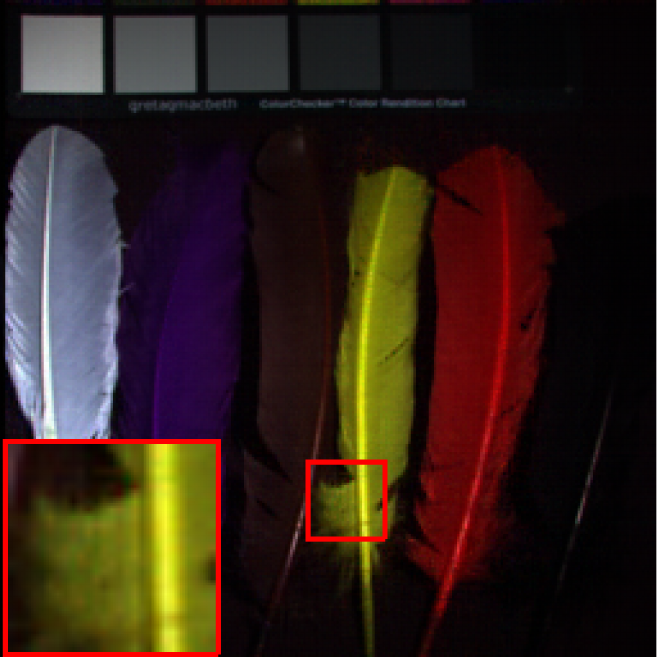}&
			\includegraphics[width=0.2\linewidth]{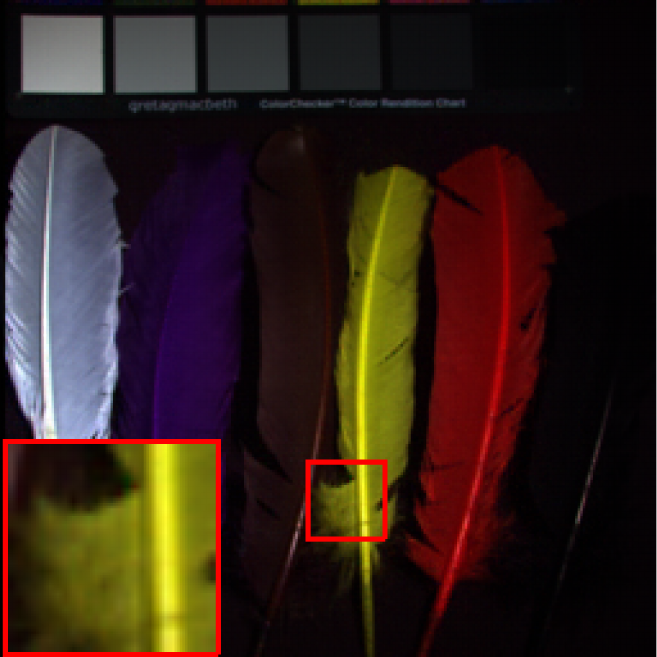}&
			\includegraphics[width=0.2\linewidth]{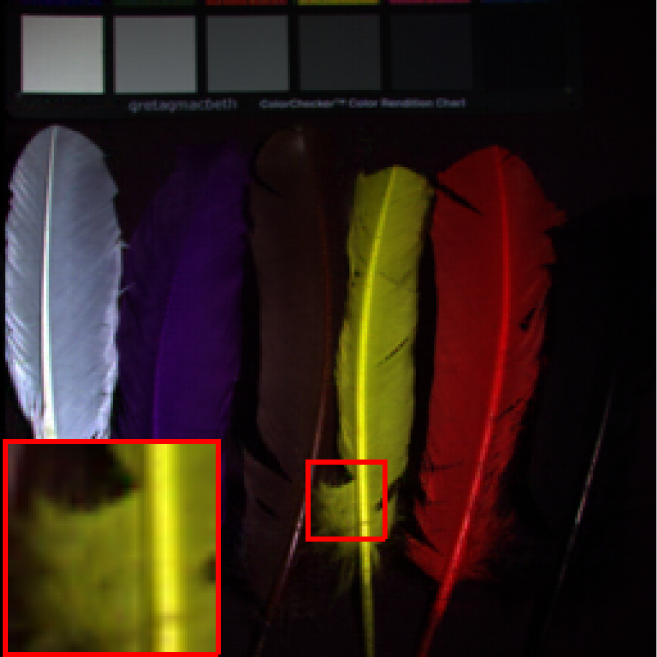}&
			\includegraphics[width=0.2\linewidth]{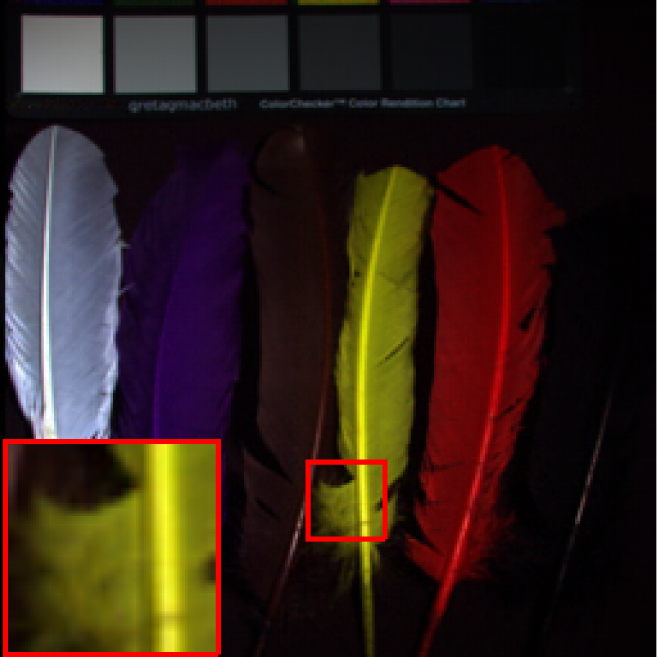}&
            \includegraphics[width=0.2\linewidth]{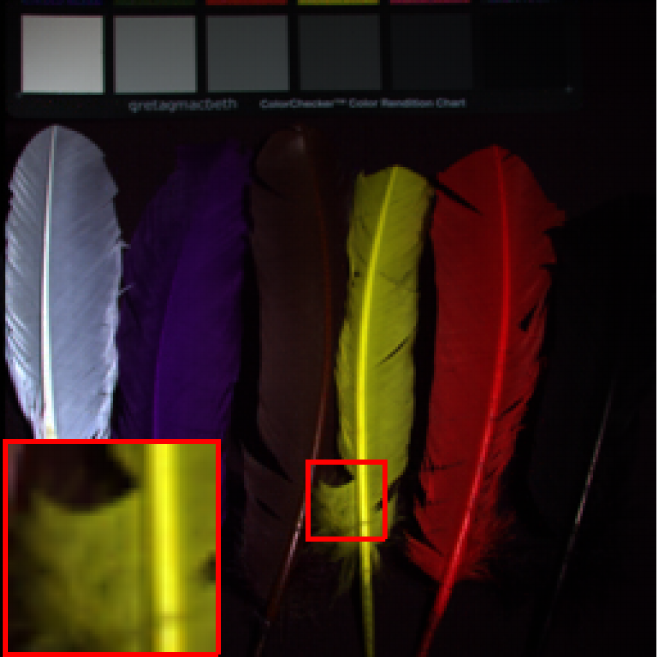}&
            \includegraphics[width=0.2\linewidth]{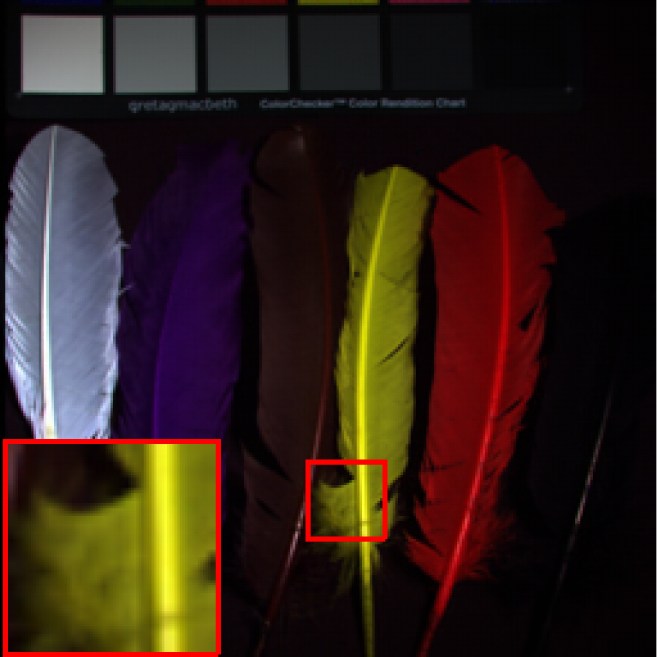}\\
			
			Original & Observed & TNN &TQRTNN  & UTNN & DTNN & LS2T2NN & HLRTF & OTLRM\\
               && \cite{lu_TNN} &\cite{TQRTNN}&\cite{TTNN_song}&\cite{DTNN_jiang}&\cite{liu2023learnable}&\cite{lrtf}&
		\end{tabular}} 

		\caption{ The selected pseudo-color images of recovery results by different methods on MSI \emph{Feathers} under different SRs.  From top to bottom: \emph{SR=0.05},  \emph{SR=0.10}, \emph{SR=0.15}, and \emph{SR=0.20}. }
		\label{MSIimshow05_feathers}
	\end{center}
\end{figure*}

\begin{table}
        \centering
        \caption{Evaluation on \textbf{MSIs} \emph{Beers} in \textbf{tensor completion}.}

        \def\arraystretch{1.0}
        \setlength{\tabcolsep}{1pt}
\scalebox{0.64}{
    \begin{tabular}{cc|ccccccc}
    \bottomrule[0.15em]
    \multirow{2}[2]{*}{\textbf{Method}} & \multirow{2}[2]{*}{\textbf{Reference}} & \multicolumn{2}{c}{\textbf{SR=0.05}} & \multicolumn{2}{c}{\textbf{SR=0.10}} & \multicolumn{2}{c}{\textbf{SR=0.15}} & \multirow{2}[2]{*}{\textbf{Time (s)}} \\
          &  & \textbf{PSNR}$\uparrow$  & \textbf{SSIM}$\uparrow$  & \textbf{PSNR}$\uparrow$  & \textbf{SSIM}$\uparrow$  & \textbf{PSNR}$\uparrow$  & \textbf{SSIM}$\uparrow$  & \\
    \hline
           Tensor $Q$-rank & ML 2021 & 37.35   & 0.96  & 42.55   & \textbf{0.99}  & 45.52   & \textbf{0.99} & 33 \\
          DTNN  &TNNLS 2023  & 35.95  & 0.95   & 41.29  & \underline{0.97}  & 45.05  & \textbf{0.99} & 415\\
          \textbf{OTLRM*} &Ours&   \underline{39.16} & \underline{0.97} & \underline{43.53}  & \textbf{0.99} & \underline{46.92}  & \textbf{0.99} & 111 \\
          \textbf{OTLRM}& Ours & \textbf{41.74}  & \textbf{0.98} & \textbf{45.09}  & \textbf{0.99}  & \textbf{47.52} & \textbf{0.99} & 118 \\
    \toprule[0.15em]
    \end{tabular}%
    } 
	\label{comparison}
\end{table}

\begin{figure*}[htb]
	\footnotesize
	\setlength{\tabcolsep}{1pt}
	\begin{center}

             \scalebox{0.51}{
		\begin{tabular}{cccccccccc}
			\includegraphics[width=0.2\textwidth]{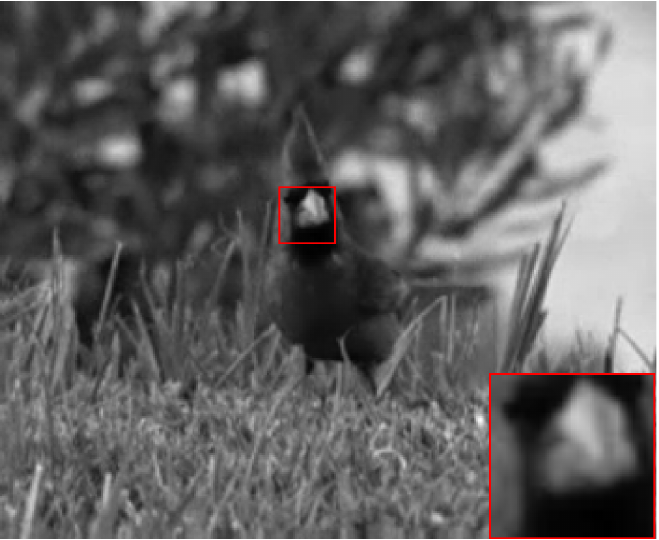} &
			\includegraphics[width=0.2\textwidth]{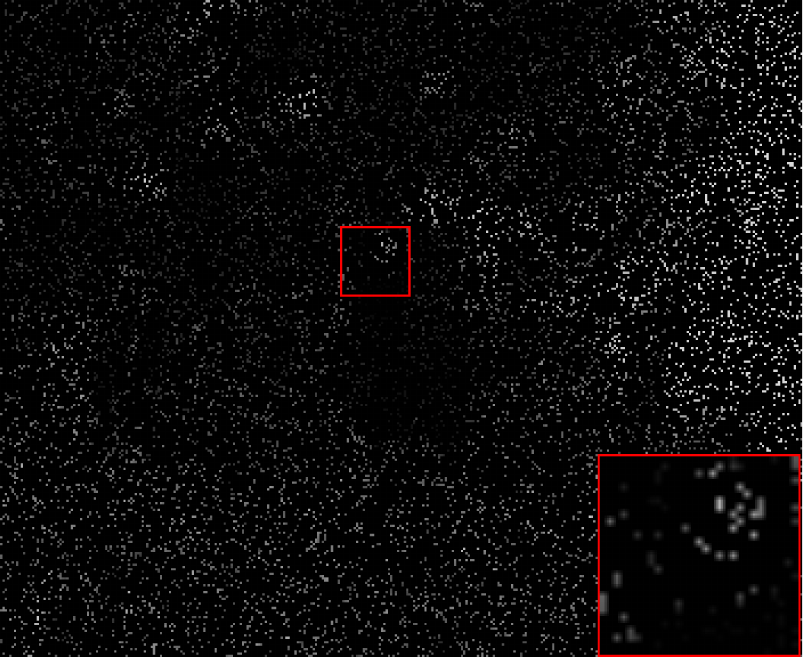}&
			\includegraphics[width=0.2\textwidth]{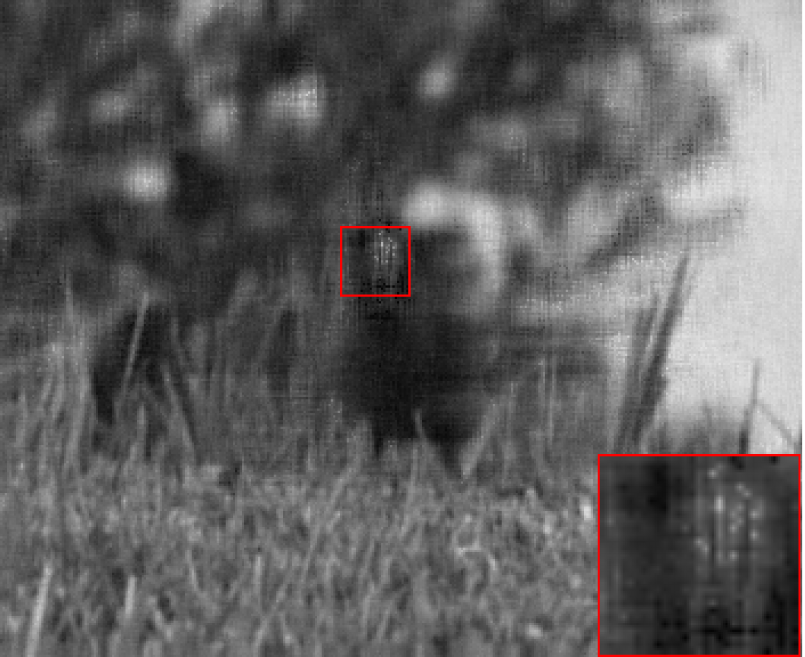}&
			\includegraphics[width=0.2\textwidth]{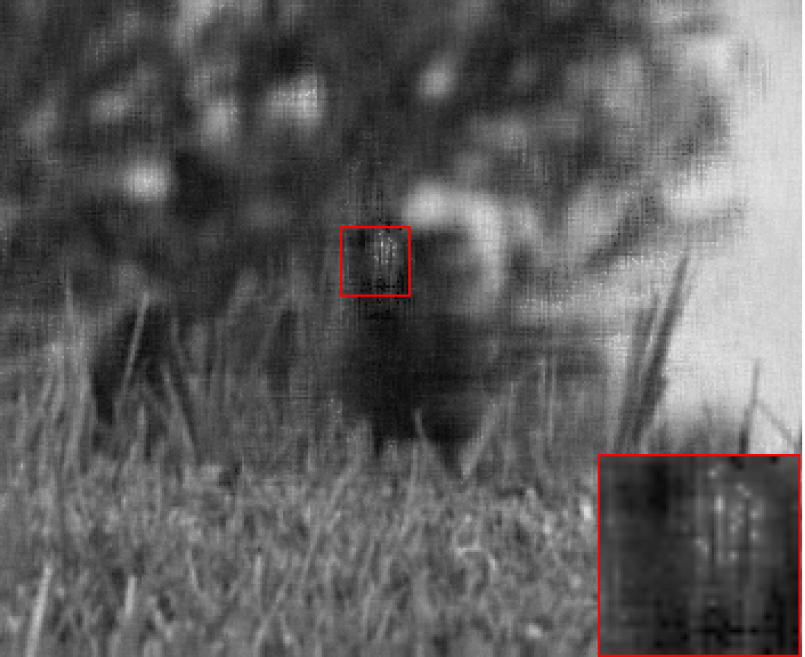}&
			\includegraphics[width=0.2\textwidth]{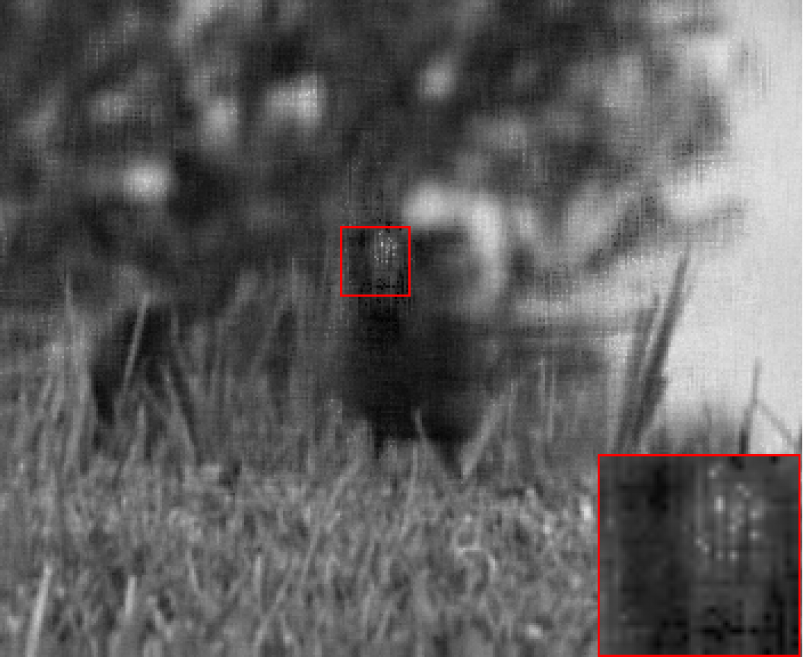}&
			\includegraphics[width=0.2\textwidth]{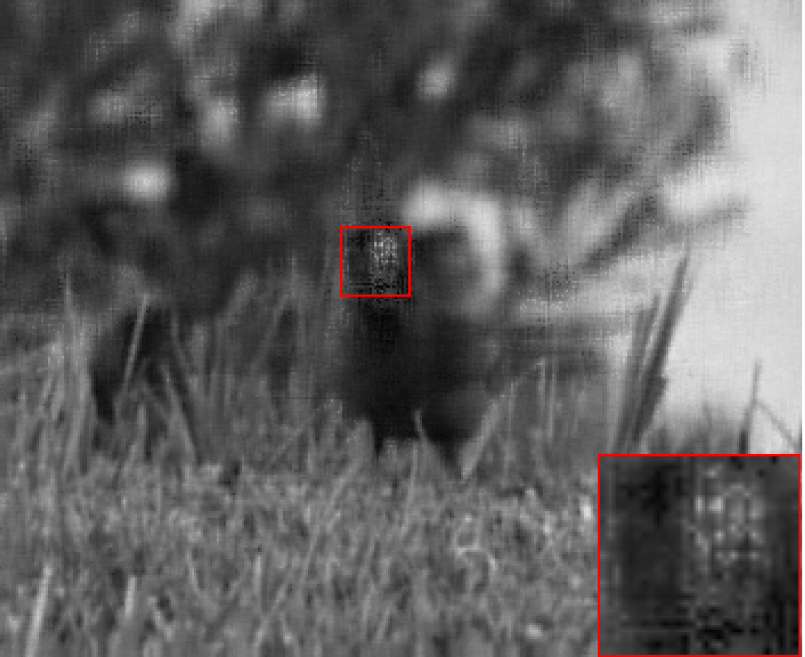}&
			\includegraphics[width=0.2\textwidth]{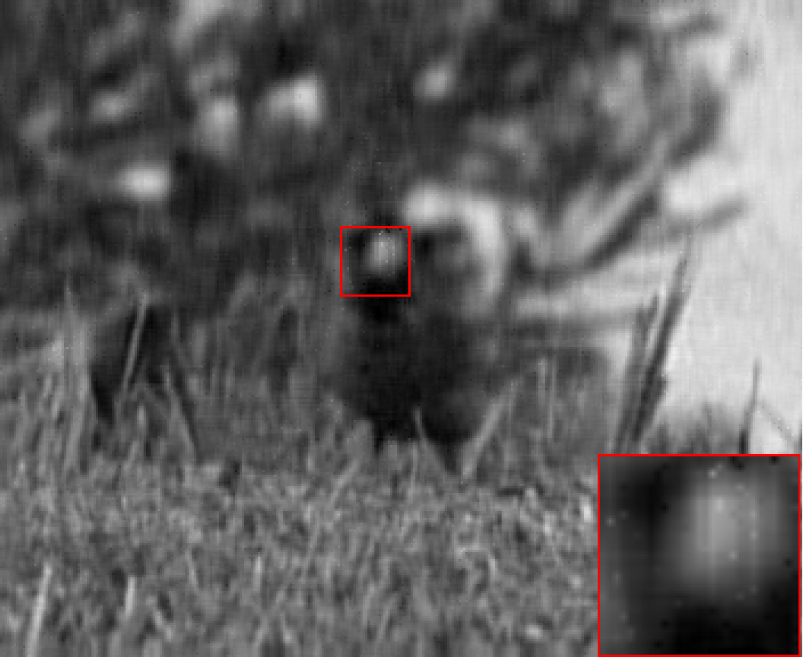}&
                \includegraphics[width=0.2\textwidth]{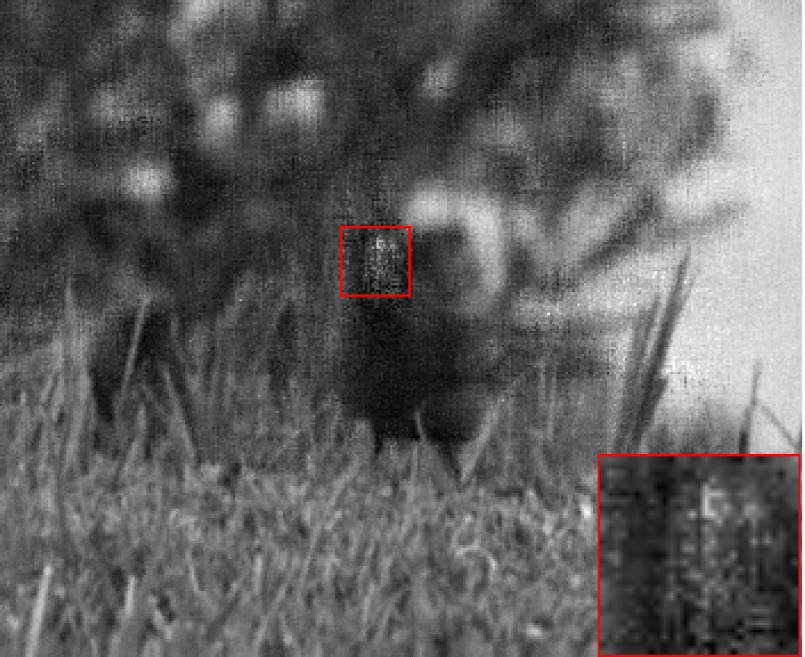}&
                \includegraphics[width=0.2\textwidth]{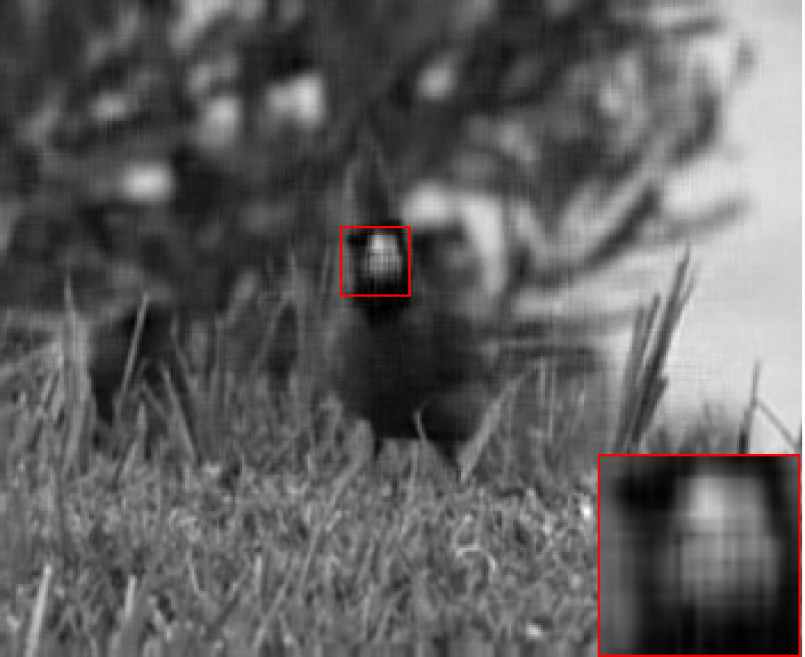}\\
			\includegraphics[width=0.2\textwidth]{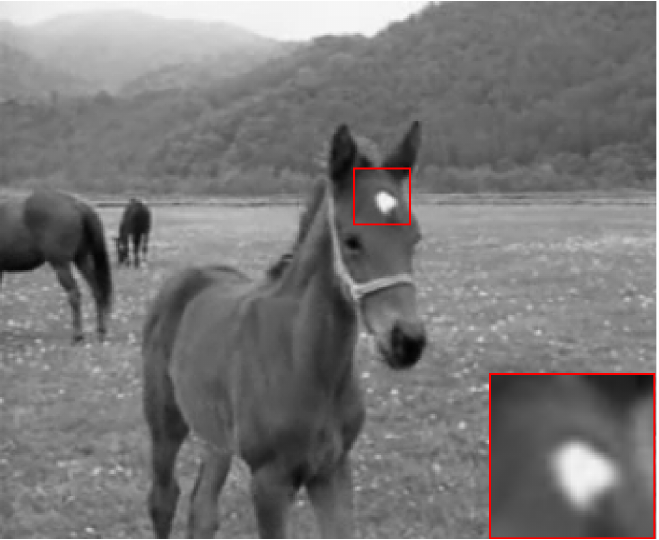} &
			\includegraphics[width=0.2\textwidth]{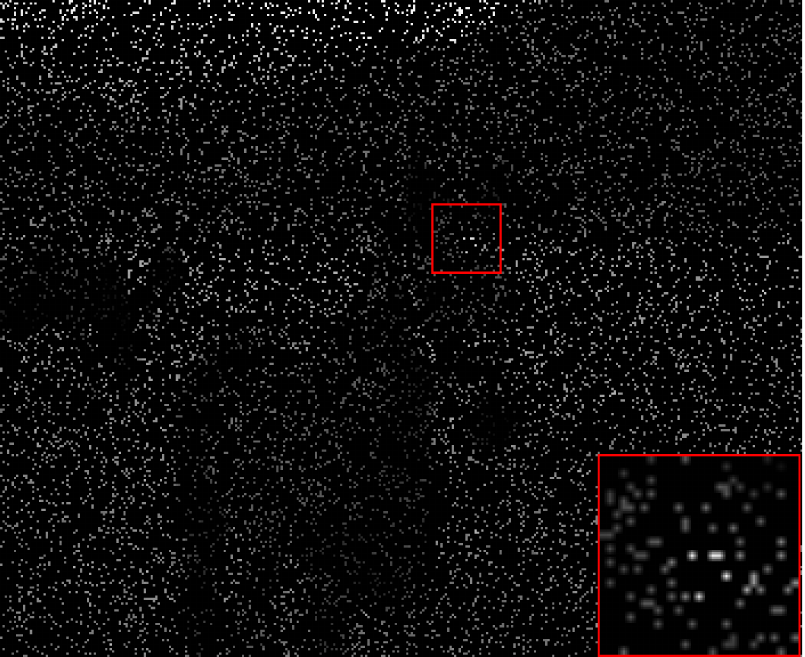}&
			\includegraphics[width=0.2\textwidth]{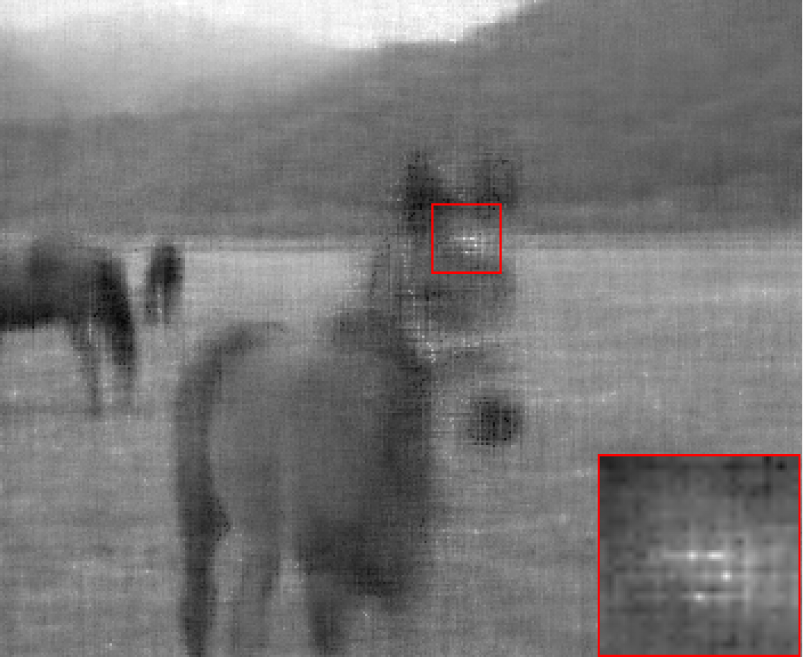}&
			\includegraphics[width=0.2\textwidth]{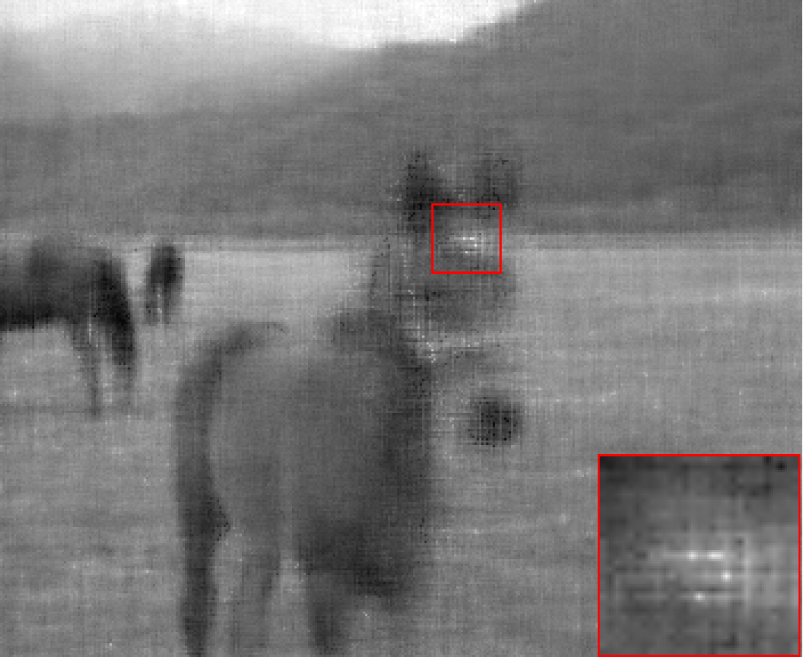}&
			\includegraphics[width=0.2\textwidth]{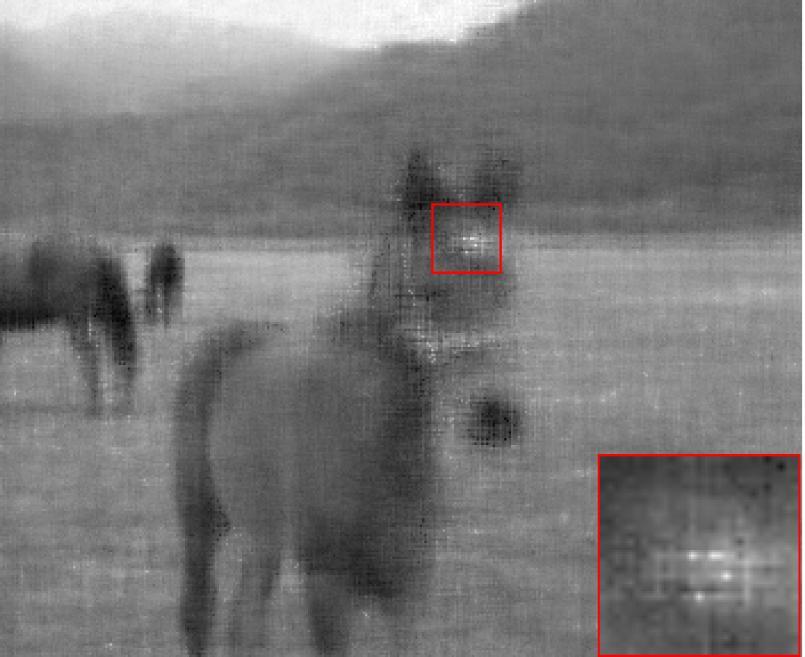}&
			\includegraphics[width=0.2\textwidth]{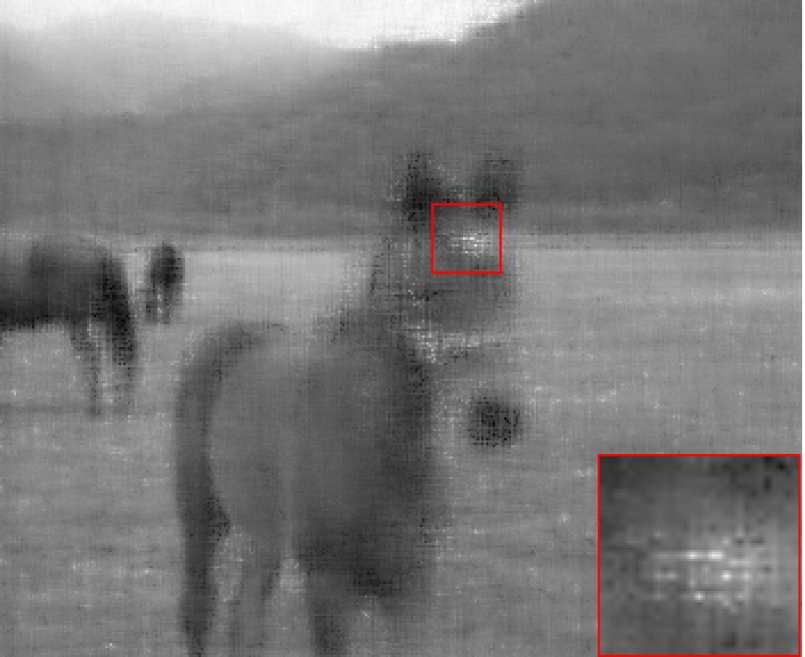}&
			\includegraphics[width=0.2\textwidth]{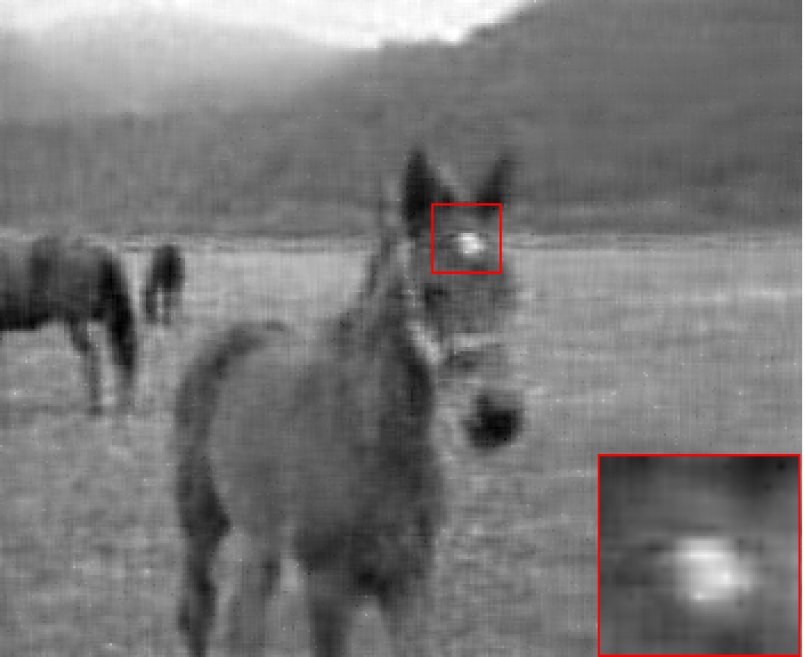}&
                \includegraphics[width=0.2\textwidth]{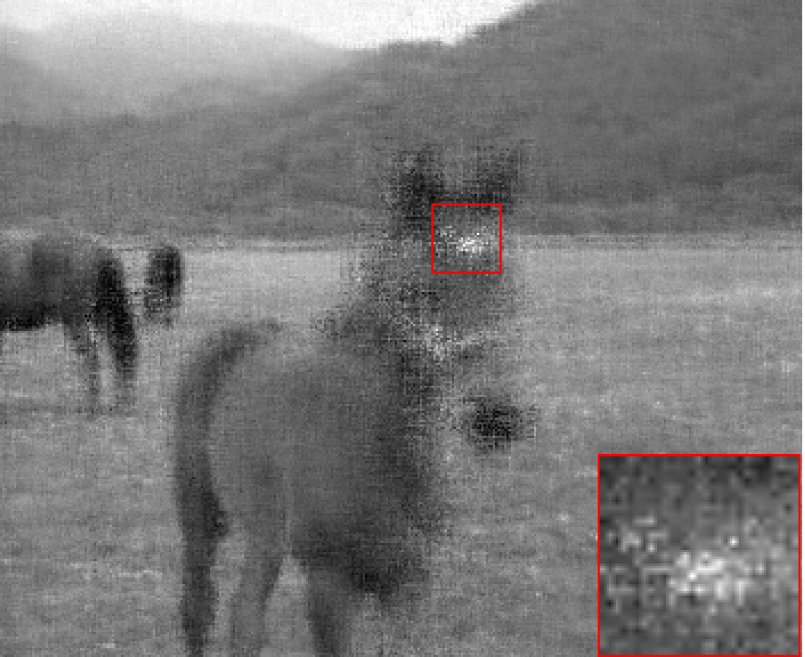}&
                \includegraphics[width=0.2\textwidth]{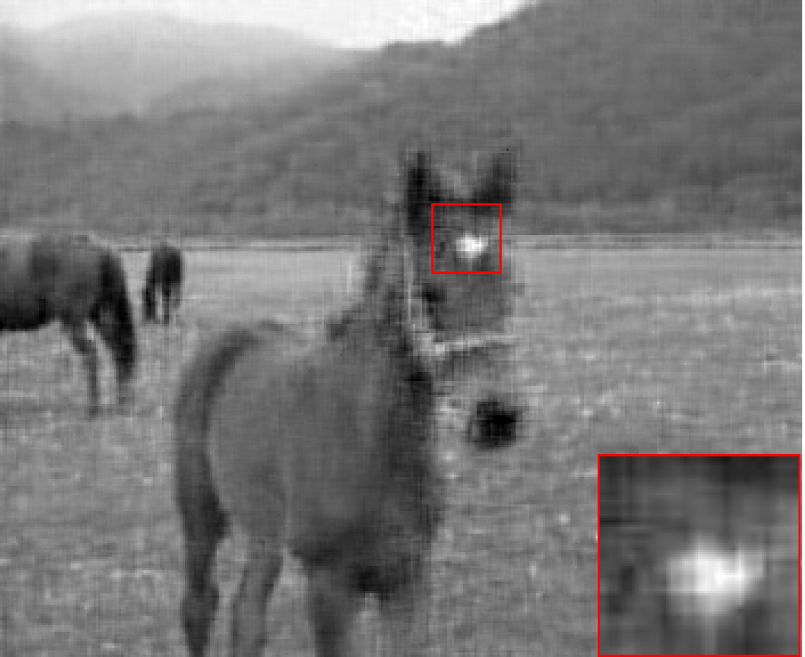}\\
			 Original & Observed & TNN &TQRTNN  & UTNN & DTNN & LS2T2NN & HLRTF & OTLRM\\
               && \cite{lu_TNN} &\cite{TQRTNN}&\cite{TTNN_song}&\cite{DTNN_jiang}&\cite{liu2023learnable}&\cite{lrtf}&
    
		\end{tabular}}

		\caption{ Selected band of recovery results by different methods on videos under \emph{SR=0.10}. From top to bottom: \emph{Bird} and \emph{Horse}. }
		\label{imagevideofig_sr0.1}
	\end{center}
\end{figure*}

For MRI completion, which is heavily reliant on inverse problem techniques and whose data have low-rank properties, we provide a \textit{Brain} data to verify the performance of our method in Table \ref{tab:MRI_completion}.
The data is selected for the first $80$ frames, which is $181 \times 217 \times 80$.
And $\mathscr{l}_1$ loss is adopted for the completion process.
Our method achieves comparable results in MRI completion.
Despite the MSIs and videos provided in the main paper, we also verify the multi-dimensional inverse problems with MRI data.

\begin{table}
    \centering
    \caption{\textbf{MRI completion} results under different SRs $\in \{ 0.1,0.2,0.3 \}$ for \textit{Brain}.}
    \begin{tabular}{ccccc}
    \toprule[0.15em]
        \textbf{Method} & $0.1$ & $0.2$ & $0.3$ & \textbf{Avg}\\
    \midrule[0.1em]
        DTNN & 26.02 & 29.50 & 30.25 & 28.59\\
        HLRTF & 27.35 & 31.52 & 33.70 & 30.87\\
        Ours & \textbf{28.85} & \textbf{32.22} & \textbf{34.60} & \textbf{31.89}\\
    \bottomrule[0.15em]
    \end{tabular}
    \label{tab:MRI_completion}
\end{table}

\subsection{MSI reconstruction in CASSI system}
We compare our method with one model-based method (DeSCI\cite{desci}), six supervised method ($\lambda$-Net\cite{miao2019net}, TSA-Net\cite{meng2020end}, HDNet\cite{hu2022hdnet}, DGSMP\cite{huang2021deep}, ADMM-Net\cite{admm-net}, GAP-Net\cite{meng2023deep}), two Zero-Shot method (PnP-CASSI \cite{zheng2021deep}, DIP-HSI\cite{meng2021self}) and one generative tensor factorization network method (HLRTF\cite{lrtf}).

We also exhibits the results of CASSI on five more scenes (\emph{scene06}-\emph{scene10}) in \emph{KAIST} in Table \ref{tab:simu1}.
And the recovery results of the selected band are demonstrated in Figure \ref{snapshot_S4}, Figure \ref{snapshot_S5}, Figure \ref{snapshot_S9} and Figure \ref{snapshot_S10}.
Compared to other state-of-the-art methods, our method indicates clearer and more fine-grained results.

\begin{figure*}[htb]
	\footnotesize
	\setlength{\tabcolsep}{1pt}
	\begin{center}
            \scalebox{0.7}{
		\begin{tabular}{ccccccccccc}
			\includegraphics[width=0.23\linewidth]{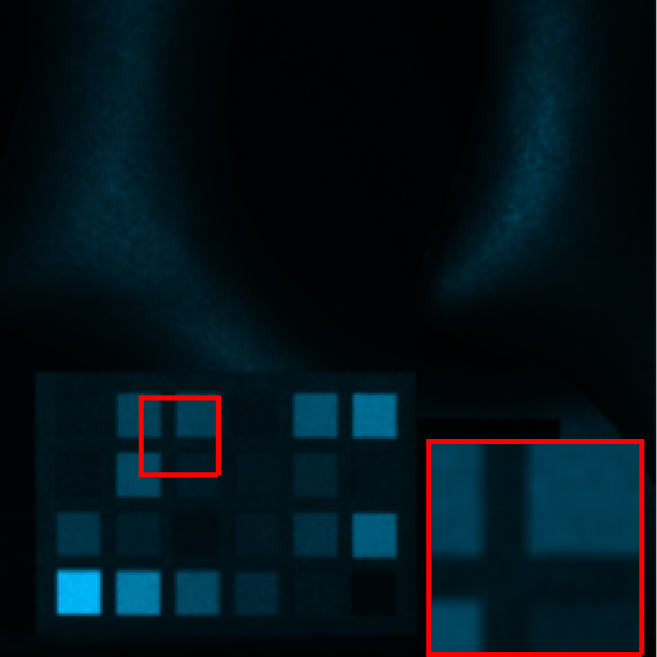}&
			\includegraphics[width=0.23\linewidth]{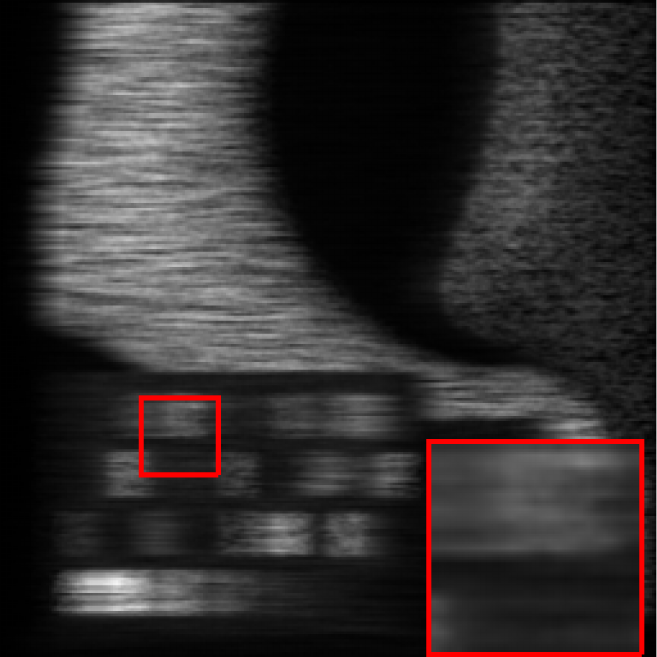}&
			\includegraphics[width=0.23\linewidth]{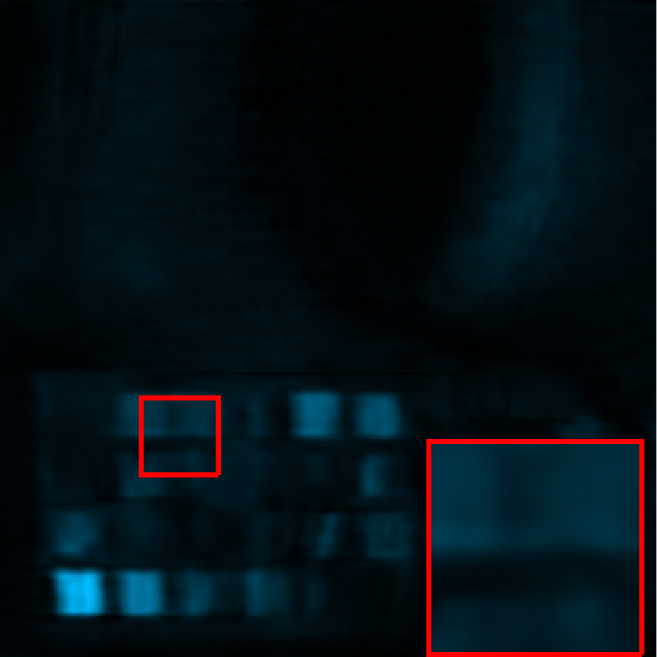}&
			\includegraphics[width=0.23\linewidth]{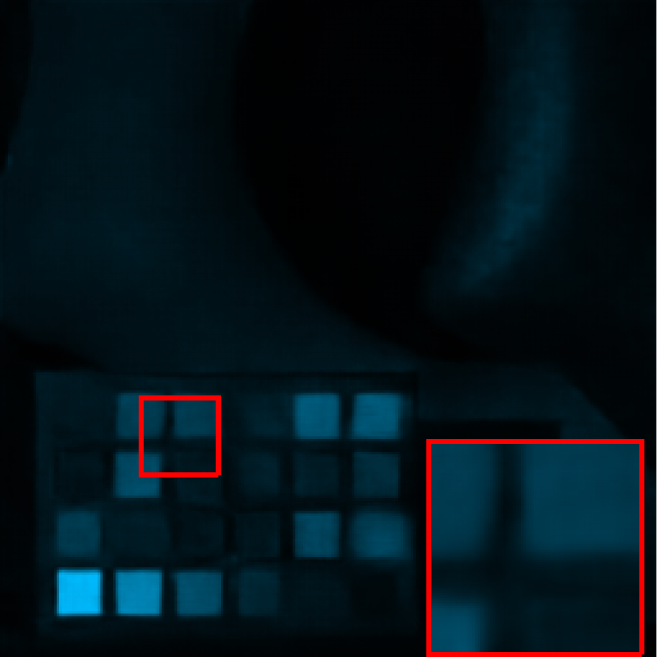}&
			\includegraphics[width=0.23\linewidth]{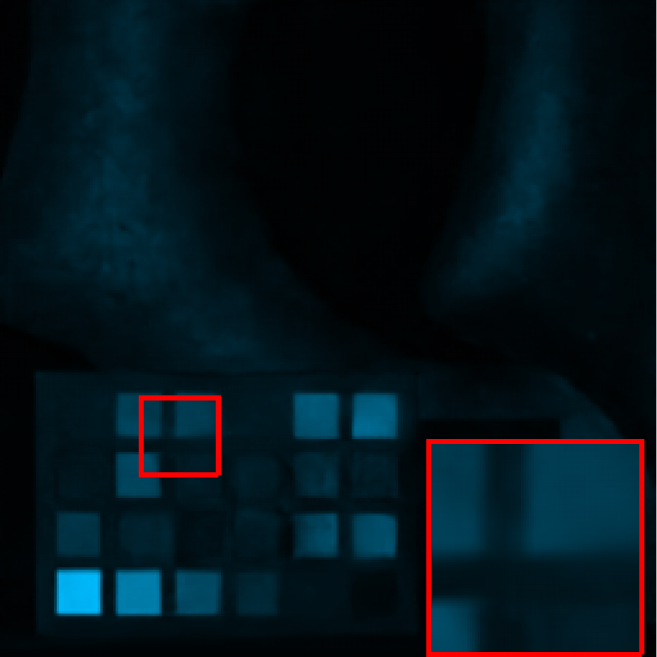}&
			\includegraphics[width=0.23\linewidth]{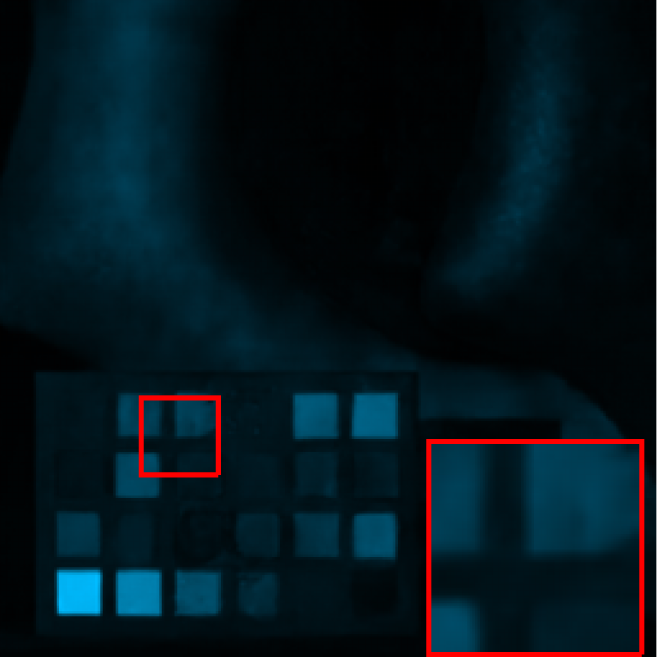}&\\
             Original & Measurement & $\lambda$-Net& TSA-Net& HDNet& DGSMP\\
              && \cite{miao2019net} &\cite{meng2020end} & \cite{hu2022hdnet} &\cite{huang2021deep}\\
			\includegraphics[width=0.23\linewidth]{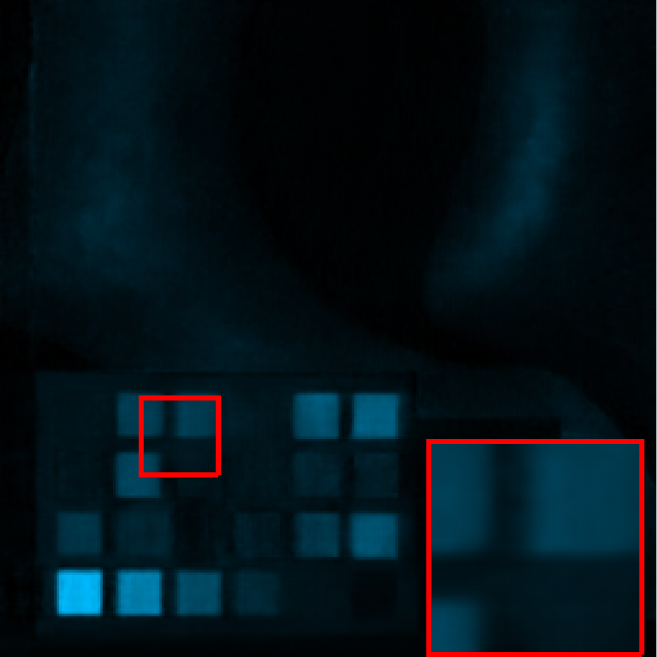}&
            \includegraphics[width=0.23\linewidth]{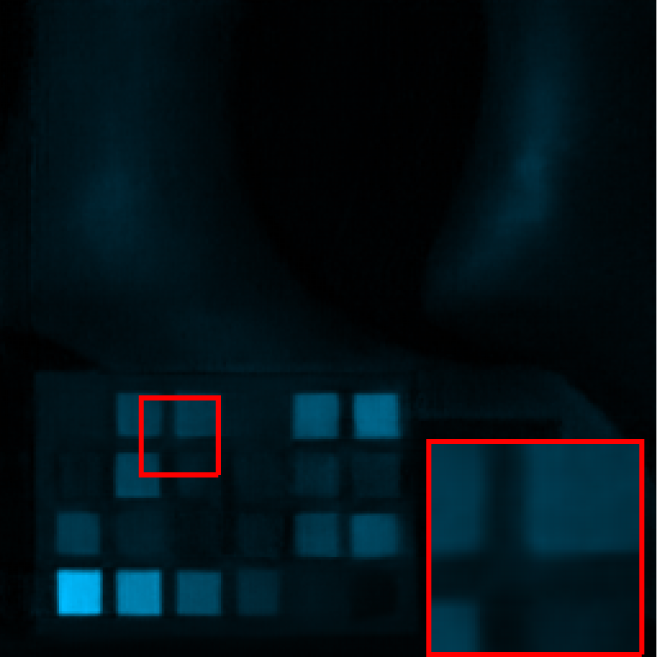}&
            \includegraphics[width=0.23\linewidth]{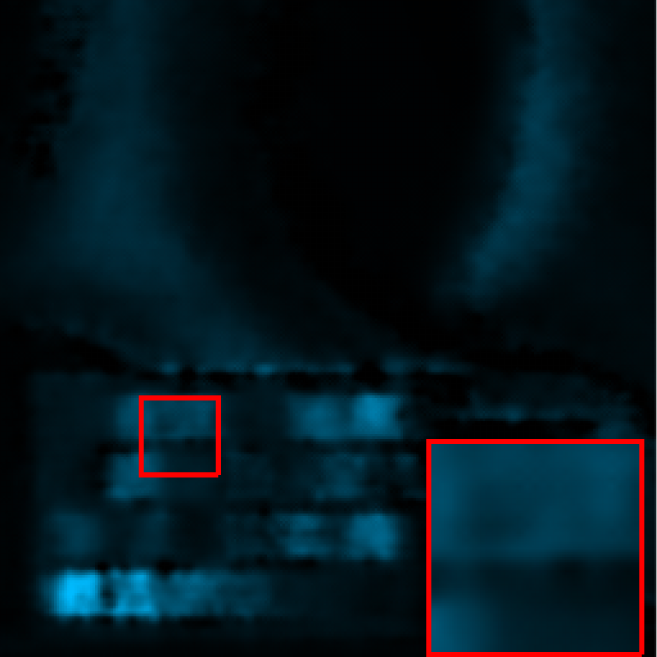}&
            \includegraphics[width=0.23\linewidth]{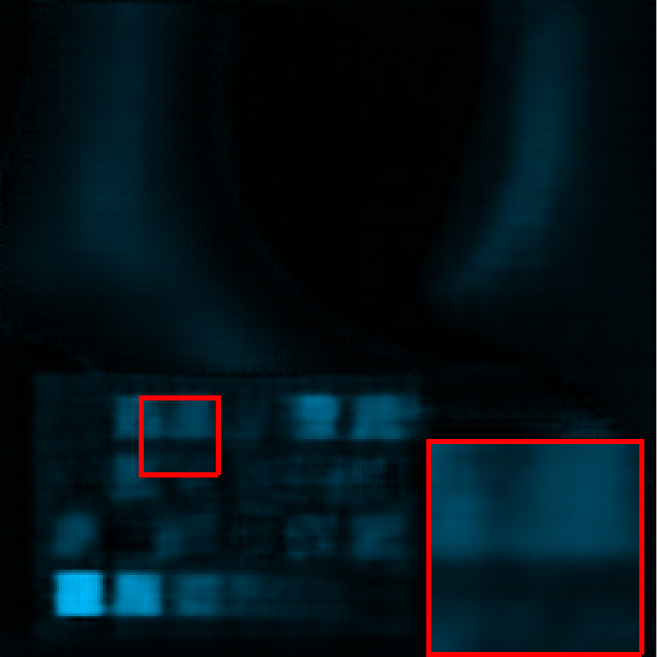}&
            \includegraphics[width=0.23\linewidth]{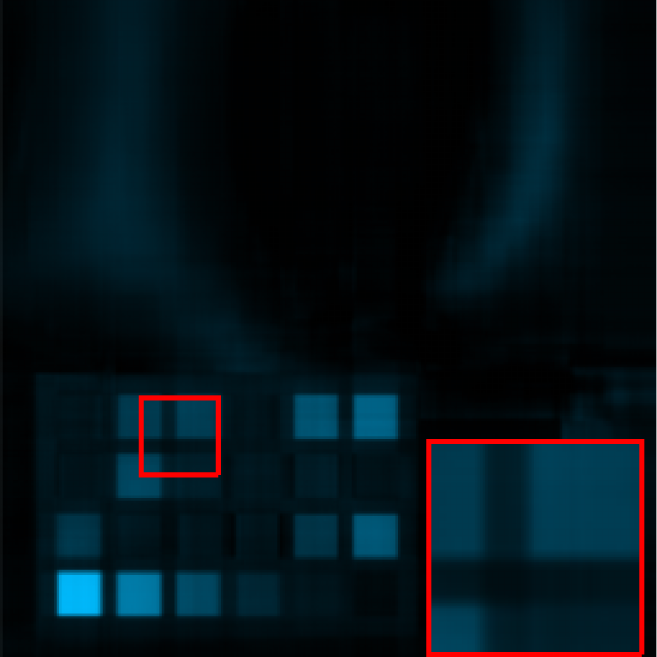}&
            \includegraphics[width=0.23\linewidth]{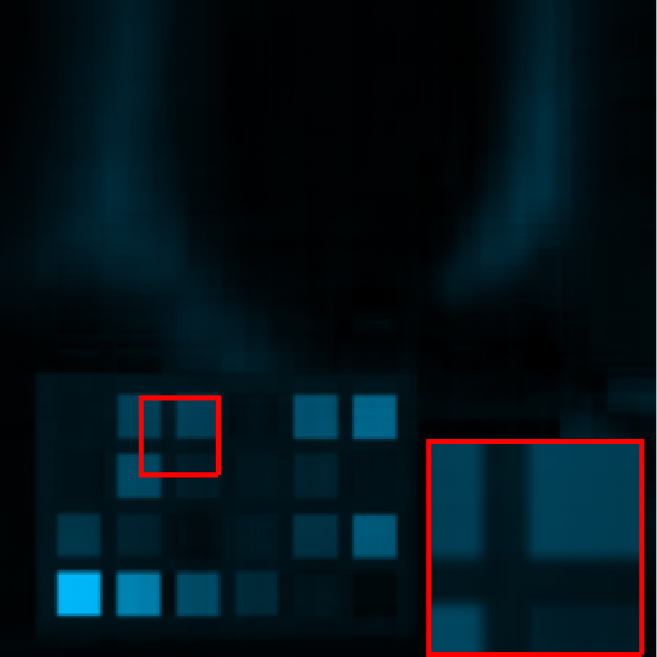}\\
            ADMM-Net & GAP-Net & PnP-CASSI & DIP-HSI & HLRTF & OTLRM\\
            \cite{admm-net}&\cite{meng2023deep}&\cite{zheng2021deep}&\cite{meng2021self}&\cite{lrtf}&
		\end{tabular}} 

		\caption{ The recovery results of \emph{scene03} by different methods in CASSI reconstruction. }
		\label{snapshot_S3}
	\end{center}
\end{figure*}
\begin{table*}[htb]
	\caption{Comparisons between the proposed model and SOTA methods on 5 simulation scenes (\emph{scene06}$\sim$\emph{scene10}) in KAIST.}

        \def\arraystretch{1}
        \setlength{\tabcolsep}{2pt}
	\centering
	\resizebox{1\textwidth}{!}
	{
		\centering
		\begin{tabular}{ccc|cccccccccccc}
			\bottomrule[0.15em]
   \multirow{2}[2]{*}{\textbf{Method}} &\multirow{2}[2]{*}{\textbf{Category}} & \multirow{2}[2]{*}{\textbf{Reference}} & \multicolumn{2}{c}{\textbf{scene06}} & \multicolumn{2}{c}{\textbf{scene07}} & \multicolumn{2}{c}{\textbf{scene08}}& \multicolumn{2}{c}{\textbf{scene09}}&\multicolumn{2}{c}{\textbf{scene10}}&\multicolumn{2}{c}{\textbf{Avg}}\\
   &  & & \textbf{PSNR}$\uparrow$  & \textbf{SSIM}$\uparrow$  & \textbf{PSNR}$\uparrow$  & \textbf{SSIM}$\uparrow$  & \textbf{PSNR}$\uparrow$  & \textbf{SSIM}$\uparrow$  &  \textbf{PSNR}$\uparrow$  & \textbf{SSIM}$\uparrow$  & \textbf{PSNR}$\uparrow$  & \textbf{SSIM}$\uparrow$  & \textbf{PSNR}$\uparrow$  & \textbf{SSIM}$\uparrow$\\
   \midrule[0.1em]
			DeSCI\cite{desci}
			& Model
               & TPAMI2019
			&24.66&0.76
			&24.96&0.73
			&24.15&0.75
			&23.56&0.70
			&24.17&0.68
			&24.30&0.72
			\\
			$\lambda$-Net\cite{miao2019net}
			& CNN (Supervised)
               & ICCV2019
			&28.64&0.85
			&26.47&0.81
			&26.09&0.83
			&27.50&0.83
			&27.13&0.82
			&27.17&0.83
			\\
			TSA-Net\cite{meng2020end}
			& CNN (Supervised)
               & ECCV2020
			&31.06&0.90
			&30.02&0.88
			&29.22&0.89
			&31.14&0.91
			&29.18&0.87
			&30.12&0.89
			\\
			HDNet\cite{hu2022hdnet}
			& Transformer (Supervised)
               & CVPR2022
			&\textbf{34.33}&\textbf{0.95}
			&33.27&\underline{0.93}
			&32.26&\textbf{0.95}
			&34.17&0.94
			&\textbf{32.22}&\textbf{0.94}
			&33.25&\textbf{0.94}
			\\
			DGSMP\cite{huang2021deep}
			& Deep Unfolding (Supervised)
               & CVPR2021
			&33.08&\underline{0.94}
			&30.74&0.89
            &31.55&\underline{0.92}
			&31.66&0.91
			&31.44&\underline{0.93}
			&31.69&\underline{0.92}
			\\
			ADMM-Net\cite{admm-net}
			& Deep Unfolding (Supervised)
               & ICCV2019
			&32.47&0.93
			&32.01&0.90
			&30.49&0.91
			&33.38&0.92
			&30.55&0.90
			&31.78&0.91
           \\
			GAP-Net\cite{meng2023deep}
			& Deep Unfolding (Supervised)
               & IJCV2023
			&32.29&0.93
			&31.79&0.90
			&30.25&0.91
			&33.06&0.92
			&30.14&0.90
			&31.51&0.91
			\\
			PnP-CASSI\cite{zheng2021deep}
			& PnP (Zero-Shot)
               & PR2021
			&26.16&0.75
			&26.92&0.74
			&24.92&0.71
			&27.99&0.75
			&25.58&0.66
			&26.31&0.72
			\\
			DIP-HSI\cite{meng2021self}
			& PnP (Zero-Shot)
               & ICCV2021
			&29.53&0.82
			&27.46&0.70
			&27.69&0.80
			&33.46&0.86
			&26.10&0.73
			&28.85&0.78
			\\
			HLRTF\cite{lrtf}
                &Tensor Network (Self-Supervised)
                & CVPR2022
			&32.81&0.92
			&\textbf{34.82}&\textbf{0.94}
			&\textbf{33.83}&0.89
			&\underline{38.73}&\underline{0.96}
			&31.52&0.90
			&34.34&\underline{0.92}
                \\
			\bf OTLRM
			& Tensor Network (Self-Supervised)
               & Ours
			&\underline{34.02}&0.88
			&\underline{34.09}&0.90
			&\underline{33.23}&0.85
			&\textbf{39.52}&\textbf{0.97}
			&\underline{32.14}&0.89
			&\textbf{34.60}&0.90
			\\
			\toprule[0.15em]
		\end{tabular}
	}
	\label{tab:simu1}
\end{table*}

\begin{figure*}[htb]

	\footnotesize
	\setlength{\tabcolsep}{1pt}
	\begin{center}
            \scalebox{0.7}{
		\begin{tabular}{ccccccccccc}
			\includegraphics[width=0.23\linewidth]{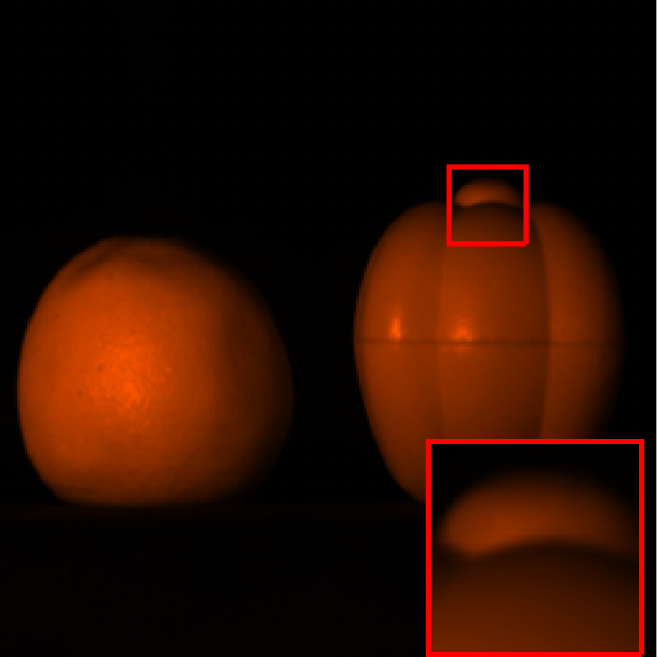}&
			\includegraphics[width=0.23\linewidth]{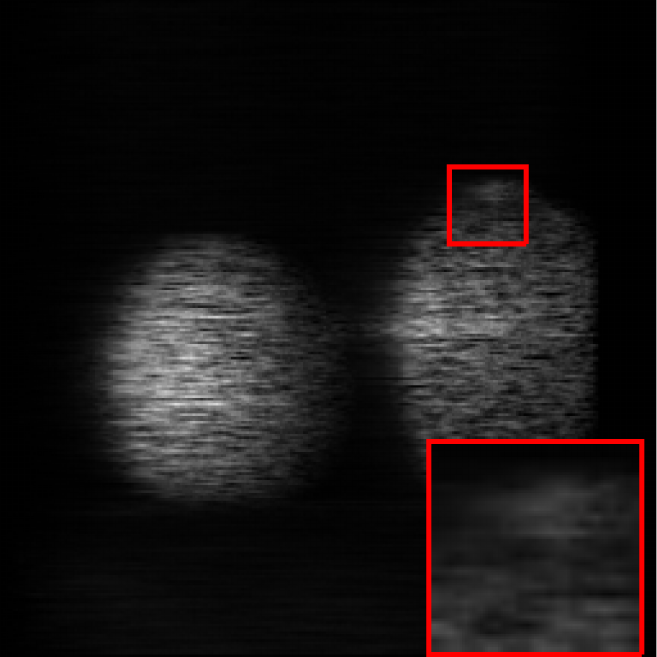}&
			\includegraphics[width=0.23\linewidth]{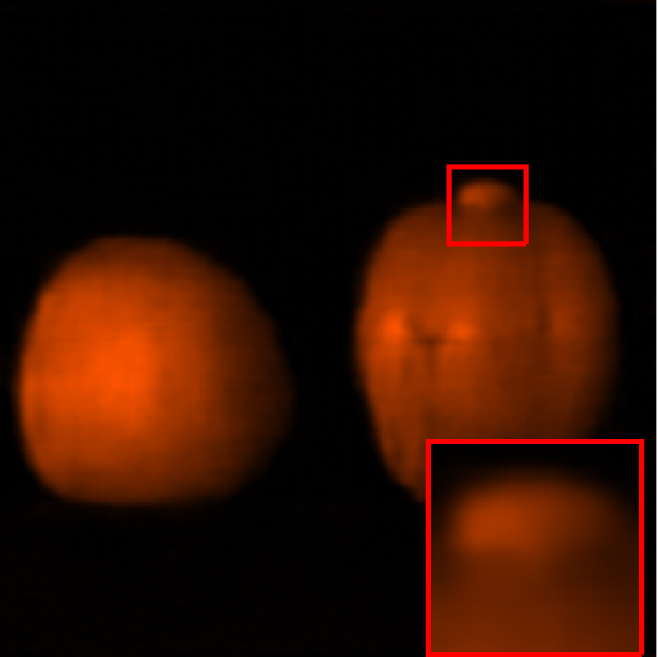}&
			\includegraphics[width=0.23\linewidth]{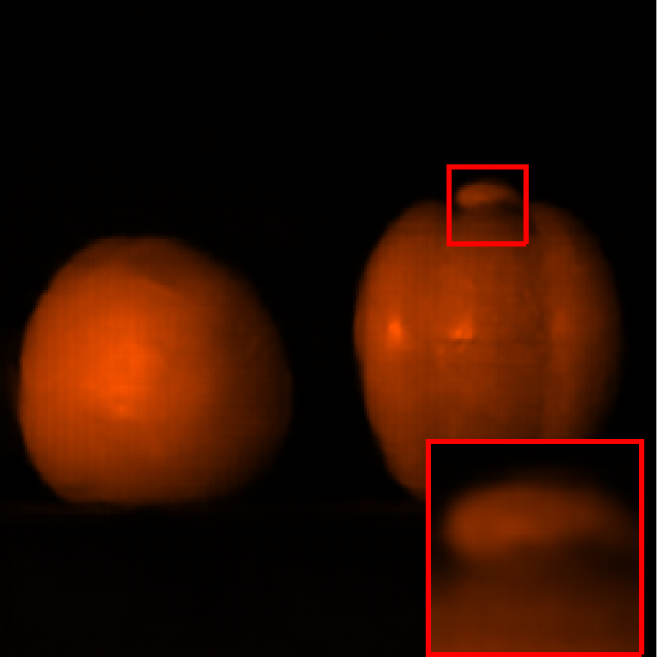}&
			\includegraphics[width=0.23\linewidth]{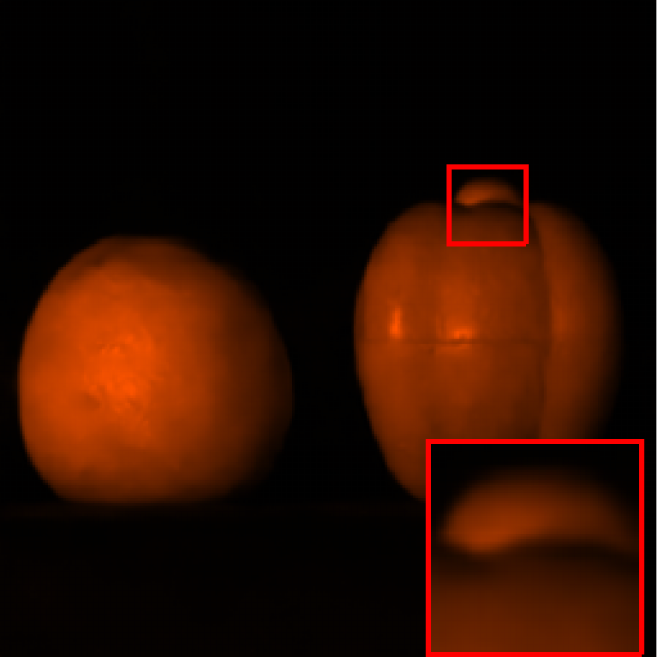}&
			\includegraphics[width=0.23\linewidth]{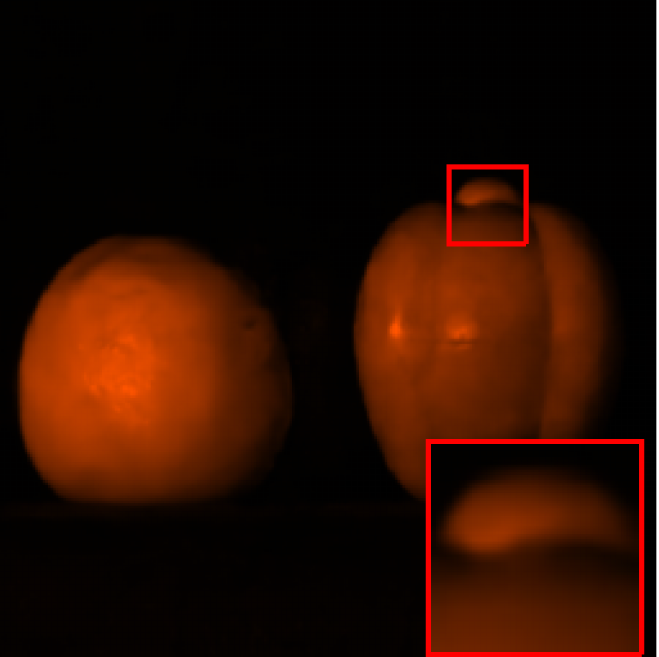}&\\
             Original & Measurement & $\lambda$-Net& TSA-Net& HDNet& DGSMP\\
              && \cite{miao2019net} &\cite{meng2020end} & \cite{hu2022hdnet} &\cite{huang2021deep}\\
			\includegraphics[width=0.23\linewidth]{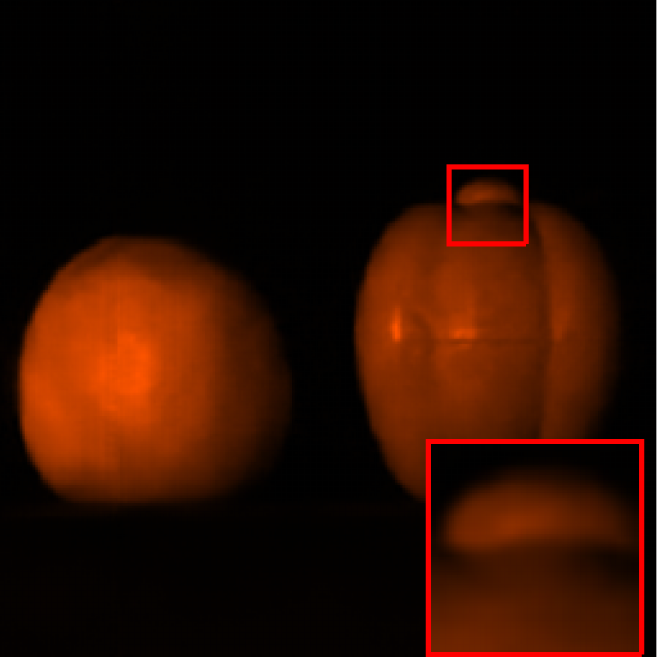}&
            \includegraphics[width=0.23\linewidth]{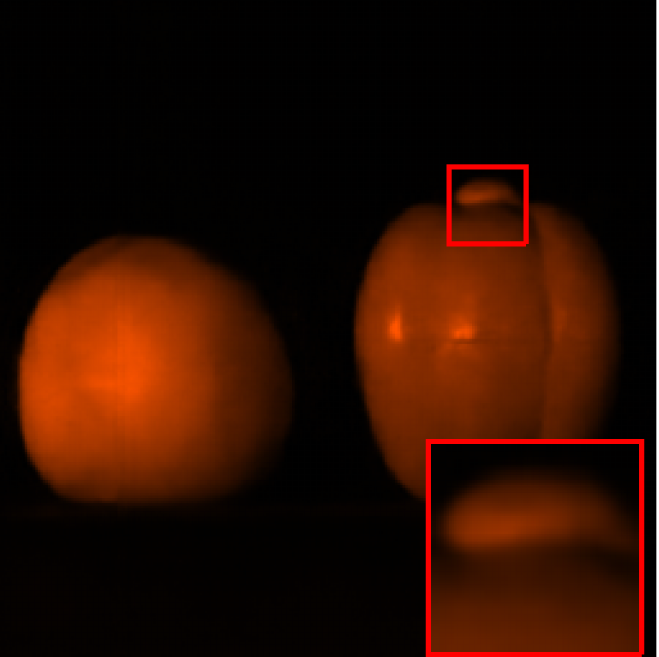}&
            \includegraphics[width=0.23\linewidth]{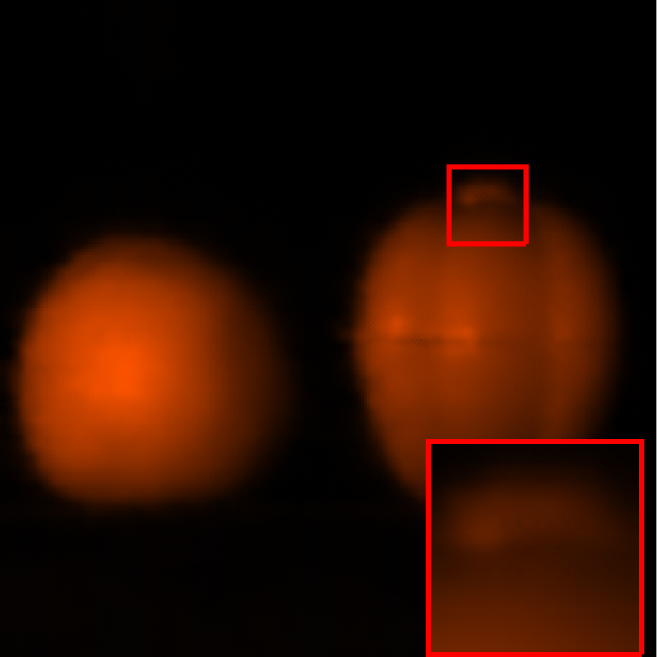}&
            \includegraphics[width=0.23\linewidth]{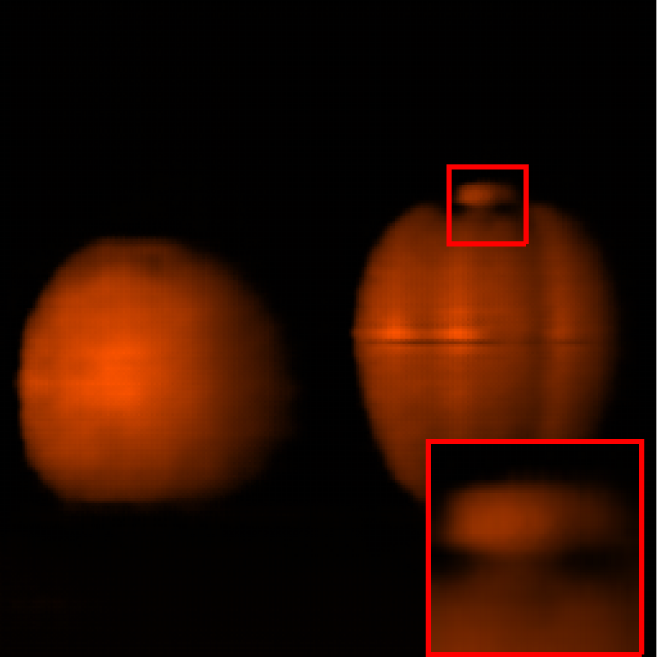}&
            \includegraphics[width=0.23\linewidth]{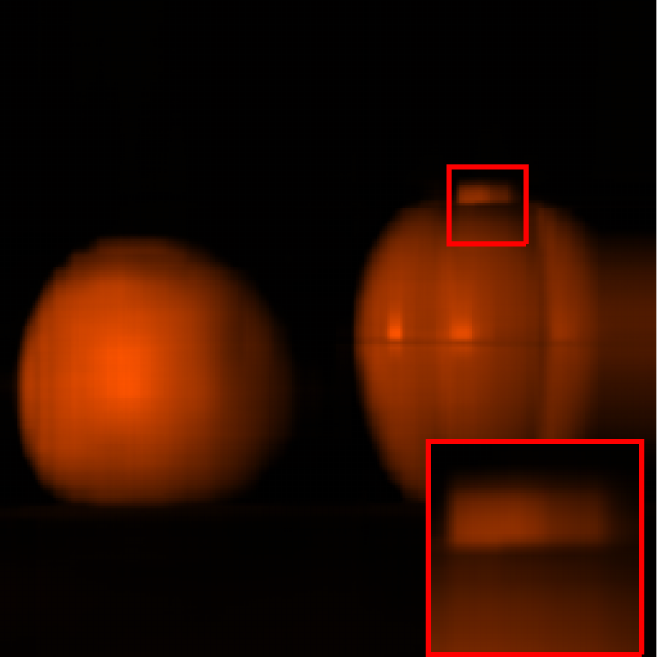}&
            \includegraphics[width=0.23\linewidth]{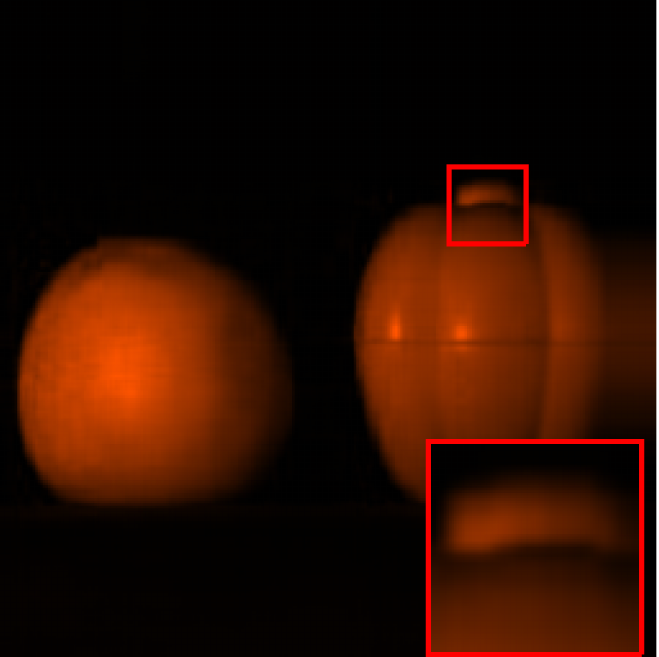}\\
            ADMM-Net & GAP-Net & PnP-CASSI & DIP-HSI & HLRTF & OTLRM\\
            \cite{admm-net}&\cite{meng2023deep}&\cite{zheng2021deep}&\cite{meng2021self}&\cite{lrtf}&
		\end{tabular}} 

		\caption{ The recovery results of \emph{scene04} by different methods in CASSI reconstruction. }
		\label{snapshot_S4}
	\end{center}
        
\end{figure*}

\begin{figure*}[htb]

	\footnotesize
	\setlength{\tabcolsep}{1pt}
	\begin{center}
            \scalebox{0.7}{
		\begin{tabular}{ccccccccccc}
			\includegraphics[width=0.23\linewidth]{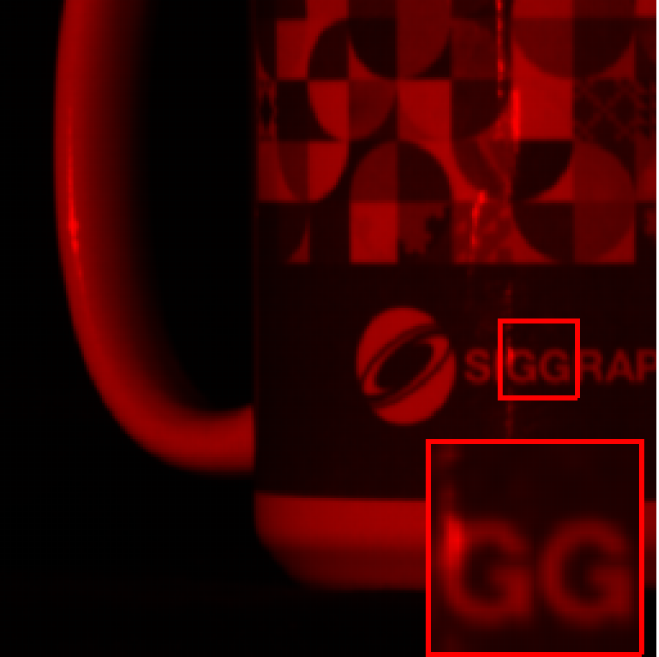}&
			\includegraphics[width=0.23\linewidth]{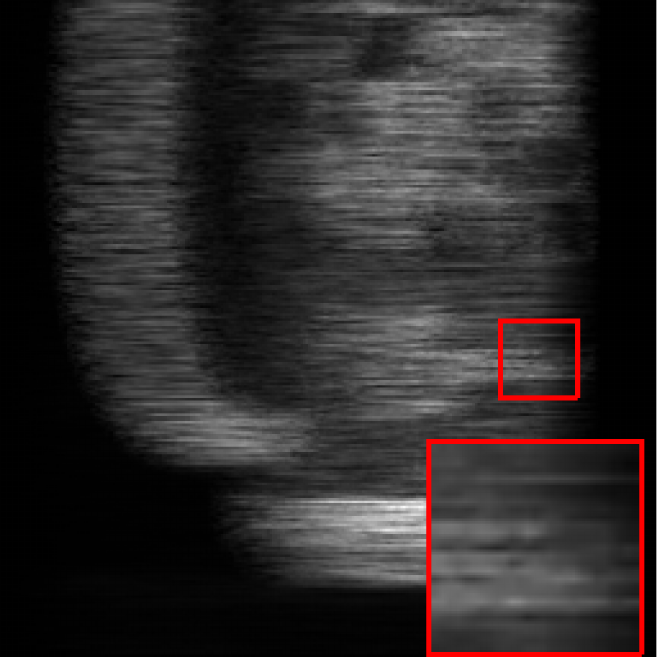}&
			\includegraphics[width=0.23\linewidth]{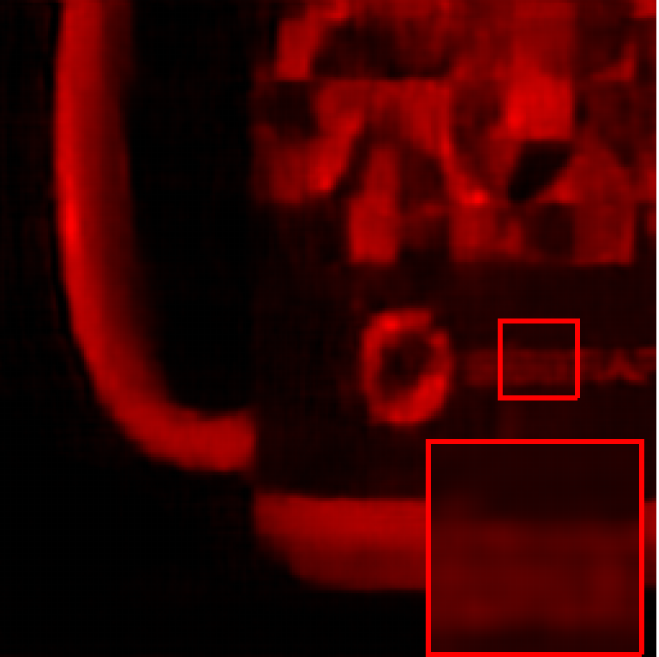}&
			\includegraphics[width=0.23\linewidth]{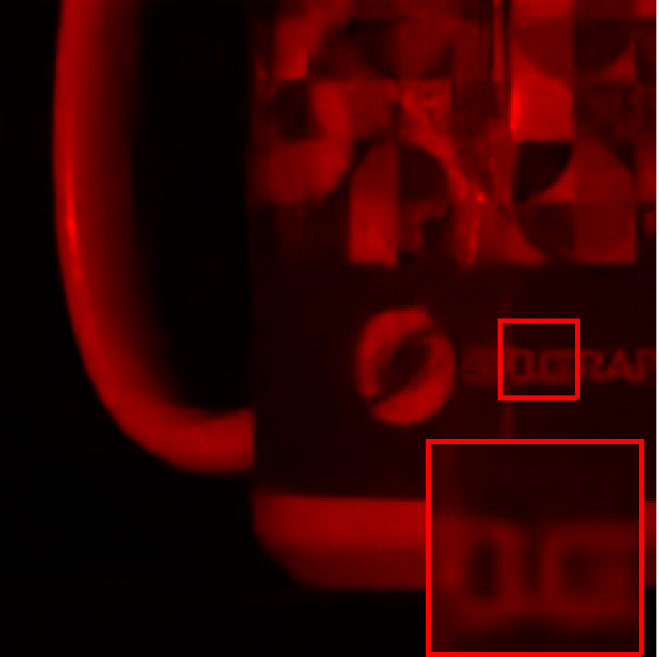}&
			\includegraphics[width=0.23\linewidth]{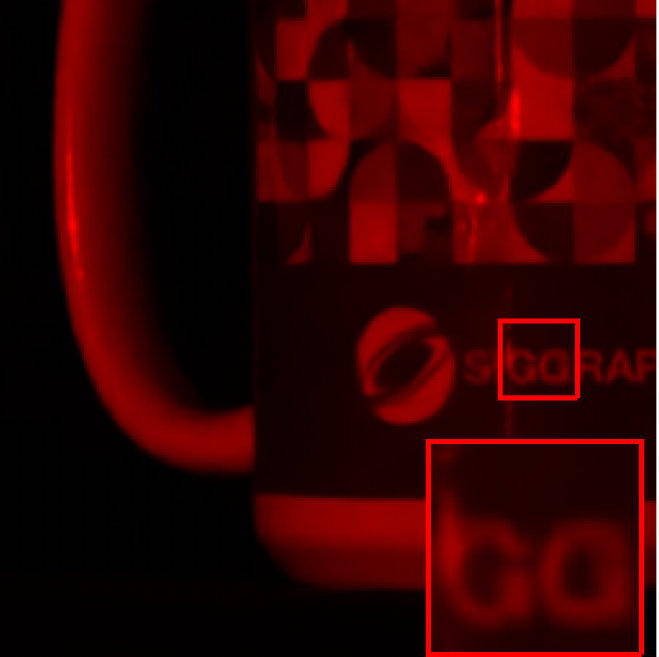}&
			\includegraphics[width=0.23\linewidth]{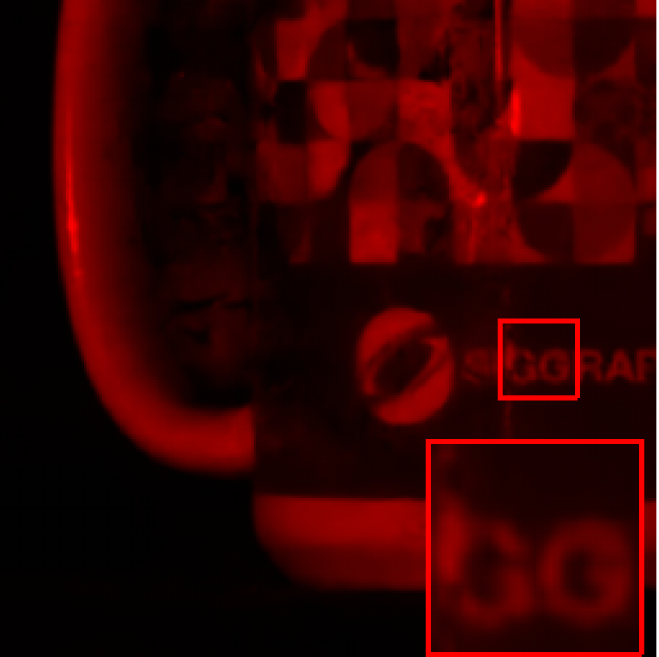}&\\
             Original & Measurement & $\lambda$-Net& TSA-Net& HDNet& DGSMP\\
              && \cite{miao2019net} &\cite{meng2020end} & \cite{hu2022hdnet} &\cite{huang2021deep}\\
			\includegraphics[width=0.23\linewidth]{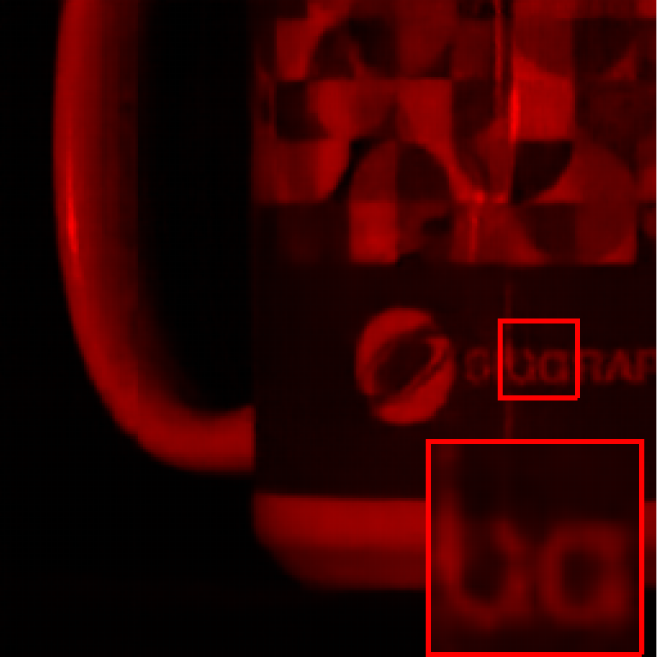}&
            \includegraphics[width=0.23\linewidth]{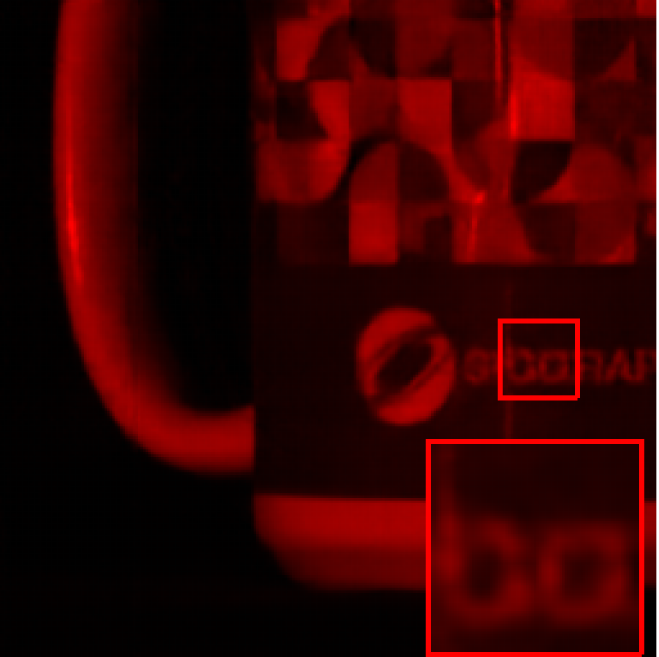}&
            \includegraphics[width=0.23\linewidth]{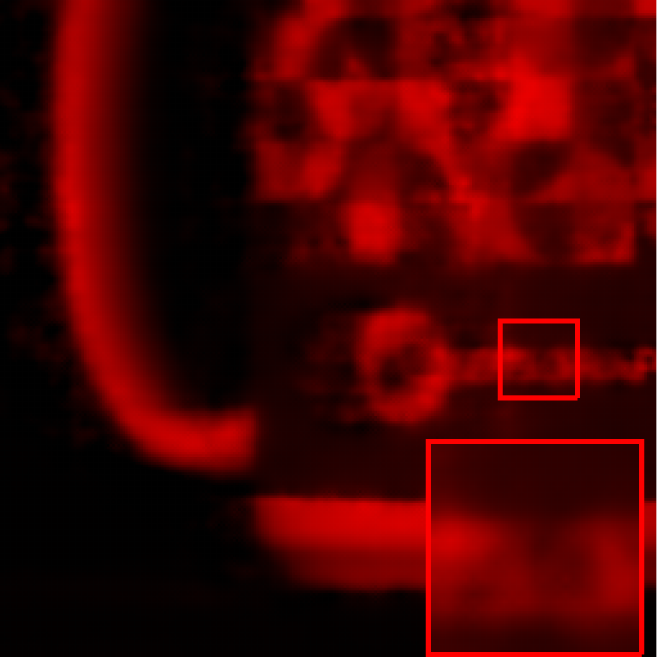}&
            \includegraphics[width=0.23\linewidth]{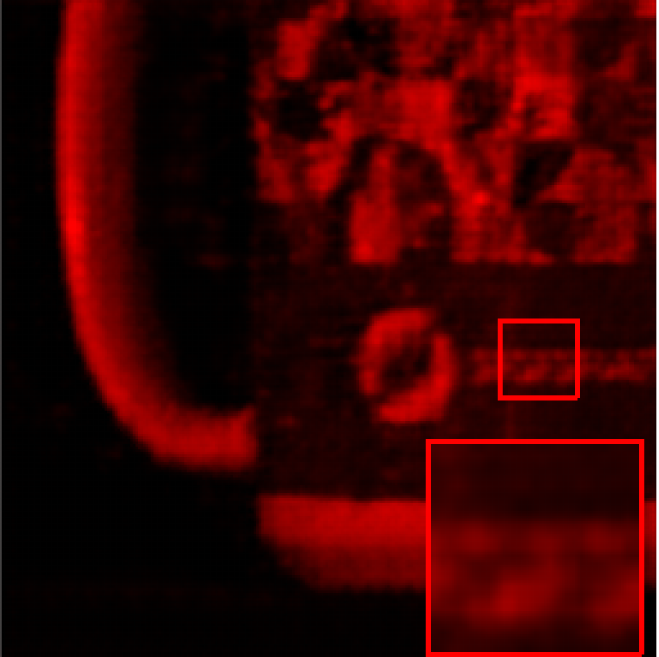}&
            \includegraphics[width=0.23\linewidth]{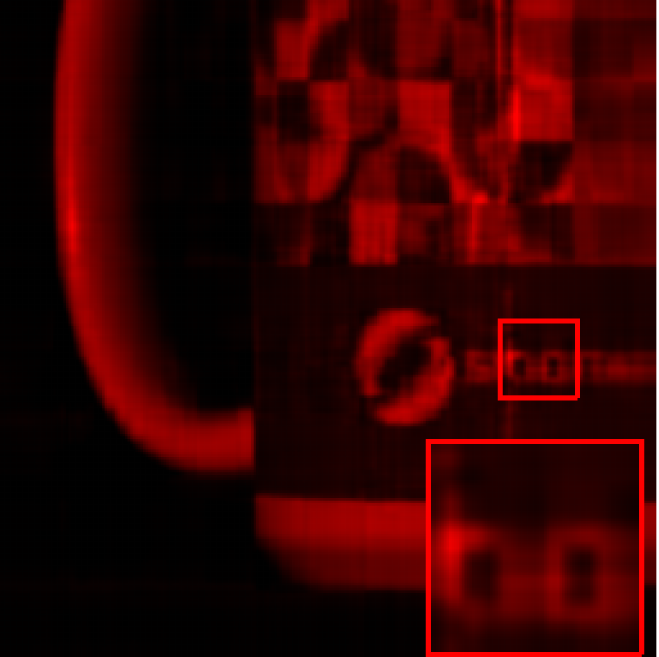}&
            \includegraphics[width=0.23\linewidth]{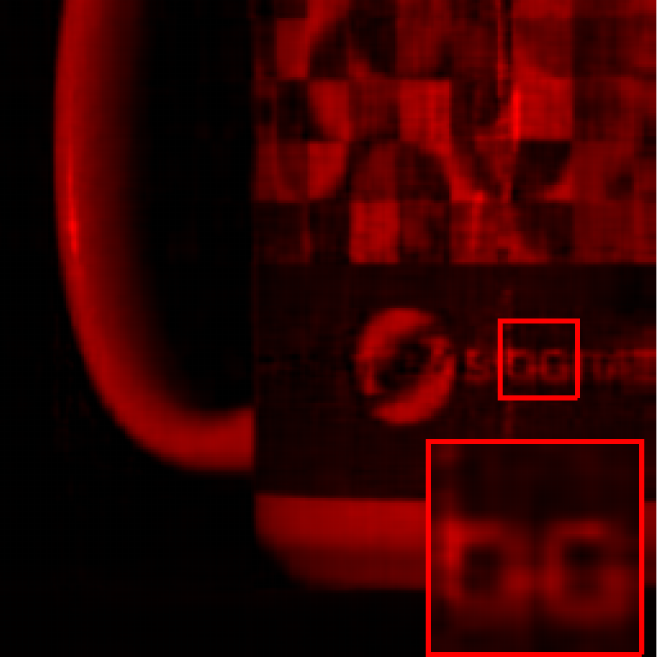}\\
            ADMM-Net & GAP-Net & PnP-CASSI & DIP-HSI & HLRTF & OTLRM\\
            \cite{admm-net}&\cite{meng2023deep}&\cite{zheng2021deep}&\cite{meng2021self}&\cite{lrtf}&
		\end{tabular}} 

		\caption{ The recovery results of \emph{scene05} by different methods in CASSI reconstruction. }
		\label{snapshot_S5}

	\end{center}
 
\end{figure*}

\begin{figure*}[htb]

	\footnotesize
	\setlength{\tabcolsep}{1pt}
	\begin{center}

            \scalebox{0.7}{
		\begin{tabular}{ccccccccccc}
			\includegraphics[width=0.23\linewidth]{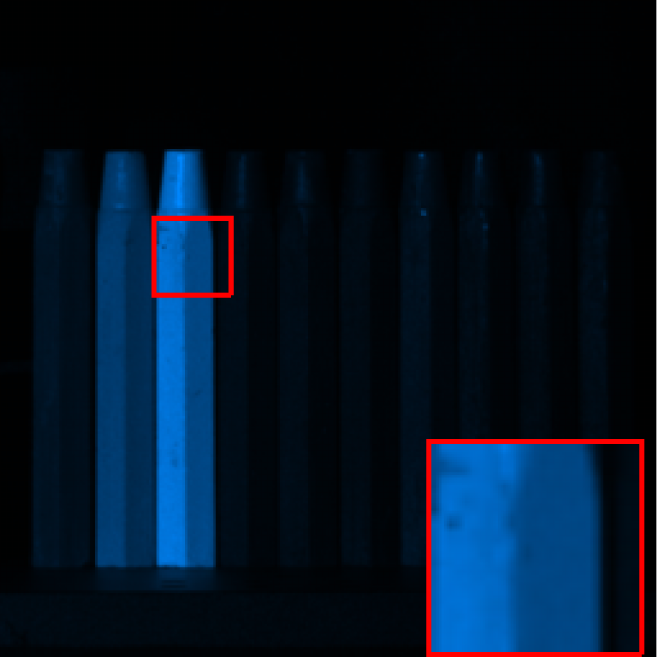}&
			\includegraphics[width=0.23\linewidth]{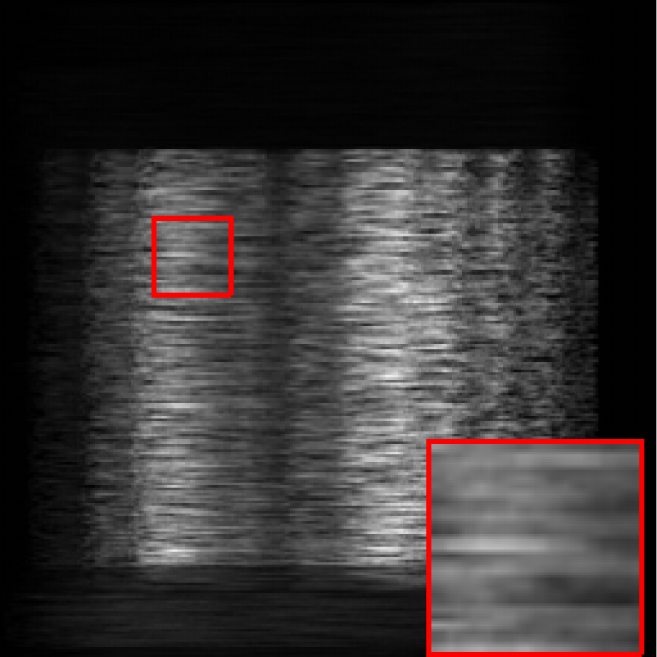}&
			\includegraphics[width=0.23\linewidth]{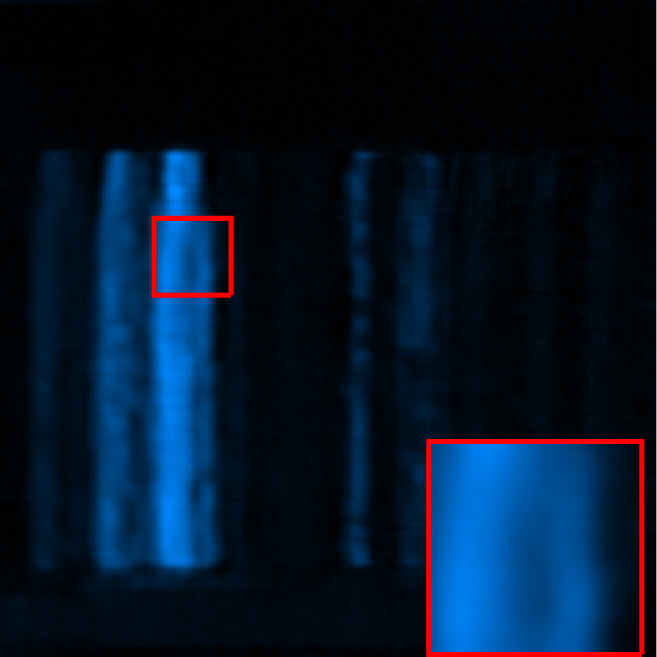}&
			\includegraphics[width=0.23\linewidth]{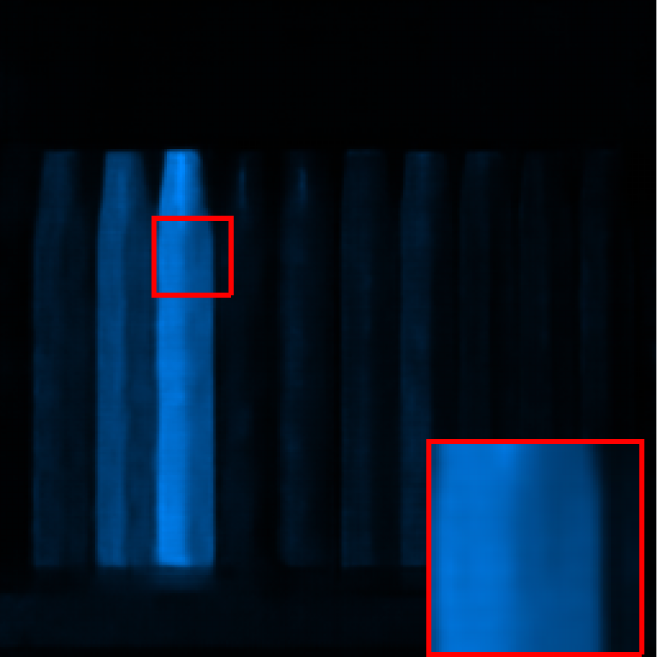}&
			\includegraphics[width=0.23\linewidth]{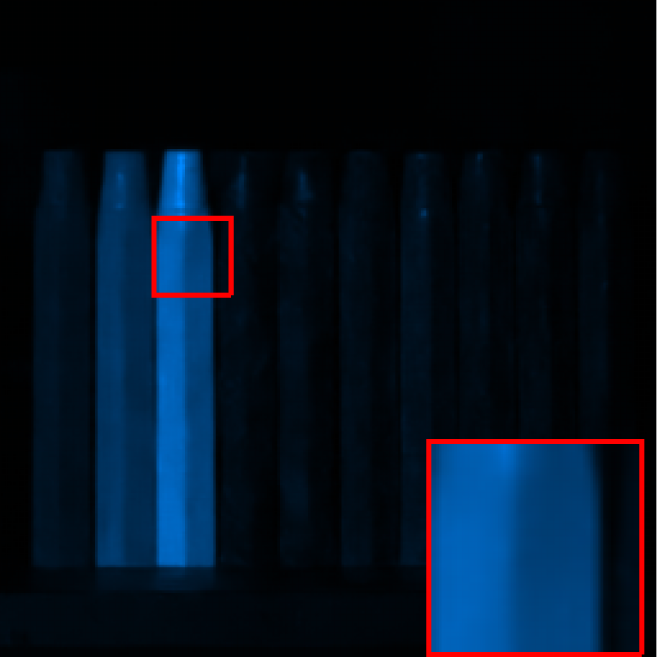}&
			\includegraphics[width=0.23\linewidth]{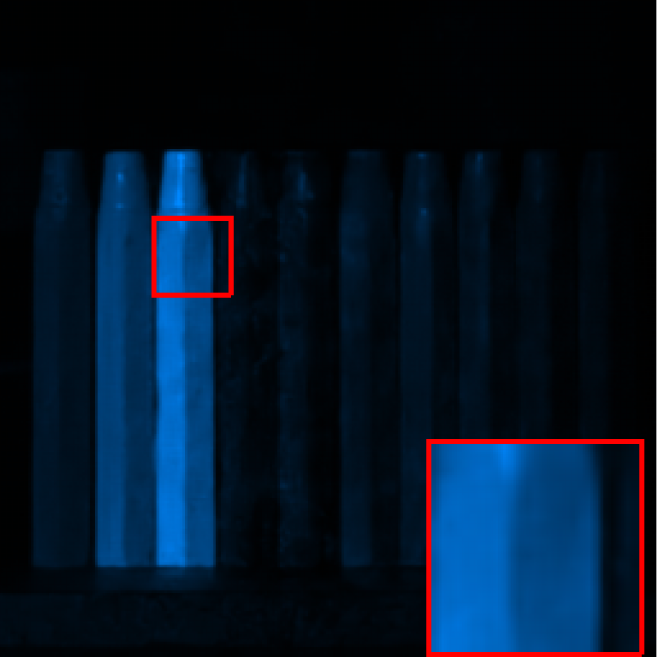}&\\
             Original & Measurement & $\lambda$-Net& TSA-Net& HDNet& DGSMP\\
              && \cite{miao2019net} &\cite{meng2020end} & \cite{hu2022hdnet} &\cite{huang2021deep}\\
			\includegraphics[width=0.23\linewidth]{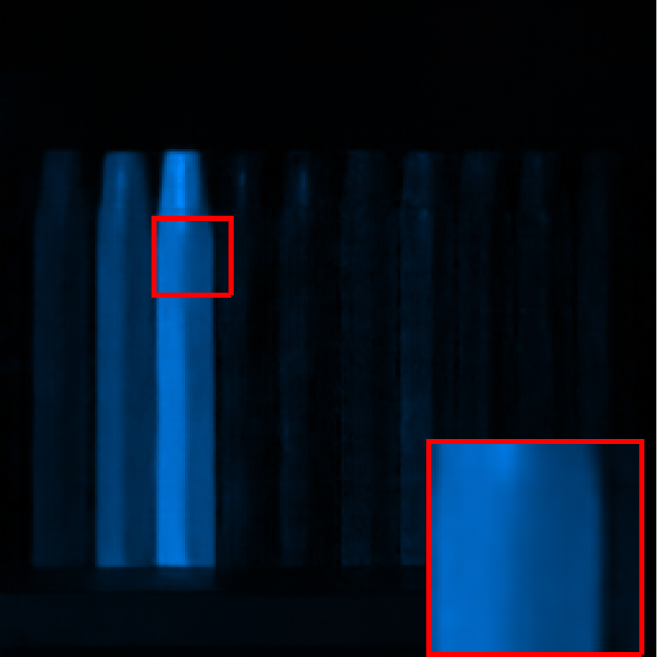}&
            \includegraphics[width=0.23\linewidth]{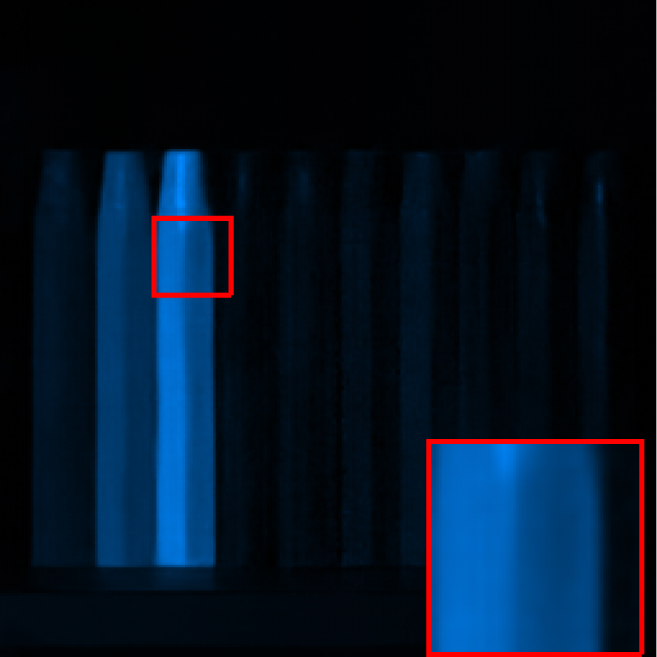}&
            \includegraphics[width=0.23\linewidth]{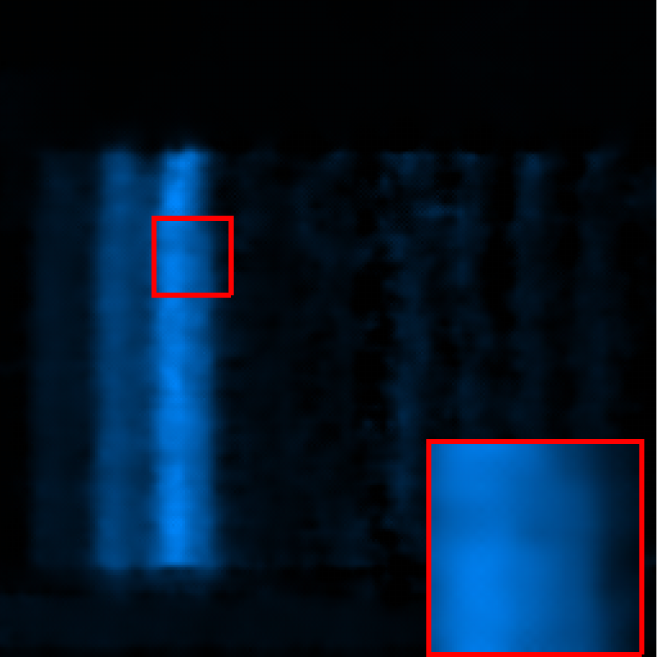}&
            \includegraphics[width=0.23\linewidth]{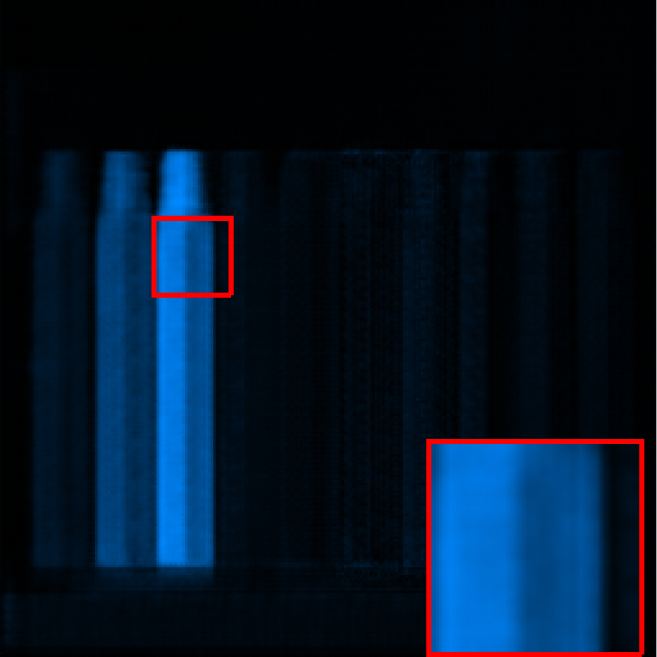}&
            \includegraphics[width=0.23\linewidth]{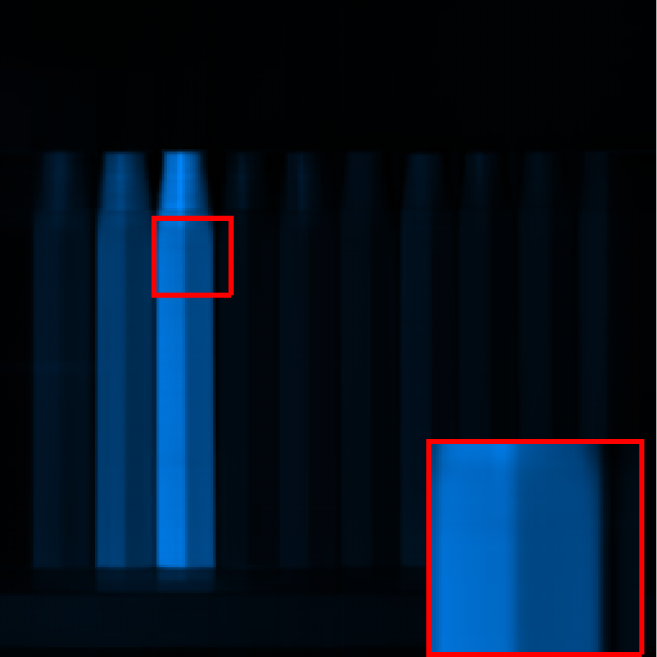}&
            \includegraphics[width=0.23\linewidth]{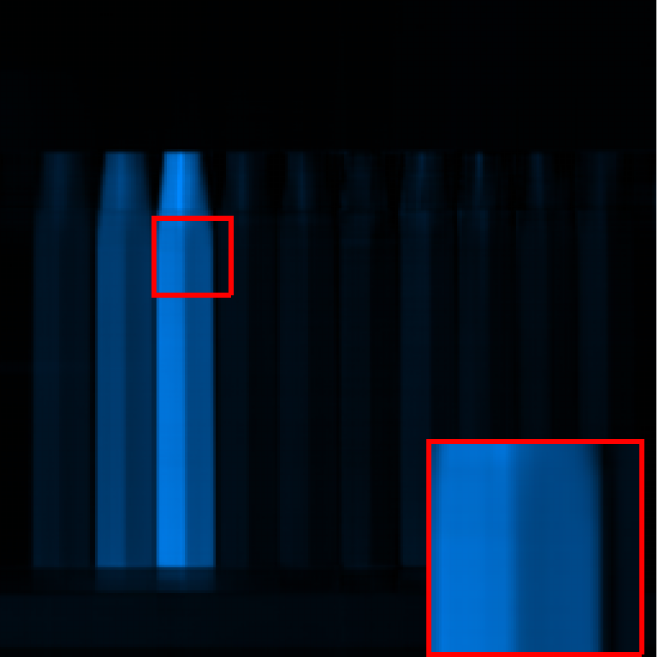}\\
            ADMM-Net & GAP-Net & PnP-CASSI & DIP-HSI & HLRTF & OTLRM\\
            \cite{admm-net}&\cite{meng2023deep}&\cite{zheng2021deep}&\cite{meng2021self}&\cite{lrtf}&
		\end{tabular}} 

		\caption{ The recovery results of \emph{scene09} by different methods in CASSI reconstruction. }
		\label{snapshot_S9}
	\end{center}
 
\end{figure*}

\begin{figure*}[htb]

	\footnotesize
	\setlength{\tabcolsep}{1pt}
	\begin{center}

            \scalebox{0.7}{
		\begin{tabular}{ccccccccccc}
			\includegraphics[width=0.23\linewidth]{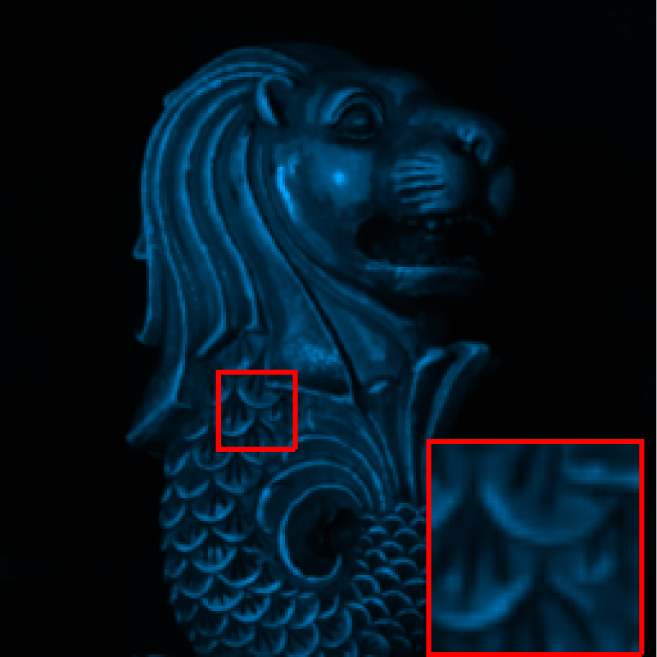}&
			\includegraphics[width=0.23\linewidth]{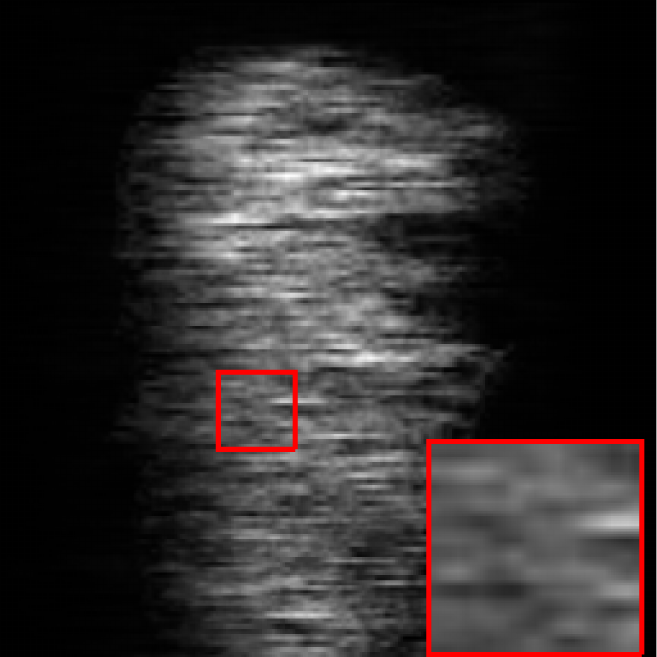}&
			\includegraphics[width=0.23\linewidth]{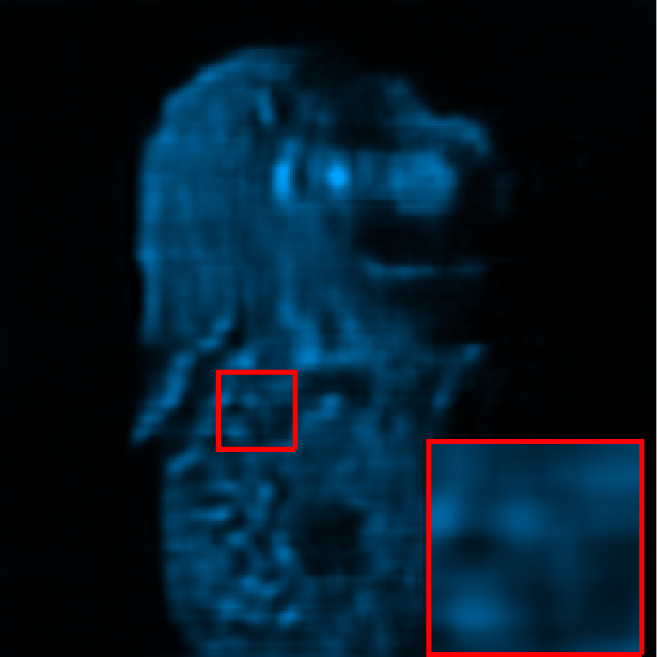}&
			\includegraphics[width=0.23\linewidth]{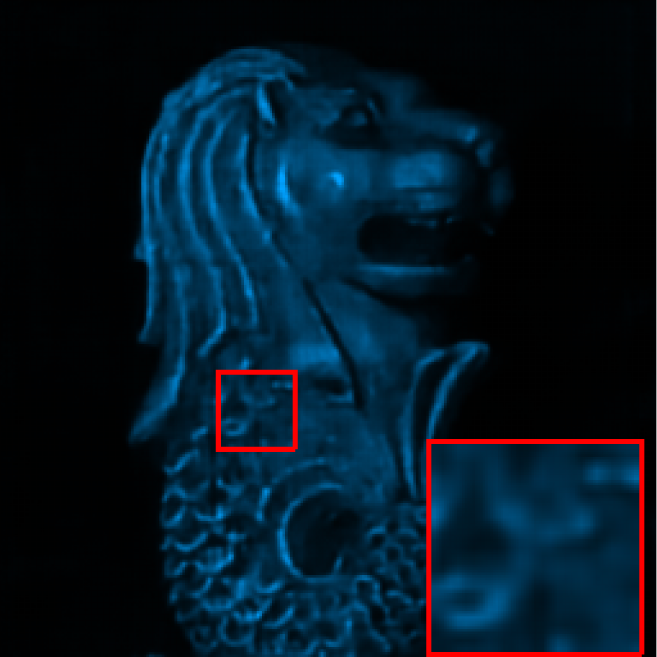}&
			\includegraphics[width=0.23\linewidth]{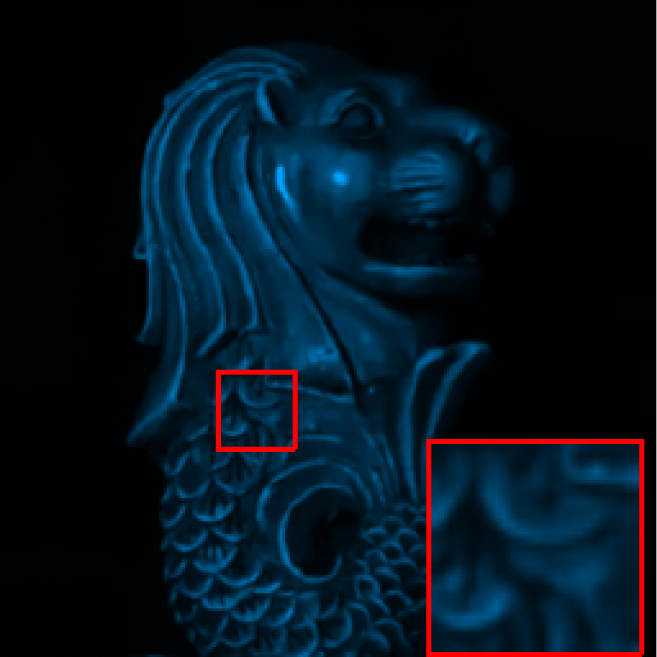}&
			\includegraphics[width=0.23\linewidth]{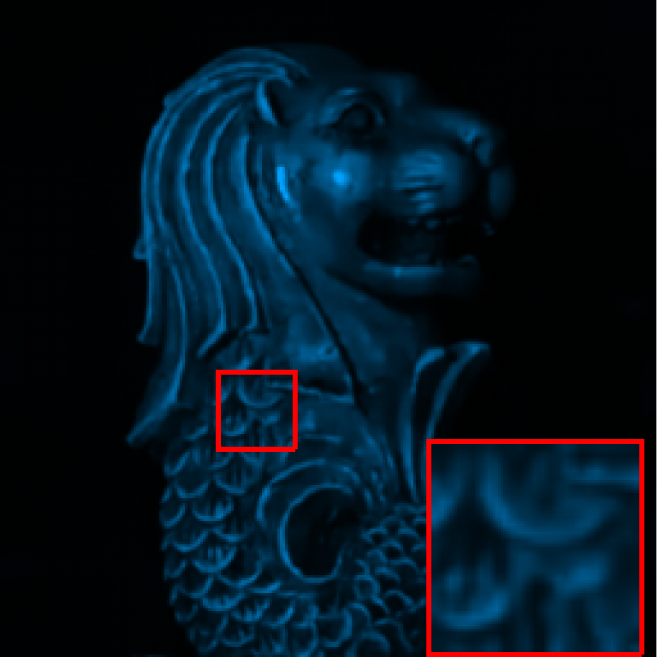}&\\
             Original & Measurement & $\lambda$-Net& TSA-Net& HDNet& DGSMP\\
              && \cite{miao2019net} &\cite{meng2020end} & \cite{hu2022hdnet} &\cite{huang2021deep}\\
			\includegraphics[width=0.23\linewidth]{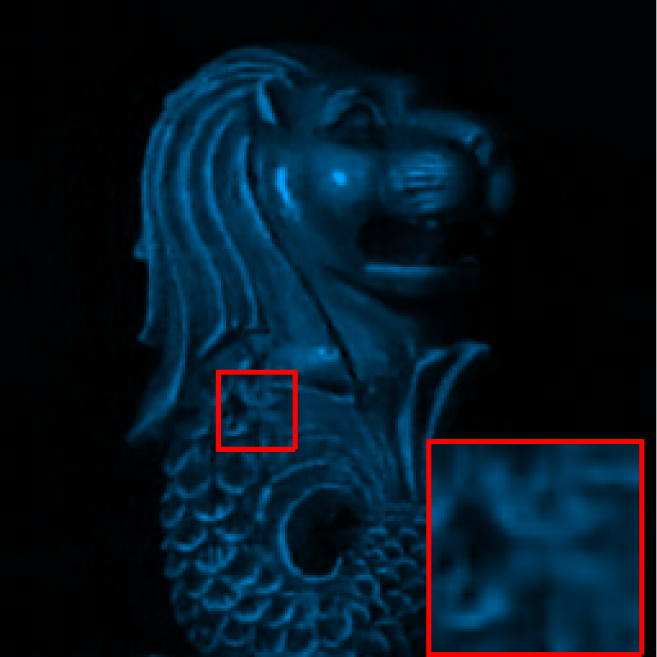}&
            \includegraphics[width=0.23\linewidth]{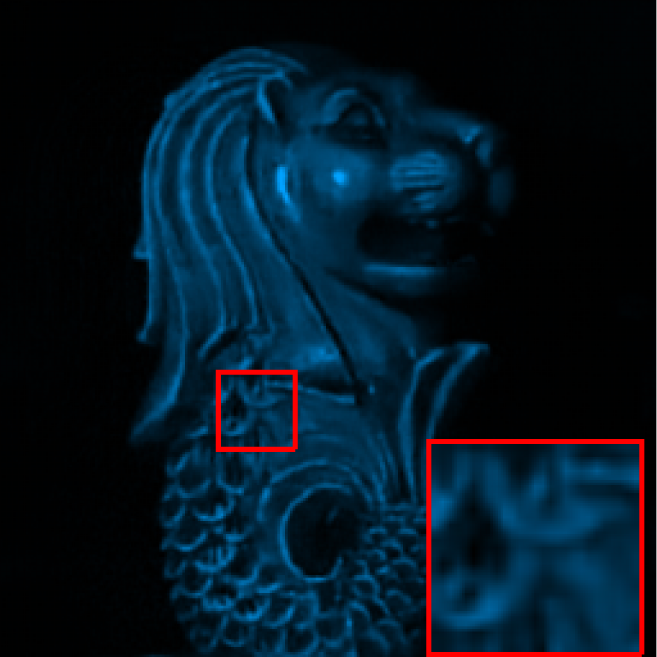}&
            \includegraphics[width=0.23\linewidth]{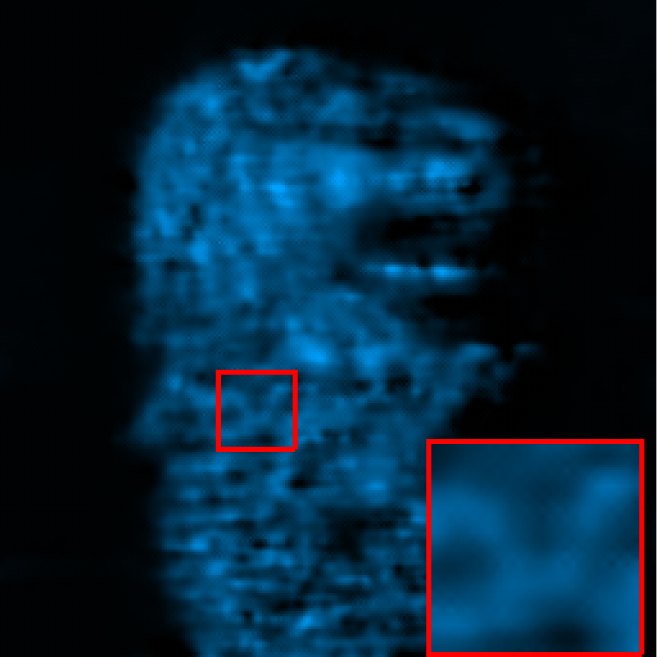}&
            \includegraphics[width=0.23\linewidth]{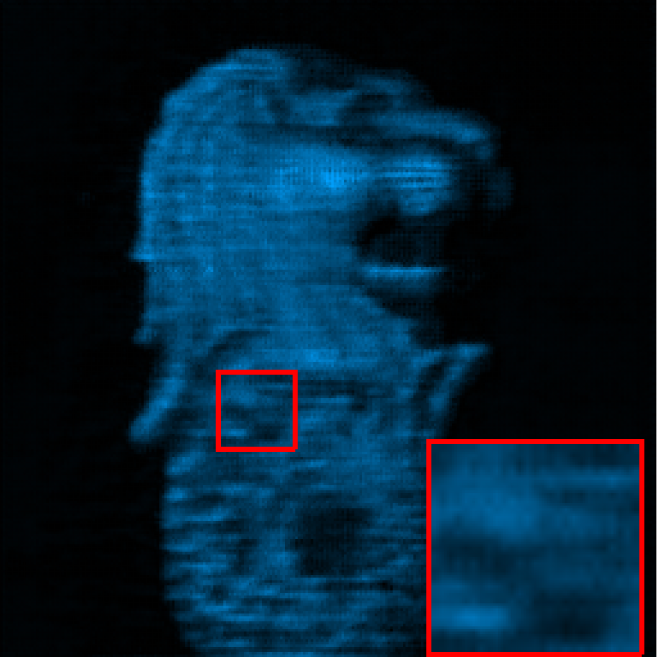}&
            \includegraphics[width=0.23\linewidth]{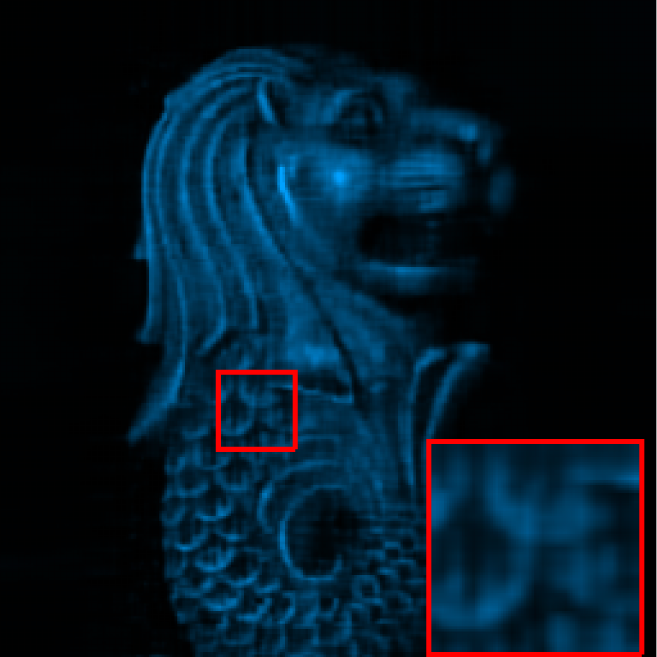}&
            \includegraphics[width=0.23\linewidth]{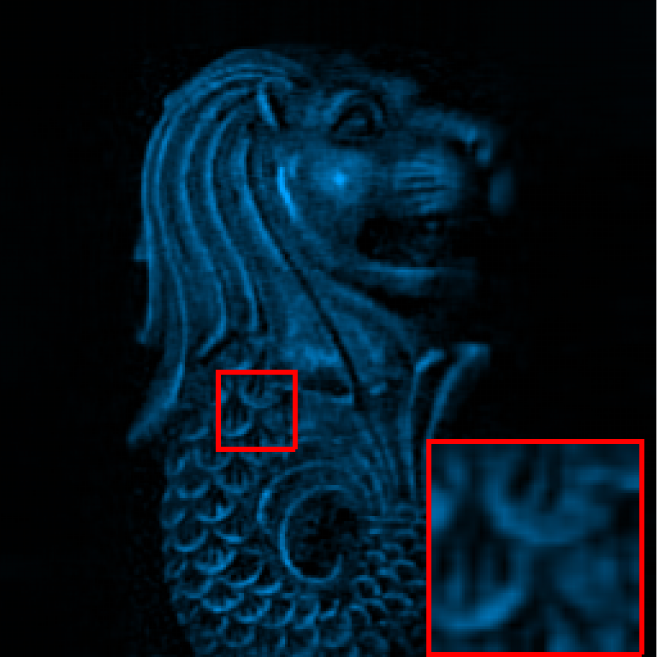}\\
            ADMM-Net & GAP-Net & PnP-CASSI & DIP-HSI & HLRTF & OTLRM\\
            \cite{admm-net}&\cite{meng2023deep}&\cite{zheng2021deep}&\cite{meng2021self}&\cite{lrtf}&
		\end{tabular}} 

		\caption{ The recovery results of \emph{scene10} by different methods in CASSI reconstruction. }
		\label{snapshot_S10}
	\end{center}
 
\end{figure*}

\subsection{MSI Denoising}
\begin{figure*}[htb]
	\footnotesize
	\setlength{\tabcolsep}{1pt}
	\begin{center}
            \scalebox{0.5}{
		\begin{tabular}{cccccccccc}
			\includegraphics[width=0.26\linewidth]{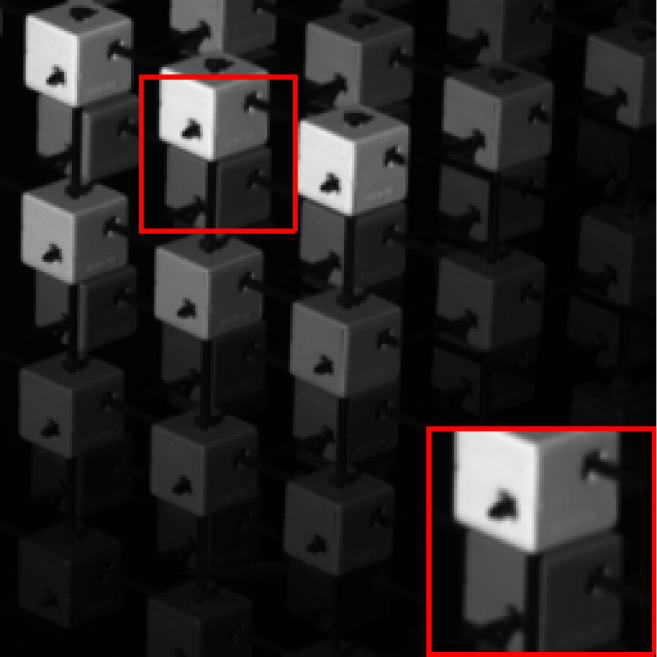}&
			\includegraphics[width=0.26\linewidth]{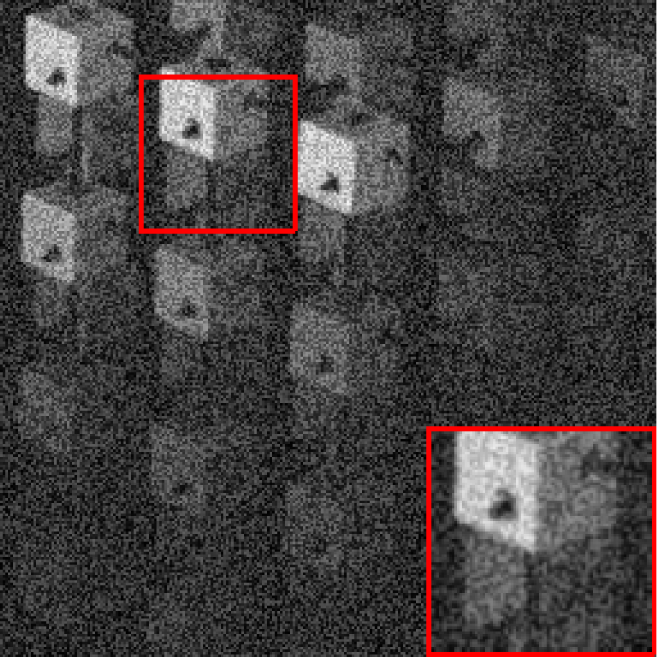}&
			\includegraphics[width=0.26\linewidth]{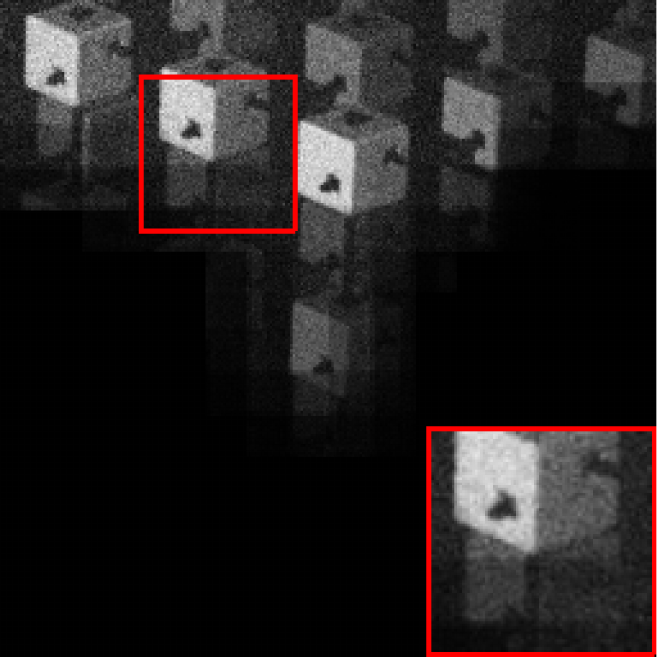}&
			\includegraphics[width=0.26\linewidth]{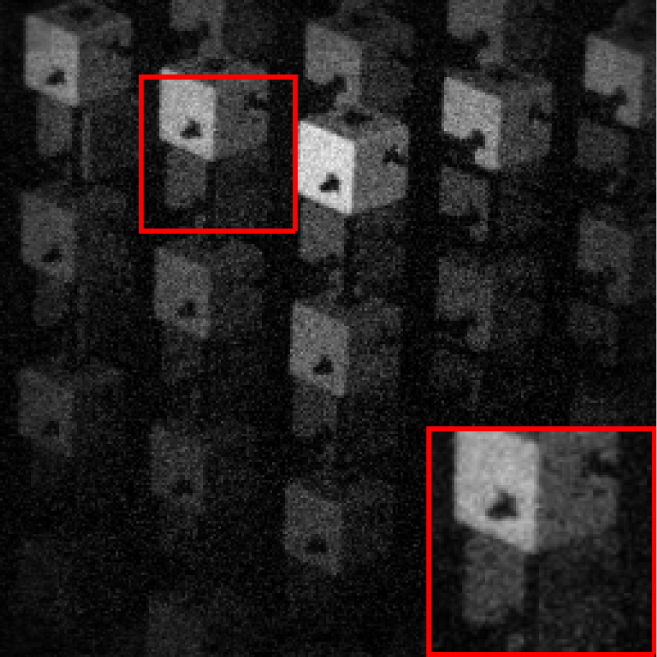}&
            \includegraphics[width=0.26\linewidth]{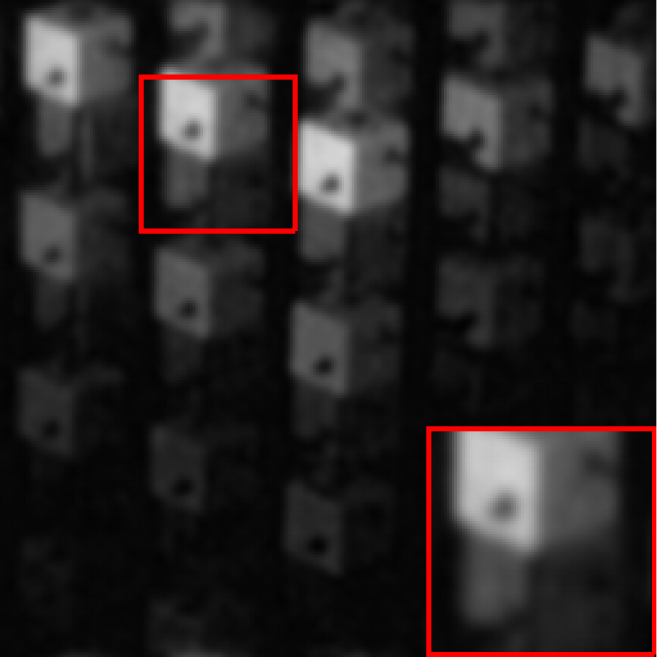}&
            \includegraphics[width=0.26\linewidth]{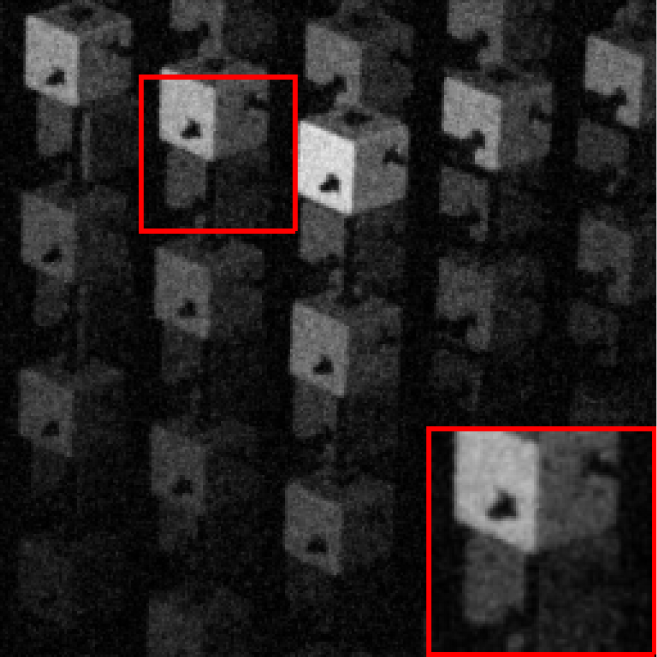}\\
            Original & Noisy image & NonLRMA & TLRLSSTV & 3DTNN & LRTDCTV \\
             &  & \cite{NonLRMA} & \cite{TLR_LSSTV} & \cite{3DTNN_FW} & \cite{LRTDCTV} \\
            \includegraphics[width=0.26\linewidth]{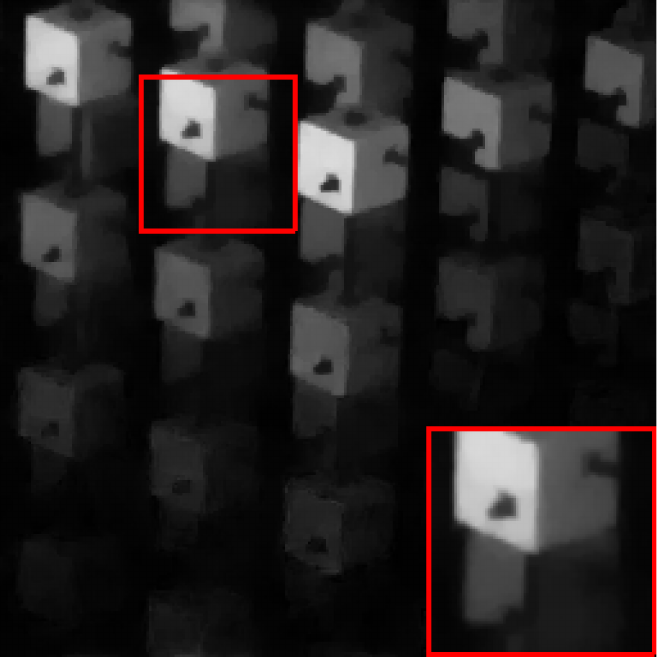}&
            \includegraphics[width=0.26\linewidth]{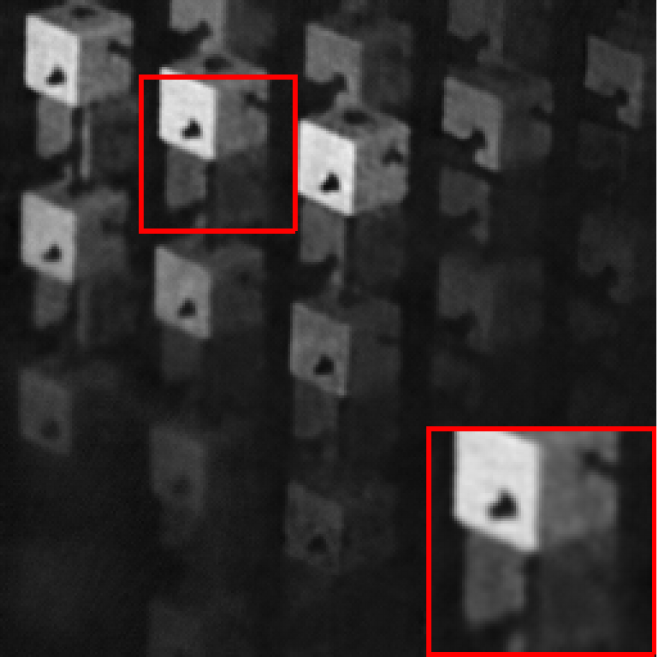}&
            \includegraphics[width=0.26\linewidth]{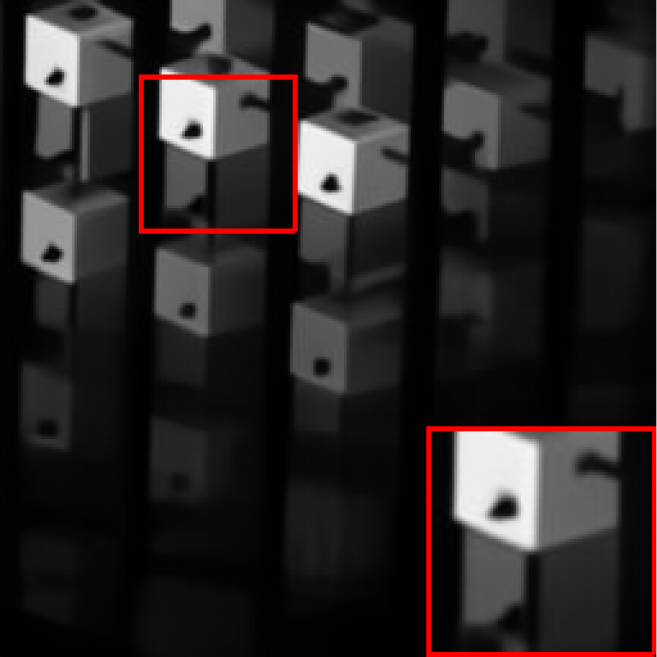}&
            \includegraphics[width=0.26\linewidth]{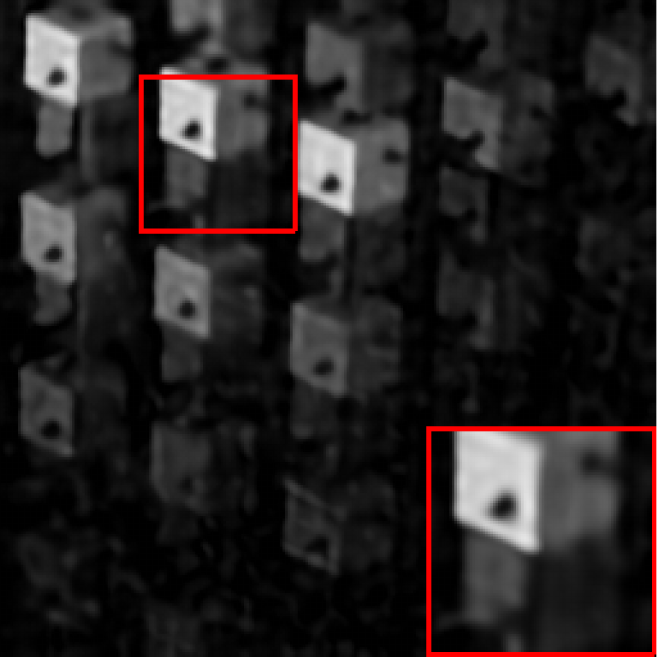}&
            \includegraphics[width=0.26\linewidth]{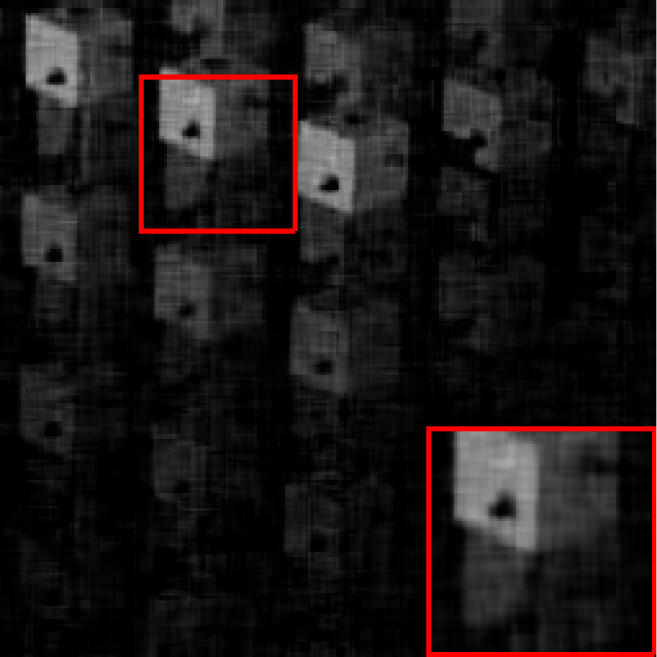}&
            \includegraphics[width=0.26\linewidth]{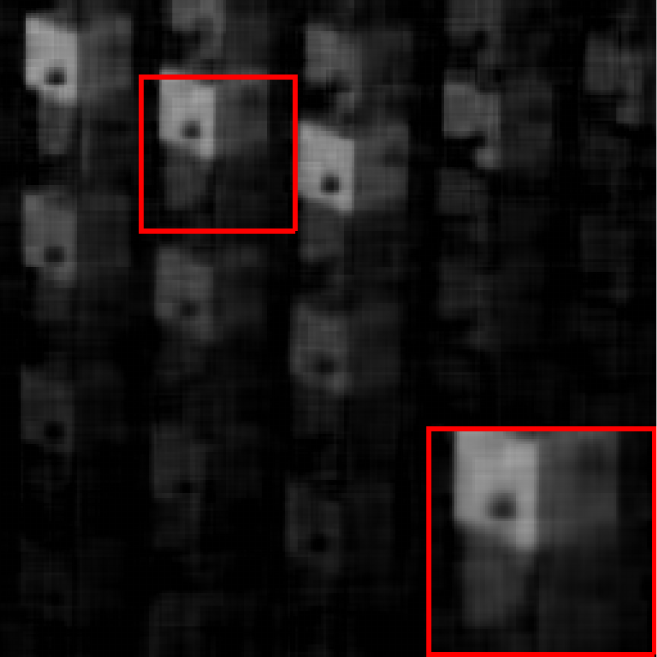}\\
            E3DTV & DIP & DDRM & DDS2M & HLRTF & OTLRM\\
            \cite{E3DTV} & \cite{dip} & \cite{LLRPnP} & \cite{ddrm} & \cite{dds2m} & \cite{lrtf} &
		\end{tabular}} 

		\caption{ The denoising results of \emph{scene02} with \emph{Case}: $\mathcal{N}(0, 0.3)$ by different methods. }
		\label{denoisng_S2}
	\end{center}
\end{figure*}

For MSI denoising, we select ten SOTA methods which contain six model-based methods (NonLRMA\cite{NonLRMA}, TLRLSSTV\cite{TLR_LSSTV}, LLxRGTV\cite{LLxRGTV}, 3DTNN\cite{3DTNN_FW}, LRTDCTV\cite{LRTDCTV}, E3DTV\cite{E3DTV}), two Plug-and-Play (PnP) based methods (DIP\cite{dip}, two diffusion based methods (DDRM\cite{ddrm}, DDS2M\cite{dds2m}) and one generative tensor factorization based method (HLRTF \cite{lrtf}).
Figure \ref{denoisng_S2} shows the denoising results of \emph{scene02} with Gaussian noise $\mathcal{N}(0, 0.3)$.
Compared with the similar tensor network-based method HLRTF, our method is more informative and smoother.

\section{Ablation results}

\subsection{Effect of $r$:}
\begin{figure*}[htb]
	\footnotesize
	\setlength{\tabcolsep}{1pt}
	\begin{center}
 
             \scalebox{1}{
		\begin{tabular}{ccc}
			\includegraphics[width=0.3\textwidth]{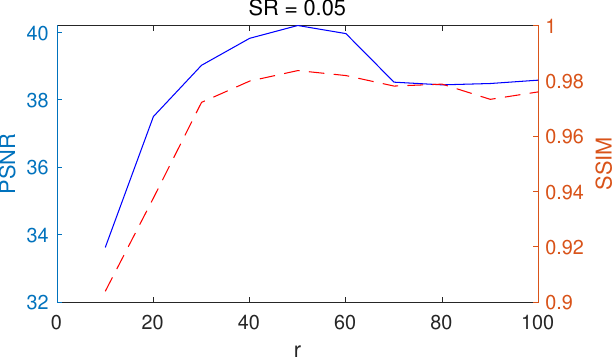} &
			\includegraphics[width=0.3\textwidth]{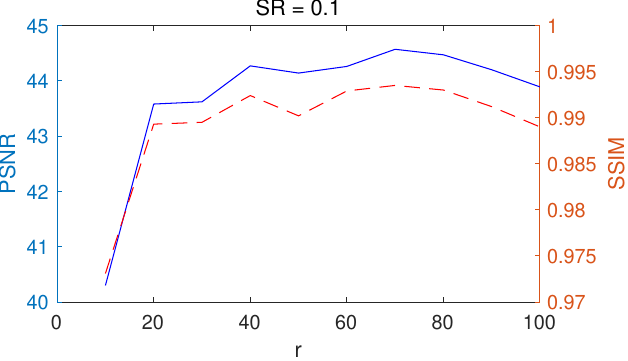} &
            \includegraphics[width=0.3\textwidth]{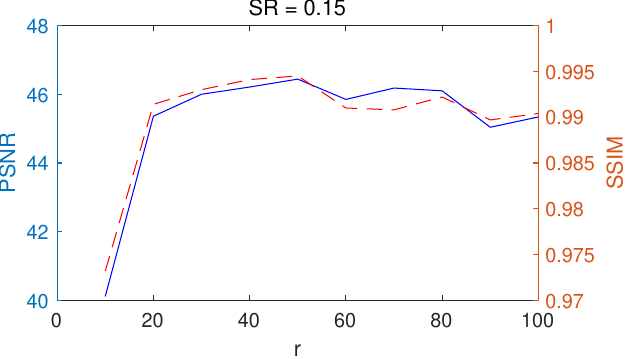}\\
		\end{tabular}} 

		\caption{Effect of $r$ in tensor completion with dataset MSI \emph{Balloons}. From left to right: \emph{SR=0.05}, \emph{SR=0.10}, \emph{SR=0.15}. }
		\label{r_ablation}
	\end{center}
\end{figure*}

Figure \ref{r_ablation} shows visual ablation results of $r$, which controls the rank of the target reconstruction tensor.

\subsection{Effect of $\lambda$:}
\begin{figure*}[htb]
	\footnotesize
	\setlength{\tabcolsep}{1pt}
	\begin{center}
             \scalebox{1}{
		\begin{tabular}{ccc}
			\includegraphics[width=0.3\textwidth]{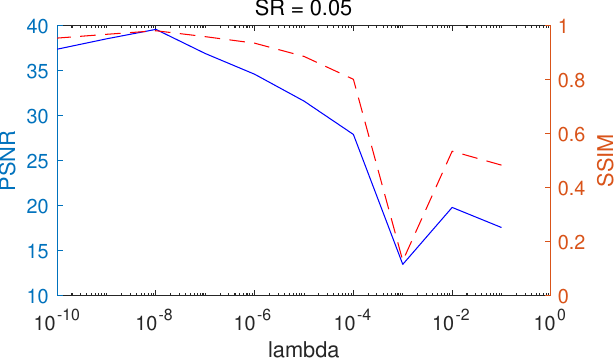} &
			\includegraphics[width=0.3\textwidth]{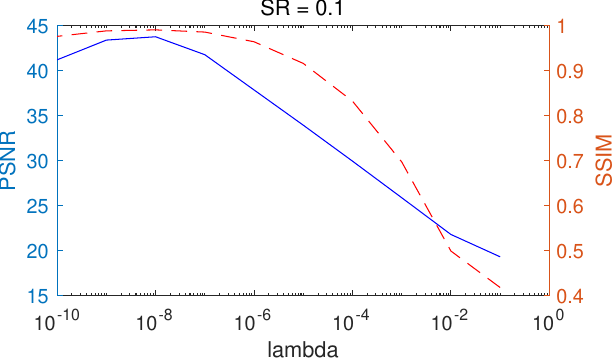} &
            \includegraphics[width=0.3\textwidth]{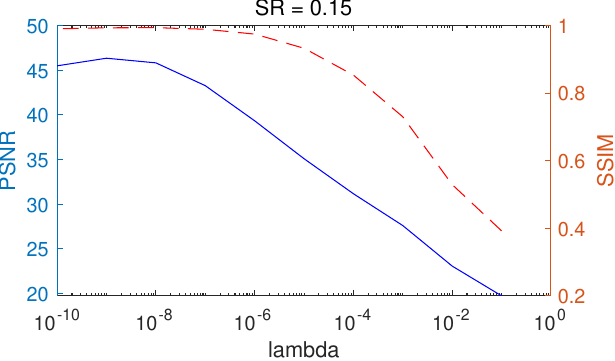}\\
		\end{tabular}} 

		\caption{Effect of $\lambda$ in tensor completion with dataset MSI \emph{Balloons}. From left to right: \emph{SR=0.05}, \emph{SR=0.10} and \emph{SR=0.15}. }
		\label{lambda_ablation}
	\end{center}
\end{figure*}

Figure \ref{lambda_ablation} shows the effect of $\lambda$ in tensor completion for different SR.
$\lambda$ is the hyperparameter which controls the $\mathrm{OTV}(\cdot)$.
The ablation experiments for $\lambda$ without regularisation on \emph{Balloons} are shown in Table \ref{lambda_no_regular}.
Probably due to the magnitude between the loss terms leading to a small $\lambda$.
Also to highlight our novelty in low-rank, we present Table \ref{tab:tv_ablation}, which provides the average PSNR under SRs $\in \{0.05,0.10,0.15\}$ without the $\mathrm{OTV}(\cdot)$ regularization, which promotes the smoothness of the tensor.

\begin{table}
        \centering
        \caption{Ablation for $\lambda$ on \emph{Balloons} in \textbf{tensor completion}.}
        \def\arraystretch{1.0}
        \setlength{\tabcolsep}{1pt}
\scalebox{0.75}{
    \begin{tabular}{c|ccccccccc}
    \bottomrule[0.15em]
    \multirow{2}[2]{*}{$\lambda$} & \multicolumn{2}{c}{\textbf{SR=0.05}} & \multicolumn{2}{c}{\textbf{SR=0.10}} & \multicolumn{2}{c}{\textbf{SR=0.15}} & \multicolumn{2}{c}{\textbf{SR=0.20}} & \multirow{2}[2]{*}{\textbf{Time (s)}}\\
        & \textbf{PSNR}$\uparrow$  & \textbf{SSIM}$\uparrow$  & \textbf{PSNR}$\uparrow$  & \textbf{SSIM}$\uparrow$  & \textbf{PSNR}$\uparrow$  & \textbf{SSIM}$\uparrow$  &
        \textbf{PSNR}$\uparrow$  & \textbf{SSIM}$\uparrow$ &\\
    \hline
         0 & 36.57& 0.94 & \underline{41.73} & \underline{0.98} & 45.28 & \textbf{0.99} & 47.06 & \textbf{0.99} & 118\\
         $10^{-10}$ & \underline{37.37} & \underline{0.95} & 41.19 & \underline{0.98} & \underline{45.49} & \textbf{0.99} & \underline{47.13} & \textbf{0.99} & 128\\
         $10^{-8}$ & \textbf{39.57} & \textbf{0.98} & \textbf{43.75} & \textbf{0.99} & \textbf{45.82} & \textbf{0.99} & \textbf{47.53} & \textbf{0.99} & 127\\
         $10^{-6}$ & 34.59 & 0.89 & 37.85 & 0.92 & 39.34 & 0.93 & 41.13 & 0.96 & 127\\
    \toprule[0.15em]
    \end{tabular}%
    } 
	\label{lambda_no_regular}
\end{table}

\begin{table}
    \centering
    \caption{Comparison with SOTA for the $\mathrm{OTV}$ regularization.}
    \scalebox{0.9}{
    \begin{tabular}{c|cccc}
    \toprule[0.15em]
        \textbf{Method} & HLRTF w/o TV & HLRTF & Ours w/o TV & Ours\\
    \midrule[0.1em]
        \textbf{PSNR} & 40.19 & 41.90 & 43.20 & \textbf{44.78}\\
    \bottomrule[0.15em]
    \end{tabular}}
    \label{tab:tv_ablation}
\end{table}

\subsection{Analysis for the semi-orthogonality of $\mathcal{U}$ and $\mathcal{V}$:}
\begin{figure*}[htb]
	\footnotesize
	\setlength{\tabcolsep}{1pt}
	\begin{center}

    \newcommand{\tabincell}[2]{\begin{tabular}{@{}#1@{}}#2\end{tabular}}
            \scalebox{0.8}{
		\begin{tabular}{ccccccc}
            \includegraphics[width=0.2\linewidth]{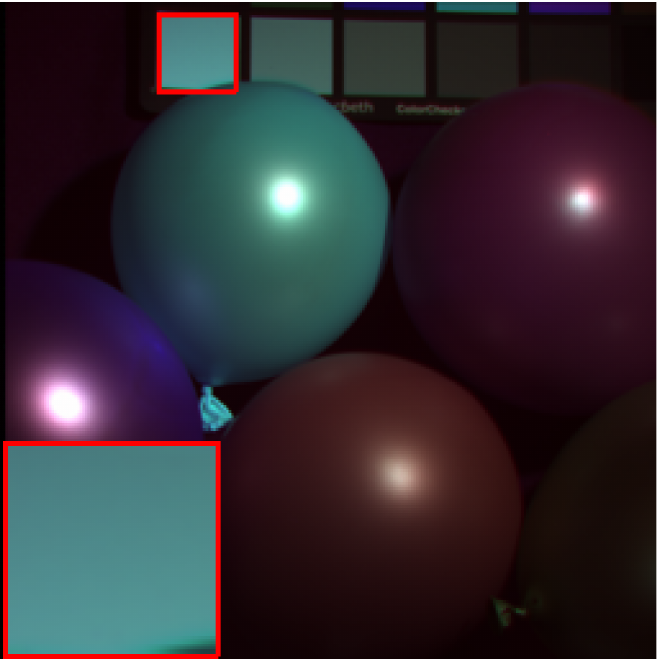}&
            \includegraphics[width=0.2\linewidth]{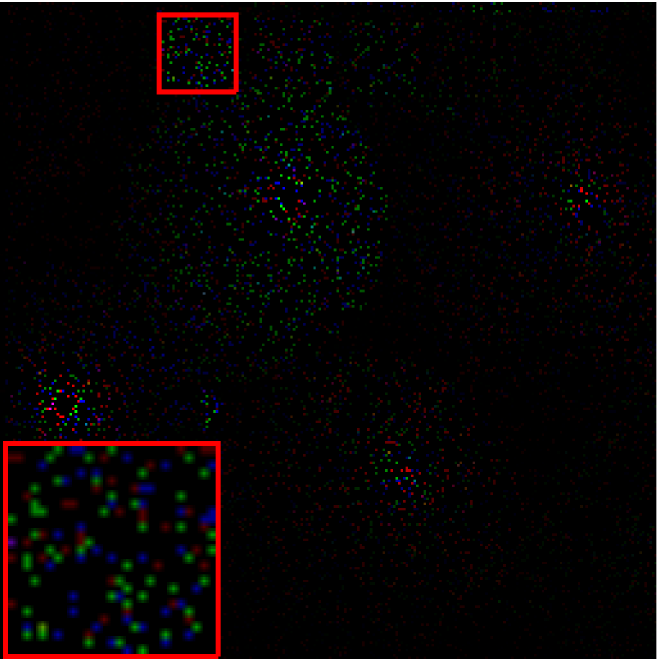}&
			\includegraphics[width=0.2\linewidth]{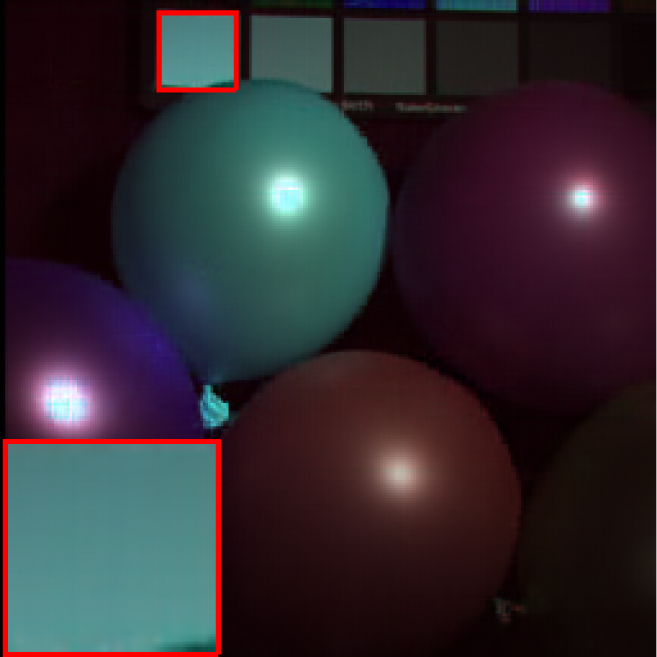}&
			\includegraphics[width=0.2\linewidth]{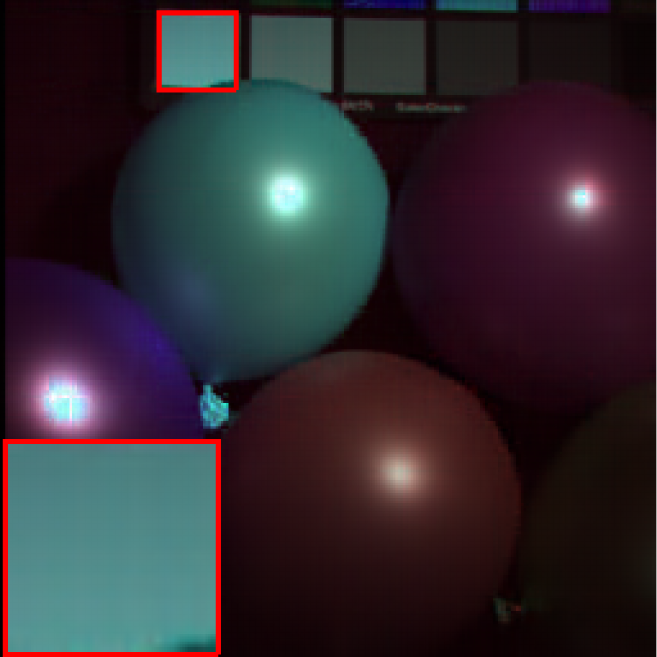}&
			\includegraphics[width=0.2\linewidth]{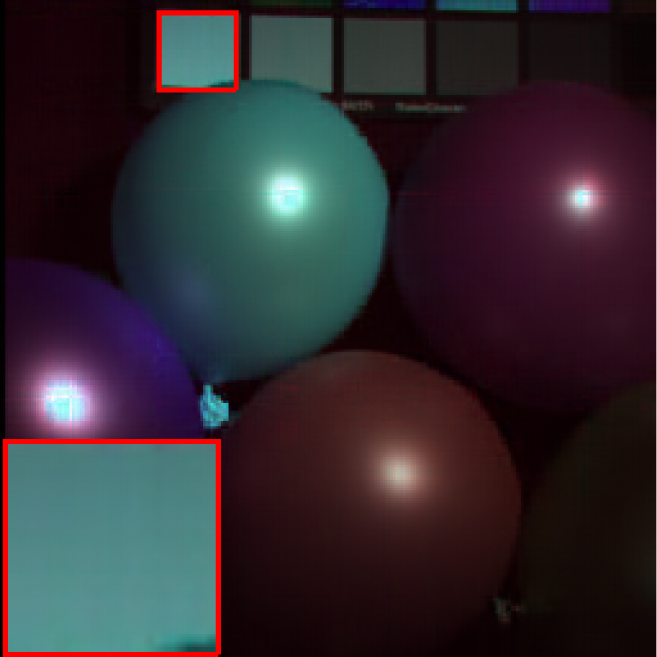}&\\
            \tabincell{c}{Original} & \tabincell{c}{Observed} & \tabincell{c}{$\beta$=0\\ psnr=40.20} & \tabincell{c}{$\beta$=1e-7\\ psnr=40.01} & \tabincell{c}{$\beta$=1e-6\\ psnr=39.72}\\
			\includegraphics[width=0.2\linewidth]{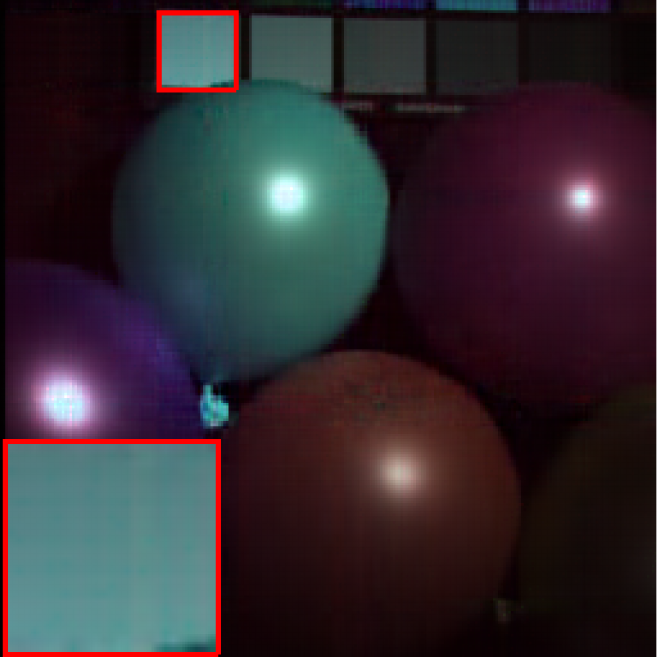}&
            \includegraphics[width=0.2\linewidth]{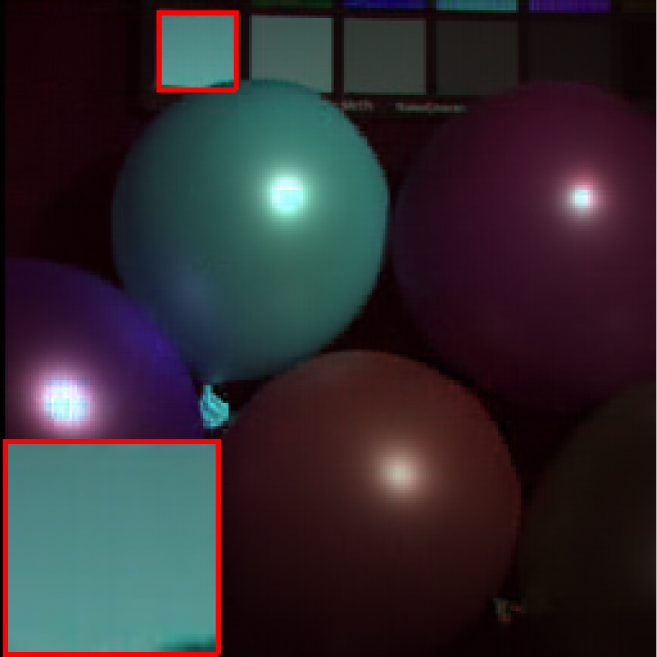}&
            \includegraphics[width=0.2\linewidth]{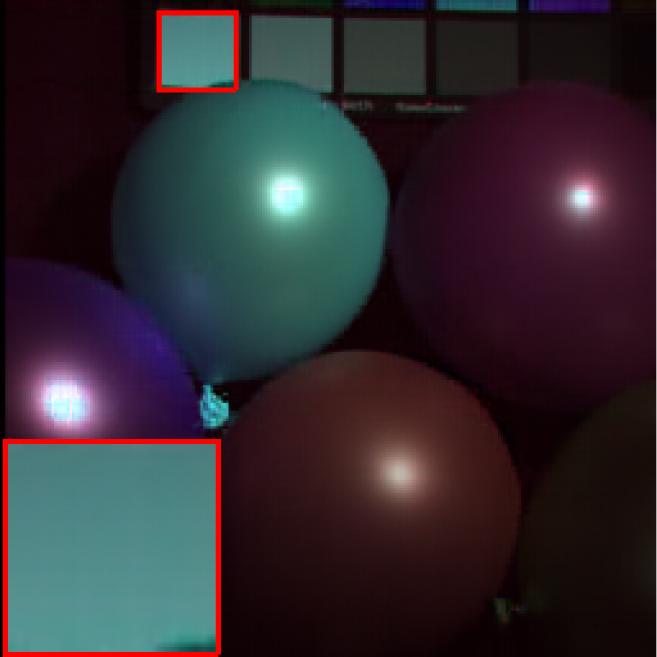}&
            \includegraphics[width=0.2\linewidth]{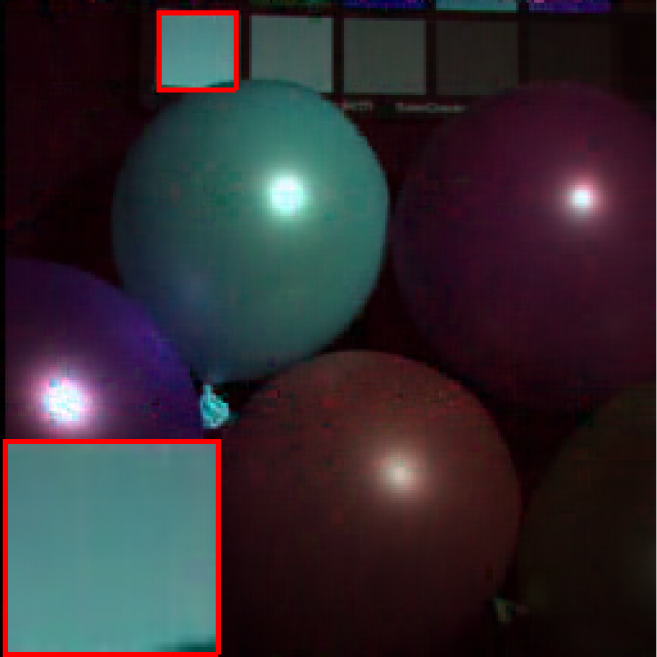}&
            \includegraphics[width=0.2\linewidth]{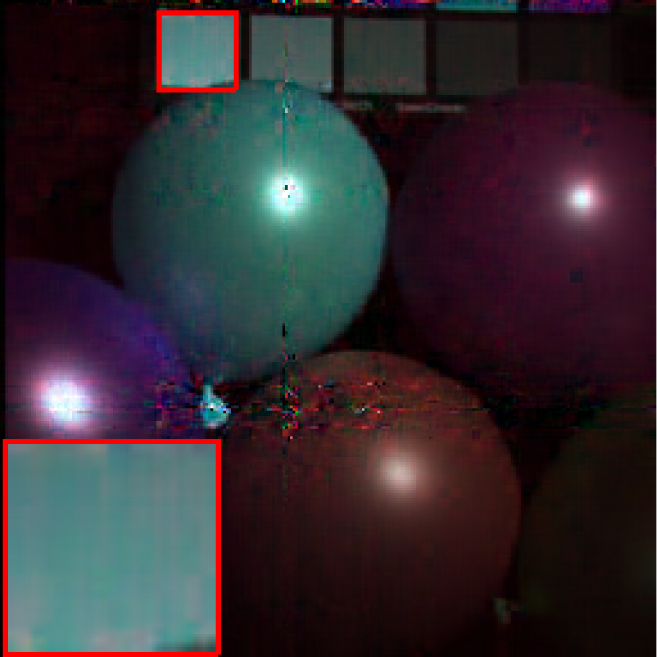}\\
            \tabincell{c}{$\beta$=1e-5\\ psnr=37.26}&\tabincell{c}{$\beta$=1e-4\\ psnr=39.83}&\tabincell{c}{$\beta$=1e-3\\ psnr=39.82}&\tabincell{c}{$\beta$=1e-2\\ psnr=38.15}&\tabincell{c}{$\beta$=1e-1\\ psnr=29.70}
		\end{tabular}} 

		\caption{ Semi-orthogonality loss results of \emph{Balloons} in tensor completion with \emph{SR=0.05}. }
		\label{ablation_utuvtv}
	\end{center}
\end{figure*}

Figure \ref{ablation_utuvtv} shows the tensor completion performance with $\beta$ which controls the strength of the semi-orthogonality of the tensor factors $\mathcal{U}$ and $\mathcal{V}$.

\subsection{Effect of $\rho(\cdot)$:}
\begin{figure*}[htb]
    \centering
    \scalebox{1}{
		\begin{tabular}{cc}
			\includegraphics[width=0.45\linewidth]{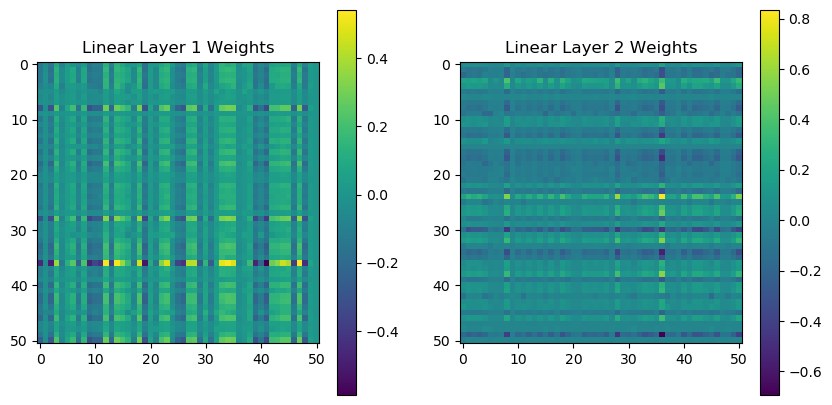}&
			\includegraphics[width=0.45\linewidth]{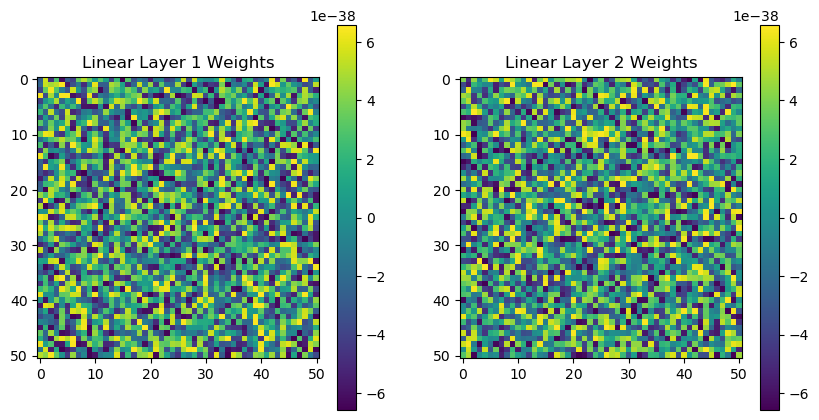}\\
            Weights after training & Weights before training\\
		\end{tabular}} 
    \caption{Visualization of the weight matrices in $\rho(\cdot)$.}
    \label{fig:weights_visualization}
\end{figure*}

For the rank information extractor $\rho(\cdot)$, the visualization results of the two weight matrices are shown in the Figure \ref{fig:weights_visualization}.

\subsection{Comparisons on orthogonal transform $\mathbf{L}$:}
Table \ref{comparison} shows the evaluation results of above methods.
We also add Table \ref{tab:ablation_transforms}, which presents the average PSNR results of different SRs ($0.05,0.10, 0.15$) for MSI tensor completion task.
The proposed learnable and orthogonal transform achieve better performance compared with the learnable but linear transform, also the version without any transform.

\begin{table}
    \centering
    \caption{Average PSNR of SR $\in \{0.05,0.10,0.15\}$ in \textbf{tensor completion} for different types of transform.}
    \begin{tabular}{ccccc}
    \toprule[0.15em]
        \textbf{Type} & $k=0$ & $k=1$ & $k=2$ & $k=3$\\
    \midrule[0.1em]
        w/o any transform & 30.20 & 30.56 & 34.50 & 35.51\\
        linear transform & 36.56 & 39.10 & 41.78 & 41.68\\
        orthogonal transform & \textbf{39.37} & \textbf{42.31} & \textbf{43.82} & \textbf{43.85}\\
    \bottomrule[0.15em]
    \end{tabular}
    \label{tab:ablation_transforms}
\end{table}

\subsection{Initialization analysis:}

We analyze different initialization strategies, and the initial number is selected from small to large, we provide the average results of SRs $\in \{0.1,0.15,0.2,0.3\}$ for each initialization in Table \ref{tab:inits}.
When the initialization is too large, the results slightly drop. 
For a more stable result, the initialization should be better between 0 and 1.

\begin{table}
    \centering
    \caption{PSNR under different initialization strategies.}
    \scalebox{0.88}{
    \begin{tabular}{cccccccc}
    \toprule[0.15em]
         \textbf{Inits}&0&1e-4&1e-2&1e-1&1&10&100 \\
    \midrule[0.1em]
         \textbf{Avg}&31.47& 31.33& 31.45& \underline{31.67}& \textbf{32.05}& 31.32& 31.19 \\
    \bottomrule[0.15em]
    \end{tabular}}
    \label{tab:inits}
\end{table}

\end{document}